\documentclass[numbers]{article}
\usepackage[final]{neurips_2022}

\usepackage{stfloats}
\usepackage{graphicx}
\usepackage{subcaption}
\usepackage{amsmath}
\usepackage{amssymb}
\usepackage{booktabs}
\usepackage{rotating}
\usepackage{pdflscape}
\usepackage{adjustbox}
\usepackage{xcolor}
\definecolor{darkblue}{rgb}{0.21,0.49,0.74}
\usepackage[pagebackref,breaklinks,colorlinks]{hyperref}
\hypersetup{
    colorlinks=true,
    linkcolor=blue,    
    citecolor=darkblue,     
    urlcolor=blue      
}
\usepackage[capitalize]{cleveref}
\usepackage{makecell}
\usepackage{multirow}
\usepackage{pifont}
\setcitestyle{square}
\setcitestyle{citesep={,}}
\usepackage{longtable}
\usepackage{ltablex}
\keepXColumns
\newcolumntype{Y}{>{\raggedright\arraybackslash}X} 
\usepackage{pdflscape}
\usepackage{longtable}
\usepackage{booktabs}
\usepackage{array}
\usepackage{makecell}
\usepackage{ragged2e}
\usepackage{hyperref} 
\usepackage{threeparttable}
\usepackage{ragged2e}
\usepackage{xspace}
\usepackage{placeins}

\usepackage[T1]{fontenc}
\renewcommand{\arraystretch}{0.92}
\crefname{section}{Sec.}{Secs.}
\Crefname{section}{Section}{Sections}
\Crefname{table}{Table}{Tables}
\crefname{table}{Tab.}{Tabs.}

\newcommand{\eg}{\textit{e.g.}\xspace}
\newcommand{\ie}{\textit{i.e.}\xspace}

\newcommand{\etc}{\textit{etc.}\xspace}

\newcounter{datasetidx} 
\newcommand{\datasetidx}{\refstepcounter{datasetidx}\thedatasetidx}
\newcounter{rownum}

\newcommand{\dataref}[1]{(\#\ref{#1})}

\title{Project Imaging-X: \\ A Survey of 1000+ Open-Access Medical Imaging Datasets for Foundation Model Development}

\author{
Zhongying Deng$^{1,5,\ast}$ \and
Cheng Tang$^{1,3,\ast}$ \and
Ziyan Huang$^{1,\ast}$ \and
Jiashi Lin$^{1,\ast}$ \and
Ying Chen$^{1,\ast}$ \and
Junzhi Ning$^{1,\ast}$ \and
Chenglong Ma$^{2,4,\ast}$ \and
Jiyao Liu$^{1,4}$ \and
Wei Li$^{1,6}$ \and
Yinghao Zhu$^7$
% ...
\\ \\
\small
\parbox{\linewidth}{\centering 
$^1$Shanghai Artificial Intelligence Laboratory; 
$^2$Shanghai Innovation Institute; 
$^3$Shanghai Institute of Optics and Fine Mechanics; 
$^4$Fudan University; 
$^5$University of Cambridge; 
$^6$Shanghai Jiao Tong University; 
$^7$The University of Hong Kong; 
% ...
}
\thanks{Equal contribution \quad $^\dagger$Corresponding author}
}

\begin{document}

\maketitle

\newpage
\begin{abstract}
Foundation models have demonstrated remarkable success across diverse domains and tasks, primarily due to the thrive of large-scale, diverse, and high-quality datasets. However, in the field of medical imaging, the curation and assembling of such medical datasets are highly challenging due to the reliance on clinical expertise and strict ethical and privacy constraints, resulting in a scarcity of large-scale unified medical datasets and hindering the development of powerful medical foundation models. 
In this work, we present the largest survey to date of medical image datasets, covering over 1,000 open-access datasets with a systematic catalog of their modalities, tasks, anatomies, annotations, limitations, and potential for integration. 
Our analysis exposes a landscape that is modest in scale, fragmented across narrowly scoped tasks, and unevenly distributed across organs and modalities, which in turn limits the utility of existing medical image datasets for developing versatile and robust medical foundation models. 
To turn fragmentation into scale, we propose a metadata-driven fusion paradigm (MDFP) that systematically integrates public datasets with shared modalities or tasks, thereby transforming multiple small data silos into larger, more coherent resources. 
Building on MDFP, we release an \href{https://tchenglv520.github.io/medical-dataset-browser/}{interactive discovery portal} that enables end-to-end, automated medical image dataset integration, and compile all surveyed datasets into a unified, structured table that clearly summarizes their key characteristics and provides reference links, offering the community an accessible and comprehensive repository.
By charting the current terrain and offering a principled path to dataset consolidation, our survey provides a practical roadmap for scaling medical imaging corpora, supporting faster data discovery, more principled dataset creation, and more capable medical foundation models for the biomedical imaging research community. Our project repository can be found at \url{https://github.com/uni-medical/Project-Imaging-X}.

\end{abstract}

\newpage
\tableofcontents
\newpage

\section{Introduction}\label{sec:intro}

\begin{figure}
    \centering
    \includegraphics[width=\linewidth]{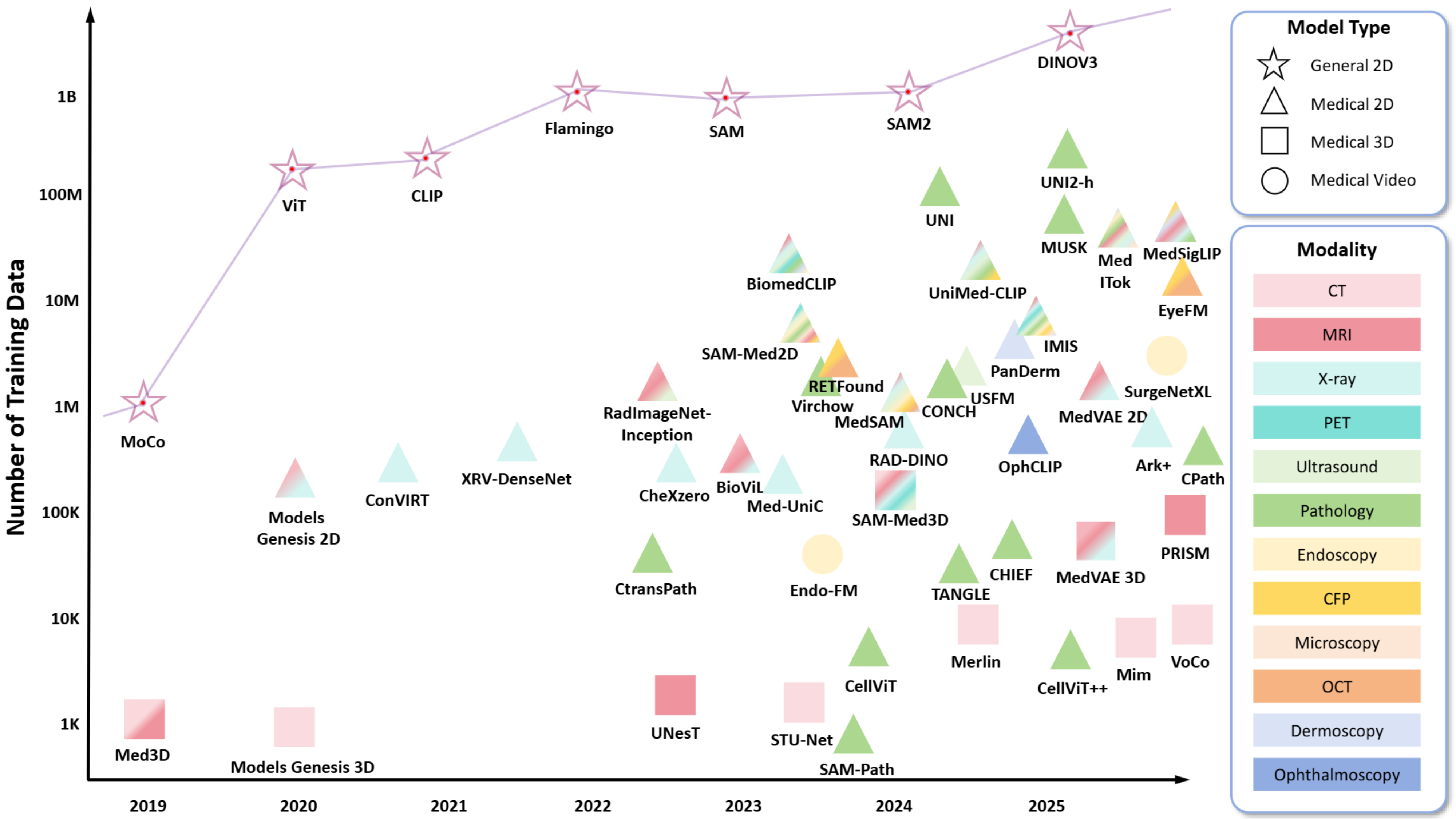}
    \caption{Evolution of medical foundation models and general domain foundation models. Medical foundation models are mostly trained using millions of images, while advanced general domain ones are trained using billions of natural images. Additionally, most medical foundation models cover only a few modalities like CT and MRI, which may introduce modality-specific bias that constrains clinical applicability.
    }
    \label{fig:medfm_overview}
\end{figure}

Medical imaging foundation models hold the promise of significantly advancing clinical decision-making by analyzing diverse medical imaging modalities and executing multiple tasks through a single, pre-trained system. This paradigm parallels the trajectory of advanced models in the domain of natural language processing~\cite{achiam2023gpt} and computer vision~\cite{clip,zhang2020contrastive,simeoni2025dinov3,sam}, 
which are trained on extensive and diverse datasets to achieve broad generalization across tasks and applications~\cite{sellergren2025medgemma,ma2024segment,chen2024uni,hu2025survey,zhang2025llms4all}, as depicted in Figure~\ref{fig:medfm_overview}.
This highlights a similar shift in medical AI from narrow, single-modality, task-specific models toward multi-modal, multi-functional foundation, which could better reflect the complexity of clinical workflows and enhance utility across specialties~\cite{moor2023foundation}. 
Despite this potential, current medical imaging foundation models, such as STUNet~\cite{huang2023stu}, MedSAM~\cite{ma2024segment}, SAM-Med3D~\cite{wang2025sam}, SAM-Brain3D~\cite{deng2025brain} and PanDerm~\cite{yan2025multimodal}, are often tailored to well-represented settings, such as a few modalities like computed tomography (CT) and magnetic resonance imaging (MRI), a narrow set of tasks (\eg, segmentation), or limited anatomical regions (\eg, brain, abdomen). 
Many clinically valuable settings remain less covered, which introduces modality-, task-, and anatomy-specific biases that constrain generalization and clinical applicability.

\begin{figure}
    \centering
    \includegraphics[width=\linewidth]{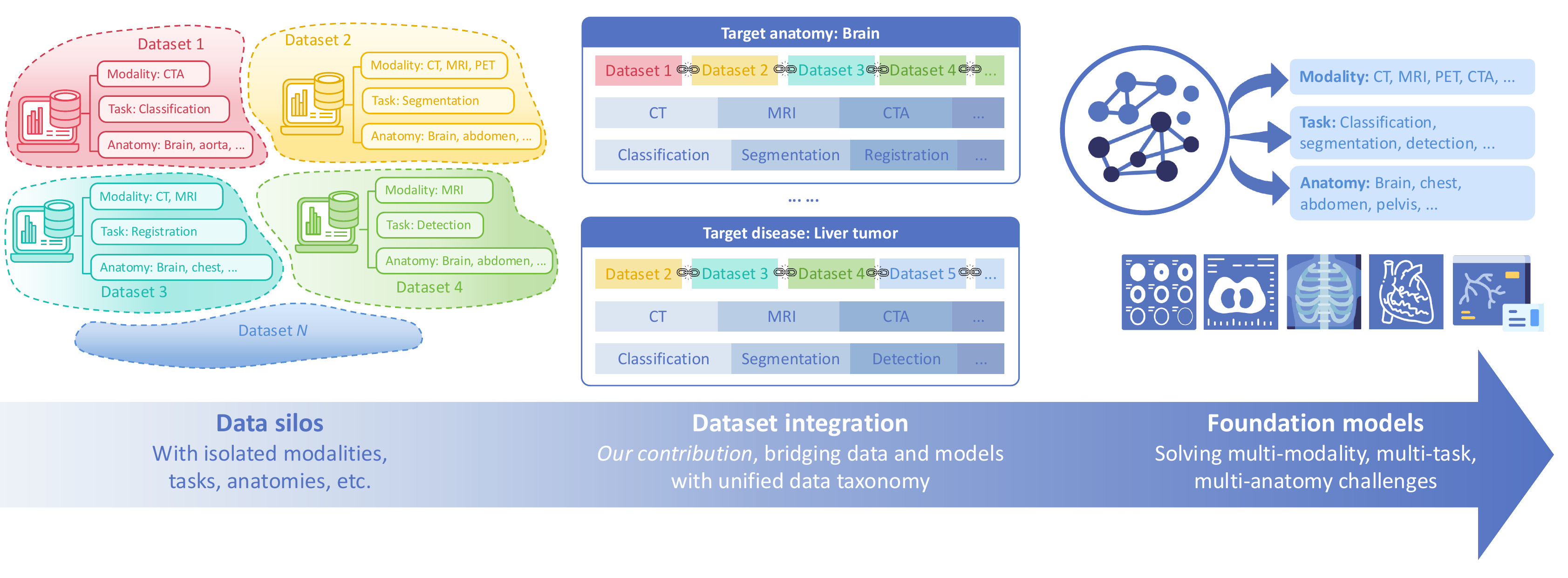}
    \caption{
    Conceptual overview of moving from fragmented medical image data to integrated resources for medical foundation models. Our survey addresses the data fragmentation issue in public medical image datasets by introducing a metadata-driven dataset integration paradigm, which is crucial for the development of advanced foundation models that can tackle multi-modality, multi-task, and multi-anatomy challenges effectively, ultimately enhancing clinical AI applications. 
    }
    \label{fig:survey_feature}
\end{figure}

The root challenge lies in data availability and diversity~\cite{islam2025foundation,bai2025intern}. Most public medical datasets contain only thousands of images, \eg, BraTS series~\cite{crimi2021brainlesion,menze2014multimodal}, which are orders of magnitude smaller than natural image datasets with billions of samples, such as Segment Anything 1 Billion (SA-1B)~\cite{sam} and LAION-5B~\cite{schuhmann2022laion}. This substantial difference in the number of training images between the natural image (or general) domain and the medical one is further depicted in Figure~\ref{fig:medfm_overview}.
Constructing large, diverse medical datasets is resource-intensive, requiring specialized imaging equipment, expert annotations, and careful navigation of ethical and privacy constraints. 
Consequently, the current dataset landscape is \emph{highly fragmented}, with data scattered across isolated, narrowly scoped collections~\cite{willemink2020preparing}. 
This fragmentation not only limits pre-training scale, but also overlooks opportunities to integrate related datasets into richer, more balanced training resources.

A promising direction emerging in recent research is dataset integration~\cite{ye2023sa,ye2024gmai}, where multiple smaller datasets with shared modalities, anatomies, or tasks are merged into unified large-scale resources. As shown in Figure~\ref{fig:survey_feature}, the merged datasets can bridge data and models, facilitating the development of foundation models~\cite{haghighi2025learning}. 
While this strategy has shown potential, existing efforts typically focus on specific imaging types or organ systems~\cite{combalia2022validation,huang2025aeval,wu2025mswal}. 
Furthermore, when lacking guidance from a comprehensive overview of available datasets, 
dataset integration risks reinforcing existing biases rather than enabling balanced, general-purpose foundation model development.

Given these challenges, a comprehensive survey of medical imaging datasets is urgently needed. Such a survey can illustrate gaps in data coverage, highlight opportunities for dataset integration, and establish a standardized framework for dataset selection and evaluation, which are crucial for the development of robust medical foundation models. 
A few prior surveys have reviewed medical imaging datasets~\cite{li2021systematic,khan2021global, wen2022characteristics, tafavvoghi2024publicly, dishner2024survey}, yet they often lack subject- and image-level statistics, omit many recently released large-scale datasets such as TotalSegmentor~\cite{totalsegmentator} and AbdomenAtlas~\cite{qu2023abdomenatlas}, and do not provide a systematic framework that links dataset characteristics to the requirements of foundation model development. 

To address these limitations, we present the most comprehensive review to date of over 1,000 open-access medical imaging datasets published between 2000 and 2025. 
We introduce a novel taxonomy to organize datasets by modality, anatomy, task, and label availability. Leveraging this taxonomy, we conduct a gap analysis to identify underrepresented modalities, tasks, and anatomies, establishing clear priorities for future dataset creation. 
Building on these insights, we further propose a metadata-driven fusion paradigm (MDFP) for integrating existing datasets, incorporating it into our interactive discovery portal\footnote{\url{https://tchenglv520.github.io/medical-dataset-browser/}} that enables end-to-end process of fine-grained search, statistical analysis, and dataset integration. 
We conclude with a forward-looking discussion on the challenges and opportunities toward building truly general-purpose medical imaging foundation models.
%

Our main contributions are summarized as follows:
\begin{itemize}
    \vspace{-1.5mm}
    \item \textbf{Comprehensive large-scale survey}: We provide the most extensive review to date, covering over 1,000 open-access medical image datasets released over the past 25 years, accompanied by standardized and detailed metadata.  
    \vspace{-1.5mm}
    \item \textbf{Integration paradigm}: We establish a structured taxonomy and present a metadata-driven fusion paradigm (MDFP), effectively scaling-up existing medical imaging data for medical foundation model development by integrating datasets with shared characteristics. 
    \vspace{-1.5mm}
    \item \textbf{Interactive discovery portal}: Based on the unified taxonomy and the MDFP, we build an interactive discovery portal that enables automated and fine-grained dataset search, integration, and statistical analyses by modality, anatomy, task, and label type. 
    \vspace{-1.5mm}
    \item \textbf{Gap analysis}: We identify underrepresented modalities, anatomical regions, and tasks, highlighting critical limitations that hinder the development of future foundation models. 
    \vspace{-1.5mm}
    \item \textbf{Accessible community resource}: We release the portal, all surveyed dataset information, related Python toolkit, and a merged large-scale dataset for public use, 
    offering a transparent and practical resource for the research community.
\end{itemize}

\begin{figure}[ht] 
    \centering  \includegraphics[width=1.0\textwidth]{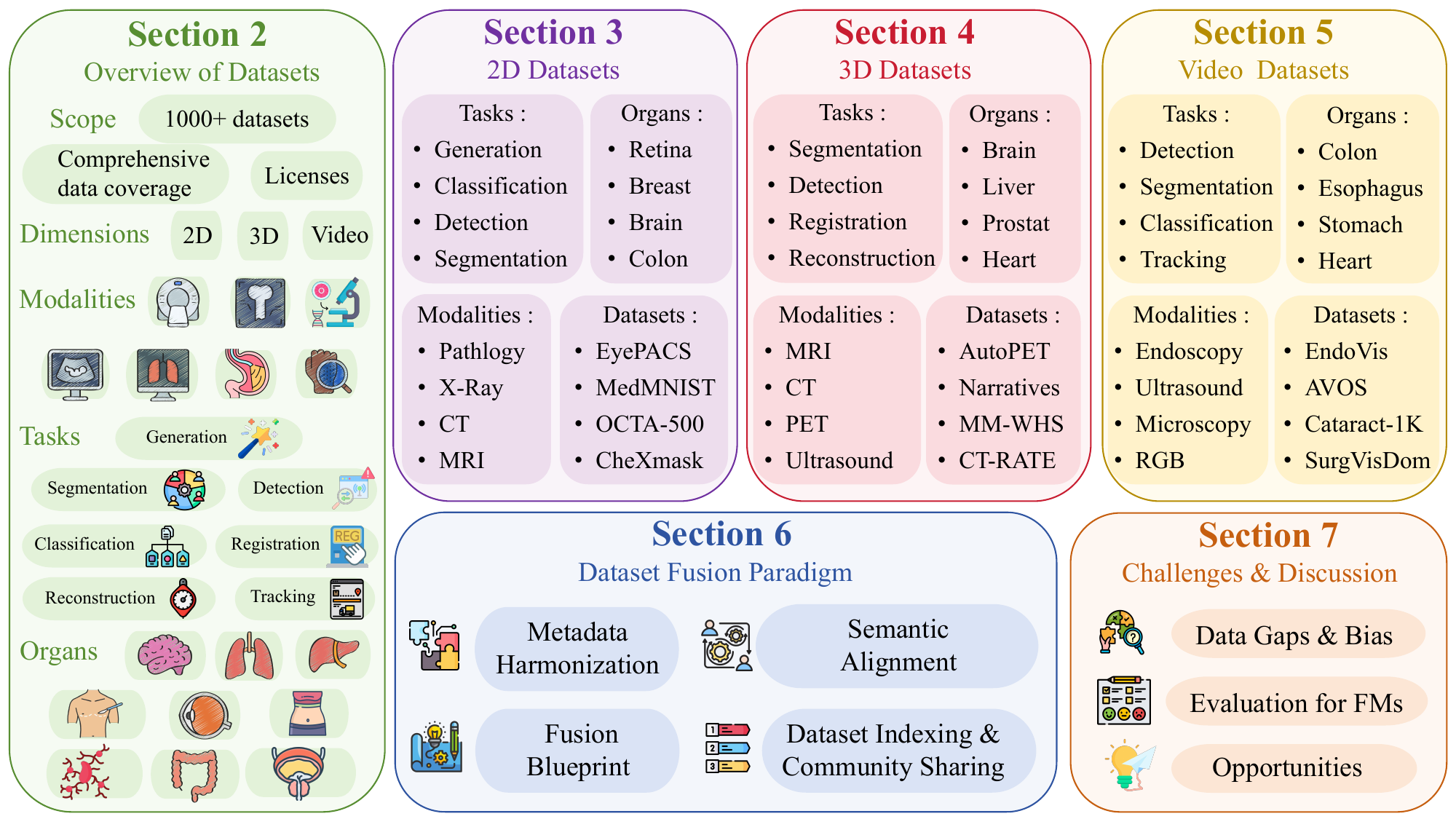} 
    \caption{Overview of the survey. We first introduce the overview of the medical imaging datasets, followed by three sections detailing 2D, 3D, and video datasets. We further implement integration strategies to merge the datasets for large-scale resources, which can potentially be leveraged for the development of foundation models. Finally, we discuss the challenges for foundation model development.  
    } 
    \label{fig:content_structure} 
\end{figure}

The remainder of this paper is organized as demonstrated in Figure~\ref{fig:content_structure}. Section~\ref{sec:overview} offers a high-level panorama of the landscape of over 1,000 open-access medical image datasets, analyzing their distribution across modalities, tasks, and anatomical regions. Section~\ref{sec:2d_data} zooms in on two-dimensional (2D) image datasets, providing a modality-specific breakdown and revealing extreme fragmentation and a long-tail distribution. Section~\ref{sec:3d_data} covers three-dimensional (3D) volumetric datasets, focusing on their unique clinical value and challenges of high cost and annotation complexity. Section~\ref{sec:video_data} reviews video datasets, highlighting their role in spatiotemporal analysis. To address the pervasive data fragmentation, Section~\ref{sec:merge_paradigm} introduces our Metadata-Driven Fusion Paradigm (MDFP), a systematic workflow for integrating disparate datasets, and the corresponding interactive discovery portal for automated and effective dataset integration. Section~\ref{sec:discussion} discusses broader challenges and future directions. 
Finally, Section~\ref{sec:conclusion} concludes the survey.

\section{An Overview of Medical Image Datasets}
\label{sec:overview}

\begin{figure}[ht] 
    \centering  \includegraphics[width=1.0\textwidth]{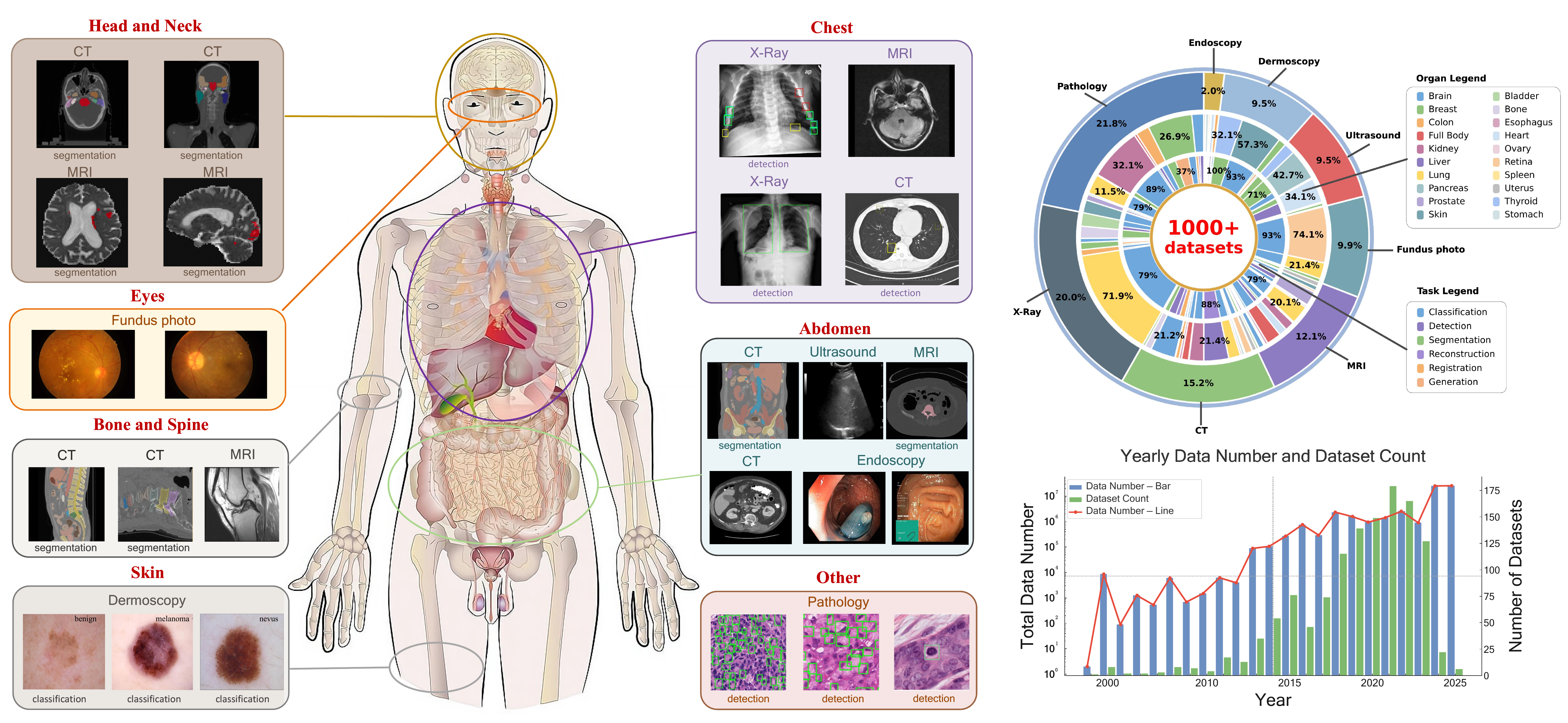} 
    \caption{Overview of medical imaging datasets: representative modalities by anatomical region (left), dataset distribution across modalities, anatomical regions, and tasks (upper right), and temporal trends in dataset numbers (lower right).
    } 
    \label{fig:overview} 
\end{figure}

This section provides an overview of 1000+ medical image datasets released between 2000 and 2025, covering diverse anatomical structures, modalities, and tasks as illustrated in Figure~\ref{fig:overview}. 
These datasets are compiled from major public repositories (The Cancer Imaging Archive\footnote{\url{https://www.cancerimagingarchive.net}}, \etc) and recent challenge sites (Grand Challenge\footnote{\url{https://grand-challenge.org}}, \etc) followed by deduplication, manual verification of landing pages/licences, and metadata normalization, ensuring comprehensive coverage. We leave details of dataset collection process in Section~\ref{sec:merge_paradigm}. 

\begin{figure}[ht] 
    \centering  \includegraphics[width=0.65\textwidth]{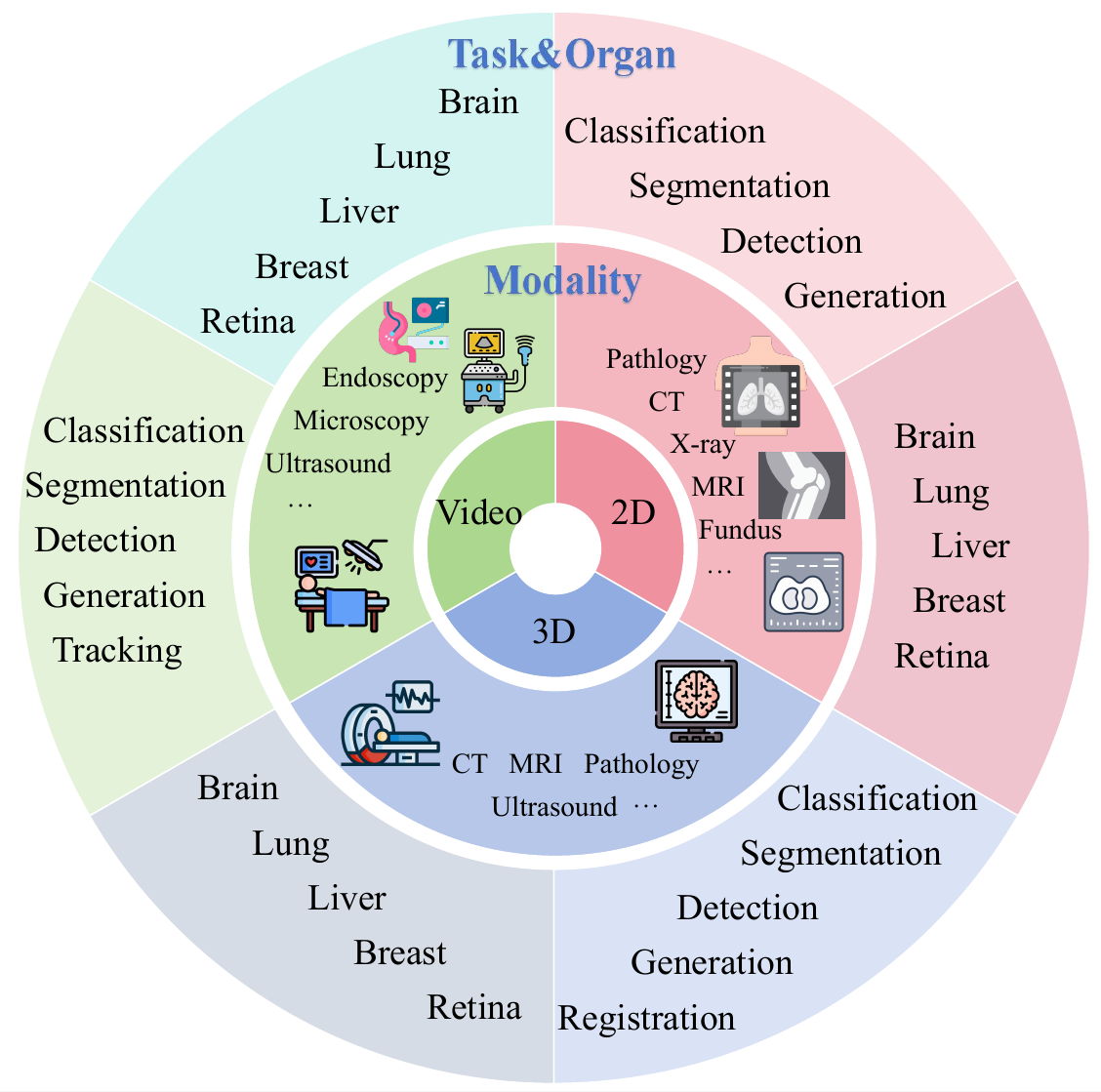} 
    \caption{Taxonomy of medical imaging datasets across data dimensions, modalities, tasks, and anatomical organs. 
    } 
    \label{fig:Taxonomy} 
\end{figure}

To better organize the landscape, we adopt the \textbf{taxonomy} in Figure~\ref{fig:Taxonomy}. Specifically, we begin by grouping medical imaging datasets by \emph{imaging dimensionality} (2D, 3D, and video). 
Within each dimensionality, we further categorize datasets by imaging \emph{modality}. Finally, within each modality, we subcategorize datasets by \emph{task} (\eg, segmentation, classification) and \emph{anatomical region}. This provides a comprehensive basis for the analyses below, and aligns well with foundation model training needs, where dimensionality influences backbone architectural design, modality reflects acquisition physics and clinical use, task determines supervision signals and anatomical diversity shapes generalization in clinical practice.

The resulting manifest underpins all figures in this section. We are particularly interested in the number of images in these datasets (see the right panel of Figure~\ref{fig:overview}), as it strongly influences the effectiveness of foundation model pre-training. Following this principle, we first present the total growth over time and then analyze distributions by imaging dimensionality, modality, task, and organ.

\begin{figure}[htbp]
  \centering
  \begin{subfigure}[b]{0.45\textwidth}
    \centering
    \includegraphics[width=\linewidth]{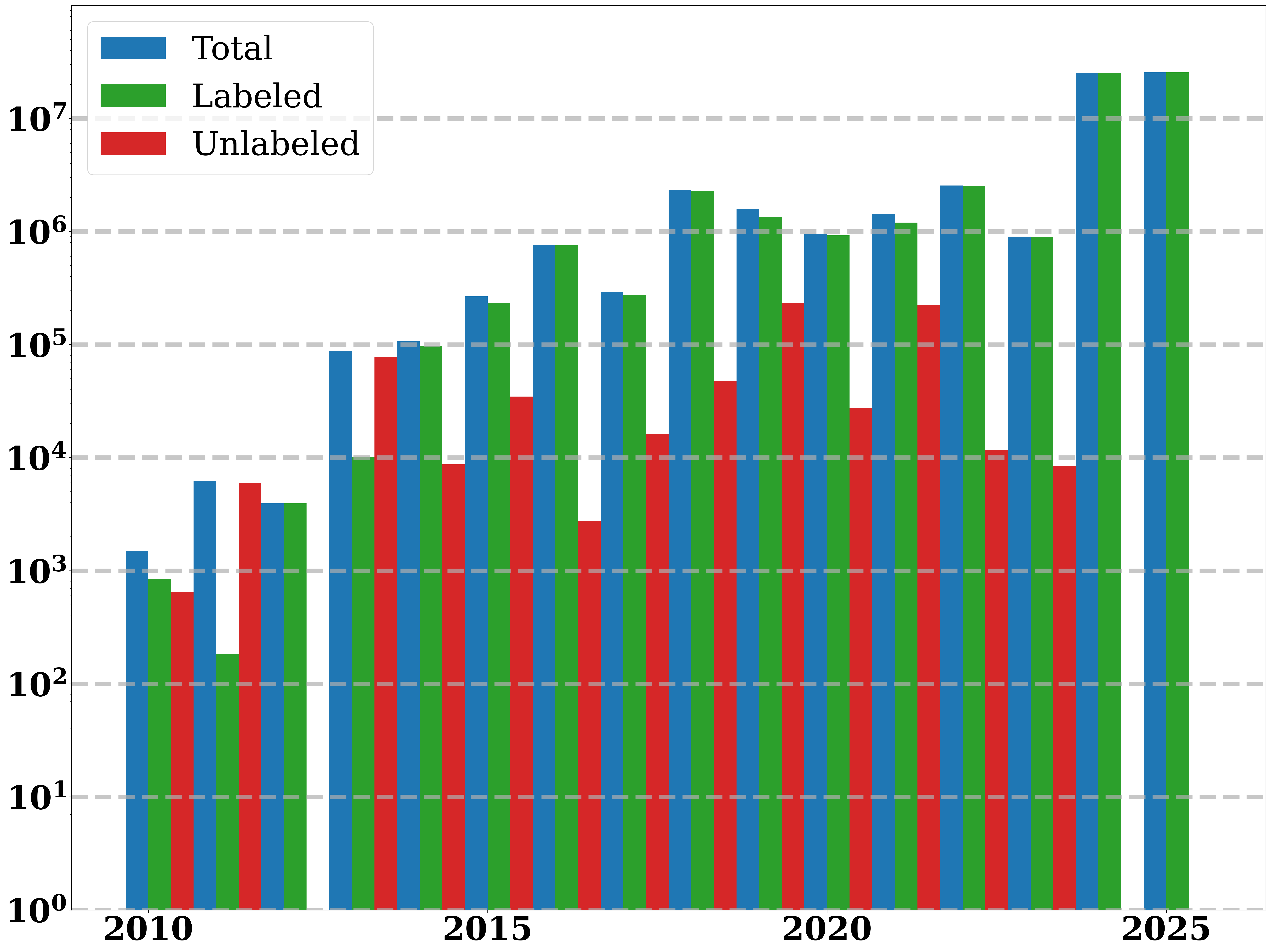}
    \caption*{\textbf{(a)}}
  \end{subfigure}
  \hfill
  \begin{subfigure}[b]{0.45\textwidth}
    \centering
    \includegraphics[width=\linewidth]{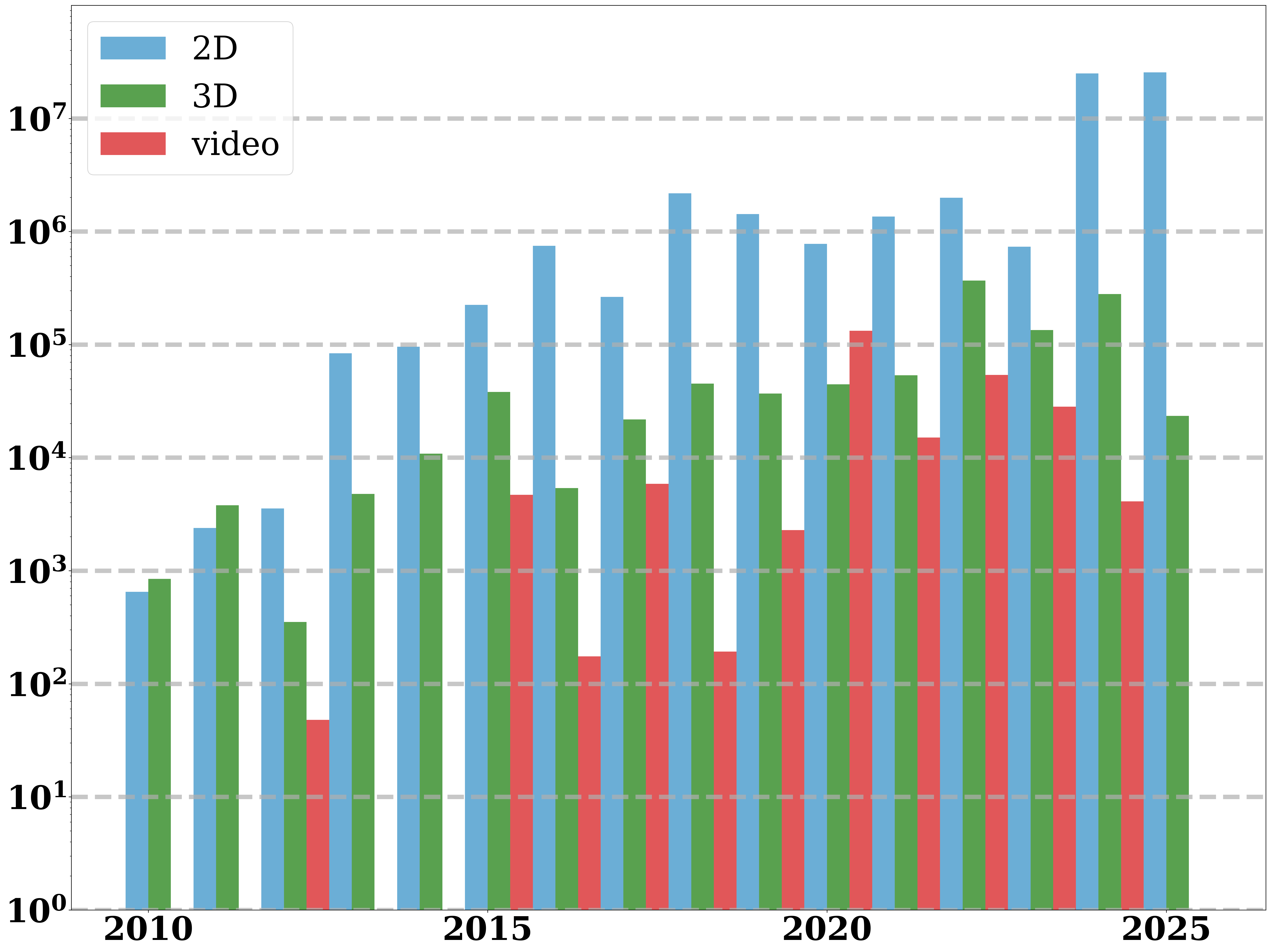}
    \caption*{\textbf{(b)}}
  \end{subfigure}

  \vspace{0.5em}

  \begin{subfigure}[b]{0.45\textwidth}
    \centering
    \includegraphics[width=\linewidth]{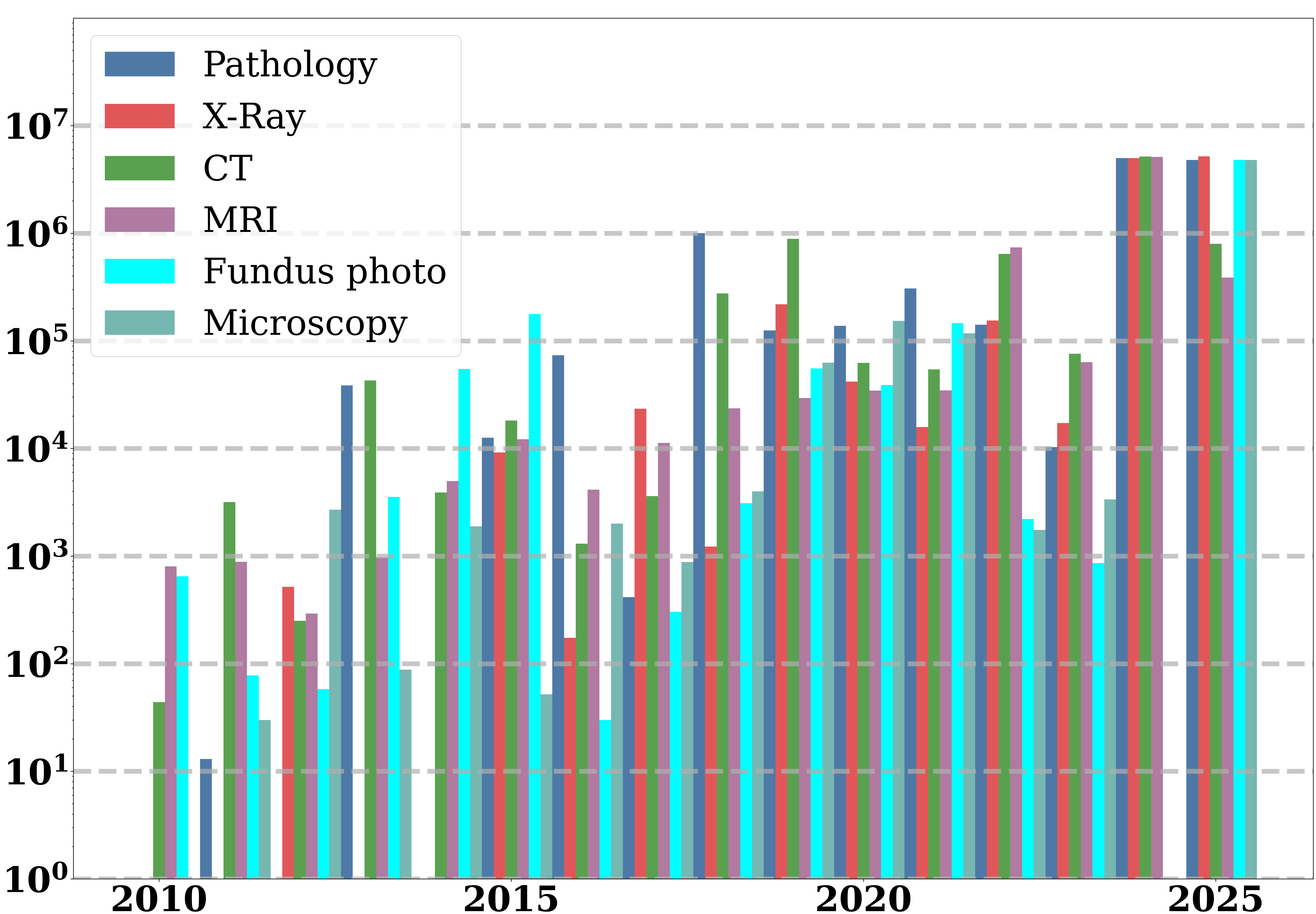}
    \caption*{\textbf{(c)}}
  \end{subfigure}
  \hfill
  \begin{subfigure}[b]{0.45\textwidth}
    \centering
    \includegraphics[width=\linewidth]{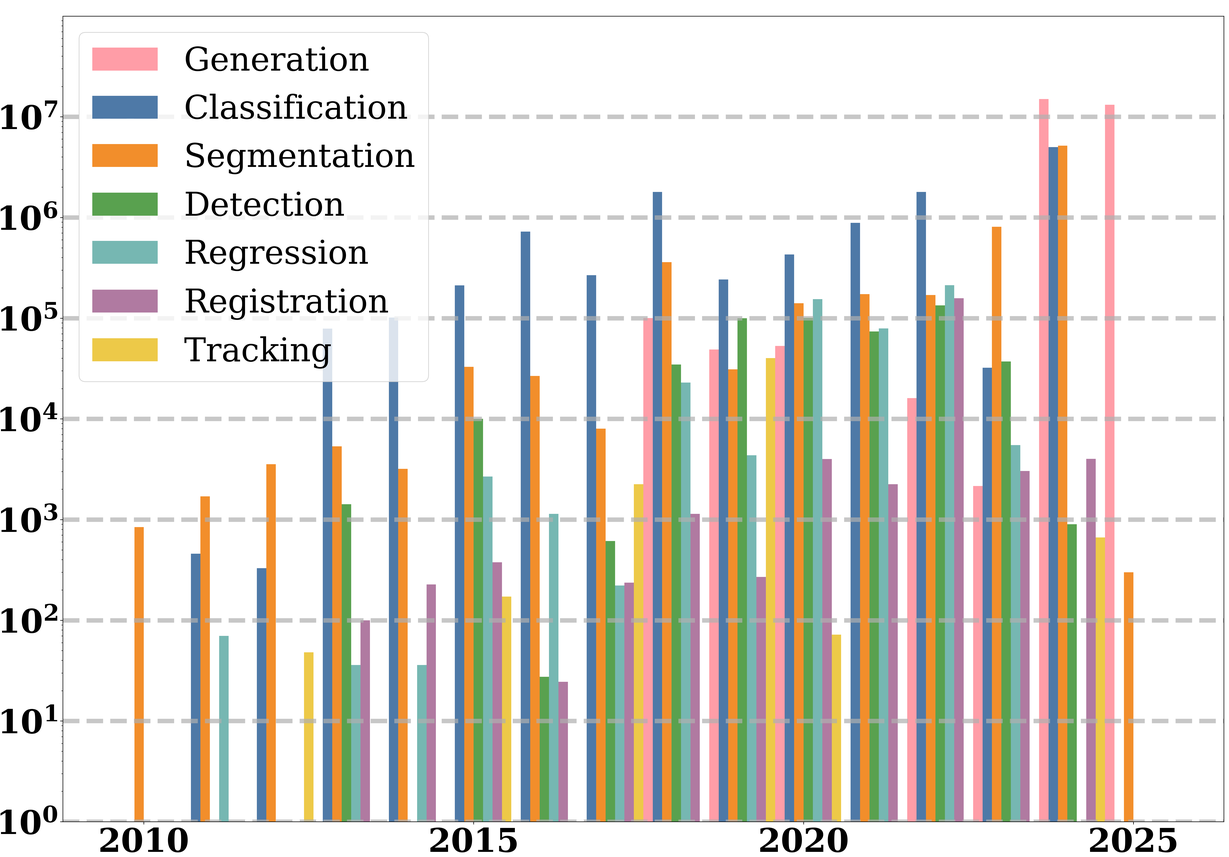}
    \caption*{\textbf{(d)}}
  \end{subfigure}

  \vspace{0.5em}

  \begin{subfigure}[b]{0.45\textwidth}
    \centering
    \includegraphics[width=\linewidth]{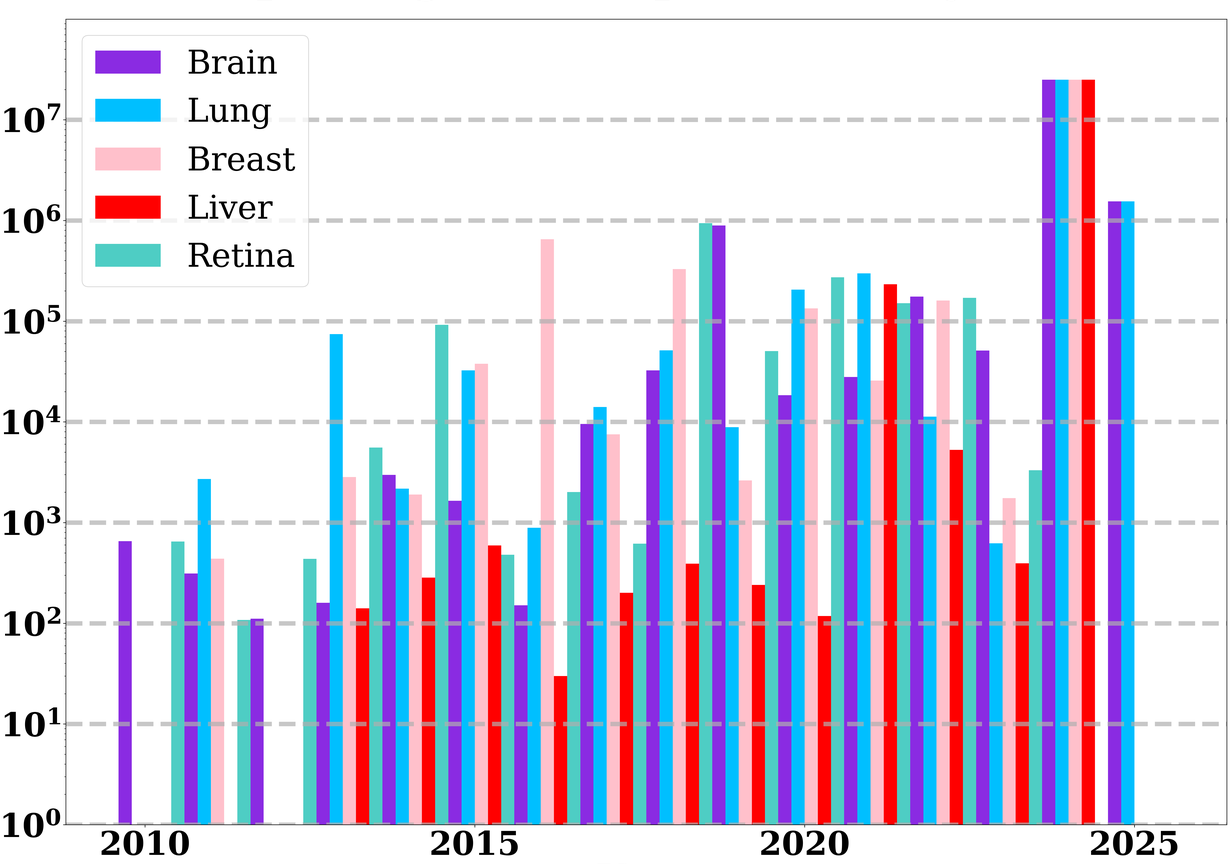}
    \caption*{\textbf{(e)}}
  \end{subfigure}
  \caption{The overview of image number in medical image datasets released from 2000 to 2025. (a) Total image number; Image number of different (b) dimensions, (c) modalities, (d) tasks, and (e) top five organs.}
\label{fig:overview_case_year}
\end{figure}


\subsection{Total Growth}

We first examine the annual count of released imaging data to gain insight into the temporal evolution of open-access medical image datasets. Figure~\ref{fig:overview_case_year}(a) illustrates the number of imaging items publicly released per year from 2000 to 2025, with clear inflection after 2012 and another surge after 2023. The first phase tracks the rise of deep learning methods~\cite{he2016deep}, which increased demand for extensive, curated training data. 
The recent surge beginning in 2023 reflect the adoption of self-supervised and large-scale foundation models, which benefits from scale even with limited labels. 
These advances highlight the centrality of massive datasets for enabling foundation-level models, motivating the medical imaging community to collect substantially larger resources in pursuit of general-purpose medical AI.
For example, AbdomenAtlas~\cite{qu2023abdomenatlas} aggregates 1.5 million 2D CT images and 5,195 3D CT volumes. CT-RATE~\cite{hamamci2024developing} introduces 25,692 non-contrast 3D chest CT scans from 21,304 unique patients. These datasets rank among the largest public medical imaging resources. 
Nonetheless, the scale of existing medical imaging datasets, particularly in terms of 3D volumes, remains orders of magnitude smaller than data resources in natural image and language domains, where training corpora typically contain trillions of tokens~\cite{grattafiori2024llama}. Given the prohibitive cost of curating trillion-scale medical datasets, an alternative and more practical strategy is to integrate multiple existing datasets into larger, heterogeneous corpora. This observation motivates our dataset fusion paradigms (Section~\ref{sec:merge_paradigm}).


\begin{figure*}[t]
  \centering
  \includegraphics[width=\textwidth]{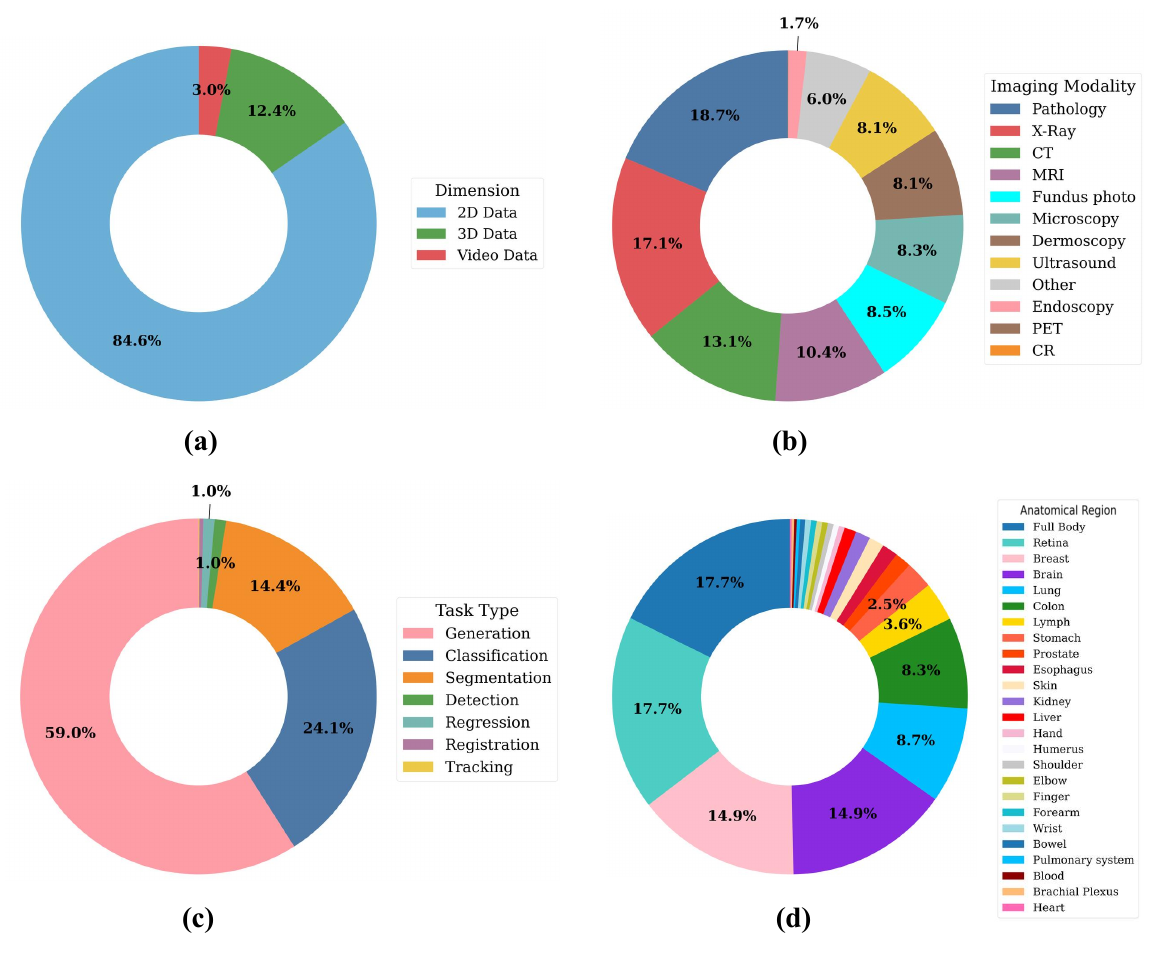}
  \caption{The distribution of (a) imaging dimensionalities, (b) modalities, (c) tasks, and (d) anatomical regions.}
  \label{fig:overview_pie_chart}
\end{figure*}

\subsection{Imaging Dimensionalities}
Two-dimensional images have height and width as their two dimensions, while three-dimensional volumes add a depth axis; videos are time-ordered 2D frames with temporal continuity. Figure~\ref{fig:overview_case_year}(b) presents the total number of 2D images, 3D volumes, and videos released between 2000 and 2025, and Figure~\ref{fig:overview_pie_chart}(a) shows the distribution of them. 

Two-dimensional images dominate in absolute scale, especially after 2023 (Figure~\ref{fig:overview_case_year}(b)), reflecting the wide use of 2D images for medical applications. This dominance has practical and methodological roots: 2D images are easier to store and share; patch extraction from histopathology whole-slide images (WSIs) multiplies sample counts~\cite{borkowski2019lung,spanhol2015dataset}; and many long-standing benchmarks target 2D tasks~\cite{veeling2018rotation,yang2023medmnist}. 
In contrast, 3D and video data remain comparatively scarce and show slower growth, largely due to higher acquisition costs and storage constraints, and the complexity of curation and annotation~\cite{tajbakhsh2020embracing}. 
Despite lower availability, 3D and video data are often more clinically informative for diagnosis and treatment planning, particularly in radiology, as they capture volumetric context and temporal dynamics that 2D cannot~\cite{williams2019we}. Increasing the availability of high-quality 3D and video datasets is therefore a priority for advancing clinically useful foundation models.



\subsection{Imaging Modalities}
\begin{figure}[ht] 
    \centering 
    \includegraphics[width=0.85\textwidth]{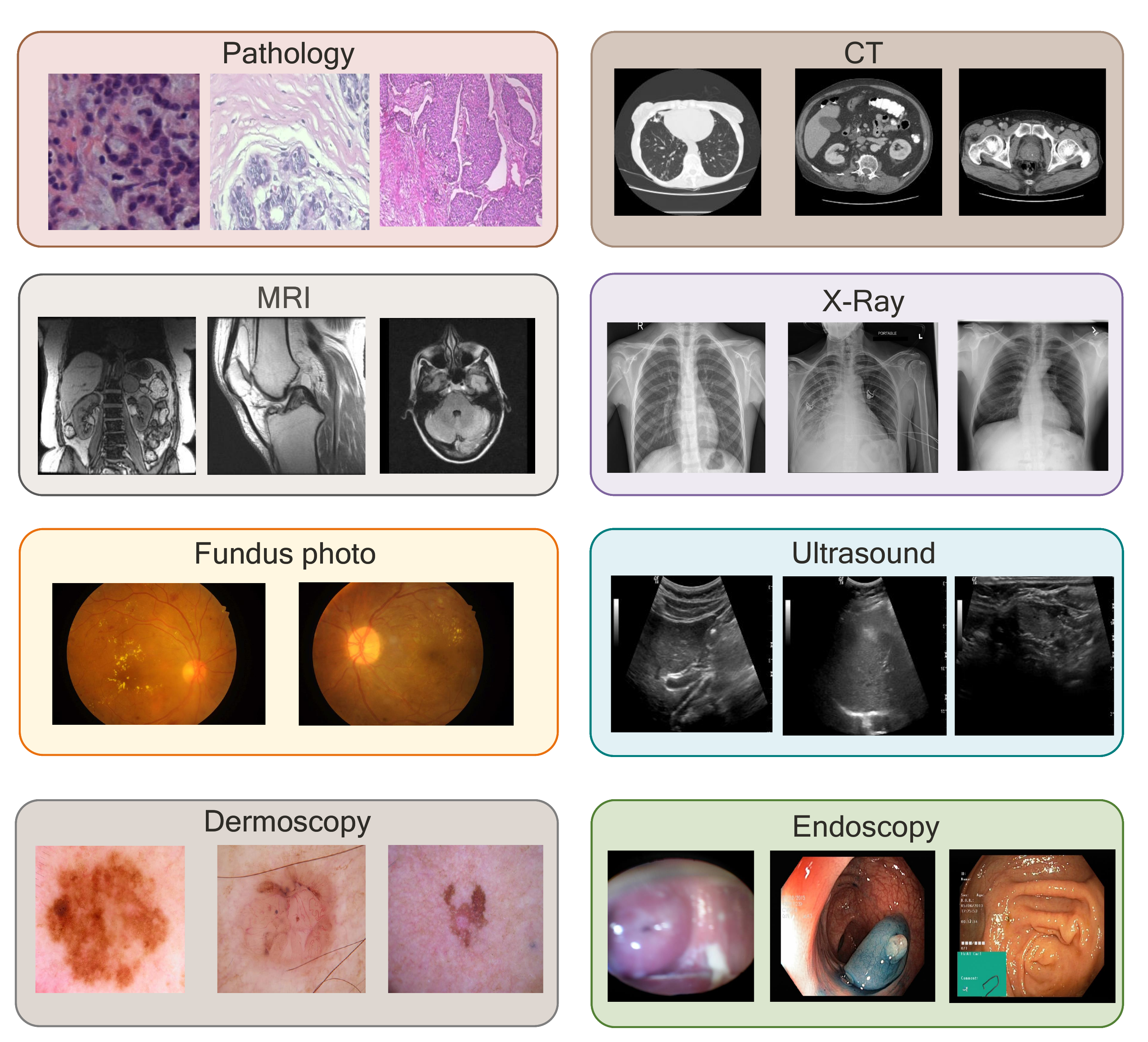} 
    \caption{Representative modalities in medical imaging datasets.} 
    \label{fig:organ-modal}
\end{figure}

Clinical practice employs different technologies and techniques to acquire medical images; these techniques are known as imaging modalities, as depicted in Figure~\ref{fig:organ-modal}. Each modality is designed to capture specific anatomical, functional, or molecular characteristics of the human anatomical regions/structures, and plays a critical role in clinical diagnosis and disease monitoring. Some of the most commonly used imaging modalities include:

\begin{itemize}
    \item \textbf{X-ray} imaging is among the oldest and most widely used techniques~\cite{rehani2020higher}, capturing 2D projections of internal structures using X-rays. It is widely applied to detect hard tissues such as bone fractures, lung infections, or dental issues.
    \item \textbf{Computed Tomography (CT)} visualizes internal structure of human body with tomographic acquisition of many slices to form 3D volumes. CT offers high spatial resolution and speed, making it particularly valuable in trauma, oncology, and cardiovascular imaging~\cite{liguori2015emerging}.
    \item \textbf{Magnetic Resonance Imaging (MRI)} generates high-contrast images of soft tissues using strong magnetic fields and radiofrequency pulses. It is frequently used in neuroimaging and visualizing internal organs such as the heart, liver, and kidneys~\cite{lohrke201625}, with sub-modalities including T1, T2, FLAIR, DWI, and fMRI.
    \item \textbf{Ultrasound} imaging leverages high-frequency sound waves to visualize soft tissues and fluid-filled structures. It is safe, portable, and economical, and is widely used in obstetrics, cardiology, and abdominal imaging.
    \item \textbf{Positron Emission Tomography (PET)} uses a radioactive tracer to detect diseased cells, featuring in functional imaging. It is commonly used for diagnosing dementia, cancers, and assessing heart conditions~\cite{wang2020anatomy}.
    \item \textbf{Pathology imaging} applies advanced microscopy and digital slide scanning to achieve ultra-high-resolution reconstruction and computational analysis of tissue. Beyond serving as the gold standard for histological classification, grading, and definitive cancer diagnosis, it is increasingly central to biomarker discovery, prognostic modeling, and AI-driven computational pathology~\cite{verghese2023computational}. 
    \item \textbf{Endoscopy} employs a mini-camera embedded in a flexible tube, which will be inserted into the gastrointestinal tract, respiratory pathways, or other body orifices to directly visualize internal organs and cavities~\cite{kurniawan2017flexible}. It is widely applied in diagnostic inspection and interventional procedures.
    \item \textbf{Fundus photography} captures detailed images of the retina at the back of the eye. It is essential in ophthalmology for diagnosing and monitoring diabetic retinopathy, age-related macular degeneration, glaucoma, and other retinal diseases~\cite{panwar2016fundus}.
    \item \textbf{Dermoscopy} provides non-invasive, magnified views of skin lesions, enabling observation of both skin surface and superficial layers. This technique reveals fine structural details of pigmented lesions and thereby improves the diagnostic accuracy of skin lesions, particularly for melanoma~\cite{kittler2002diagnostic}.
    \item Other modalities include \textbf{mammography} for breast screening, \textbf{fundus fluorescein angiography (FFA)} and \textbf{optical coherence tomography (OCT)} for visualizing internal structures of the eyes, as well as non-imaging modalities that record electrical activity such as the \textbf{electrocardiogram (ECG)} for the heart, the \textbf{electroencephalogram (EEG)} for the brain, and \textbf{electromyography (EMG)} for muscle response.
\end{itemize}

Figure~\ref{fig:overview_case_year}(c) shows the image number of the top six modalities from 2000 to 2025. Prior to 2023, CT, pathology, and MRI account for the majority of images. Post-2023 growth is especially pronounced in pathology imaging, X-ray, fundus photography, and microscopy. 
As shown in Figure~\ref{fig:overview_pie_chart}(b), pathology datasets contain substantially more images than other imaging modalities because the gigapixel-scale WSIs are often divided into thousands of patches, each used as a separate image for analysis~\cite{chetan2021deep}. The inherently multi-scale nature of pathology, spanning cellular morphology to tissue-level architecture, further increases patch generation by requiring sampling at multiple magnifications. Moreover, the diversity of staining protocols and specimen types adds further heterogeneity and volume~\cite{tellez2019quantifying}. These factors provide pathology with an unmatched reservoir of fine-grained image data that underpins the training of foundation models~\cite{verghese2023computational}. 

X-ray and CT also benefit from clinical ubiquity and high throughput~\cite{liguori2015emerging}. MRI accounts for about 10.4\% of the total number of images due to its effectiveness in visualizing soft tissues. Despite being radiation-free, MRI imaging data grows relatively slowly due to cost, longer acquisition, and complex multi-sequence labeling. 
In addition, fundus photography, microscopy, dermoscopy, and ultrasound are widely used and produce a significant number of images. However, other modalities like PET, mammography, and endoscopy remain comparatively less available in open data, which may limit the ability of foundation models trained on public corpora to fully address modality-specific clinical tasks.

\subsection{Tasks}
Medical image datasets can be collected and curated to address a wide range of tasks, each targeting specific aspects of image analysis and interpretation in computer-aided diagnosis and clinical workflows. These tasks include, but are not limited to, segmentation, classification, registration, generation, detection, and tracking. 

\textbf{Segmentation} tasks involve assigning a class label to each individual pixel in 2D images or voxel in 3D volumes. The goal is to delineate anatomical structures or pathological regions of interest, such as organs, tumors, and lesions, allowing for precise spatial localization and quantitative analysis. For example, in abdominal MRI, segmentation can distinguish between the liver and kidneys, facilitating downstream analysis like volume estimation or disease monitoring.

\textbf{Classification} tasks aim to categorize an entire medical image or a specific region within it into predefined classes. This could involve distinguishing between healthy and diseased states, grading the severity of a condition, or identifying the presence of particular disease types. For instance, in brain MRI, classification might involve determining whether an image corresponds to a cognitively normal subject, someone with mild cognitive impairment, or a patient with Alzheimer's disease. 

\textbf{Registration} refers to the process of aligning two or more images into a common coordinate system. This is particularly important when comparing scans from different time points (longitudinal analysis), modalities (\eg, MRI and PET), or subjects (for population studies). Registration techniques compute spatial transformations to ensure that anatomical structures in one image accurately correspond to those in another. Accurate registration is essential for tasks like image fusion, growth tracking, or mapping patient data to standardized anatomical atlases.

\textbf{Generation} tasks typically use models to synthesize new medical images, often conditioned on specific attributes or constraints. This can help augment training datasets, simulate rare disease appearances, or recover missing modalities. 

\textbf{Detection} focuses on efficient identification and localization of specific pathological findings with bounding boxes, such as lung nodules in CT, surgical instruments in endoscopic videos, or cancer cells in pathology slides. 

\textbf{Tracking} monitors the movement or evolution of anatomical structures or lesions across image sequences or time-series data, which is critical for assessing disease progression or treatment response. 

\textbf{Reconstruction} transforms incomplete or indirect raw data obtained from sensors into a meaningful image, which often involves solving inverse problems and addressing low-level vision tasks. 

\textbf{Regression} predicts continuous, quantitatively meaningful targets from medical images or sequences, such as physiologic indices (ejection fraction), image-derived biomarkers (arteriolar–venular ratio), severity/quality scores, or voxel-wise physical fields (dose), supporting precise monitoring, prognosis, and treatment planning. 

\textbf{Localization} seeks to identify specific anatomical points or landmarks in an image, such as corners of bones or key organ boundaries, to support diagnosis, measurement, registration, or treatment planning. 

Beyond these, vision-language tasks are emerging thanks to the rapid development of multimodal large language models. For instance, 
\textbf{visual question answering (VQA)} aims to answer questions about given images in natural language, while \textbf{captioning} and \textbf{report generation} produce free-text or structured descriptions from images or image series, in different levels of detail. 

In Figure~\ref{fig:overview_case_year}(d), from the task-wise statistics, classification and segmentation account for the largest share of released images over the past decade, while datasets tagged for generation exhibit a marked uptick after 2023, showing the strong community interest in applying general-purpose generative AI for advanced medical image analysis. In contrast, other tasks like registration, detection, and tracking, remain a relatively small number of images over time. We stress, however, that the apparent imbalance is more indicative of practical constraints than of community priorities, because counts are shaped by mixed factors of label economics (\eg, per-image classification labels are comparatively inexpensive) as well as acquisition and annotation burden (\eg, tracking requires videos with temporal labels~\cite{al2019cataracts}; registration often lacks easily verifiable ground truth and may depend on multi-timepoint/multimodal data~\cite{brock2017use}).

Figure~\ref{fig:overview_pie_chart}(c) further presents the imbalanced distribution of these tasks, where generation, classification, and segmentation tasks are more extensively studied compared to the remaining tasks. This imbalance may not be ideal for training a general-purpose AI excelling in these less-represented tasks.

\begin{figure}[tp]
    \centering
    \includegraphics[width=\textwidth]{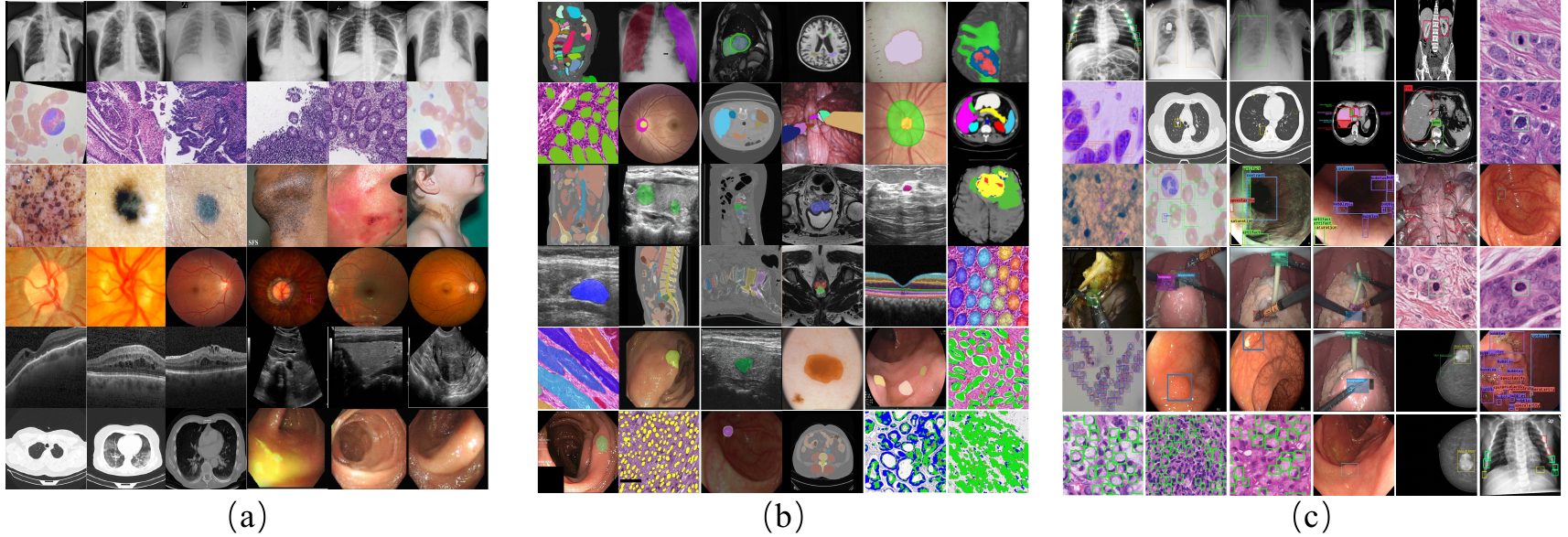}
    \caption{Representative samples of three medical image analysis tasks: 
    (a) classification, (b) segmentation, and (c) detection. 
    Each column shows images from diverse modalities and clinical applications, illustrating the characteristic outputs of the respective task types.
    }
    \label{fig:task_examples}
\end{figure}

Visual examples in Figure~\ref{fig:task_examples} intuitively demonstrate the distinct outputs and clinical relevance of each task, highlighting how the same imaging modalities can serve different analytical purposes depending on the problem at hand.

\subsection{Anatomical Regions}


\begin{figure}[ht] 
    \centering 
    \includegraphics[width=1.0\textwidth]{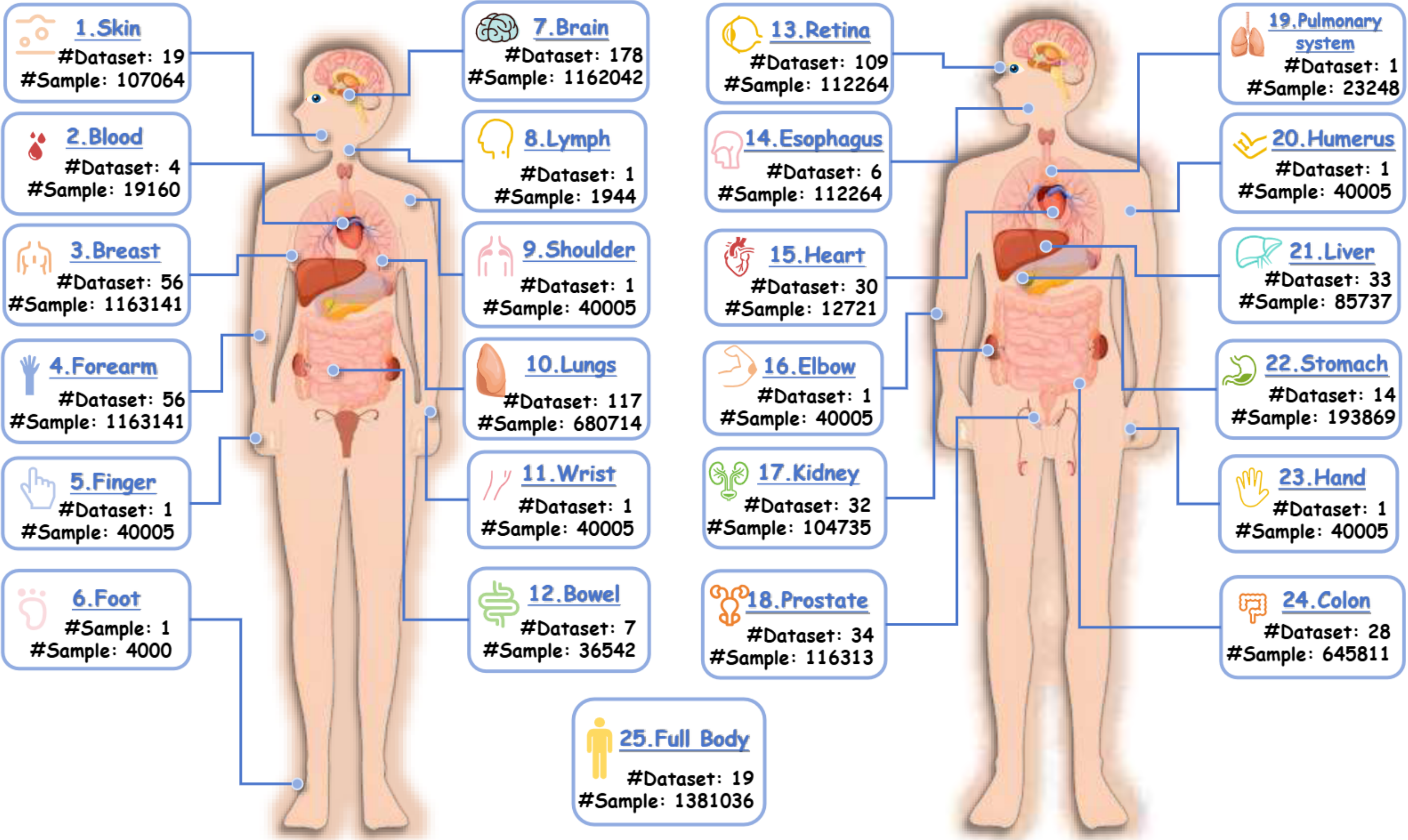} 
    \caption{Anatomical structures of medical image datasets. We also show the total number of datasets and images for each anatomical structure. 
    } 
    \label{fig:organ-body} 
\end{figure}

Anatomical regions, illustrated in Figure~\ref{fig:organ-body}, are specific, named areas of the human anatomical regions used to organize and describe structures. 
This subsection analyzes the datasets from a compact set of anatomical regions, identified in two steps. First, we align each dataset's native labels with standard medical vocabularies. Second, we group those mapped labels into a concise set of anatomical regions/structures that are consistently reported across public datasets and align with major clinical workload and benchmark. This choice favors comparability and coverage across sources. While finer-grained systems are possible, they are unevenly annotated in the open-access datasets.

Figure~\ref{fig:overview_case_year}(e) tracks medical imaging data counts for common target organs. We observe that brain and lung contribute the largest volumes of images before 2023. Starting in 2023, there is an abrupt and pronounced surge in several modalities, most notably brain, liver, lung, and breast, while retina does not exhibit a comparable surge. This pattern highlights shifting research priorities toward clinically significant organs. 


Figure~\ref{fig:overview_pie_chart}(d) shows the distribution of anatomical regions. Notably, the number of fully body, retina, breast, brain, lung, and colon images significantly exceeds that of other regions, highlighting a strong research emphasis on with high clinical and societal impact, including Alzheimer’s disease, diabetic retinopathy, and common cancers such as breast, lung, and colorectal cancer. 
In contrast, other anatomical regions are less represented, such as the foot, blood, heart, bowel, shoulder, huerus, forearm, \etc
Such undercoverage often stems from practical challenges such as limited accessibility, lower disease prevalence, or the complexity of imaging certain anatomical sites.

\subsection{Summary}

The public landscape of open medical imaging is distinctly long-tailed: many small, tightly scoped datasets coexist with a smaller set of large hubs, with pronounced skew toward 2D images, a subset of modalities (notably pathology, X-ray, CT, MRI), and a handful of organs (brain, lung, liver, breast). Task labels are likewise imbalanced where classification and segmentation dominate while other tasks are comparatively underrepresented, largely reflecting practical constraints such as label economics, acquisition burden, and the scarcity of ground truth for certain tasks. 
These patterns imply that scale for general-purpose models is attainable, but not via nai\"ve concatenation: it requires careful normalization of counting conventions, balanced sampling across modalities and organs, and task-aware objectives to avoid amplifying existing biases.

Given this heterogeneity and uneven distribution, it is increasingly crucial to effectively utilize all these datasets for training medical foundation models. 
Specifically, recent medical foundation models across subdomains have been trained by integrating multiple public datasets within a modality or organ. Examples include backbone model for ophthalmology~\cite{zhou2023retfound,wu2025eyecare}, histopathology~\cite{xiang2025musk}, radiography~\cite{perez2025raddino,ma2025ark}, segmentation-focused families in 2D~\cite{ma2024segment} and 3D~\cite{wang2025sam,deng2025brain}, and even autoencoders for generative models~\cite{ma2025meditok,varma2025medvae}. In short, the very imbalances mapped above become design signals for assembling data and objectives. With the proposed taxonomy and metadata-driven fusion, this paper provides a principled path from fragmented public datasets to scalable, diverse, and clinically relevant training distributions for medical foundation models.

\section{2D Medical Image Datasets}
\label{sec:2d_data}

We have collected 502 2D medical image datasets. In aggregate data count, 2D image far exceeds 3D volumes and video frames.  
We partition them into 475 labeled and 45 unlabeled datasets. Labeled datasets are analyzed by modality, task, and anatomical focus; for unlabeled datasets which lack explicit task definitions, we summarize modality and anatomy.

\subsection{Overview}
\begin{figure}
    \centering
    \includegraphics[trim=0 0 0 0, clip, width=\linewidth]{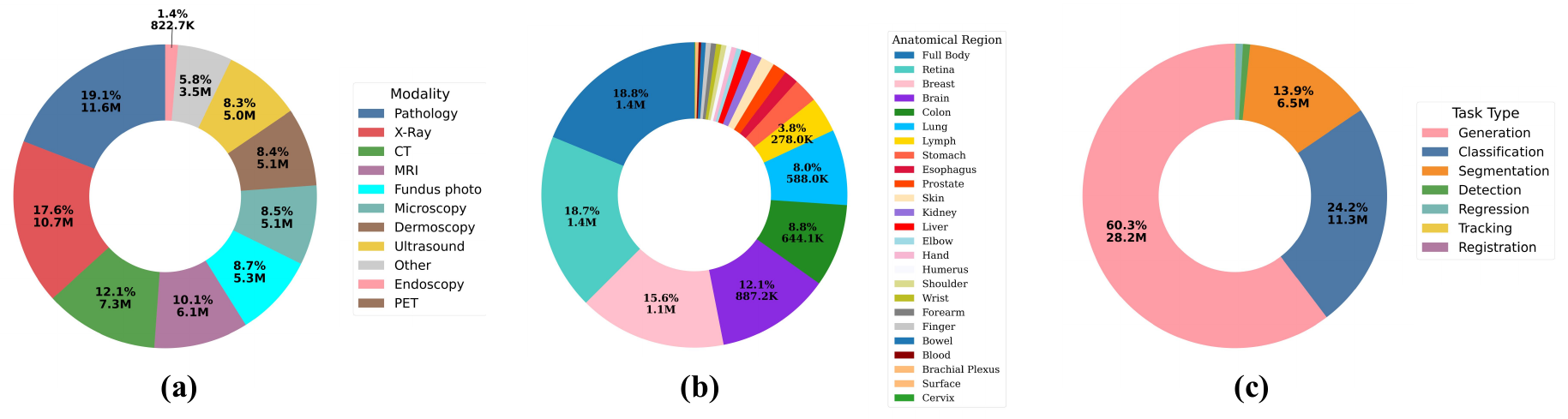}
    \caption{The distribution of different (a) modalities, (b) anatomical structures, and (c) tasks for for 2D labeled datasets. Each slice of the pie chart shows the percentage and the actual number of images.}
    \label{fig:2d_datasets_overview}
\end{figure}

Figure~\ref{fig:2d_datasets_overview} shows the distributions of different modalities, anatomical regions/structures, and tasks for 2D labeled images, which represent clear long-tail distributions. 

\begin{figure}[!tp]
    \centering
    \includegraphics[width=\textwidth]{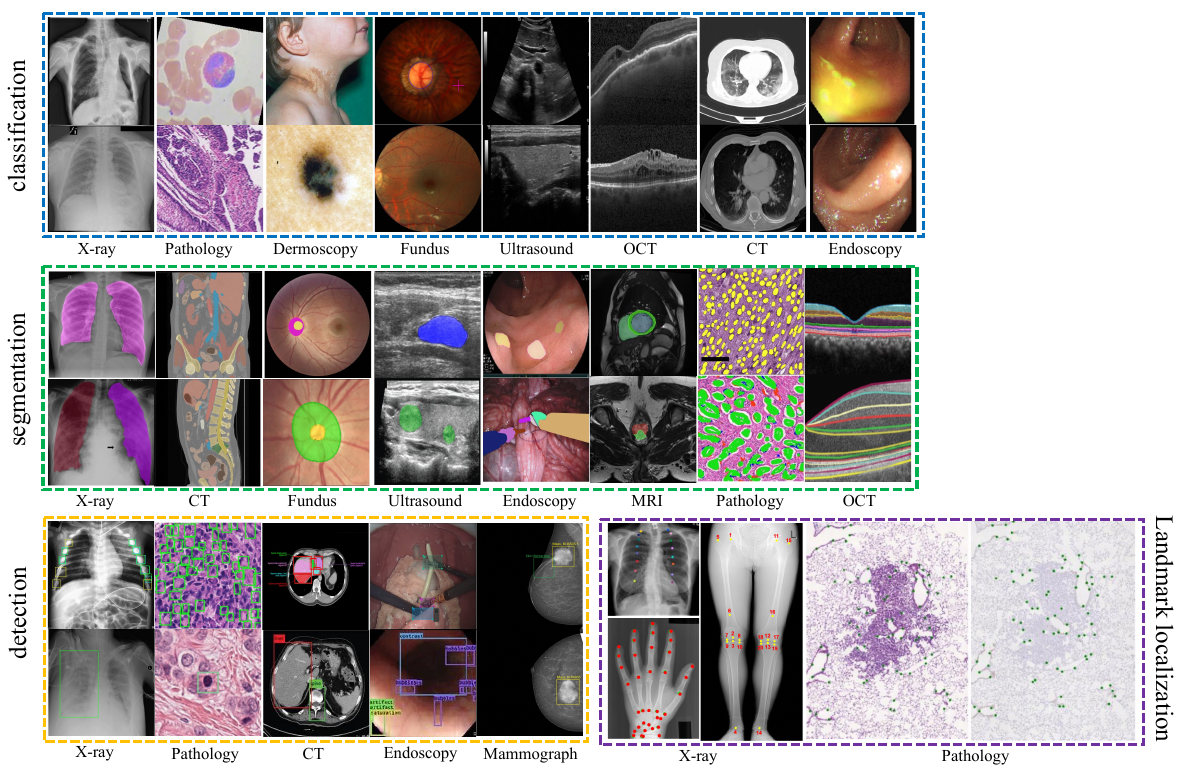}
    \caption{Demonstration of the collected 2D medical datasets across different modalities and anatomical regions.}
    \label{fig:2d_datasets_demo}
\end{figure}

In terms of modality, pathology and X-ray dominate, followed by CT, MRI, and fundus photography. Together these account for the majority of images. Other modalities such as endoscopy are less representative. In terms of anatomy, large shares concentrate on full-body/multi-structure views and a few organs with mature screening pipelines, \eg, retina, breast, and brain, followed by lung and colon. By contrast, datasets targeting uterus, heart, esophagus, limb joints, and small substructures (\eg, nodules) remain underrepresented, indicating opportunities for targeted curation.
%
The dominating tasks include generation, classification, segmentation, regression, and detection. However, other tasks, \eg, registration, tracking, localisation, reconstruction, and visual question answering, have much fewer images. Figure \ref{fig:2d_datasets_demo} demonstrates representative examples of the collected 2D medical image datasets across different modalities and anatomical regions.

\subsection{CT Modality}
CT is a cornerstone of radiological imaging, providing detailed cross-sectional views of the body. In 2D datasets, CT images are typically axial, sagittal, or coronal slices extracted from 3D volumes. A predominant characteristic of CT datasets is their extensive use in segmentation tasks. These tasks can be broadly categorized into delineating anatomical structures, such as organs for surgical planning, and identifying pathologies, such as tumors or hemorrhages for diagnosis and treatment monitoring. This makes CT datasets highly valuable for a wide range of clinical applications. Among 2D labeled CT datasets, 39 provide CT slices (see \Cref{tab:2d_ct_datasets}), totaling approximately \textbf{1.4 million} images. Scale varies dramatically: from small, specialized collections such as The Visible Human Project~\dataref{data:the-visible-human-project_ct} 
with only 2 images, to large-scale resources like RSNA Intracranial Hemorrhage Detection~\dataref{data:rsna-intracranial-hemorrhage-detection} with 874,000 images.
\paragraph{CT Datasets by Anatomical Regions/Structures.}


%
A clear trend in the distribution of CT datasets is the focus on specific anatomical regions. Datasets related to the brain are the most represented in terms of image volume, primarily due to a single large-scale dataset~\dataref{data:rsna-intracranial-hemorrhage-detection}. Lung-related datasets are the most numerous, driven by research in COVID-19 and cancer screening. Conversely, data for abdominal and other structures remains relatively scarce, highlighting potential gaps in data availability for developing models for those areas.

\emph{1) Lung} (11 datasets, $\sim$60,400 images). A significant portion of the datasets is dedicated to the lungs, focusing on tasks like cancer classification in the National Lung Screening Trial~\dataref{data:national-lung-screening-trial} and COVID-19 classification in datasets such as COVID-19-CT SCAN IMAGES~\dataref{data:covid-19-ct-scan-images} and SARS-COV-2 Ct-Scan Dataset~\dataref{data:sars-cov-2-ct-scan-dataset}. Segmentation is also a key task, as seen in CT Medical Images~\dataref{data:ct-medical-images}.
These datasets are characterized by distinct visual patterns, such as ground-glass opacities for COVID-19 and well-defined nodules in cancer screening, making them ideal for developing specialized classifiers.

\emph{2) Brain} (5 datasets, $\sim$874,400 images). Brain datasets constitute the largest collection by image count, dominated by the RSNA Intracranial Hemorrhage Detection dataset~\dataref{data:rsna-intracranial-hemorrhage-detection} for localization tasks. Other datasets like Brain CT Images with ICH Masks~\dataref{data:brain-ct-images-with-ich-masks} focus on segmentation, while smaller sets like Cranium Image Dataset~\dataref{data:cranium-image-dataset} are used for detection.

\emph{3) Abdomen/Pelvis} (7 datasets, $\sim$1,500 images). This category covers organs such as the kidney, pancreas, colon, and prostate. Key tasks include segmentation and classification of tumors in datasets like CMB-CRC~\dataref{data:cmb-crc} for colorectal cancer and segmentation of kidneys and pancreas in the QUBIQ challenges~\dataref{data:qubiq2020}. These datasets are typically small, limiting their use for training large-scale deep learning models.
They often feature multiple organs with subtle boundaries and variable shapes, making multi-organ segmentation a significant challenge despite limited data availability.

\emph{4) Full-Body/Multistructure} (5 datasets, $\sim$454,400 images). These datasets provide data from multiple anatomical regions or cell structures, making them suitable for pre-training generalizable models. Notable examples include RadImageNet~\dataref{data:radimagenet-subset-ct}, a large-scale classification dataset with 34 anatomic categories, and MedMNIST~\dataref{data:medmnist}, which contains diverse 2D slices for educational and research purposes.
Their diversity across different anatomical regions/structures helps models learn a more generalized representation of CT imaging characteristics, reducing the risk of overfitting to a specific anatomy.

\emph{5) Others} (11 datasets, $\sim$11,000 images). This group comprises datasets for various other anatomical regions/structures or those without a specified structure. It includes specialized collections such as 5K+ CT Images on Fractured Limbs~\dataref{data:5k-ct-images-on-fractured-limbs} for limb fracture segmentation and Head CT Image Data~\dataref{data:head-ct-image-data} for classification. Datasets with non-specific structures, like RIDER Phantom PET-CT~\dataref{data:rider-phantom-pet-ct} for calibration, are also in this category.

\paragraph{CT Datasets by Tasks.}
The distribution of datasets is heavily skewed towards classification, which accounts for a large volume of images. Detection and localization tasks are dominated by a single large dataset, while segmentation and reconstruction datasets are generally smaller in scale.

\emph{1) Classification} (12 datasets, $\sim$513,900 images). Classification is the most common task, especially for pulmonary applications spurred by the COVID-19 pandemic, with datasets like COVID-CT~\dataref{data:covid-ct-covid-ct} and SARS-COV-2 Ct-Scan Dataset~\dataref{data:sars-cov-2-ct-scan-dataset}. Large multi-purpose datasets like RadImageNet~\dataref{data:radimagenet-subset-ct} and MedMNIST~\dataref{data:medmnist} also contribute significantly to this category. Oncology is another major focus, with datasets such as the National Lung Screening Trial~\dataref{data:national-lung-screening-trial} for lung cancer.
These tasks often involve distinguishing between different diseases or staging disease severity from a single representative slice.

\emph{2) Segmentation} (9 datasets, $\sim$2,100 images). Segmentation datasets are diverse but generally small. They cover organ segmentation, such as in the QUBIQ challenges~\dataref{data:qubiq2020}, lesion segmentation in Brain CT Images with ICH Masks~\dataref{data:brain-ct-images-with-ich-masks}, and quantitative imaging in Finding and Measuring Lungs in CT Data~\dataref{data:finding-and-measuring-lungs-in-ct-data}.
Segmentation in CT is crucial for quantitative analysis, such as measuring tumor volume or assessing organ health, moving beyond simple qualitative assessment.

\emph{3) Detection/Localization} (2 datasets, $\sim$874,100 images). This task category is dominated by the RSNA Intracranial Hemorrhage Detection dataset~\dataref{data:rsna-intracranial-hemorrhage-detection}, which contains 874,000 slices with hemorrhage annotations. The only other dataset in this category is the much smaller Cranium Image Dataset~\dataref{data:cranium-image-dataset}, also for hemorrhage detection.
This task is often a precursor to segmentation and is critical in large-scale screening programs where anomalies need to be quickly identified.

\emph{4) Reconstruction} (1 dataset, 28 images). The LoDoPaB-CT dataset~\dataref{data:lodopab-ct-table} is the sole entry dedicated to reconstruction, specifically for sparse-view reconstruction challenges.

\emph{5) Multi-task datasets} (3 datasets, $\sim$500 images). A few small datasets are designed for multiple tasks. For example, CMB-CRC~\dataref{data:cmb-crc} provides data for both segmentation and classification of colorectal cancer, while CMB-PCA~\dataref{data:cmb-pca} is for classification and prediction in prostate cancer.

\emph{6) Others} (12 datasets, $\sim$11,200 images). The remaining datasets are for other specific tasks or have no specified task. This includes AREN0534~\dataref{data:aren0534-ct} for estimation and LDCTIQAC2023~\dataref{data:ldctiqac2023} for registration. A significant number of datasets, such as those from the TCIA archive like CPTAC-LSCC\_CT\_PET~\dataref{data:cptac-lscc-ct-pet} and Prostate-MRI~\dataref{data:prostate-mri}, have no explicit task listed and may be used for a variety of research purposes.

\subsection{MRI Slices}
MRI offers superior soft-tissue contrast compared to CT and does not introduce ionizing radiation, making it ideal for neurological, musculoskeletal, and oncological imaging. A key feature of MRI datasets is their multi-contrast nature; a single study often includes multiple sequences (\eg, T1-weighted, T2-weighted, FLAIR) that highlight different tissue properties. This multi-channel information provides a rich basis for tasks like tumor segmentation and tissue characterization, though it also presents a challenge in fusing the information effectively.
Our analysis of 24 diverse MRI and multimodal imaging datasets (\Cref{tab:2d_mri_datasets}) reveals important trends in dataset development across modalities and clinical applications. In total, there are approximately \textbf{722,400 images}, with significant variations in scale: from small, specialized collections, such as The Visible Human Project~\dataref{data:the-visible-human-project_mri} with only 2 images, to large-scale resources like RadImageNet (Subset: MR)~\dataref{data:radimagenet-subset-mr} with 673,000 images.

\paragraph{MRI Datasets by Anatomical Regions/Structures.}
A clear trend in the distribution of MRI datasets is the focus on specific anatomical regions, alongside a growing number of large-scale, multi-structure collections suitable for pre-training generalizable models. Datasets related to the brain are a common focus, though typically smaller in scale. Abdominal and pelvic datasets are also present but limited in image volume. Conversely, data for other specific regions like the heart or spine is available, highlighting diverse clinical applications.

\emph{1) Brain} (2 datasets, 220 images). Datasets focused on the brain are represented by two small-scale collections for segmentation tasks: braimMRI~\dataref{data:braimmri} and Brain-MRI~\dataref{data:brain-mri}, each containing 110 images for analyzing brain tumors and diseases.

\emph{2) Abdomen/Pelvis} (4 datasets, $\sim$560 images). This category covers organs such as the colon and prostate. Key tasks include segmentation and classification of tumors in datasets like CMB-CRC~\dataref{data:cmb-crc-mri} for colorectal cancer and multiple datasets for prostate cancer analysis, including CMB-PCA~\dataref{data:cmb-pca-mri}, Prostate Fused-MRI-Pathology~\dataref{data:prostate-fused-mri-pathology}, and Prostate-MRI~\dataref{data:prostate-mri-mri}. These datasets are typically small, with a combined total of around 560 images.

\emph{3) Full-Body/Multistructure} (8 datasets, $\sim$704,900 images). These datasets provide data from multiple anatomical regions/structures, making them suitable for pre-training generalizable models. This category is dominated by RadImageNet (Subset: MR)~\dataref{data:radimagenet-subset-mr}, a large-scale classification dataset with 673,000 images. Other notable examples include ImageCLEF 2016~\dataref{data:imageclef-2016} with 31,000 images and multi-organ challenge datasets like the QUBIQ series~\dataref{data:qubiq2020-mri},~\dataref{data:qubiq2021-2d-mr}.

\emph{4) Others} (10 datasets, $\sim$16,700 images). This group comprises datasets for various other anatomical regions/structures or those without a specified structure. It includes specialized collections such as Cardiac Atrial Images~\dataref{data:cardiac-atrial-images} for heart segmentation with 8,000 images, SpinalDisease2020~\dataref{data:spinaldisease2020} for spine analysis, and KNOAP2020~\dataref{data:knoap2020-mri} for knee osteoarthritis. It also includes several datasets from The Cancer Imaging Archive where the specific structure is not listed, such as APOLLO-5~\dataref{data:apollo-5} and ICDC-Glioma (GLIOMA01)\_3D-MR~\dataref{data:icdc-glioma-3d-mr}.

\paragraph{MRI Datasets by Tasks.}
The distribution of datasets is heavily skewed towards classification, which accounts for the vast majority of images due to one large-scale collection. Segmentation is the next most common task, though the corresponding datasets are significantly smaller. A number of datasets are provided without a specific task, offering resources for various research purposes.

\emph{1) Classification} (3 datasets, 704,000 images). Classification is the most represented task by image volume, dominated by RadImageNet (Subset: MR)~\dataref{data:radimagenet-subset-mr} (673,000 images) and ImageCLEF 2016~\dataref{data:imageclef-2016} (31,000 images). ImageCLEF 2015~\dataref{data:imageclef-2015} also falls into this category, although it contains no images.

\emph{2) Segmentation} (6 datasets, $\sim$8,900 images). Segmentation datasets are more numerous but contain far fewer images in total. They cover various organs, including the heart in Cardiac Atrial Images~\dataref{data:cardiac-atrial-images} (8,000 images), the brain in braimMRI~\dataref{data:braimmri} and Brain-MRI~\dataref{data:brain-mri}, and multiple abdominal organs in the QUBIQ challenges~\dataref{data:qubiq2020-mri},~\dataref{data:qubiq2021-2d-mr}.
These datasets often require precise delineation of soft tissues with subtle intensity differences, a task for which MRI is uniquely suited.

\emph{3) Multi-task datasets} (2 datasets, $\sim$500 images). A couple of small datasets are designed for multiple tasks. CMB-CRC~\dataref{data:cmb-crc-mri} provides data for both segmentation and classification of colorectal cancer, while CMB-PCA~\dataref{data:cmb-pca-mri} is designed for classification and prediction in prostate cancer.

\emph{4) Others} (13 datasets, $\sim$9,000 images). The remaining 13 datasets cover a range of other tasks or have no specified task. This includes SpinalDisease2020~\dataref{data:spinaldisease2020} for detection (150 images), KNOAP2020~\dataref{data:knoap2020-mri} and CMB-MML~\dataref{data:cmb-mml} for prediction, and AREN0534~\dataref{data:aren0534-mri} for estimation (239 images). A significant number of datasets (9) are provided without an explicit task, such as APOLLO-5~\dataref{data:apollo-5} and the ICDC-Glioma series~\dataref{data:icdc-glioma-3d-mr}, 
making them flexible resources for exploratory research.

\subsection{PET Slices}

PET is a functional imaging modality that visualizes metabolic processes, often by tracking the uptake of a radioactive tracer. 2D PET slices are typically used in conjunction with anatomical imaging like CT or MRI for accurate localization of metabolic activity. Therefore, a common characteristic of PET datasets is their multi-modal nature (PET/CT or PET/MR). The primary tasks involve detecting and quantifying regions of high metabolic activity, which are often indicative of cancer, inflammation, or neurological disorders.
We have collected 13 PET imaging datasets, a majority of which are sourced from The Cancer Imaging Archive (TCIA), as detailed in \Cref{tab:2d_pet_datasets}. These collections often include multiple modalities alongside PET. Compared to CT and MRI datasets, they span less diverse tasks and anatomic regions, focusing primarily on brain and abdominal imaging for segmentation and classification tasks. In total, these datasets comprise approximately \textbf{41,942} images. The scale varies significantly, from small collections like CMB-GEC~\dataref{data:cmb-gec} with only 14 images to the large-scale ImageCLEF 2016~\dataref{data:imageclef-2016-pet} dataset, which contains 31,000 images.

\paragraph{PET Datasets by Anatomical Regions/Structures.}
The distribution of PET datasets shows a concentration in specific anatomical areas, with a significant number of datasets lacking explicit structural information. Datasets with multi-structure or full-body scope contribute the largest volume of images, primarily due to one large collection.

\emph{1) Brain} (2 datasets, $\sim$269 images). Brain-related PET datasets are represented by CMB-GEC~\dataref{data:cmb-gec} and CMB-MEL~\dataref{data:cmb-mel}. These datasets focus on the detection and segmentation of cerebral microbleeds in melanoma patients. However, their small sample sizes limit their suitability for training large-scale deep learning models.

\emph{2) Abdomen/Pelvis} (1 dataset, 472 images). This category contains a single dataset, CMB-CRC~\dataref{data:cmb-crc-pet}, which provides images of the colon for research on colorectal cancer. The limited size of this collection may constrain its use for developing complex models.

\emph{3) Full-Body/Multistructure} (2 datasets, $\sim$31,200 images). This category is dominated by the large-scale ImageCLEF 2016 dataset~\dataref{data:imageclef-2016-pet}, containing 31,000 images across skin, cell, and breast structures. The other dataset, AREN0534~\dataref{data:aren0534-pet-2d}, provides 239 images of the kidney and lung.

\emph{4) Others} (8 datasets, $\sim$10,000 images). The majority of the collected PET datasets do not specify an anatomical region. This category includes collections for various diseases, such as AREN0532~\dataref{data:aren0532-pet-2d} for Wilms Tumor research. While diverse, many of these datasets, such as CMB-MML~\dataref{data:cmb-mml-pet} (60 images), have limited numbers of images. This category also includes larger collections like APOLLO-5~\dataref{data:apollo-5-pet} with 6,200 images.

\paragraph{PET Datasets by Tasks.}
The tasks are unevenly distributed, with classification datasets providing the vast majority of images. A significant number of datasets lack explicit task labels, making them candidates for unsupervised or semi-supervised learning approaches.

\emph{1) Classification} (1 dataset, 31,000 images). The classification task is represented by a single, large-scale dataset, ImageCLEF 2016~\dataref{data:imageclef-2016-pet}, which contains 31,000 images and is designed for classification challenges.

\emph{2) Segmentation} (1 dataset, 255 images). The sole dataset dedicated purely to segmentation is CMB-MEL~\dataref{data:cmb-mel}, which provides 255 images for melanoma-related cerebral microbleed segmentation.

\emph{3) Multi-task datasets} (2 datasets, 486 images). Two small datasets are designed for multiple tasks. CMB-CRC~\dataref{data:cmb-crc-pet} (472 images) supports both segmentation and classification for colorectal cancer, while CMB-GEC~\dataref{data:cmb-gec} (14 images) is annotated for the same tasks in the context of cerebral microbleeds.

\emph{4) Others} (9 datasets, $\sim$10,200 images). The remaining nine datasets are intended for other specific tasks or have no defined task ('NA'). This group includes AREN0534~\dataref{data:aren0534-pet-2d} for estimation and CMB-MML~\dataref{data:cmb-mml-pet} for prediction. The majority, however, are general-purpose collections without specified tasks, such as APOLLO-5~\dataref{data:apollo-5-pet} and AREN0532~\dataref{data:aren0532-pet-2d}, which can be valuable for developing and testing unsupervised models or for a variety of bespoke research questions.

\subsection{Ultrasound (US) Images}

Ultrasound imaging is a real-time, non-invasive, and portable modality, making it widely used for various applications from fetal monitoring to cardiac assessment. A key characteristic of ultrasound datasets is the inherent image noise (speckle) and operator-dependent variability, which pose significant challenges for automated analysis. Common tasks include segmentation of anatomical structures (\eg, cardiac chambers, fetal head) and classification of lesions (\eg, benign vs. malignant breast tumors).
As presented in \Cref{tab:2d_us_datasets}, we have collected 19 major ultrasound imaging datasets from various sources including TCIA and Kaggle. The datasets include approximately \textbf{457,663} images in total, with RadImageNet-US~\dataref{data:radimagenet_us} contributing the vast majority (390k images).

\paragraph{Ultrasound Datasets by Anatomical Regions/Structures.}
The available datasets cover a wider range of anatomical regions/structures including the skull, breast, heart, thyroid, and liver, in addition to full-body imaging, though some TCIA collections (APOLLO-5~\dataref{data:apollo5} and CMB-LCA~\dataref{data:cmblca}) lack anatomic specifications. Following the guideline that datasets containing multiple organs are categorized separately, RadImageNet-US~\dataref{data:radimagenet_us} represents the most comprehensive full-body coverage with 390k images, while other datasets remain relatively small-scale.

\emph{1) Breast} (2 dataset, $\sim$803 images). The BUSI~\dataref{data:busi} and BreastMNIST~\dataref{data:breastmnist} datasets focus on breast ultrasound for cancer detection, providing segmented images with binary classification labels. This small-scale collection may support basic supervised learning applications.

\emph{2) Skull} (1 dataset, 1,344 images). HC18~\dataref{data:hc18} targets fetal head circumference measurement through skull ultrasound imaging. As a challenge dataset with CC BY 4.0 license, it facilitates standardized benchmarking.

\emph{3) Full-Body} (1 dataset, 390k images). RadImageNet-US~\dataref{data:radimagenet_us} dominates the ultrasound category with extensive coverage of 15 abdominal structures, though its commercial license may restrict accessibility.

\emph{4) Multi-structure} (2 datasets, 31,239 images). Two datasets, including ImageCLEF 2016~\dataref{data:imageclef2016} and AREN0534~\dataref{data:aren0534-us-2d}, cover multiple structures such as skin, breast, kidney, and lung.

\emph{5) Others} (12 datasets, $\sim$31,200 images). The remaining datasets focus on specific organs like the heart (CAMUS~\dataref{data:camus}), thyroid (TN-SCUI2020~\dataref{data:tnscui2020}), and brachial plexus (Ultrasound Nerve Segmentation~\dataref{data:uns}), or lack detailed anatomic descriptions. These multi-modal collections currently provide 6,203 images from APOLLO-5~\dataref{data:apollo5}, while CMB-LCA~\dataref{data:cmblca} has no available images.

\paragraph{Ultrasound Datasets by Tasks.}
Ultrasound datasets show several distinct task types represented, namely measurement, segmentation, and classification, along with tracking, estimation, and reconstruction. Among classification datasets, RadImageNet (US)~\dataref{data:radimagenet_us} has the largest image count, while ImageCLEF 2016~\dataref{data:imageclef2016} offers more classes (30).

\emph{1) Classification} (5 dataset, $\sim$421,500 images). RadImageNet-US~\dataref{data:radimagenet_us} offers large-scale multi-class classification across 15 abdominal categories.

\emph{2) Segmentation} (8 dataset, $\sim$26,300 images). Multiple datasets including BUSI~\dataref{data:busi}, CAMUS~\dataref{data:camus}, and the Ultrasound Nerve Segmentation~\dataref{data:uns} dataset provide pixel-level annotations for organ and tumor segmentation, supporting computer-aided diagnosis development.
The main challenge in these datasets is dealing with weak boundaries and acoustic shadowing artifacts.

\emph{3) Measurement} (1 dataset, 1,300 images). HC18~\dataref{data:hc18} specializes in biometric measurement tasks, particularly fetal head circumference calculation.

\emph{4) Unlabeled datasets} (3 datasets). The TCIA datasets (APOLLO-5~\dataref{data:apollo5}, CMB-LCA~\dataref{data:cmblca}, and AREN0532~\dataref{data:aren0532-us-2d}) currently lack labels; APOLLO-5~\dataref{data:apollo5} contains 6,203 images and AREN0532~\dataref{data:aren0532-us-2d} contains 1,021 images, while CMB-LCA~\dataref{data:cmblca} has none available, though their multi-modal nature may enable future fusion studies.

\subsection{X-Ray Images}

As one of the oldest and most common medical imaging techniques, 2D X-ray (radiography) provides a projectional view of anatomical structures, excelling at visualizing bone and air-filled spaces like the lungs. X-ray datasets are characterized by their large volume, particularly for chest imaging, driven by routine screening for diseases like pneumonia and tuberculosis. The primary tasks are classification of pathologies and segmentation or localization of abnormalities, though the overlapping of anatomical structures in the 2D projection can make these tasks challenging.
\Cref{tab:2d_xray_datasets} shows the 61 major X-ray imaging datasets from diverse sources, including TCIA, Grand Challenges, and open data platforms. These collections comprise approximately \textbf{1,657,000} images in total. The CheXmask~\dataref{data:chexmask} dataset dominates the quantity with 676,800 images for lung segmentation, followed by the CheXpert~\dataref{data:chexpert} and VICTRE~\dataref{data:victre} datasets, while most other datasets range from hundreds to thousands of samples, presenting a long-tail distribution common in medical imaging.

\paragraph{X-Ray Datasets by Anatomical Regions/Structures.}
The collected X-ray datasets cover diverse anatomical regions, with a strong emphasis on thoracic imaging due to its clinical prevalence in pulmonary and cardiac diagnostics. Approximately 46\% of the datasets focus on the chest/lung region, reflecting the widespread use of X-rays for respiratory disease screening (\eg, COVID-19, pneumonia). Other anatomical regions/structures are less represented, with limited datasets for musculoskeletal, neurological, and abdominal applications.

\emph{1) Thorax/Lung} (28 datasets, $\sim$537,900 images). This category dominates the X-ray collections, including large-scale datasets like NIH Chest X-ray 14~\dataref{data:nih_chestxray14} (112,100 images) and CheXpert~\dataref{data:chexpert} (224,300 images). These datasets are notable for their multi-label classification tasks, where a single image can be associated with multiple pathologies. The ChestX-Det~\dataref{data:chestxdet} series (3,600 images) provides detailed annotations for lung pathologies, while MIDRC-RICORD-1c~\dataref{data:midrc_ricord1c} (1,300 images) supports COVID-19 research. Smaller datasets like JSRT~\dataref{data:jsrt} (247 images) focus on pneumonia and pulmonary nodules.

\emph{Breast / Mammography} (3 dataset, $\sim$248,300 images). VICTRE~\dataref{data:victre} dominates this category. VICTRE’s~\dataref{data:victre} massive scale underscores breast imaging’s importance but lacks disease annotations. Mammography datasets are characterized by the need to detect subtle signs of cancer, such as microcalcifications and masses, in dense breast tissue.

\emph{2) Musculoskeletal} (8 datasets, $\sim$15,681 images). Musculoskeletal datasets include spine (AASCE~\dataref{data:aasce}, 609 images), clavicle (CRASS~\dataref{data:crass}, 518 images), and pelvic bone (PENGWIN2024-Task2~\dataref{data:pengwin2024}, 150 images) studies. The TCB-Challenge~\dataref{data:tcb_challenge} (174 images) targets osteoporosis detection via bone radiographs, highlighting X-ray’s role in orthopedic diagnostics. A common task in these datasets is fracture detection and classification.

\emph{3) Brain/Head} (2 datasets, $\sim$1,400 images). Brain datasets are limited to DENTEX~\dataref{data:dentex} (1,000 images) for dental imaging and Cephalometric X-ray Image~\dataref{data:ceph_xray} (400 images) for cephalometric analysis, indicating a gap in neurological X-ray datasets compared to CT/MRI.

\emph{4) Multi-structure} (5 datasets, $\sim$186,200 images). This category includes datasets spanning multiple distinct anatomical regions, such as MedMNIST~\dataref{data:medmnist_xray} (100,000 images) and MURA~\dataref{data:mura} (40,000 images).

\emph{5) Others} (8 datasets, $\sim$700,000 images). Includes generic collections like the CheXmask~\dataref{data:chexmask} (676,800 images) and X-ray Pneumonia Image Dataset~\dataref{data:chestxray_pneumonia} (5,900 images) without detailed anatomic labels.

\paragraph{X-Ray Datasets by Tasks.}
The datasets exhibit clear task specialization, with classification being the most prevalent application scenario. Notably, 31\% of the collections (19/61) provide pixel-level annotations or detection labels, reflecting the clinical demand for precise localization in diagnostic imaging.

\emph{1) Classification} (30 datasets, $\sim$670,100 images). This category represents the largest task group, predominantly focusing on pulmonary and COVID-19 related diagnoses. Key collections include CheXpert~\dataref{data:chexpert} (224,300 images), NIH Chest X-ray 14~\dataref{data:nih_chestxray14} (112,100 images), and RANZCR CLiP~\dataref{data:ranzcr_clip} (30,100 images, catheter classification). The JSRT~\dataref{data:jsrt} dataset, though small (247 images), provides valuable multi-class annotations for both pneumonia and pulmonary nodules.

\emph{2) Segmentation} (10 datasets, $\sim$708,500 images). These datasets emphasize anatomical structure delineation, with CheXmask~\dataref{data:chexmask} (676,800 images) and Pneumothorax Masks X-Ray~\dataref{data:pneumothorax_masks} (12,000 images) being the most substantial. The Pulmonary Chest X-Ray~\dataref{data:pulmonary_seg} dataset (800 images) specifically targets lung abnormality segmentation, while CRASS~\dataref{data:crass} (518 images) focuses on clavicle identification for orthopedic applications.

\emph{3) Detection/Localization} (9 datasets, $\sim$59,700 images). Emerging needs for surgical planning are addressed by DENTEX~\dataref{data:dentex} (1,005 brain images) and CL-Detection2023~\dataref{data:cl_detection2023} (555 images). The CEPHA29~\dataref{data:cepha29} dataset (1,000 images) stands out for cephalometric landmark localization, despite its current data accessibility issues.

\emph{4) Others} (5 datasets, $\sim$36,100 images). Unique applications include AASCE's~\dataref{data:aasce} spinal curvature regression (609 images), CoronARe's~\dataref{data:coronare} vascular reconstruction, and RSNA Bone Age's~\dataref{data:rsna_bone_age} bone age estimation (14,200 images). These demonstrate X-ray's versatility beyond conventional diagnostic roles.

\subsection{Optical Coherence Tomography (OCT) Images}

OCT provides micrometer-resolution, cross-sectional images of biological tissues in real-time. It is analogous to "optical ultrasound," using light instead of sound. Its primary application is in ophthalmology for imaging the layers of the retina. Consequently, OCT datasets are highly specialized, focusing on tasks like retinal layer segmentation for thickness mapping and classification of retinal diseases based on layer morphology.
\Cref{tab:2d_oct_datasets} provides 22 major optical coherence tomography (OCT) imaging datasets from diverse sources, including Kaggle, Grand Challenges, and academic institutions. These collections demonstrate remarkable specialization in retinal imaging, comprising approximately \textbf{221k} images in total. Two large public classification benchmarks — OCT2017~\dataref{data:oct2017} (about 83.5k images) and MedMNIST~\dataref{data:medmnist_oct} (100k images) — account for the majority of images in the corpus. In contrast, most other datasets range from hundreds to thousands of samples, presenting a typical long-tail distribution in medical imaging resources. 

\paragraph{OCT Datasets by Anatomical Regions/Structures.}
Notably, almost all of the datasets focus exclusively on retinal applications, reflecting OCT's primary clinical use in ophthalmology. The only exception is MedMNIST~\dataref{data:medmnist_oct}, which can also be applied to breast and lung. As such, we do not break down to introduce the anatomical regions/structures.

\paragraph{OCT Datasets by Tasks.}
The datasets exhibit clear task specialization, with classification and segmentation being the most prevalent application scenarios. The classification task has the largest number of images, though the number of datasets for classification is less than that of the segmentation task. Segmentation datasets account for approximately 50\% of the datasets, providing pixel-level annotations for precise anatomical analysis. 

\emph{1) Classification} (5 datasets, $\sim$210,200 images). This category represents the largest task group in terms of image number, predominantly focusing on diabetic retinopathy and glaucoma detection. Key collections include OCT2017~\dataref{data:oct2017} (83,484 images), Retinal OCT-C8~\dataref{data:retinal_oct_c8} (24,000 images), and MedMNIST~\dataref{data:medmnist_oct} (100,000 images combining multiple modalities). The core task in these datasets is to distinguish diseases based on morphological changes in retinal layers, such as the presence of drusen or intraretinal fluid. The iChallenge-AGE19~\dataref{data:ichallenge_age19} dataset (1,600 images) specifically targets glaucoma classification with detailed angle closure annotations.

\emph{2) Segmentation} (11 datasets, $\sim$2,600 images). These datasets emphasize retinal layer delineation, with SinaFarsiu-009~\dataref{data:sinafarsiu009} (840 images) and SinaFarsiu-018~\dataref{data:sinafarsiu018} (784 images) providing the most substantial annotations. The DRAC22~\dataref{data:drac22} dataset (174 images) specializes in diabetic retinopathy lesion segmentation, while iChallenge-GOALS~\dataref{data:ichallenge_goals} (300 images) offers three-layer retinal segmentation crucial for thickness measurements.

\emph{3) Prediction} (3 datasets, $\sim$8,500 images). The APTOS series (APTOS-2021~\dataref{data:aptos2021}, APTOS Cross-Country Stage 1~\dataref{data:aptos_stage1}, and APTOS Cross-Country Stage 2~\dataref{data:aptos_stage2}) total 8,500 images for diabetic retinopathy severity prediction, using the International Clinical Diabetic Retinopathy scale. These datasets demonstrate OCT's growing role in quantitative disease progression monitoring.

\subsection{Fundus Images}

Fundus photography captures high-resolution color images of the retina, making it a cornerstone of ophthalmology. A key characteristic of fundus datasets is their similarity to natural RGB images in terms of data format, which allows for the direct application and transfer learning of models developed for general computer vision. However, the content is highly specialized, featuring unique anatomical landmarks like the optic disc, fovea, and a complex network of blood vessels. Common tasks revolve around detecting and grading pathologies such as diabetic retinopathy and glaucoma. The challenge lies in identifying these subtle, often minute, pathological features within a complex anatomical background.
\Cref{tab:2d_fundus_datasets} shows 75 major fundus photography datasets from diverse sources, including Grand Challenges, Kaggle, and academic institutions. These collections demonstrate remarkable specialization in retinal imaging, comprising approximately \textbf{412,400} images in total. The AIROGS~\dataref{data:fundus_airogs} dataset dominates the quantity with 101,400 images, while most other datasets range from hundreds to thousands of samples, presenting a typical long-tail distribution in medical imaging resources. Notably, almost all of the datasets focus exclusively on retinal applications, reflecting fundus photography's primary clinical use in ophthalmology diagnostics.

\paragraph{Fundus Photography Datasets by Anatomical Regions/Structures.}
The collected datasets exclusively focus on retinal imaging, reflecting fundus photography's specialized application in ophthalmology. All 75 datasets target the retina, with varying emphasis on specific anatomical structures or pathological features. This extreme specialization contrasts with other modalities like CT or MRI that cover multiple body regions.

\paragraph{Fundus Photography Datasets by Tasks.}
The datasets exhibit clear task specialization, with classification being the most prevalent application scenario. Approximately 30\% of the collections provide pixel-level annotations or detection labels, enabling precise anatomical analysis crucial for diagnostic applications.

\emph{1) Classification} (42 datasets, $\sim$304,200 images). This category represents the largest task group, predominantly focusing on diabetic retinopathy and glaucoma detection. Key collections include OIA-ODIR~\dataref{data:fundus_oia_odir} (10,000 images), APTOS 2019~\dataref{data:fundus_aptos2019_kaggle} (5,590 images for diabetic retinopathy grading), and Yangxi~\dataref{data:fundus_yangxi} (20,394 images for eye axis classification). These datasets are pivotal for developing automated screening systems for prevalent eye diseases, often framed as multi-class grading problems based on the number and type of lesions present. The JSIEC~\dataref{data:fundus_jsiec} dataset (1,000 images) stands out for its comprehensive coverage of 38 fundus disease categories, though sample sizes per category remain limited.

\emph{2) Segmentation} (21 datasets, $\sim$5,300 images). These datasets emphasize retinal structure delineation, with RIM-ONE~\dataref{data:fundus_rim_one} (485 images) and GAMMA CFP~\dataref{data:fundus_gamma_task3_cfp} (200 images) providing optic disc/cup annotations crucial for glaucoma assessment. The HRF Seg~\dataref{data:fundus_hrf_seg} dataset (45 images) offers high-resolution vessel segmentation, while AO-SLO~\dataref{data:fundus_ao_slo} (840 images) specializes in photoreceptor mapping. The iChallenge-GAMMA series (\dataref{data:fundus_ichallenge_gamma_2dfundus}, \dataref{data:fundus_ichallenge_gamma_3doct}) demonstrates growing interest in multi-modal retinal analysis. Segmentation tasks are critical for quantitative analysis, focusing on delineating blood vessels to assess vascular health, the optic disc and cup to measure glaucomatous changes, and lesions like exudates or hemorrhages to quantify disease severity.

\emph{3) Regression} (6 datasets, $\sim$2,300 images). The INSPIRE series (\dataref{data:fundus_inspire_stereo}, \dataref{data:fundus_inspire_avr}) (70 images combined) focuses on arteriovenous ratio measurement, while DeepDR-Task2~\dataref{data:fundus_deepdr_task2} (2,000 images) addresses disease progression prediction. These datasets highlight fundus photography's expanding role in quantitative disease monitoring.

\subsection{Dermoscopy Images}

Dermoscopy involves imaging the skin with a specialized magnifying lens to visualize subsurface structures not visible to the naked eye. These datasets are crucial for the early detection of skin cancer, particularly melanoma. The images are typically high-resolution RGB photos of skin lesions. Key tasks include the segmentation of lesion boundaries and the classification of lesions into categories (\eg, benign nevus, melanoma, basal cell carcinoma).
There are 17 major dermoscopy imaging datasets in our collection, as shown in \Cref{tab:2d_dermoscopy_datasets}. They are collected from various sources, including ISIC challenges, CVPR competitions, and independent research collections. These datasets predominantly focus on skin imaging. They primarily address segmentation and classification tasks, with a strong emphasis on skin lesion analysis. These datasets include approximately \textbf{167,300} images in total, with Monkeypox~\dataref{data:dermo_monkeypox} having the largest single collection (40,200 images) and ISIC20~\dataref{data:dermo_isic20} (33,100 images), ISIC19~\dataref{data:dermo_isic19} (25,300 images), and Fitzpatrick17k~\dataref{data:dermo_fizpatrick17k} (16,600 images) also providing substantial sample sizes for training medical imaging models.

\paragraph{Dermoscopy Datasets by Anatomical Regions/Structures.} 
The vast majority of these datasets focus on skin imaging, though a few cover other anatomical regions. \emph{1) Skin} (13 datasets, $\sim$133,600 images). This dominant category includes all ISIC challenge datasets (ISIC16-20~\dataref{data:dermo_isic16, data:dermo_isic17, data:dermo_isic18, data:dermo_isic19, data:dermo_isic20}), Fitzpatrick17k~\dataref{data:dermo_fizpatrick17k}, MED-NODE~\dataref{data:dermo_mednode}, PH2~\dataref{data:dermo_ph2}, and others. The largest collections are Monkeypox~\dataref{data:dermo_monkeypox} (40,200 images), ISIC20~\dataref{data:dermo_isic20} (33,100 images), and ISIC19~\dataref{data:dermo_isic19} (25,300 images). These datasets demonstrate strong clinical focus on melanoma detection and skin lesion analysis. \emph{2) Foot} (1 dataset, 2,000 images). DFUC2020~\dataref{data:dermo_dfuc2020} specifically targets foot imaging for diabetic foot ulcer analysis. \emph{3) Thyroid} (1 dataset, 637 images). DDTI focuses on thyroid nodule segmentation. \emph{4) Multi-structure} (1 datasets, $\sim$31,000 images). ImageCLEF2016~\dataref{data:dermo_imageclef_a} covers skin, cell, and breast imaging with 31,000 images.

\paragraph{Dermoscopy Datasets by Tasks.} 
The collected datasets show clear task specialization, with most providing high-quality labels suitable for supervised learning. \emph{1) Segmentation} (5 datasets, $\sim$9,400 images). Key collections include ISIC16~\dataref{data:dermo_isic16} (1,279 images), ISIC17~\dataref{data:dermo_isic17} (2,750 images), ISIC18~\dataref{data:dermo_isic18} (2,694 images), and DDTI (637 images). These typically focus on precise lesion boundary delineation. \emph{2) Classification} (10 datasets, $\sim$157,500 images). Major collections include Monkeypox~\dataref{data:dermo_monkeypox} (40,200 images), ISIC20~\dataref{data:dermo_isic20} (33,100 images), ImageCLEF 2016~\dataref{data:dermo_imageclef_a} (31,000 images), and ISIC19~\dataref{data:dermo_isic19} (25,300 images). These datasets often provide multi-class categorization of skin lesions. \emph{Unlabeled dataset} (1 dataset, 368 images). Vitiligo~\dataref{data:dermo_vitiligo} is the only unlabeled collection, potentially useful for unsupervised learning.

\subsection{Histopathology}

Histopathology is the microscopic examination of tissues to study the manifestations of disease. Digital pathology datasets, particularly those based on Whole Slide Images (WSIs), possess unique characteristics. WSIs are gigapixel-resolution images, often exceeding 100,000$\times$100,000 pixels, which makes it computationally infeasible to process them directly. Consequently, a standard preprocessing pipeline involves patch extraction or tiling, where the WSI is divided into thousands of smaller, manageable patches. Common tasks include patch-level classification (\eg, identifying tumorous vs. normal tissue), object-level segmentation or detection (\eg, delineating nuclei, glands, or mitotic figures), and WSI-level classification for diagnosis. The challenges in this modality stem from the massive image size, significant variations in staining and preparation, and the need to aggregate patch-level predictions into a coherent slide-level diagnosis.
\Cref{tab:2d_histopathology_datasets_part1,tab:2d_histopathology_datasets_part2}
 present 117 major histopathology imaging datasets from diverse sources, including grand challenges (MICCAI, ISBI), open data platforms (TCGA, TCIA, OpenDataLab), and research collections. These datasets predominantly utilize hematoxylin and eosin (H\&E) staining, with some incorporating immunohistochemistry (IHC). They collectively contain approximately \textbf{2.22 million images} (comprising $\sim$2.15 million patch images and $\sim$67,000 WSI), with the Quilt-1M~\dataref{data:histo_quilt1m} (1,000,000 images) and PatchCamelyon (PCam)~\dataref{data:histo_patchcamelyon} (328,000 images) being the largest collections. Notably, 82\% of datasets provide high-quality labels suitable for supervised learning.
 The prohibitive cost of large-scale WSI annotation catalyzed a shift towards SSL, enabling the rise of Pathology Foundation Models from vast unlabeled data archives. Initial development centered on algorithmic innovations using public datasets like TCGA. A subsequent "scale revolution" utilized massive, private "real-world" datasets, powering models like UNI (trained on over 100,000 WSIs) and Prov-GigaPath (trained on over 171,000 WSIs). This addressed the "domain shift" limitations of public data, proving that dataset scale is now a primary engine of progress in the field.

\paragraph{Histopathology Datasets by Anatomical Regions/Structures.} 
The datasets show a strong clinical focus on cancer diagnosis across multiple anatomical sites. 
\emph{1) Breast} (25 datasets, $\sim$53,000 images). Major collections include BRIGHT~\dataref{data:histo_bright} (5,086 images), BRCA-M2C~\dataref{data:histo_brca_m2c} (120 images), and the BreakHis series 
(\#\ref{data:histo_breakhis_40x}, \#\ref{data:histo_breakhis_100x},  \#\ref{data:histo_breakhis_100x}, \#\ref{data:histo_breakhis_400x})
(combined 35,236 images across magnifications). These primarily address tumor classification and segmentation. 
\emph{2) Prostate} (9 datasets, $\sim$42,000 images). PANDA~\dataref{data:histo_panda_main} (10,616 images) and SICAPv2~\dataref{data:histo_sicapv2} (18,783 images) are the largest, focusing on Gleason grading. 
\emph{3) Colon/Rectum} (12 datasets, $\sim$113,000 images). CRC100K~\dataref{data:histo_crc100k} (100,000 images) and CoNIC2022~\dataref{data:histo_conic2022} (4,981 images) provide extensive data for colorectal cancer analysis. 
\emph{4) Multi-organ} (17 datasets, $\sim$1.18 million images). Quilt-1M~\dataref{data:histo_quilt1m} (1,000,000 images) and MedMNIST~\dataref{data:histo_medmnist} (100,000 images) cover multiple cancer types.
\emph{5) Others} include lung (7 datasets, $\sim$38,000 images), lymph nodes (9 datasets, $\sim$537,000 images), and blood (5 datasets, $\sim$53,000 images).

\paragraph{Histopathology Datasets by Tasks.}
The datasets demonstrate specialized task distributions. Emerging trends include increased WSI adoption (32\% of recent datasets) and multi-task collections combining segmentation with classification or counting.

\emph{1) Classification} (38 datasets, $\sim$709,000 images). Key datasets include LC25000~\dataref{data:histo_lc25000} (25,000 images, lung/colon classification) and Histopathologic Cancer Detection~\dataref{data:histo_cancer_detect_kaggle} (220,000 images). The BreakHis series (\#\ref{data:histo_breakhis_40x}, \#\ref{data:histo_breakhis_100x}, \#\ref{data:histo_breakhis_200x}, \#\ref{data:histo_breakhis_400x})
provides multi-magnification classification (40$\times$-400$\times$). A key challenge is handling intra-class variation and inter-class similarity at the cellular level, making fine-grained classification difficult.
\emph{2) Segmentation} (31 datasets, $\sim$368,000 images). Notable collections are GlaS~\dataref{data:histo_glas} (165 images, colorectal glands) and CRAG~\dataref{data:histo_crag} (213 images, extended from GlaS). Segmentation targets range from macro-structures like tumor regions to micro-structures like individual nuclei or glands, which are essential for quantitative pathology.
\emph{3) Detection} (6 datasets, $\sim$14,000 images). MIDOG2021~\dataref{data:histo_midog2021} (200 images) focuses on mitotic figure detection.
\emph{4) Multi-task} (4 datasets, $\sim$14,000 images). PanNuke combines segmentation and classification (PanNuke (Seg)~\dataref{data:histo_pannuke_seg}, 7,901 images), while CoNIC2022~\dataref{data:histo_conic2022} adds counting tasks.
\emph{5) Specialized tasks} include registration (ANHIR~\dataref{data:histo_anhir}, 481 images), generation (BCI~\dataref{data:histo_bci}, 4,900 images) and VQA (Quilt-1M~\dataref{data:histo_quilt1m}, 1,000,000 images).

\subsection{Microscopy Imaging}

\Cref{tab:2d_microscopy_datasets} summarizes 34 major microscopy imaging datasets. These datasets predominantly utilize brightfield and fluorescence microscopy, with a strong focus on cellular and subcellular imaging. They collectively contain approximately \textbf{1.8 million images}, with the CellTracking2019~\dataref{data:micro_celltrack2019} dataset (1.44 million images), DLBCL-Morph~\dataref{data:micro_dlbcl_morph} (152,200 images), and Kaggle-HPA~\dataref{data:micro_kaggle_hpa} (89,460 images) being the largest collections. Unlike histopathology which focuses on tissue architecture, these microscopy datasets often center on the morphology, count, and behavior of individual cells or microorganisms. Notably, most datasets provide high-quality labels suitable for supervised learning, covering a wide range of biological scales from single molecules to whole organisms.

\paragraph{Microscopy Datasets by Anatomical Regions/Structures.} 
The datasets demonstrate specialized focus on specific anatomical structures:
\emph{1) Cellular} (8 datasets, $\sim$1.51M images). Key collections include CellTracking2019~\dataref{data:micro_celltrack2019} (16,042 sequences, 1.44M frames), Kaggle-HPA~\dataref{data:micro_kaggle_hpa} (89,460 images), and OCCISC (\dataref{data:micro_occisc_semseg}, \dataref{data:micro_occisc_instseg}) (945 images). These primarily address cell segmentation and tracking.
\emph{2) Ocular} (5 datasets, $\sim$153,000 images). The corneal series (CornealNerve~\dataref{data:micro_corneal_nerve}, NerveTortuosity~\dataref{data:micro_corneal_tortuosity}, CornealEndothelial~\dataref{data:micro_corneal_endothelial}) and DLBCL-Morph~\dataref{data:micro_dlbcl_morph} (152,200 images) focus on eye microstructure analysis.
\emph{3) Breast} (1 dataset, 400 images). ICIAR2018~\dataref{data:micro_iciar2018_micro} provides histopathology images for breast cancer classification.
\emph{4) Blood} (3 datasets, $\sim$28,500 images). Blood Cell Images~\dataref{data:micro_blood_cell_tianchi} (12,500 images) and Leukemia Classification~\dataref{data:micro_leukemia_tianchi} (15,100 images) analyze blood cell morphology.
\emph{5) Multi-structure} (2 datasets, $\sim$31,500 images). ImageCLEF2016~\dataref{data:micro_imageclef2016} (31,000 images) covers multiple tissue types.

\paragraph{Microscopy Datasets by Tasks.} 
There is an increased use of deep learning benchmarks (Kaggle-HPA~\dataref{data:micro_kaggle_hpa}) and integration of multiple tasks (CBC series~\dataref{data:micro_cbc_count, data:micro_cbc_detect} combining counting and detection). The datasets show clear specialization in analysis tasks:
\emph{1) Segmentation} (11 datasets, $\sim$99,000 images). Kaggle-HPA~\dataref{data:micro_kaggle_hpa} (89,500 images), CREMI~\dataref{data:micro_cremi}, and OCCISC-Seg~\dataref{data:micro_occisc_semseg} (945 images) provide precise cellular boundary delineation. A common challenge is accurately separating densely clustered or overlapping cells.
\emph{2) Classification} (12 datasets, $\sim$81,400 images). ImageCLEF 2016~\dataref{data:micro_imageclef2016} (31,000 images), B-ALL Classification~\dataref{data:micro_ball_cls} (15,100 images), and ICIAR2018~\dataref{data:micro_iciar2018_micro} (400 images) enable morphological categorization.
\emph{3) Detection} (3 datasets, $\sim$2,600 images). BloodCell~\dataref{data:micro_blood_cell_detect_heywhale} (874 images) and Tuberculosis~\dataref{data:micro_tuberculosis_heywhale} (1,265 images) localize specific cellular features.
\emph{4) Tracking} (1 datasets, $\sim$1.4M images). CellTracking2019~\dataref{data:micro_celltrack2019} dominates this category with 1.4 million time-lapse frames.
\emph{5) Specialized tasks} include regression (DLBCL-Morph~\dataref{data:micro_dlbcl_morph}, 152.2k images; CBC-Count~\dataref{data:micro_cbc_count}, 420 images) and protein localization (Kaggle-HPA~\dataref{data:micro_kaggle_hpa}).

\subsection{Infrared Imaging}

Infrared imaging in medicine captures thermal patterns or reflectance properties not visible in the normal spectrum. In the context of the collected datasets, it is primarily used in ophthalmology to image retinal structures with different light wavelengths. This modality is non-invasive and can provide unique contrast for features like the retinal pigment epithelium. The tasks often revolve around image quality assessment or classification based on specific features visible in the infrared spectrum.
\Cref{tab:2d_infrared_datasets} includes 6 major infrared reflectance imaging datasets. These collections focus exclusively on ocular imaging, particularly retinal analysis, using infrared reflectance technology. The datasets contain approximately \textbf{424,532 images} in total, with the MRL Eye series (\dataref{data:ir_mrl_glasses, data:ir_mrl_state, data:ir_mrl_reflections, data:ir_mrl_quality, data:ir_mrl_sensor}) (combined 424,490 images across 5 sub-datasets) representing the largest collection. All datasets provide high-quality labels suitable for supervised learning, with a strong emphasis on classification tasks (5/6 datasets).

\paragraph{Infrared Datasets by Anatomical Regions/Structures.} 
Infrared imaging remains highly specialized, with 100\% of datasets focusing on retinal applications, and all created since 2018, suggesting growing interest in this modality. Specifically, \emph{Retina} (six datasets, $\sim$424,532 images). The MRL Eye series (\dataref{data:ir_mrl_glasses, data:ir_mrl_state, data:ir_mrl_reflections, data:ir_mrl_quality, data:ir_mrl_sensor}) (84,898 images per sub-dataset) provides comprehensive coverage of various retinal features. This extreme specialization in retinal imaging contrasts with other modalities that typically cover multiple anatomical regions.

\paragraph{Infrared Datasets by Tasks.} 
The datasets show clear task specialization: \emph{1) Classification} (5 datasets, $\sim$424,490 images). The MRL Eye series addresses multiple classification tasks: glasses detection (MRL-Eye-Glasses~\dataref{data:ir_mrl_glasses}), eye state (MRL-Eye-State~\dataref{data:ir_mrl_state}), reflection analysis (MRL-Eye-Reflections~\dataref{data:ir_mrl_reflections}), image quality assessment (MRL-Eye-Quality~\dataref{data:ir_mrl_quality}), and sensor type identification (MRL-Eye-Sensor~\dataref{data:ir_mrl_sensor}). \emph{2) Segmentation} (one dataset, 42 images). RAVIR~\dataref{data:ir_ravir} is the only segmentation dataset, focusing on retinal blood vessel delineation with three classes (background, arteries, veins).

\subsection{Endoscopy Imaging}

Endoscopy provides direct real-time video visualization of internal organs and cavities through a flexible tube with a camera. Datasets are often composed of individual frames extracted from these videos. A key characteristic is the high variability in appearance due to camera motion, lighting changes, specularity, and physiological artifacts (\eg, bubbles, debris). Common tasks include polyp detection and segmentation for cancer screening, tool tracking for surgical navigation, and classification of tissue abnormalities.
We provide an overview of endoscopy imaging datasets in \Cref{tab:2d_endoscopy_datasets}, where 41 major ones are collected from diverse sources, \eg, ISBI and MICCAI. These datasets predominantly feature endoscopic imaging (39/41), with a few incorporating multi-modal data (2/41). They cover diverse anatomical regions and tasks, totaling approximately \textbf{322,200 images and videos}, with EndoSlam~\dataref{data:endo_endoslam} being the largest collection (76,837 images). Notably, 39\% of datasets (16/41) contain over 1,000 images, making them potentially suitable for training medical vision models.

\paragraph{Endoscopy Datasets by Anatomical Regions/Structures} 
The datasets cover several major anatomical regions, with strong emphasis on gastrointestinal tract examination:

\emph{1) Colon/Bowel} (8 datasets, $\sim$109,400 images): This represents the most extensively examined region, featuring large-scale datasets like SUN\_SEG~\dataref{data:endo_sun_seg} (49,136 images), SARAS-ESAD~\dataref{data:endo_saras_esad} (33,398 images), and Kavsir~\dataref{data:endo_kavsir} (14,000 images) for polyp segmentation and detection. The CVC series (CVC-ClinicDB~\dataref{data:endo_cvc_clinicdb}, CVC-ColonDB) provide high-quality annotations for polyps, while EndoCV2020~\dataref{data:endo_endocv2020_sub1} and EndoVis15~\dataref{data:endo_endovis15} focus on artifact detection.

\emph{2) Esophagus} (1 datasets, 157 images): Focused on Barrett's esophagus detection, with AIDA-E\_2~\dataref{data:endo_aida_e2} (157 images) providing a specialized benchmark.

\emph{3) Multi-structure gastrointestinal tract} (6 datasets, $\sim$86,000 images): Comprehensive collections like EndoSlam~\dataref{data:endo_endoslam} (76k images) cover the entire gastrointestinal tract including esophagus, stomach, and colon. These are particularly valuable for developing generalizable endoscopic AI systems.

\emph{5) Other Regions}: Includes specialized collections for uterus (FetReg~\dataref{data:endo_fetreg}, 2.7k images), gallbladder (m2cai16-tool~\dataref{data:endo_m2cai16_tool}, 15 videos), and prostate (SARAS-MESAD~\dataref{data:endo_saras_mesad}, 50k images). While clinically important, these generally have smaller sample sizes.

\paragraph{Endoscopy Datasets by Tasks} 
The datasets demonstrate a progression from single-task to multi-task benchmarks:

\emph{Segmentation} (17 datasets, $\sim$20,000 images): Forms a large task category, with Kvasir-SEG~\dataref{data:endo_kvasir_seg} (8,000 images), FetReg~\dataref{data:endo_fetreg} (2,718 images), and EndoVis 2018 - RSS~\dataref{data:endo_endovis18_rss} (2,840 images) providing high-quality segmentation masks. Most focus on polyp segmentation, while specialized targets include surgical tools (EndoVis 2018-RSS~\dataref{data:endo_endovis18_rss}) and placental vasculature (FetReg~\dataref{data:endo_fetreg}).

\emph{Detection} (6 datasets, $\sim$86,600 images): SARAS-MESAD~\dataref{data:endo_saras_mesad} (50,284 images) and SARAS-ESAD~\dataref{data:endo_saras_esad} (33,398 images) are notable for bounding box annotations of abnormalities and instruments. The m2cai series~\dataref{data:endo_m2cai16_tool} provide instrument detection benchmarks.

\emph{Classification} (10 datasets, $\sim$77,700 images): Ranges from binary classification (MedFM2023) to fine-grained categorization (ImageCLEF~\dataref{data:endo_imageclef2016}). AIDA series (E1-E3)~\dataref{data:endo_aida_e1, data:endo_aida_e2, data:endo_aida_e3} provide histology classification benchmarks.

\emph{Multi-task datasets} (5 datasets, ~156k images): HyperKvasir~\dataref{data:endo_hyperkvasir} (captioning, classification, localization), SUN\_SEG~\dataref{data:endo_sun_seg} (segmentation, detection, classification), and Endo-FM~\dataref{data:endo_endofm} combine multiple annotation types, reflecting recent trends towards comprehensive benchmarks.

\emph{Others}: Includes reconstruction and depth estimation (EndoSlam~\dataref{data:endo_endoslam}) and registration (P2ILF~\dataref{data:endo_p2ilf}). Some of these tasks, like in the EndoSlam~\dataref{data:endo_endoslam} dataset (76,837 images), are supported by a large number of samples.

\subsection{Other Modalities}

Finally, we introduce all the 2D datasets of other modalities that are not listed in the previous sub-sections. This section consolidates datasets from a variety of imaging modalities that, while less numerous than the major categories, represent important and often specialized clinical applications. \Cref{tab:2d_others_datasets} summarize the information of these modalities, spanning diverse modalities, including Mammography (4 datasets), X-Ray (3), Fundus (2), Colposcopy (2), and others. These datasets collectively contain approximately \textbf{858,000 images}, with the Digital Mammography~\dataref{data:other_digital_mammo} dataset being the largest (640,000 images), followed by MRL Eye Gender~\dataref{data:other_mrl_gender} (84,898 images) and ADDI ALZHEIMER'S DETECTION CHALLENGE~\dataref{data:other_addi_alz} (34,614 images). The datasets demonstrate a strong emphasis on classification tasks (75\%) and cover all major anatomical regions, though with uneven distribution across modalities.

\paragraph{Datasets by Anatomical Regions/Structures.} 
The datasets cover comprehensive anatomical structures with a particular concentration on thoracic and retinal imaging. 
\emph{1) Thoracic/Lung} (2 datasets, $\sim$27,000 images). This category includes collections like VinDr-CXR~\dataref{data:other_vindr_cxr} (18,000 images) and VinDr-PCXR~\dataref{data:other_vindr_pcxr} (9,125 images) for lung abnormalities.
\emph{2) Retina} (3 datasets, $\sim$88,000 images). Retinal imaging features collections like MRL Eye Gender~\dataref{data:other_mrl_gender} (84,898 images) and specialized datasets for various ophthalmic diseases.
\emph{3) Breast} (4 datasets, $\sim$663,000 images). The Digital Mammography~\dataref{data:other_digital_mammo} dataset dominates this category with 640,000 images, supplemented by specialized collections like CMMD~\dataref{data:other_cmmd} (1,775) and VinDr-Mammo~\dataref{data:other_vindr_mammo} (19,992).
\emph{4) Brain/Head} (2 datasets, $\sim$5,000 images). While smaller in quantity, these include important collections like Br35H~\dataref{data:other_br35h} (3,060) for brain tumors.
\emph{5) Whole-body/Multi-structure} collections like OralCancer~\dataref{data:other_oralcancer} (131 images) provide cross-anatomical coverage.

\paragraph{Datasets by Tasks.} 
The datasets demonstrate clear task specialization across modalities. 
\emph{1) Classification} (15 datasets, $\sim$798,000 images): Mammography datasets like The Digital Mammography DREAM Challenge~\dataref{data:other_digital_mammo} and retinal collections (MRL Eye Gender~\dataref{data:other_mrl_gender}) dominate this category.
\emph{2) Segmentation} (4 datasets, $\sim$2,300 images): Notable collections include CDD-CESM~\dataref{data:other_cdd_cesm} (2,006 images).
\emph{3) Multi-task} datasets like CDD-CESM~\dataref{data:other_cdd_cesm} (segmentation+classification) provide versatile training opportunities.
\emph{4) Emerging tasks} like reconstruction (BigNeuron~\dataref{data:other_bigneuron}) demonstrate expanding research frontiers.









\subsection{Challenge and Opportunity}

The landscape of 2D medical imaging datasets presents a distinct duality. On one hand, its sheer volume, particularly in modalities like histopathology and radiography, offers a scale for model pre-training that is unparalleled in the medical domain. On the other hand, this abundance is coupled with significant fragmentation, heterogeneity, and the inherent limitations of two-dimensional representations, posing unique challenges for the development of robust and generalizable foundation models.

\paragraph{Key Challenges in 2D Medical Imaging Datasets.}
The primary obstacles stem from the diversity and nature of 2D data acquisition and annotation practices. \emph{Extreme fragmentation and heterogeneity} represent a major barrier. The vast number of 2D datasets are scattered across numerous independent repositories and challenges, often with inconsistent imaging protocols, varying resolutions, and non-standardized metadata. This leads to significant domain shifts between datasets of the same modality, complicating large-scale integration efforts. For instance, histopathology slides exhibit wide variations in staining and preparation, while chest X-rays differ in projection and exposure settings.

\emph{Pervasive data imbalance and long-tail distributions} introduce substantial biases. As our analysis reveals, modalities like pathology, X-ray, and fundus photography dominate the data landscape, while clinically vital modalities such as endoscopy and ultrasound remain underrepresented. This imbalance extends to anatomical regions and tasks; for example, over 80\% of images in our collection come from just thoracic and breast datasets, leaving other regions (\eg, abdominal organs) critically underserved. This also creates modality-specific limitations; for instance, X-Ray datasets in this collection average only $\sim$12.5K images per dataset. Foundation models pre-trained on such skewed data may fail to generalize to less common modalities or pathologies, limiting their clinical utility.

Furthermore, \emph{annotation quality and scalability} present a persistent challenge. The creation of large-scale 2D datasets often relies on weak supervision, such as labels extracted from radiology reports, which can be noisy and imprecise. While pixel-level annotations are the gold standard, they are labor-intensive and scarce at scale. The lack of a unified annotation ontology across datasets makes it difficult to harmonize labels for multi-dataset training, hindering the creation of truly comprehensive benchmarks.

Finally, the \emph{inherent limitation of 2D representation} is a fundamental constraint. A single 2D image, whether a projection like an X-ray or a slice from a volume, provides only a partial view of the underlying three-dimensional anatomy. This loss of spatial context can be a critical handicap for diagnosing complex diseases that require volumetric understanding, such as assessing tumor morphology or subtle structural changes.

\paragraph{Opportunities for Advancement.}
Despite these challenges, the 2D medical imaging domain offers exceptional opportunities to advance foundation models. The \emph{unprecedented scale for self-supervised pre-training} is the most significant advantage. With millions of available images, thoracic imaging (pathology and chest radiography) has achieved a critical mass for large-scale AI training. This scale, alongside exceptionally standardized large collections (such as the >80K retinal image datasets), enables the effective application of self-supervised learning paradigms, such as masked auto-encoding and contrastive learning, to build foundational backbones that can be fine-tuned for a multitude of downstream tasks.

The \emph{rich diversity of modalities enables powerful multi-modal learning}. The breadth of 2D imaging, spanning from macroscopic radiographic images to microscopic pathology slides, provides a fertile ground for developing models that can reason across different biological scales and data sources. A particularly promising avenue is the integration of imaging data with unstructured clinical text. Large datasets paired with radiology reports, such as MIMIC-CXR~\cite{johnson2019mimic} and CheXpert~\cite{irvin2019chexpert}, unlock the potential for vision-language pre-training, allowing models to learn semantically rich representations that align visual features with clinical narratives.

Moreover, the widespread clinical use and lower cost of 2D imaging modalities create opportunities for \emph{high-impact, scalable clinical applications}. Foundation models trained on common 2D data like X-rays, fundus, or dermoscopy images can be deployed for large-scale screening programs in resource-constrained settings. This can democratize access to expert-level diagnostics for conditions like tuberculosis, diabetic retinopathy, and skin cancer, addressing critical global health challenges.

In summary, while the path to building generalist 2D medical foundation models is fraught with challenges of data heterogeneity and annotation quality, the opportunities are immense. Strategic dataset consolidation, prioritization of balanced anatomical coverage, and the development of standardized multi-task annotations, coupled with advanced self-supervised and multi-modal learning techniques, can harness the vast scale of 2D data to create transformative AI tools for global healthcare.
\section{3D Medical Image Datasets}
\label{sec:3d_data}


We have collected 591 3D medical image datasets, comprising \textbf{1,242,022+ volumes} in total. Although the total number of volumes is considerably smaller than that of 2D datasets, 3D datasets provide richer spatial information that is essential for volumetric analysis and clinical decision-making. We categorize these 3D datasets according to their modalities, tasks, and body parts. The labeled datasets dominate the collection, while unlabeled datasets provide additional opportunities for self-supervised learning approaches.

\begin{figure}
  \centering
  \includegraphics[trim=0 0 0 0, clip, width=\linewidth]{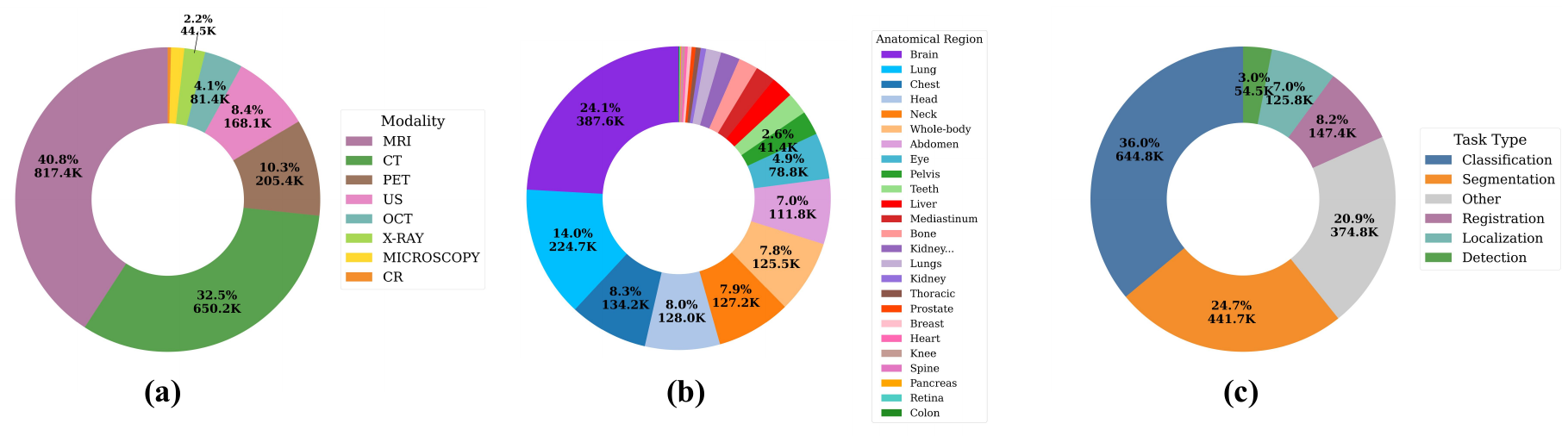}
\caption{The distribution of different (a) modalities, (b) anatomical structures, and (c) tasks for 3D datasets. Each slice of the pie chart shows the percentage and the actual number of images.}
\label{fig:3d_datasets_overview}
\end{figure}

\subsection{Overview}
\begin{figure}
    \centering
    \includegraphics[trim=0 0 0 0, clip, width=\linewidth]{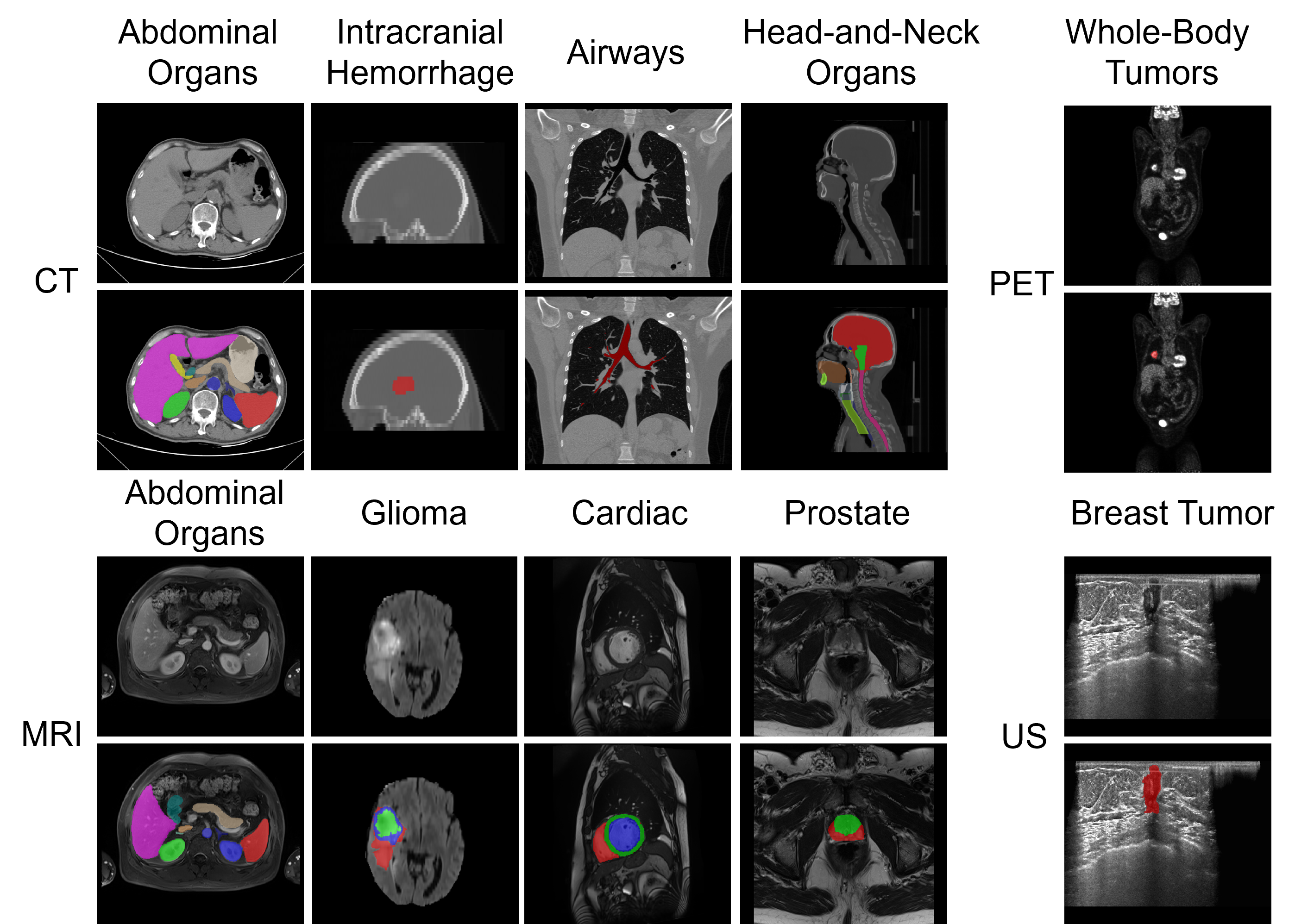}
    \caption{3D visualization examples of medical imaging datasets across different modalities and anatomical structures.}
    \label{fig:3d_visualization}
\end{figure}

We first provide an overview of 3D medical image datasets. Figure~\ref{fig:3d_datasets_overview} shows the distributions of different modalities, anatomical structures, and tasks for 3D datasets, which represent clear long-tail distributions. In terms of modality, MRI and CT are the most popular, while other modalities, like PET, ultrasound, and OCT, are less representative. From the perspective of anatomical structures, the brain, abdomen, and lung have the largest number of datasets, while the prostate, teeth, and other structures are still limited in their dataset numbers. The dominating tasks include classification, segmentation, and other tasks. However, other tasks, \eg, registration, localization, and detection, have much fewer datasets. Figure~\ref{fig:3d_visualization} demonstrates representative examples of the collected 3D medical image datasets across different modalities and anatomical regions.

\subsection{CT Volumes}

CT is a widely used imaging modality that employs X-rays to visualize internal structures in three dimensions. We identify 252 3D CT datasets comprising approximately 516,087 volumes in total, as summarized in Table~\ref{tab:3d_ct_datasets}. These datasets diverge considerably in scale and annotation quality, from small, domain-specific collections (for example, 3D-IRCADb \cite{ircad2010dircadb} with 20 liver volumes) to large, multi-center compilations such as CT-RATE \cite{hamamci2024developing} (50,188 volumes). Large collections like CT-RATE and M3D \cite{bai2024md} aim to cover a wide range of acquisition protocols but often depend on semi-automated or weak supervision for annotations, while curated challenge datasets like TotalSegmentator \cite{wasserthal2023totalsegmentator} (1,204 volumes) deliver expert-verified labels across 104 anatomical structures. Annotation consistency remains a persistent challenge: manual lesion delineation is laborious, operator-dependent, and subject to inter-observer variability, as illustrated by segmentation benchmarks such as LiTS \cite{bilic2019the}. Regarding clinical representativeness, CT datasets range from broad population-based cohorts like NLST \cite{team2013data} to small, specific single-center collections (e.g. 3D-IRCADb), whereas multi-institution benchmarks like AMOS \cite{ji2022amos} (500 CT + 100 MRI scans, collected across multiple centers and vendors) better reflect real-world diversity in scanner types and imaging protocols \cite{ji2022amos}.

\paragraph{CT Datasets by anatomical structures.}
CT datasets show strong concentration in lung/chest applications, driven by large-scale screening programs and COVID-19 research. Whole-body datasets represent an emerging trend for foundation model development, while traditional abdominal and bone imaging remain important clinical applications.

\emph{1) Lung/Chest} (96 datasets, 279,285 volumes). This dominant category reflects CT's primary clinical role in thoracic imaging. Major applications include COVID-19 analysis (STOIC2021 \dataref{data:stoic2021} with 10,735 volumes, COV19-CT-DB \dataref{data:cov19ctdb} with 7,750 volumes), lung cancer screening (NLST \dataref{data:nationallungscreenin} with 26,254 volumes), chest abnormalities detection (CT-RATE \dataref{data:ctrate} with 50,188 volumes), and nodule detection (LUNA16 \dataref{data:luna16} with 888 volumes, LIDC-IDRI \dataref{data:lidcidri} with 1,018 volumes). The category benefits from extensive public health initiatives and automated screening demands.

\emph{2) Whole-body} (7 datasets, 123,557 volumes). An emerging category driven by foundation model development needs. Key datasets include M3D \dataref{data:m3d} (120,000 volumes), TotalSegmentator \dataref{data:totalsegmentatordata} (1,204 volumes), and AutoPET series \dataref{data:autopet-ctpet} (2,233 volumes combined). These comprehensive collections enable multi-organ segmentation and cross-anatomical learning.

\emph{3) Abdomen} (55 datasets, 46,305 volumes). Traditional CT application focusing on multi-organ segmentation and tumor analysis. Notable collections include AbdomenAtlas \dataref{data:abdomenatlas} (20,460 volumes), FLARE series \dataref{data:flare21} (7,311 volumes combined), AbdomenCT-1K \dataref{data:abdomenct1k} (1,062 volumes), and specialized organ datasets like KiTS series \dataref{data:kits19} for kidney analysis (1,329 volumes combined). These datasets support both organ-specific and comprehensive abdominal analysis.

\emph{4) Bone/Spine} (10 datasets, 41,641 volumes). Specialized orthopedic applications including CTPelvic1K \dataref{data:ctpelvic1k} (1,184 volumes), CTSpine1K \dataref{data:ctspine1k} (1,005 volumes), VerSe series \dataref{data:verse19} (460 volumes combined), and RibFrac2020 \dataref{data:ribfrac2020} (660 volumes). Tasks focus on bone segmentation, fracture detection, and spinal analysis.

\emph{5) Head and Neck} (21 datasets, 8,969 volumes). Applications in radiation therapy planning and head/neck cancer treatment. Key datasets include HECKTOR series \dataref{data:hecktor2020} (1,462 volumes combined), SegRap2023 \dataref{data:segrap2023} (400 volumes), and various structural segmentation challenges.

\emph{6) Brain} (19 datasets, 4,887 volumes). CT brain imaging focuses on emergency applications including stroke detection (ISLES 2024 \dataref{data:isles2024-XXX} with 250 volumes), hemorrhage analysis (InSTANCE2022 \dataref{data:instance2022} with 200 volumes), and trauma assessment. Most brain imaging utilizes MRI, with CT serving specialized acute care roles.

\paragraph{CT Datasets by Tasks.}
CT datasets demonstrate strong task diversity, with segmentation dominating due to CT's excellent structural contrast. Classification applications leverage large-scale screening datasets, while specialized tasks like reconstruction and registration support advanced imaging workflows.

\emph{1) Segmentation} (150 datasets, 266,862 volumes). Segmentation represents the dominant task category, reflecting CT's strength in structural imaging. Applications include multi-organ segmentation (TotalSegmentator \dataref{data:totalsegmentatordata} with 1,204 volumes, AbdomenAtlas \dataref{data:abdomenatlas} with 20,460 volumes, M3D \dataref{data:m3d} with 120,000 volumes), organ-specific segmentation (KiTS series \dataref{data:kits19} with 1,329 volumes for kidneys, LiTS \dataref{data:lits} with 201 volumes for liver), and specialized targets like airway segmentation (AIIB23 \dataref{data:aiib23} with 312 volumes) and fracture detection (RibFrac2020 \dataref{data:ribfrac2020} with 660 volumes).

\emph{2) Classification} (93 datasets, 206,483 volumes). Classification tasks focus on disease screening and diagnostic applications. Major datasets include chest abnormalities detection (CT-RATE \dataref{data:ctrate} with 50,188 volumes), COVID-19 severity assessment (STOIC2021 \dataref{data:stoic2021} with 10,735 volumes), lung cancer screening (NLST \dataref{data:nationallungscreenin} with 26,254 volumes), and various cancer staging applications across TCGA collections. These datasets enable automated diagnosis and population-level screening.

\emph{3) Reconstruction} (5 datasets, 130,668 volumes). Emerging task category driven by dose reduction and image enhancement needs. Key datasets include M3D \dataref{data:m3d} (120,000 volumes) for multi-modal reconstruction, LDCT-and-Projection-data \dataref{data:ldctandprojectiondat} (299 volumes) for low-dose reconstruction, and specialized synthesis applications.

\emph{4) Localization} (6 datasets, 124,107 volumes). Localization tasks primarily support workflow automation and anatomical reference. The M3D dataset \dataref{data:m3d} (120,000 volumes) provides comprehensive localization annotations across multiple organs and structures.

\emph{5) Registration} (20 datasets, 123,382 volumes). Registration applications focus on longitudinal analysis and multi-modal fusion. Key datasets include Learn2Reg series \dataref{data:learn2reg-lungct} for lung CT (450 volumes) and abdomen CT-CT/MR-CT registration \dataref{data:learn2reg-abdomenctct} (172 volumes), supporting motion correction and atlas construction.

\emph{6) Detection} (20 datasets, 52,542 volumes). Detection tasks target specific anatomical structures and pathological findings. Notable applications include pulmonary nodule detection (LUNA16 \dataref{data:luna16} with 888 volumes, LIDC-IDRI \dataref{data:lidcidri} with 1,018 volumes), pulmonary embolism detection (RSNA STR \dataref{data:rsnastrpulmonaryembo} with 12,195 volumes), and lesion detection across various organs.

\subsection{MRI Volumes}

Magnetic Resonance Imaging (MRI) provides rich soft-tissue contrast and diverse sequence types for volumetric analysis. We identified 231 3D MRI datasets comprising approximately 523,847 volumes in total, as summarized in Table~\ref{tab:3d_mr_datasets}. These datasets span a wide range of sequences (T1, T2, FLAIR, DWI) and specialized protocols, varying greatly in scale and focus, from small studies such as MRBrainS13 \dataref{data:mrbrains13} for brain tissue segmentation\cite{mendrik2015mrbrains} to large-scale resources like OpenMind \dataref{data:openmind}\cite{wald2024an}. Dataset quality is shaped by sequence heterogeneity and scanner variability: standardized collections such as BraTS \cite{menze2015the} enforce uniform preprocessing across four canonical sequences, whereas multi-site datasets like OASIS-3 \dataref{data:oasis3}\cite{lamontagne2019oasis} include heterogeneous acquisition protocols and magnetic field strengths. Annotation consistency remains challenging; peritumoral or edema-related boundaries are known to be more ambiguous than enhancing or core regions in brain tumor tasks, contributing to inter-observer variability. In terms of clinical representativeness, MRI datasets range from healthy young adult cohorts in HCP \dataref{data:humanconnectomeproje}\cite{vanessen2013the} to elderly dementia populations in ADNI \dataref{data:adni}\cite{mueller2005ways}, while multi-vendor datasets such as M\&Ms \dataref{data:mms} (Siemens, Philips, GE, Canon) capture broader scanner and protocol diversity and highlight persistent cross-vendor generalization gaps.

\paragraph{MRI Datasets by anatomical structures.}
MRI is predominantly used in neuroimaging, with brain datasets dominating the 3D MRI landscape. Cardiac and abdominal applications show more limited representation, though they provide valuable specialized resources.

\emph{1) Brain/Neuro} (155 datasets, 356,751 volumes). The brain represents the most studied anatomy in 3D MRI, featuring major collections including BraTS series for tumor segmentation (BraTS 2023 \dataref{data:brats2023} with 5,880 volumes, UPENN-GBM \dataref{data:upenngbm} with 3,680 volumes), Alzheimer's research datasets (OASIS-3 \dataref{data:oasis3} with 5,699 volumes, ADNI \dataref{data:adni} with 2,500 volumes, TADPOLE \dataref{data:tadpole} with 1,667 volumes), stroke studies (ISLES 2022 \dataref{data:isles2022} with 400 volumes), and multiple sclerosis research (MSSEG-2 \dataref{data:msseg2} with 100 volumes). Brain datasets dominate the 3D MRI landscape in both dataset count and total volumes.

\emph{2) Head and Neck} (3 datasets, 114,643 volumes). Dominated by the OpenMind collection \dataref{data:openmind} (114,570 volumes), which represents a breakthrough in large-scale MR data collection. Other datasets include specialized head and neck cancer applications (AAPM-RT-MAC \dataref{data:aapmrtmac} with 55 volumes).

\emph{3) Prostate} (15 datasets, 3,704 volumes). Prostate MRI represents a well-established clinical application, with notable collections including PI-CAI \dataref{data:picai} (1,500 volumes), Prostate-MR-US-Biopsy \dataref{data:prostatemriusbiopsy-mr} (1,151 volumes for fusion imaging), PROSTATEx \dataref{data:prostatex} (204 volumes for classification), Prostate-MR-Segmentation \dataref{data:prostatemrisegmentat} (116 volumes), and PROMISE12 \dataref{data:promise12} (50 volumes for segmentation). These datasets support cancer diagnosis, treatment planning, and MR-US fusion workflows.

\emph{4) Breast} (11 datasets, 3,262 volumes). Breast MRI applications include Duke-Breast-Cancer-MR \dataref{data:dukebreastcancermri} (922 volumes), I-SPY1 \dataref{data:ispy1acrin6657} (847 volumes), I-SPY2 \dataref{data:ispy2trial} (719 volumes), ACRIN-Contralateral-Breast-MR \dataref{data:acrincontralateralbr} (984 volumes), and specialized collections. These datasets support cancer diagnosis, treatment response assessment, and radiomics research.

\emph{5) Cardiac} (13 datasets, 2,991 volumes). Cardiac MRI datasets focus on ventricular/myocardial segmentation and functional quantification. Key collections include M\&Ms \dataref{data:mms} (375 volumes), M\&Ms-2 \dataref{data:mms2} (360 volumes), LAScarQS++ 2024 \dataref{data:lascarqs2024} (200+ volumes), MyoPS++ 2024 \dataref{data:myops2024} (200+ volumes), ACDC \dataref{data:acdc} (150 volumes), and EMIDEC \dataref{data:emidec} (150 volumes). These datasets support automated cardiac analysis and multi-center validation studies.

\emph{6) Knee} (2 datasets, 1,823 volumes). Include MRNet \dataref{data:mrnet} (1,370 volumes for knee abnormalities detection) and SKI10 \dataref{data:ski10} (150 volumes for cartilage segmentation), supporting orthopedic applications and sports medicine research.

\emph{7) Others} (17 datasets, 1,539 volumes). Include liver applications (LLD-MMR2023 \dataref{data:lldmmri2023} with 498 volumes), spine imaging, gastrointestinal tract studies, and various specialized anatomical regions.

\emph{8) Whole-body} (2 datasets, 1,016 volumes). Include TotalSegmentator MRI \dataref{data:totalsegmentatormri} (616 volumes) and UW-Madison GI Tract \dataref{data:uwmadisongitractimag} (467 volumes), providing comprehensive anatomical coverage for foundation model development.

\paragraph{MRI Datasets by Tasks.}
3D MRI datasets are predominantly designed for segmentation and classification tasks, reflecting MRI's strength in soft-tissue contrast and anatomical delineation. The task distribution aligns with MRI's clinical applications in detailed tissue analysis and disease characterization.

\emph{1) Classification} (80 datasets, 322,508 volumes). Classification represents the largest category by total volumes, dominated by the OpenMind collection \dataref{data:openmind} (114,570 volumes) and large-scale neuroimaging studies. Major applications include Alzheimer's disease classification (OASIS-3 \dataref{data:oasis3} with 5,699 volumes, ADNI \dataref{data:adni} with 2,500 volumes, TADPOLE \dataref{data:tadpole} with 1,667 volumes), population studies (Human Connectome Project \dataref{data:humanconnectomeproje} with 1,206 volumes, Brain Genomics Superstruct Project \dataref{data:braingenomicssuperst} with 1,570 volumes), and cancer staging (PROSTATEx \dataref{data:prostatex} with 204 volumes for prostate cancer). These datasets enable automated diagnosis, disease staging, and population-level brain research.

\emph{2) Segmentation} (114 datasets, 151,433 volumes). Segmentation represents the largest category by dataset count, leveraging MRI's excellent soft-tissue contrast. Major applications include brain tumor delineation (BraTS 2023 \dataref{data:brats2023} with 5,880 volumes, UPENN-GBM \dataref{data:upenngbm} with 3,680 volumes, MSD01\_BrainTumor \dataref{data:task01braintumour} with 750 volumes), cardiac segmentation (M\&Ms \dataref{data:mms} with 375 volumes, ACDC \dataref{data:acdc} with 150 volumes), prostate segmentation (PI-CAI \dataref{data:picai} with 1,500 volumes, Prostate-MR-US-Biopsy \dataref{data:prostatemriusbiopsy-mr} with 1,151 volumes), and whole-body segmentation (TotalSegmentator MRI \dataref{data:totalsegmentatormri} with 616 volumes). The diversity in anatomical targets reflects MRI's versatility in tissue delineation.

\emph{3) Reconstruction} (15 datasets, 127,464 volumes). MR reconstruction focuses on acceleration techniques and image enhancement. Key datasets include fastMR \dataref{data:fastmri} (1,594 volumes), CMRxRecon \dataref{data:cmrxrecon} (300 volumes for cardiac reconstruction), and OpenMind \dataref{data:openmind} which also supports reconstruction tasks. This category addresses critical clinical needs for faster MR acquisition and improved image quality.

\emph{4) Registration} (31 datasets, 17,808 volumes). Registration applications include multi-timepoint studies, atlas construction, and multi-modal fusion. Notable datasets include Learn2Reg series \dataref{data:learn2reg-oasis} (OASIS, Hippocampus, LUMIR), CuRIOUS series \dataref{data:curious2018mrflair-mr} for MR-US registration, and various longitudinal studies for disease progression monitoring. These datasets enable temporal analysis and cross-modal alignment.

\emph{5) Tracking} (5 datasets, 1,855 volumes). Motion tracking applications primarily in cardiac MRI, including STACOM 2011 \dataref{data:motiontrackingchalle} (1,158 volumes) for cardiac motion analysis and various diffusion tractography studies. These datasets support dynamic analysis and fiber tracking applications.

\emph{6) Detection} (4 datasets, 1,245 volumes). Detection tasks focus on automated identification of anatomical landmarks and pathological structures, including aneurysm detection (ADAM2020 \dataref{data:adam2020} with 255 volumes) and various brain pathology identification tasks.

\subsection{Ultrasound Volumes}

3D ultrasound provides real-time volumetric imaging widely used for interventional guidance and multi-modal fusion. We identify 27 ultrasound-related 3D datasets containing approximately 56,609 volumes, as summarized in Table~\ref{tab:3d_us_datasets}. Most of these datasets appear within multi-modal collections (e.g., US/MR or US/CT), reflecting ultrasound’s predominant role in image-guided and fusion-based clinical workflows rather than as a standalone modality. Data quality is strongly operator-dependent, with clinical acquisitions showing higher variability compared to controlled research settings (e.g., the CuRIOUS series \dataref{data:curious2018mrflair}\cite{xiao2019evaluation}). In terms of representativeness, existing 3D ultrasound datasets are primarily derived from high-end interventional systems, underrepresenting handheld or point-of-care imaging scenarios common in real-world clinical practice.

\paragraph{Ultrasound Datasets by anatomical structures.}
The available 3D ultrasound datasets span diverse anatomical regions, with multi-modal combinations being particularly common for registration and fusion applications.

\emph{1) Brain} (9 datasets, $\sim$500 volumes). Brain ultrasound datasets focus primarily on US-MR registration for neurosurgical guidance. The CuRIOUS series \dataref{data:curious2018mrflair} (2018, 2019, 2022) provides datasets for brain tumor applications, while Learn2Reg LUMIR \dataref{data:learn2reg-lumir} (269 volumes) supports multi-modal registration research. These datasets enable US-guided brain interventions and intraoperative navigation.

\emph{2) Cardiac} (3 datasets, $\sim$1,400 volumes). Cardiac ultrasound datasets include STACOM 2011 \dataref{data:motiontrackingchalle} (1,158 volumes for motion tracking), CETUS2014 \dataref{data:cetus2014} (45 volumes), and MVSeg-3DTEE2023 \dataref{data:mvseg3dtee2023} (175 volumes for mitral valve segmentation). These datasets support automated echocardiography, cardiac function quantification, and structural heart analysis.

\emph{3) Prostate} (2 datasets, $\sim$1,300 volumes). Prostate datasets focus on US-MR fusion for biopsy guidance and treatment planning. Prostate-MR-US-Biopsy \dataref{data:prostatemriusbiopsy} (1,151 volumes) and $\mu$-RegPro2023 \dataref{data:regpro2023} (108 volumes) support fusion imaging applications critical for prostate cancer diagnosis and intervention.

\emph{4) Kidney} (4 datasets, $\sim$1,400 volumes). Pediatric kidney datasets from the AREN series \dataref{data:aren0532} (AREN0532, AREN0533, AREN0534) provide multi-modal collections including ultrasound for Wilms tumor research, supporting both classification and segmentation tasks in pediatric oncology.

\emph{5) Breast} (1 dataset, 200 volumes). TDSC-ABUS2023 \dataref{data:tdscabus2023} provides automated breast ultrasound data for breast cancer detection, supporting segmentation, classification, and detection tasks in breast imaging screening workflows.

\emph{6) Other Abdominal Organs} (8 datasets, $\sim$1,100 volumes). Include pancreas (CPTAC-PDA \dataref{data:cptacpda}), liver (AHEP0731 \dataref{data:ahep0731}), uterus (CPTAC-UCEC \dataref{data:cptacucec}), and other organs from multi-modal cancer imaging collections, primarily supporting classification tasks for oncological applications.

\paragraph{Ultrasound Datasets by Tasks.}
3D ultrasound datasets are dominated by registration applications, reflecting the modality's role in multi-modal image fusion and guidance systems.

\emph{1) Registration} (15 datasets, $\sim$2,000 volumes). Registration represents the dominant task category, reflecting ultrasound's critical role in real-time guidance and multi-modal fusion. Major applications include US-MR brain registration (CuRIOUS series \dataref{data:curious2018mrflair}), prostate fusion imaging (Prostate-MR-US-Biopsy \dataref{data:prostatemriusbiopsy}, $\mu$-RegPro2023 \dataref{data:regpro2023}), cardiac motion tracking (STACOM 2011 \dataref{data:motiontrackingchalle}), and multi-modal brain registration (Learn2Reg LUMIR \dataref{data:learn2reg-lumir}). This dominance reflects ultrasound's primary clinical value in providing real-time guidance for interventions and fusion with other imaging modalities.

\emph{2) Classification} (10 datasets, $\sim$1,800 volumes). Classification tasks focus primarily on cancer staging and diagnosis across multiple organs, including kidney tumors (AREN series \dataref{data:aren0532}), pancreatic cancer (CPTAC-PDA \dataref{data:cptacpda}), liver cancer (AHEP0731 \dataref{data:ahep0731}), and other malignancies. These applications leverage ultrasound's accessibility for screening and staging workflows.

\emph{3) Segmentation} (8 datasets, $\sim$800 volumes). Segmentation applications target organ and structure delineation for cardiac analysis (CETUS2014 \dataref{data:cetus2014}, MVSeg-3DTEE2023 \dataref{data:mvseg3dtee2023}), tumor segmentation (AREN0533-Tumor-Annotations \dataref{data:aren0533}, AREN0534 \dataref{data:aren0534}), and breast lesion detection (TDSC-ABUS2023 \dataref{data:tdscabus2023}). These datasets support automated measurement and volumetric analysis critical for clinical assessment.

\emph{4) Detection} (1 dataset, 200 volumes). Detection tasks focus on automated lesion identification, exemplified by TDSC-ABUS2023 \dataref{data:tdscabus2023} for breast cancer screening, supporting computer-aided detection workflows in clinical practice.

\subsection{PET Volumes}

Positron Emission Tomography (PET) provides functional information complementary to anatomical imaging. Public PET volumes are scarce and often appear in multi-modality collections (\eg, PET/CT, PET/MR). We identify 65 PET-related 3D datasets with \textbf{95,456 volumes} in total, as presented in Table~\ref{tab:3d_pet_datasets}. These collections span diverse anatomic regions with a strong focus on oncology applications, particularly in lung/chest (15 datasets), head and neck (11 datasets), and brain (8 datasets) regions. This significant expansion largely comes from comprehensive cancer imaging archives, multi-center studies, and large-scale neuroimaging initiatives.

\paragraph{PET Datasets by anatomical structures.}
These datasets use PET primarily for oncology applications across various anatomical regions, though multi-modal combinations are the norm rather than the exception. The distribution shows clear preferences for certain anatomical regions where PET imaging provides the most clinical value.

\emph{1) Lung/Chest} (15 datasets, $\sim$55,000+ volumes). This represents the largest category by dataset count, reflecting PET's critical role in lung cancer diagnosis and staging. Key collections include QIDW \dataref{data:qidw} (52,000 volumes for quality assurance), Lung-PET-CT-Dx \dataref{data:lungpetctdx} (355 volumes), CPTAC-LUAD \dataref{data:cptacluad} (244 volumes), ACRIN-NSCLC-FDG-PET \dataref{data:acrinnsclcfdgpetacri-pet} (242 volumes), CPTAC-LSCC \dataref{data:cptaclscc} (212 volumes), and NSCLC-Radiogenomics \dataref{data:nsclcradiogenomics} (211 volumes). The dominance of lung-related datasets demonstrates PET's established clinical utility in pulmonary oncology.

\emph{2) Head and Neck} (11 datasets, $\sim$4,200 volumes). Head and neck cancers represent a major application area for PET imaging, with notable collections including HECKTOR 2022 \dataref{data:hecktor2022} (883 volumes), HNSCC \dataref{data:hnscc} (627 volumes), TCGA-HNSC \dataref{data:tcgahnsc} (479 volumes), HECKTOR 2021 \dataref{data:hecktor2021} (325 volumes), Head-Neck-PET-CT \dataref{data:headneckpetct} (298 volumes), QIN-HEADNECK \dataref{data:qinheadneck} (279 volumes), and ACRIN-HNSCC-FDG-PET-CT \dataref{data:acrinhnsccfdgpetctac-pet} (260 volumes). These datasets support both tumor segmentation and treatment response assessment.

\emph{3) Brain} (8 datasets, $\sim$13,300 volumes). Brain PET datasets focus primarily on neurodegenerative diseases and provide the largest individual dataset volumes. Major collections include OASIS-3 \dataref{data:oasis3} (5,699 volumes), ADNI \dataref{data:adni} (2,500 volumes), TADPOLE \dataref{data:tadpole} (1,667 volumes), and PPMI \dataref{data:parkinsonsprogressio-pet} (683 volumes) for Alzheimer's and Parkinson's disease research, alongside smaller oncology-focused datasets like ACRIN-FMISO-Brain \dataref{data:acrinfmisobrainacrin-pet} (45 volumes).

\emph{4) Abdominal Organs} (7 datasets, $\sim$1,400 volumes). Include specialized datasets for liver, pancreas, and kidney imaging. Notable collections include AREN0532 \dataref{data:aren0532} (544 volumes) and AREN0534 \dataref{data:aren0534} (239 volumes) for pediatric kidney tumors, AHEP0731 \dataref{data:ahep0731} (190 volumes) for liver cancer, and CPTAC-PDA \dataref{data:cptacpda} (168 volumes) for pancreatic cancer.

\emph{5) Whole-body/Multi-organ} (3 datasets, $\sim$2,300 volumes). Comprehensive whole-body PET datasets include AutoPET II \dataref{data:autopetii-pet} (1,219 volumes), AutoPET \dataref{data:autopet-ctpet-pet} (1,014 volumes), and fastPET-LD \dataref{data:fastpetld-pet} (68 volumes), providing valuable resources for pan-cancer detection and segmentation tasks.

\emph{6) Breast} (3 datasets, $\sim$240 volumes). Specialized breast cancer datasets include BREAST-DIAGNOSIS \dataref{data:breastdiagnosis} (88 volumes), ACRIN-FLT-Breast \dataref{data:acrinfltbreastacrin6-pet} (83 volumes), and QIN-Breast \dataref{data:qinbreast} (68 volumes), supporting breast cancer diagnosis and treatment monitoring.

\paragraph{PET Datasets by Tasks.}
PET datasets reflect the modality's primary clinical applications in oncology and neurology, with task distribution strongly aligned with PET's role in functional and metabolic imaging for disease diagnosis, staging, and treatment monitoring.

\emph{1) Classification} (45 datasets, $\sim$60,000+ volumes). Classification represents the dominant task category, reflecting PET's core clinical utility in disease staging, treatment response assessment, and diagnostic classification. Oncology applications span multiple cancer types, including lung cancer datasets (CPTAC-LUAD \dataref{data:cptacluad} with 244 volumes, ACRIN-NSCLC-FDG-PET \dataref{data:acrinnsclcfdgpetacri-pet} with 242 volumes, TCGA-LUSC \dataref{data:tcgalusc-pet} with 37 volumes), head and neck cancer studies (TCGA-HNSC \dataref{data:tcgahnsc} with 479 volumes, ACRIN-HNSCC-FDG-PET-CT \dataref{data:acrinhnsccfdgpetctac-pet} with 260 volumes), and various other malignancies across different anatomical sites. Neurological applications focus on neurodegenerative diseases, particularly Alzheimer's disease classification (OASIS-3 \dataref{data:oasis3} with 5,699 volumes, ADNI \dataref{data:adni} with 2,500 volumes, TADPOLE \dataref{data:tadpole} with 1,667 volumes) and Parkinson's disease research (PPMI \dataref{data:parkinsonsprogressio-pet} with 683 volumes). The dominance of classification tasks aligns with PET's clinical role in providing metabolic information for staging and prognosis.

\emph{2) Segmentation} (20 datasets, $\sim$25,000 volumes). Segmentation tasks focus primarily on tumor delineation and organ-at-risk identification for radiation therapy planning. Major collections include AutoPET II \dataref{data:autopetii-pet} (1,219 volumes), AutoPET \dataref{data:autopet-ctpet-pet} (1,014 volumes), HECKTOR 2022 \dataref{data:hecktor2022} (883 volumes), and HNSCC \dataref{data:hnscc} (627 volumes). These datasets support automated tumor volume definition, which is critical for radiotherapy planning and treatment monitoring. The emphasis on head and neck, lung, and whole-body segmentation reflects PET's established role in oncology workflow integration.

\emph{3) Multi-task datasets} (8 datasets, $\sim$5,000 volumes). Several datasets provide annotations for multiple tasks, enabling comprehensive analysis approaches. Examples include Head-Neck-PET-CT \dataref{data:headneckpetct} (298 volumes for both segmentation and classification), NSCLC-Radiogenomics \dataref{data:nsclcradiogenomics} (211 volumes for segmentation and classification), and ACRIN-FMISO-Brain \dataref{data:acrinfmisobrainacrin-pet} (45 volumes for segmentation and classification). This multi-task approach reflects the clinical reality where PET images are used for multiple diagnostic and therapeutic purposes simultaneously.

\emph{4) Detection} (3 datasets, $\sim$400 volumes). Detection tasks focus on lesion identification and localization, exemplified by Lung-PET-CT-Dx \dataref{data:lungpetctdx} (355 volumes for classification and detection) and fastPET-LD \dataref{data:fastpetld-pet} (68 volumes for detection). While less common than classification, detection tasks are important for automated screening and lesion characterization in clinical workflows.

\emph{5) Registration} (3 datasets, $\sim$1,200 volumes). Registration applications appear primarily in the HECKTOR series \dataref{data:hecktor2021} (2021 and 2022), supporting multi-timepoint analysis for treatment response assessment. This reflects PET's growing role in longitudinal monitoring of therapy effects and disease progression.

\subsection{Other 3D Volumes}

Beyond the major modalities, we collect 26 3D datasets from specialized imaging techniques with \textbf{5,381+ volumes} in total, as presented in Table~\ref{tab:3d_other_datasets}. These modalities serve specific clinical niches and emerging applications, with OCT dominating the collection due to large-scale ophthalmology datasets. The diversity reflects the evolution of medical imaging technology and specialized clinical needs.

\paragraph{Other Modalities According to Imaging Technology.}
Each modality addresses specific clinical applications and anatomical targets, with OCT leading in volume due to comprehensive retinal imaging datasets.

\emph{1) Optical Coherence Tomography (OCT)} (14 datasets, 4,288+ volumes). OCT dominates this category, primarily targeting retinal and ophthalmologic applications. The OLIVES dataset \dataref{data:olives} alone contributes 1,268 volumes for diabetic condition analysis, while the newly added OCTA-500 dataset \dataref{data:octa500} provides 500 volumes for comprehensive retinal OCTA analysis. Specialized collections include GAMMA \dataref{data:gamma} (300 volumes for glaucoma analysis), RETOUCH \dataref{data:retouch} (112 volumes for retinal disease segmentation), and various Duke University datasets for age-related macular degeneration and diabetic macular edema. The OCTA2024 dataset \dataref{data:octa2024} supports advanced OCT to OCTA translation research. Tasks primarily focus on classification, segmentation, and reconstruction of retinal pathologies, supporting automated screening for eye diseases.

\emph{2) Digital Subtraction Angiography (3D DSA)} (4 datasets, 454 volumes). DSA applications focus on cerebrovascular imaging, particularly aneurysm detection and analysis. Key datasets include CADA series \dataref{data:cada} for cerebral aneurysm detection (372 volumes combined) and SHINY-ICARUS \dataref{data:isbi2023challengeshi} for internal carotid artery aneurysm segmentation (82 volumes). These datasets support critical neurovascular intervention planning and risk assessment.

\emph{3) Cone-beam CT (CBCT)} (4 datasets, 581 volumes). CBCT serves specialized applications in dental imaging and treatment planning. Notable collections include ToothFairy2023 \dataref{data:toothfairy2023} for dental surgery planning (443 volumes), pancreatic CT-CBCT registration datasets (40 volumes), and pelvic reference data for prostate cancer treatment (58 volumes). These datasets bridge diagnostic and interventional imaging workflows.

\emph{4) 3D Microscopy} (3 datasets, 54 volumes). Microscopy datasets target cellular and subcellular analysis, including MitoEM \dataref{data:mitoem} for mitochondrial ultrastructure (2 volumes), platelet ultrastructure analysis (2 volumes), and prostate cancer pathology (50 volumes). Though small in volume, these datasets enable high-resolution structural analysis at the cellular level.

\paragraph{Other Modalities According to Tasks.}
Task distribution reflects the specialized nature of these modalities, with classification dominating due to large-scale OCT screening applications.

\emph{1) Classification} (11 datasets, 3,914 volumes). Classification tasks predominantly target disease screening and diagnosis, especially in ophthalmology. Major applications include diabetic condition screening (OLIVES \dataref{data:olives} with 1,268 volumes), glaucoma detection (OCT Glaucoma Detection \dataref{data:octglaucomadetection} with 1,110 volumes), and various retinal disease classification tasks. These datasets enable automated screening systems for population health initiatives.

\emph{2) Segmentation} (17 datasets, 1,987 volumes). Segmentation applications span multiple modalities and anatomical targets, from retinal layer segmentation in OCT to aneurysm delineation in DSA and dental structure segmentation in CBCT. The diversity of targets reflects the specialized nature of each modality's clinical applications.

\emph{3) Registration} (3 datasets, 498 volumes). Registration tasks primarily support treatment planning and longitudinal analysis, including CBCT-CT registration for radiation therapy and structural-functional alignment in ophthalmology.

\emph{4) Reconstruction/Translation} (1 dataset, TBD volumes). Advanced reconstruction and translation tasks include OCT to OCTA image translation, enabling cross-modal analysis and synthetic data generation for retinal imaging applications.

These specialized modalities complement major imaging modalities by addressing specific clinical needs and emerging applications, contributing to the comprehensive landscape of 3D medical imaging datasets.

\subsection{Challenges and Opportunities}

The 3D medical imaging landscape presents unique challenges and opportunities that distinguish it from 2D medical imaging. Despite providing richer spatial information essential for volumetric analysis and clinical decision-making, 3D datasets remain significantly constrained by fundamental limitations in data acquisition, annotation complexity, and resource allocation.

\paragraph{Key Challenges in 3D Medical Imaging Datasets.}
The primary challenges stem from the inherent complexity and cost of 3D data acquisition and processing. \emph{High acquisition and annotation costs} represent the most significant barrier, as 3D imaging requires expensive specialized equipment (CT, MRI, and PET scanners) and expert radiologists for volumetric annotation, resulting in the modest growth observed compared to 2D datasets. This economic constraint directly impacts data availability and diversity.

\emph{Dataset overlap and duplication} presents another critical challenge that researchers must be aware of when conducting external validation studies. Some datasets in our collection contain overlapping or identical data under different names, particularly when larger datasets consolidate multiple smaller collections. For instance, the OCT2017 dataset and MedMNIST OCT dataset contain identical retinal OCT images, as MedMNIST integrates multiple publicly available datasets including OCT2017. Similar overlaps exist across other modalities where comprehensive datasets merge smaller specialized collections. Researchers should exercise caution when selecting datasets for external validation to avoid inadvertently using overlapping data that could lead to overly optimistic performance estimates and compromised generalizability assessments.

\emph{Complexity and cost.}
On the data side, challenges are multifaceted. Acquisition costs remain prohibitively high due to the expense of imaging hardware, long scanning times, and patient compliance issues. Storage costs escalate rapidly as each volumetric scan can range from hundreds of megabytes to several gigabytes, requiring robust archiving infrastructure. Annotation costs are substantial because volumetric segmentation demands time-consuming, slice-by-slice delineation by expert radiologists.
On the model side, these data characteristics translate into significant computational challenges. The high dimensionality of 3D medical images substantially increases memory consumption and processing time during training and inference, often necessitating specialized hardware and optimization strategies. Moreover, the low signal-to-noise ratio of many volumetric acquisitions and the small size of pathological regions further complicate feature extraction and model generalization. Together, these factors underscore the intricate interplay between data and model complexity in 3D medical imaging research.

\emph{Modality and anatomical imbalances} create substantial gaps in representation. While CT (261 datasets, ~753,421 volumes) and MRI (231 datasets, ~523,847 volumes) dominate the landscape, critical modalities like ultrasound (27 datasets, 56,609 volumes), PET (65 datasets, 95,456 volumes), and emerging volumetric techniques remain underrepresented relative to their clinical importance. Anatomically, while the concentration on brain and abdomen/liver regions has expanded significantly, cardiac, musculoskeletal, and certain specialized applications still have relatively limited resources, though recent large-scale initiatives are beginning to address these gaps.

\emph{Task-specific limitations} further constrain the utility of existing 3D datasets. The overwhelming dominance of classification and segmentation tasks, while clinically important, reflects the field's incomplete transition from task-oriented to foundation-oriented data engineering paradigms. Registration and reconstruction tasks remain underrepresented despite their critical importance for longitudinal studies and treatment monitoring. Additionally, the scarcity of multi-task datasets limits the development of versatile clinical AI systems capable of handling complex, real-world diagnostic workflows.

\paragraph{Opportunities for Advancement.}
Despite these challenges, the 3D medical imaging domain presents remarkable opportunities for transformative advancement. The substantial collection of \emph{unlabeled 3D volumes} (219 datasets with hundreds of thousands of volumes) offers unprecedented potential for self-supervised learning and contrastive pretraining. Large repositories like TCIA for CT and HCP for MRI provide the scale necessary for foundation model pretraining, while multi-sequence MR data enables sophisticated cross-modal consistency training and modality dropout techniques.

\emph{Foundation model-driven data augmentation} emerges as a particularly promising direction. Well-trained generative foundation models can participate in semi-supervised learning frameworks, generating synthetic 3D volumes that reflect real clinical presentations while addressing privacy constraints. This approach is especially valuable for rare diseases and underrepresented anatomical regions where data acquisition remains challenging.

\emph{Multimodal integration} presents opportunities to leverage complementary information across imaging modalities. PET/CT and PET/MR combinations demonstrate the clinical value of multimodal approaches, while the emergence of vision-language datasets that combine 3D volumes with clinical reports and radiology texts opens new possibilities for cross-modal reasoning and clinical context understanding. Beyond traditional imaging modality combinations, innovative cross-domain multimodal approaches are emerging, such as integrating macroscopic imaging (CT/MRI) with microscopic pathology data. These pathology-imaging combinations offer unique opportunities to bridge the gap between radiological findings and histological ground truth, enabling AI systems to learn from both macroscopic anatomical structures and microscopic tissue characteristics. Such approaches can significantly enhance diagnostic accuracy by combining CT's ability to detect and localize lesions with pathology's role as the diagnostic gold standard, creating more robust and clinically-relevant AI systems. Advances in cross-modal alignment techniques enable more sophisticated fusion strategies that can favorably enhance diagnostic capabilities across these diverse data types.

\emph{Multi-task learning paradigms} offer transformative potential for 3D medical imaging, analogous to the "one-for-all" paradigm exemplified by ChatGPT in natural language processing. Rather than training separate models for individual tasks, integrated frameworks can simultaneously address multiple tasks (\eg, segmentation, classification, and detection) within unified architectures. This approach not only improves computational efficiency but also enables knowledge transfer across related tasks, particularly valuable given the limited scale of individual 3D datasets. Multi-task datasets that provide diverse annotation types for the same volumetric data can unlock synergistic learning effects, where performance on individual tasks benefits from joint optimization across multiple objectives. The vision of a unified diagnostic and generative model that can handle multiple clinical tasks simultaneously represents a paradigm shift toward more versatile and efficient clinical AI systems, similar to how foundation models have revolutionized natural language understanding and generation.

Looking forward, the transition toward foundation-oriented data engineering paradigms demands fundamental changes in how 3D medical datasets are conceptualized and structured. Future dataset designs should prioritize adaptability and extensibility, enabling researchers to derive new tasks and applications from existing resources. Strategic dataset consolidation through systematic metadata harmonization, combined with advances in self-supervised learning and cross-modal reasoning, positions the 3D medical imaging domain for significant breakthroughs in clinical AI applications.

\section{Medical Video Datasets}
\label{sec:video_data}

Medical video datasets are crucial resources for developing algorithms that leverage spatiotemporal information in dynamic clinical scenarios, such as minimally invasive surgery, medical education, and video-based diagnosis. 
In contrast to static image datasets, video data facilitates the modeling of motion patterns, procedural workflows, and temporal consistencies, which are essential for tasks such as surgical instrument tracking, cross-frame anatomical structure segmentation, or physiological motion estimation. This survey identifies 77 medical video datasets, comprising a total of 166,691 samples. These datasets span a diverse range of tasks, imaging modalities, and anatomical structures. All video datasets are illustrated in the Tab. \ref{tab:video_alldata_v2}.

\subsection{Overview}
\begin{figure}
    \centering
    \includegraphics[trim=0 0 0 0, clip, width=\linewidth]{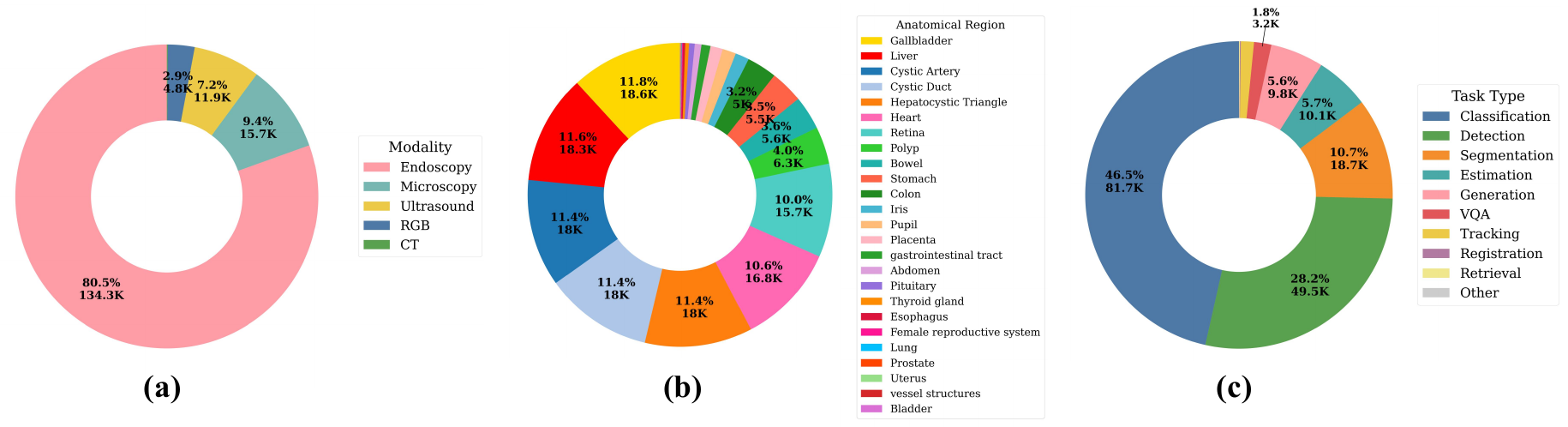}
    \caption{The distribution of different (a) modalities, (b) anatomical structures, and (c) tasks for video datasets}
  \label{fig:video_datasets_overview}
\end{figure}

~\cref{fig:video_datasets_overview} illustrates the distribution of video datasets across different anatomical structures, imaging modalities, and tasks. 
The most prevalent anatomical structures are the stomach, colon, and esophagus, with each category individually accounting for about 30\% of the total videos. 
In contrast, other anatomical structures, such as the retina, heart, pupil, and iris, are significantly underrepresented, each  constituting less than 2\% of the total collection. 
Similarly, the distribution across imaging modalities is highly skewed, with endoscopy alone accounting for a substantial 85.9\% of the videos. 
Consequently, modalities such as ultrasound microscopy, and RGB remain scarce, which highlights a critical need for larger-scale datasets to mitigate potential modality bias, particularly in the development of foundation models. 
In contrast to the severe imbalances observed across modalities, the task distribution is more moderate, though still demonstrably long-tailed.
Classification, detection, and segmentation represent the most common tasks, followed by estimation, generation, and VQA, whereas tracking, retrieval, and registration are notably underrepresented, warranting further investigation.
Given the severe modality imbalance, task-level information helps distinguish the properties of the collected video datasets. 
Therefore, the organizational structure of this section deviates from that of Sections~\ref{sec:2d_data} and~\ref{sec:3d_data}.
Specifically, the video datasets are introduced primarily based on their associated tasks rather than modality, with a supplementary analysis of the corresponding modalities and anatomical structures.

\subsection{Task}
\begin{figure}[!tp]
    \centering
    \includegraphics[width=\textwidth]{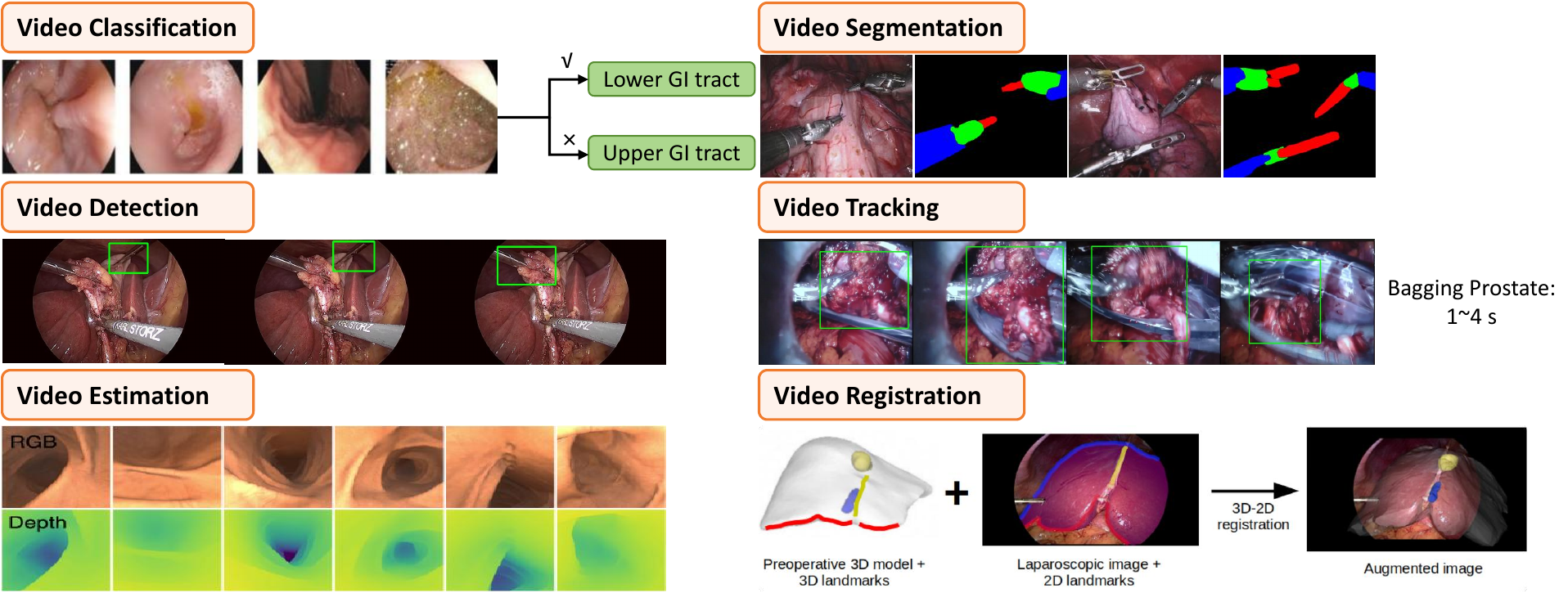}
    \caption{Demonstration of the collected video datasets from different tasks. The figure for the video estimation task is from the EndoVis 2022-SimCol-to-3D dataset \dataref{data:EndoVis 2022-SimCol-to-3D}, and the figure for the video registration task is from the EndoVis 2022-P2ILF dataset \dataref{data:EndoVis22-P2ILF}.}
    \label{fig:video_tasks}
\end{figure}

Below we introduce the major tasks in the collected video datasets. Figure \ref{fig:video_tasks} demonstrates common tasks in medical video datasets, including video classification, video segmentation, video detection, video tracking, video estimation, and video registration.

\subsubsection{Classification}
Classification in medical videos involves assigning categorical labels to entire sequences or specific temporal segments by leveraging spatial-temporal features. 
This task is fundamental to a wide range of clinical and surgical applications, such as surgical phase recognition, skill assessment, and disease diagnosis. 
To date, 40 datasets comprising 81,701 samples have been identified for this purpose, with endoscopy dominating as the primary modality. Performance is commonly evaluated using metrics such as accuracy, F1-score, and the Area Under the ROC Curve (AUC), while temporal metrics like the segmental edit score are also employed to assess sequence-level consistency.


Representative datasets span diverse surgical domains. 
Cholec80 \dataref{data:Cholec80} and its derivatives, including CholecT50 \dataref{data:CholecT50} and the CholecTriplet  challenges \dataref{data:CholecTriplet 2021},\dataref{data:CholecTriplet2022}, provide laparoscopic cholecystectomy videos annotated with surgical phases, instrument presence, and fine-grained <instrument–verb–target> triplets. These resources serve as benchmarks for workflow analysis and activity recognition. 
SurgVisDom \dataref{data:SurgVisDom} contains 488 bowel surgery videos with phase annotations, enabling cross-domain generalization studies. 
As the largest public dataset for gastrointestinal endoscopy, HyperKvasir \dataref{data:HyperKvasir} is the largest publicly available gastrointestinal endoscopy dataset, comprising 373 videos and over 110,000 video frames annotated for anatomy and pathology, supporting classification, localization, and captioning. 
In the field of ophthalmic surgery, the CATARACTS \dataref{data:CATARACTS} and Cataract-1K datasets \dataref{data:Cataract-1K} provide microscopy videos annotated for surgical phases, instruments, and pixel-level segmentation, facilitating multi-task modeling. 
Other influential resources include the EndoVis Workflow and Skill Assessment (SWSA) series \dataref{data:EndoVis 2019-SWSA} for phase and skill classification, and SAR-RARP50 \dataref{data:EndoVis22-SAR-RARP50}, the first public robot-assisted radical prostatectomy dataset with synchronized action and instrument annotations.

\subsubsection{Segmentation}
Medical video segmentation involves the frame-by-frame delineation of anatomical structures, pathological regions, or surgical instruments to enable precise spatio-temporal analysis. 
This task is crucial for applications including real-time surgical guidance, quantitative organ motion tracking, and automated assessment of lesion dynamics. 
Our review encompasses 32 datasets tailored for segmentation, containing a total of 18,739 video instances. Performance is typically evaluated using metrics such as the Dice similarity coefficient, Intersection-over-Union (IoU), and pixel-level accuracy. 

Representative datasets highlight both surgical and microscopic domains. The Robotic Instrument Segmentation (RIS) \dataref{data:Robotic Instrument Segmentation} and Kidney Boundary Detection (KBD) \dataref{data:KBD} datasets introduced pixel-level annotations for robotic surgical tools and anatomical boundaries, establishing early benchmarks for intraoperative vision. 
In ophthalmology, Cataract-1K \dataref{data:Cataract-1K} combines phase annotations with 2,256 manually segmented frames for cataract surgery, enabling joint analysis of workflow and fine-grained structures. 
The HyperKvasir dataset \dataref{data:HyperKvasir}, while primarily used for gastrointestinal classification, also includes segmentation masks for anatomical landmarks and pathological findings across 373 endoscopic videos.
More recent challenges extend segmentation to complex multi-modal and 3D contexts, For instance, P2ILF \dataref{data:EndoVis22-P2ILF}  combines laparoscopic video and CT for liver landmark delineation, while SAR-RARP50 \dataref{data:EndoVis22-SAR-RARP50} is the first public dataset of robot-assisted radical prostatectomy videos with synchronized instrument segmentation and action recognition.

\subsubsection{Detection}
Video detection aims to identify and localize target objects, such as lesions, instruments, or anatomical landmarks, within individual frames of a video while leveraging temporal continuity to improve robustness. 
This capability is crucial for early disease screening, intraoperative navigation, and automated procedural quality assessment. 
We identified 27 datasets for the detection task. Commonly used evaluation metrics include precision, recall, mean Average Precision (mAP), and frame-level F1-core.  
In our survey, these 28 datasets comprise  49,507 samples emphasize detection. 
Evaluations typically report precision, recall, mean Average Precision (mAP) at bounding-box or mask-level IoU thresholds, and frame-level F1. For temporally aggregated predictions, some studies additionally report video-mAP or track-aware scores to penalize fragmented detections.

Representative resources span lesion, artifact, and instrument detection across multiple surgical domains. GIANA \dataref{data:GIANA} and EndoCV \dataref{data:EndoCV 2021} provide endoscopic polyp detection benchmarks with bounding-box or mask annotations, stressing generalization across centers and devices.
Instrument-centric datasets include the m2cai16-tool-locations \dataref{data:m2cai16-tool-locations} set and the large-scale SurgToolLoc challenges (2022–2023) \dataref{data:EndoVis 2022-SurgToolLoc},\dataref{data:EndoVis23-SurgToolLoc} with tens of thousands of annotated frames for robotic and laparoscopic tools, enabling strong baselines for real-time instrument awareness and downstream workflow understanding. 
Beyond the abdomen, ophthalmic datasets such as CATARACTS or Cataract-1K and LensID \dataref{data:LensID} support tool and structure detection in cataract surgery, while PitVis \dataref{data:EndoVis23-PitVis} focuses on transsphenoidal neurosurgery with step- and instrument-level labels. Broader clinical coverage is offered by AVOS \dataref{data:AVOS}, a multi-procedure open-surgery corpus with dense annotations that enables cross-procedure detection, tracking, and localization. 
Recent multi-domain collections such as SARAS-MESAD \dataref{data:SARAS-MESAD} further test robustness by mixing real and phantom data under shared action or instrument vocabularies. Across these datasets, annotation granularity ranges from sparsely sampled frames to densely labeled clips, with boxes, instance masks, or keypoints. Emerging trends include spatiotemporal tube proposals, joint detection-tracking protocols, and robustness benchmarks under realistic corruptions, which together move detection from frame-wise recognition toward reliable, clinically usable video understanding.

\subsubsection{Tracking}
Tracking in medical videos entails following the spatiotemporal trajectories of objects of interest, such as surgical tools or anatomical landmarks, across consecutive frames. This task is fundamental to applications such as workflow analysis, motion quantification, and dynamic process monitoring. 
Our survey identified 8 datasets with 2,420 samples dedicated to tracking.
The tracking task usually employs metrics such as Multiple Object Tracking Accuracy (MOTA), Multiple Object Tracking Precision (MOTP), identity switches (IDSW), and track purity. 

Representative datasets focus on the surgical domain. 
For example, the m2cai16-tool-locations dataset \dataref{data:m2cai16-tool-locations} provided laparoscopic tool-tip trajectories, while the EndoVis tracking challenges expanded to encompass tracking, tissue motion estimation, and joint detection–tracking tasks. 
SurgT \dataref{data:SurgT: Surgical Track} and SARAS-MESAD \dataref{data:SARAS-MESAD} further incorporated stereoscopic views, soft-tissue tracking, and phantom–real domain variations. 
Beyond endoscopy, STIR \dataref{data:STIR} provided infrared–visible paired videos for surgical tissue tracking, and the large-scale dataset AVOS delivered dense per-frame annotations across 47 hours of open surgery from 23 procedure types. 
Specialized datasets such as HiSWA-RLLS \dataref{data:HiSWA-RLLS} for robotic liver resection and EgoSurgery \dataref{data:Egosurgery} with egocentric video plus gaze data highlight emerging subfields, reflecting a recent trend toward multi-task benchmarks that unify detection, segmentation, and temporal association for comprehensive spatiotemporal understanding.

\subsubsection{Estimation} 
Estimation tasks in medical video analysis aim to derive quantitative variables from temporal sequences, such as depth maps, motion fields, or physiological parameters. Applications include 3D reconstruction from monocular endoscopic videos, camera pose estimation for navigation, respiratory motion estimation, and surgical skill scoring. Our survey identified two datasets dedicated to this task. 
The SimCol-to-3D \dataref{data:EndoVis 2022-SimCol-to-3D} dataset contains simulated colonoscopy videos for depth prediction and camera pose estimation, with 15 sequences annotated for both simulated and real procedures, enabling evaluation under controlled and clinical conditions. 
The challenge also includes a Colposcopy subset with 30 videos for depth estimation in gynecological imaging. 
The Endovis 2019-SCRE \dataref{data:Endovis 2019-SCRE} dataset contains videos from 9 medical sites for the task of dense depth estimation. The corresponding depth maps were obtained from structured light data captured using porcine cadavers.
Evaluation metrics are task-specific, including mean absolute error (MAE), endpoint error (EPE), and correlation coefficients. Moreover, recent works have increasingly adopted multi-task formulations that jointly estimate depth, pose, and motion to improve downstream surgical navigation and workflow understanding.

\subsubsection{Registration} 
Registration in medical video analysis involves aligning multimodal data, such as 2D video endoscopy with 3D computed tomography (CT), to establish a consistent spatial correspondence across imaging modalities. This process is crucial for intraoperative guidance, anatomical structure mapping, and enhanced visualization of surgical fields. In our survey, one datasets with a total of 167 samples were identified for registration tasks. 

The P2ILF dataset \dataref{data:EndoVis22-P2ILF} provides paired endoscopy videos and CT scans, and is designed for evaluating multimodal registration methods. 
The dataset included 25 cases (10 for training, 10 for validation, and 5 for testing) with both 3D model and video-endoscopic data, supporting cross-modality alignment and benchmarking registration accuracy. 
The registration is performed between the landmarks of the 3D model and those extracted from the videos.


Evaluation metrics for registration commonly include Target Registration Error (TRE), the Dice Similarity Coefficient (DSC) for segmented structures, and success rates within clinically acceptable error thresholds. Together, the P2ILF dataset form a comprehensive benchmark for developing and validating multimodal registration approaches in minimally invasive liver surgery.



\subsection{Modalities}
Among the 77 medical video datasets identified in our survey, endoscopy constitutes the vast majority with 56 datasets, underscoring its central role in documenting dynamic intraoperative and diagnostic procedures. 
This prevalence is attributable to several factors: the routine integration of video recording systems in surgical suites, the ease of acquiring high-resolution footage during standard procedures, and the relatively straightforward annotation of visible anatomical structures or surgical tools without requiring complex multi-view reconstruction. Longstanding community initiatives, such as the EndoVis challenges, have further accelerated dataset generation and standardization, fostering a virtuous cycle between benchmark availability and method development.

\begin{figure}[!tp]
    \centering
    \includegraphics[width=\textwidth]{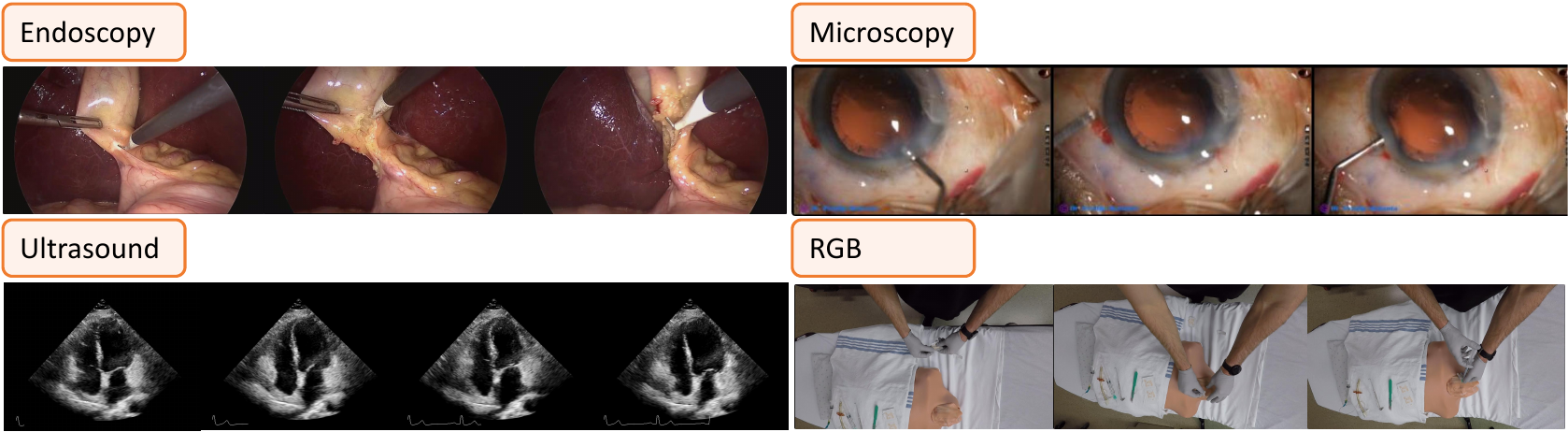}
    \caption{Illustration of four major modalities in the collected video datasets.}
    \label{fig:video_modality}
\end{figure}

In contrast, other imaging modalities are notably underrepresented.  
Microscopy videos (10 datasets) are often recorded during ophthalmic surgery to demonstrate detailed anatomical structures of the eye, such as the retina, iris, and pupil.
Ultrasound (7 dataset) are infrequently acquired as continuous cine sequences  due to clinical workflow constraints and the need for specialized protocols, such as dynamic perfusion studies or echocardiography loops. 
The RGB refers to videos captured with a camera in open environments and is most often associated with non-surgical scenarios, such as instructional recordings for emergency care, nursing, or simulated surgical procedures.
Figure \ref{fig:video_modality} illustrates four major modalities in the collected video datasets.


\subsection{Anatomical Structures}

Most of the video datasets (40 datasets) focus on abdominal anatomical structures, with a large proportion related to the gallbladder (13 datasets) and the colon (6 datasets). 
This is because many of these datasets were collected during procedures such as cholecystectomy and endoscopy. 
Seven datasets focus on eye-related anatomical structures, including the iris and pupil, with most of the videos collected during cataract surgery. 
The remaining video datasets cover a wide range of anatomical structures across the body, including the thyroid (1 dataset), pituitary (2 dataset) and placenta (1 dataset). 
A portion of the videos were also collected from non-human structures, such as artificial blood vessels (1 dataset) and porcine cadavers (1 dataset).


\subsection{Challenges and Opportunities}

The development of medical video datasets has enabled substantial progress across segmentation, detection, tracking, and registration, yet the field continues to face enduring \textbf{challenges}. 

\emph{Annotation quality}. 
On one hand, generating pixel- or frame-level ground truth requires extensive expert labor, particularly in domains such as surgical tool segmentation and landmark tracking, where precision and temporal consistency are essential \cite{giana2017challenge,hattab2021kidneyedge,maska2023celltracking,li2025ophora,}. 
Few datasets except the CaDIS dataset provide fine-grained, frame-level annotations covering full scenes in videos.
On the other hand, there may be variations in annotation quality across different videos, even within the same dataset, due to differences in surgeon skill when multiple annotators are involved.
Sparse or weak labels have been proposed as a compromise, but they often limit the reliability of downstream evaluation. 
Semi-supervised and synthetic data augmentation approaches show promise, though their acceptance in clinical research requires rigorous validation. 

\emph{Data privacy}. Unlike natural video, medical recordings inherently encode sensitive patient information. 
De-identification is particularly challenging in endoscopy and surgery, where anatomical context itself can serve as a patient identifier. 
Consequently, dataset releases are frequently restricted in scale or geographic scope, hampering the establishment of broadly generalizable benchmarks \cite{borgli2020hyperkvasir,goodman2024analyzing,hu2025towards}. Addressing this requires technical advances in anonymization as well as standardized regulatory and ethical frameworks that enable secure multi-center data sharing. 

\emph{Domain shift} represents a further persistent issue. 
Substantial variability arises from differences in imaging devices, acquisition protocols, and surgical practices, often causing models trained on one dataset to fail when applied to another. 
This problem has been observed across lesion detection, artifact removal, and instrument recognition benchmarks \cite{Bernal2017GIANA,ali2019endoscopy,bawa2021saras,yan2025make}. While phantom or synthetic data help isolate algorithmic behavior, bridging these controlled conditions with the complexity of real clinical environments remains an open research frontier. 
Robust domain adaptation, self-supervised pretraining, and benchmark designs that explicitly incorporate cross-institutional variation are therefore pressing needs. 

\emph{Computational burden.} From a computational standpoint, the scale of medical video poses formidable demands. High-resolution intraoperative recordings can span hours, making storage, annotation, and real-time analysis resource-intensive. Real-time deployment, for instance in robotic surgery or intraoperative navigation, requires methods that balance accuracy with computational efficiency \cite{zia2025surgvu}. Furthermore, emerging benchmarks increasingly combine multiple tasks—detection, segmentation, and tracking—placing pressure on algorithm design to unify spatiotemporal reasoning under constrained latency.

These challenges, however, also motivate transformative \textbf{opportunities}. The rise of multimodal datasets such as P2ILF and SAR-RARP50 opens pathways toward comprehensive scene understanding, aligning 2D video streams with 3D imaging modalities and enabling clinically relevant multimodal registration \cite{ali2022p2ilf,psychogyios2023sar}. The integration of large pre-trained models and foundation architectures has the potential to mitigate annotation bottlenecks and improve generalization across institutions, provided that interpretability and domain alignment are addressed. Longstanding community initiatives, such as the EndoVis series challenges, further underscore the importance of standardized evaluation protocols for reproducibility and clinical translation. Clinically, the opportunities are profound. Accurate lesion detection and temporal localization can support early diagnosis in screening procedures, while reliable instrument tracking and workflow analysis enable intraoperative decision support and skill assessment \cite{ghamsarian2023cataract1k,Borgli2020}. 
More broadly, the convergence of diverse datasets, robust benchmarking, and advanced learning paradigms is steering medical video analysis from narrow research prototypes toward clinically indispensable technologies.

In summary, medical video datasets face inherent challenges in annotation, privacy, domain robustness, and scalability, but these limitations are also drivers of innovation. With sustained progress in dataset diversity, federated evaluation, and integration with large-scale learning systems, the field is positioned to deliver clinically impactful solutions in the coming decade.

\begin{table}[t]
\centering
\scriptsize
\setlength{\tabcolsep}{4pt}
\begin{tabularx}{\textwidth}{c l X}
\toprule
ID & Field & Brief description \\
\midrule
1  & \texttt{dataset\_name} &
Official name or commonly used short name of the dataset. \\

2  & \texttt{release\_date} &
First public release date (YYYY-MM or YYYY-MM-DD; use \texttt{NA} if unknown). \\

3  & \texttt{homepage\_url} &
Stable URL or DOI for the dataset homepage, paper, or repository. \\

4  & \texttt{organization} &
Institution(s) releasing or hosting the dataset; multiple entries allowed, separated by commas. \\

5  & \texttt{challenge\_series} &
Name of the associated challenge or benchmark series; \texttt{NA} if not challenge-based. \\

6  & \texttt{license} &
Data usage license or access policy as specified by the download agreement. \\

7  & \texttt{dataset\_description} &
Short free-text summary of source, modality, tasks, scale, and key characteristics. \\

8  & \texttt{modality\_primary} &
Primary imaging modality or modalities (e.g., CT, MR, X-ray, Fundus). \\

9  & \texttt{modality\_secondary} &
Subtype or sequence within the primary modality (e.g., MR:T1, CT:CTA; \texttt{NA} if unspecified). \\

10 & \texttt{anatomical\_structure} &
Target organ, region, or lesion; multiple structures allowed. \\

11 & \texttt{disease} &
Disease or clinical condition(s) represented; \texttt{NA} for non disease-specific datasets. \\

12 & \texttt{data\_volume} &
Total size and split, preferably as JSON (e.g., \texttt{\{"total":..., "train":...\}}). \\

13 & \texttt{valid\_image\_n} &
Usable sample count after cleaning, optionally in the same JSON format as \texttt{data\_volume}. \\

14 & \texttt{label\_presence} &
Annotation availability: \texttt{labeled}, \texttt{unlabeled}, or \texttt{mixed}. \\

15 & \texttt{task\_type} &
Supported computational tasks (e.g., segmentation, detection, classification, VQA). \\

16 & \texttt{num\_classes\_per\_task} &
JSON describing, per task, the number of classes/targets and relevant settings. \\
\bottomrule
\end{tabularx}
\caption{Definition of \texttt{data\_meta} fields for dataset-level metadata.}
\label{tab:data-meta-schema}
\end{table}
\section{Paradigm for Dataset Fusion}
\label{sec:merge_paradigm}


\begin{figure}
    \centering
    \includegraphics[width=0.95\linewidth]{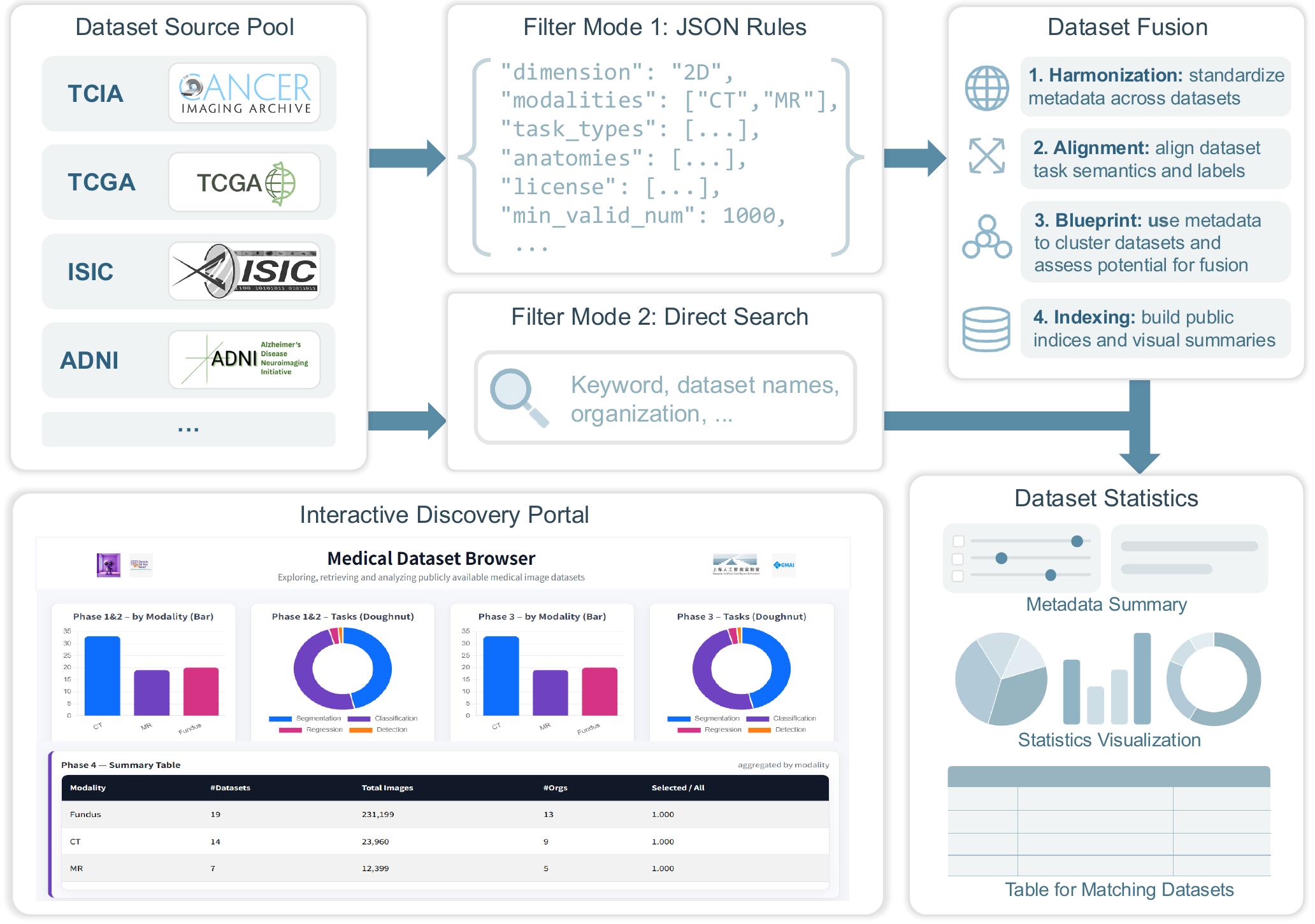}
    \caption{Pipeline of our dataset collection, processing, fusion, and summarization system based on the proposed dataset fusion paradigm, implemented in our interactive discovery portal.}
    \label{fig:search_pipeline}
\end{figure}

\begin{figure}
    \centering
    \includegraphics[width=\linewidth]{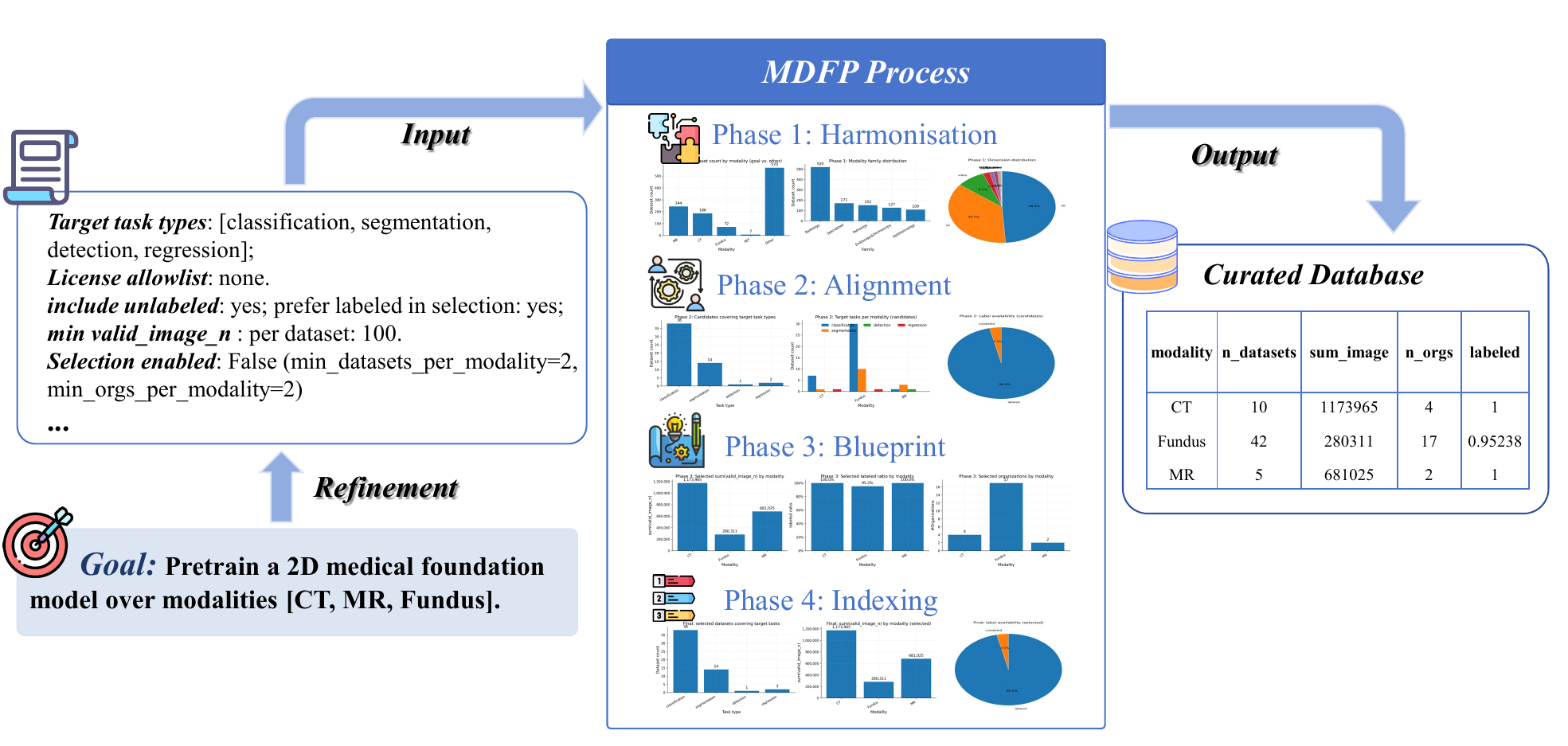}
    \caption{Detailed process of the proposed metadata-driven fusion paradigm (MDFP).}
    \label{fig:mdfp}
\end{figure}

Despite the abundance of public medical imaging datasets, their fragmentation significantly hinders their effective use in large-scale model training. To address this, we propose the \emph{Metadata-Driven Fusion Paradigm} (MDFP), grounded in our comprehensive collection and curation of medical imaging datasets, offering an efficient, scalable, and metadata-centric strategy to systematize discovery, auditing, and composition of multiple datasets. By operating primarily on metadata rather than raw pixels, MDFP reduces handling overhead and privacy risk, improves reproducibility and auditability, and enables rapid goal-conditioned dataset assembly. 
Based on MDFP, we build an interactive discovery portal that supports fine-grained dataset search, integration, and statistical analysis. Figure~\ref{fig:search_pipeline} overviews the full system, and Figure~\ref{fig:mdfp} details MDFP. 

The remainder of this section proceeds in the following order: we describe dataset collection and processing (Section~\ref{sec:data_col_pro}); introduce MDFP and its four phases (Section~\ref{sec:mdfp}); incorporate the aforementioned components to form the interactive discovery protal (Section~\ref{sec:web-retrieval}).

\subsection{Dataset Collection and Processing}
\label{sec:data_col_pro}
All datasets included in this study were obtained from publicly accessible web-based repositories, such as The Cancer Imaging Archive (TCIA)\footnote{\url{https://www.cancerimagingarchive.net}}, Grand Challenge\footnote{\url{https://grand-challenge.org}}, OpenNeuro\footnote{\url{https://openneuro.org}}, Kaggle\footnote{\url{https://www.kaggle.com/}}, NeuroImaging Tools and Resources Collaboratory (NITRC)\footnote{\url{https://www.nitrc.org}}, Synapse\footnote{\url{https://www.synapse.org}}, CodaLab\footnote{\url{https://codalab.lisn.upsaclay.fr}}, GitHub\footnote{\url{https://github.com}}, \etc 
After collecting the medical imaging datasets from these sources, we organize them into a multi-dimensional database that serves as a comprehensive overview table. This database categorizes each dataset by multiple attributes, including dimension, modality, anatomical structure, number of cases, label availability, and task type, along with other essential metadata. Such an organization enables flexible querying and filtering, allowing researchers to quickly retrieve datasets that match specific research needs, \ie, training a 3D foundation model for CT, MRI, and PET.


\begin{table}[t]
\centering
\scriptsize
\setlength{\tabcolsep}{4pt}
\begin{tabularx}{\textwidth}{l l X}
\toprule
Block & Role & Representative fields and examples \\
\midrule
\texttt{record} &
Sample identifiers and cross-references &
\texttt{dataset\_name}, \texttt{image\_path}, optional sample IDs used to locate the underlying media file and join annotations across tasks. \\[0.3em]

\texttt{context} &
Clinical and textual context &
Subject- and acquisition-level metadata (e.g., subject ID, age, sex, site, modality, anatomy), extensible \texttt{extra} dictionary, free-text descriptions or reports. \\[0.3em]

\texttt{media\_geometry} &
Media attributes and geometry &
Media-level task configuration and spatio-temporal metadata, including \texttt{task\_type}, \texttt{leaf\_task}, \texttt{annotation\_type}, \texttt{dimension}, pixel spacing, orientation, slice/frame indices, timestamps, camera parameters. \\[0.3em]

\texttt{tasks} &
Task-specific annotation payloads &
Structured labels for the predefined tasks, such as segmentation masks, detection boxes, class labels, polygons, or keypoints, grouped by \texttt{dimension} and \texttt{schema\_variant}. \\
\bottomrule
\end{tabularx}
\caption{Four logical blocks---\texttt{record}, \texttt{context}, \texttt{media\_geometry}, and \texttt{tasks}---structuring \texttt{annotations\_\{task\}.jsonl}. The blocks separate sample identity, context, media-level geometry, and task-specific annotation payloads.}
\label{tab:ann-schema-blocks}
\end{table}

For each individual dataset, we preserve the original directory layout as much as possible and enrich it with two JSONL files. The file \texttt{data-meta.jsonl} records dataset-level information such as release date, imaging modality, homepage URL, and license, using 16 well-defined fields (summarized in Table~\ref{tab:data-meta-schema}). The file \texttt{annotations-\{task\}.jsonl} stores per-media, task-specific annotations (for example, mask file paths for segmentation tasks or bounding boxes for detection tasks). Each JSON object describes a single annotated media item for a particular task and is decomposed into four logical information blocks: \texttt{record}, \texttt{context}, \texttt{media\_geometry}, and \texttt{tasks}. The \texttt{record} block contains stable identifiers such as \texttt{dataset\_name} (dataset identifier) and \texttt{image\_path} (path or key of the underlying image, volume, or video file within that dataset), which link the annotation back to the original media file and allow annotations for the same sample to be joined across tasks. The \texttt{context} block collects optional subject- and acquisition-level metadata and free-text descriptions (e.g., subject ID, age, sex, site, modality, anatomy, and an extensible \texttt{extra} dictionary for dataset-specific fields). The \texttt{media\_geometry} block captures media-level attributes and spatial/temporal geometry that are shared by all annotations on the same media item, including the high-level \texttt{task\_type} (one of 12 predefined task categories such as segmentation, detection, or classification), the dataset-specific \texttt{leaf\_task}, the \texttt{annotation\_type} (e.g., binary masks, bounding boxes, polygons), the \texttt{dimension} (2D, 3D, or video), and imaging metadata such as pixel spacing, orientation, slice or frame indices, timestamps, and camera parameters. Finally, the \texttt{tasks} block contains the structured task-specific annotation payloads themselves (e.g., mask references, box coordinates, class labels, keypoints), organized in a schema that is consistent across datasets for the same task type. A compact overview of these four blocks and their representative fields is given in Table~\ref{tab:ann-schema-blocks}.

Building on the standardized directory structure, we further unify file formats with a focus on preserving quantitative information and metadata. Specifically, all volumetric imaging data (\eg, CT, MRI, PET) are converted to NIfTI (.nii.gz), with voxel spacing, orientation (qform/sform), and intensity scaling (slope/intercept or equivalent) preserved; dynamic PET is stored as 4D NIfTI with companion JSON/TSV files for frame timing and calibration (\eg, SUV factors). When appropriate, de-identified source DICOMs are retained as an optional raw layer. For 2D modalities (\eg, radiographs, ultrasound frames), we preserve full dynamic range using lossless 16-bit formats (TIFF or 16-bit PNG) together with sidecar JSON for essential metadata (pixel spacing, orientation, window settings, modality-specific tags). For video data (\eg, endoscopy, ultrasound cine, surgical recordings), we store sequences in compressed formats (MP4 or AVI with H.264/H.265 encoding) while maintaining 8-bit color depth for most clinical videos; specialized applications requiring higher dynamic range (\eg, fluorescence microscopy, high-speed recordings) are preserved as 16-bit sequences when available. Video metadata, including frame rate, resolution, acquisition timestamps, and equipment parameters, are recorded in companion JSON files to ensure reproducibility and temporal consistency. Optional 8-bit PNG thumbnails may be generated solely for visualization and documentation; these are never used as training inputs when quantitative intensity matters. This preprocessing and standardization workflow maintains dataset fidelity and compatibility, facilitates seamless integration into model-training pipelines, and enables reproducible and comparable benchmarking across studies.

\subsection{MDFP}\label{sec:mdfp}

MDFP systematizes discovery, auditing, and composition through four sequential phases that operate primarily on structured metadata, strengthening privacy, auditability, and reproducibility while avoiding raw-pixel handling. 

Table~\ref{tab:mdfp-overview} presents these phases of MDFP, outlining their core objectives and associated metadata fields. These phases are tightly aligned with our systematic metadata collection framework, ensuring consistency, completeness, and interoperability across heterogeneous datasets. Below, we detail each phase.

\begin{table}
\centering
\caption{MDFP Workflow Overview with Key Metadata Fields}
\label{tab:mdfp-overview}
\small
\begin{tabular}{@{}p{2cm}p{5cm}p{6cm}@{}}
\toprule
\textbf{Phase} & \textbf{Objective} & \textbf{Metadata Fields} \\ 
\midrule
1. Harmonization & Standardize modalities, tasks, and anatomy. & \texttt{modality\_primary}, \texttt{dimension}, \texttt{anatomical\_structure}, \texttt{organization}, \texttt{challenge\_series} \\[0.5ex]
\midrule
2. Alignment & Align semantic labels and tasks across datasets. & \texttt{task\_type}, \texttt{modality\_secondary}, \texttt{label\_presence}, \texttt{notes} \\[0.5ex]
\midrule
3. Blueprint & Cluster datasets; assess integrative potential and data scale. & \texttt{data\_volume}, \texttt{valid\_image\_n}, \texttt{storage\_size\_gb} \\[0.5ex]
\midrule
4. Indexing & Create public metadata indices and visualization tools for easy access. & \texttt{dataset\_name}, \texttt{release\_date}, \texttt{homepage\_url}, \texttt{license} \\ 
\bottomrule
\end{tabular}
\end{table}

\subsubsection{Phase 1: Metadata Harmonization}
\label{sec:phase1}
Phase 1 resolves semantic heterogeneity by enforcing a rigorously defined metadata schema. Rather than creating a new vocabulary, we ground our schema in authoritative medical terminologies, such as the Unified Medical Language System (UMLS) and Medical Subject Headings (MeSH)~\cite{bodenreider2004unified}. This process is semi-automated, leveraging API-driven searches against these ontologies, followed by LLM-based refinement to programmatically align disparate dataset descriptors into a consistent, machine-readable form. Concretely, we:

\begin{itemize}
    \item \textbf{Standardize primary modality} (\texttt{modality\_primary}): Mapped to an enumerated set including CT, MR, PET, US, X-ray, \etc, with niche modalities deterministically aligned to this taxonomy.
    \item \textbf{Normalize data dimensionality} (\texttt{dimension}): Parsed directly from dataset metadata to determine whether the data is 2D, 3D, or video (2D + time).
    \item \textbf{Establish hierarchical classification}: Instead of a simple anatomical tag, we implement a multi-level classification system based on standard medical ontologies (\eg, UMLS, MeSH). This provides rich, hierarchical context. For example, a dataset on cataracts would be classified under \textit{Eye Diseases} $\rightarrow$ \textit{Lens Diseases} $\rightarrow$ \textit{Cataract}.
    \item \textbf{Record provenance and context} (\texttt{organization}, \texttt{challenge\_series}): Identifying the originating institution and any associated benchmark or competition series.
    \item \textbf{Document annotation granularity} (\texttt{annotation\_type}): Explicitly cataloging the type and granularity of labels provided, including landmark coordinates, pixel-level segmentation masks, region-level bounding boxes, image-level classification labels, or multi-modal annotations. This metadata enables researchers to identify datasets compatible with their task requirements and facilitates appropriate fusion strategies during retrieval.
\end{itemize}

This harmonization step yields a uniform, richly annotated metadata table that transforms a fragmented corpus into an interoperable resource, thereby enabling reliable cross-dataset comparison, reproducible filtering, and seamless integration.

\subsubsection{Phase 2: Semantic Alignment}

Phase 2 mitigates inconsistencies by mapping abstract machine learning tasks to their concrete clinical significance. This crucial step involves a systematic review of dataset documentation, source publications, and official guidelines to understand the intended real-world application. By doing so, we align heterogeneous labeling conventions and evaluation objectives with tangible clinical goals. Specifically, we:

\begin{itemize}
    \item \textbf{Define downstream tasks} (\texttt{downstream\_task}): We standardize ML tasks and explicitly map them to their clinical applications. For example:
    \begin{itemize}
        \item A \texttt{classification} task might correspond to clinical \textit{diagnosis} (\eg, malignant vs. benign tumor), \textit{severity grading} (\eg, staging diabetic retinopathy), or \textit{treatment response prediction}.
        \item A \texttt{segmentation} task may be used for \textit{lesion delineation} (\eg, outlining a tumor boundary), \textit{volumetric quantification} (\eg, measuring organ volume to track disease progression), or \textit{radiotherapy/surgical planning}.
\item A \texttt{detection} task is often used for \textit{disease screening} (\eg, identifying candidate pulmonary nodules in a chest CT).
        \item A \texttt{regression} task can quantify \textit{clinical biomarkers} (\eg, predicting bone mineral density from a CT scan or cardiac ejection fraction from an ultrasound video).
    \end{itemize}
    \item \textbf{Indicate label availability} (\texttt{label\_presence}): Denoting whether ground-truth annotations are provided (\texttt{labeled}) or not (\texttt{unlabeled}).
    \item \textbf{Specify secondary imaging modalities} (\texttt{modality\_secondary}): Capturing finer-grained protocol-level distinctions under each primary modality, such as T1 or T2 sequences for MR.
    \item \textbf{Document special considerations} (\texttt{notes}): Capturing dataset-specific nuances, assumptions, or known limitations in free-text form.
\end{itemize}

This alignment phase yields a clinically-grounded task vocabulary that supports meaningful interpretation, goal-oriented filtering, and enhances the reliability of cross-dataset benchmarking.

\subsubsection{Phase 3: Fusion Blueprints }
This phase leverages harmonized metadata to design strategic dataset integration plans.
Specifically, we perform grouping and categorization based on combinations of primary and secondary imaging modalities (\texttt{modality\_primary}, \texttt{modality\_secondary}), clinical tasks (\texttt{task\_type}), and anatomical coverage (\texttt{anatomical\_structure}). This grouping process consolidates datasets with similar or identical attributes into unified groups, providing a structured foundation for designing fusion blueprints that guide principled dataset integration. Quantitative evaluations are systematically derived from metadata, encompassing the following aspects:

\begin{itemize}
    \item \textbf{Data Volume (\texttt{data\_volume})}: Assess total images available, along with explicit training, validation, and testing splits.
    \item \textbf{Valid Image Counts (\texttt{valid\_image\_n})}: Determine precisely how many images have reliable and validated annotations, critical for training supervised models. These quantitative statistics are obtained directly from official dataset documentation, README files, published papers, and other authoritative public resources provided by dataset curators.
    \item \textbf{Storage Estimation (\texttt{storage\_size\_gb})}: Evaluate practical storage requirements, essential for infrastructure planning.
    \item \textbf{Anatomical and Task Diversity (\texttt{anatomical\_structure}, \texttt{task\_type})}: Quantify anatomical breadth and task variety within each fusion cluster, ensuring coverage diversity crucial for generalization.
\end{itemize}

This structured assessment produces a principled basis for scalable dataset merging, balancing quantity, annotation quality, and content diversity to support robust foundation model training. During fusion blueprint design, we explicitly account for data heterogeneity captured in Phase 1, including variations in imaging protocols (\eg, differences in CT reconstruction parameters or MRI field strengths), image resolutions, and annotation granularities. Our tool systematically identifies and flags datasets with incompatible annotation types (\eg, mixing pixel-level segmentation with bounding-box detection) and imaging protocol differences, alerting researchers to potential integration challenges. These metadata annotations help researchers make informed decisions about whether harmonization preprocessing, protocol-aware sampling strategies, or domain-adaptive training approaches are needed to ensure cross-dataset compatibility and model robustness.

\subsubsection{Phase 4: Dataset Indexing and Community Sharing}

Phase 4 transforms the harmonized metadata into a structured, publicly accessible dataset index to support community-scale discovery and reuse. We consolidate key metadata elements for each dataset, including:

\begin{itemize}
    \item \textbf{Dataset name} (\texttt{dataset\_name}): the canonical name of the dataset for standardized referencing;
    \item \textbf{Release date} (\texttt{release\_date}): official publication or release timestamp, enabling temporal filtering;
    \item \textbf{Homepage URL} (\texttt{homepage\_url}): direct access link to dataset documentation or hosting platform;
    \item \textbf{License} (\texttt{license}): clearly defined usage permissions, ensuring legal compliance and reproducibility.
\end{itemize}

This indexed representation facilitates rapid dataset discovery, promotes responsible reuse, and provides the infrastructure foundation for large-scale model pretraining, benchmarking, and open collaboration.

\subsubsection{Case Study: Goal-Conditioned Fusion via MDFP }

\begin{table}[t]
\centering
\caption{MDFP-derived composition for the 2D CT/MR/Fundus goal. \texttt{sum\_image} is \texttt{sum(valid\_image\_n)}; \texttt{labeled\_ratio} is the fraction of datasets that are labeled. }
\label{tab:mdfp-case-2d-ct-mr-fundus}
\small
\begin{tabular}{@{}lrrrr@{}}
\toprule
\textbf{modality} & \textbf{n\_datasets} & \textbf{sum\_image} & \textbf{n\_orgs} & \textbf{labeled\_ratio} \\
\midrule
CT     & 10 & 1{,}173{,}965 &  4 & 1.000 \\
MR     &  5 &   681{,}025   &  2 & 1.000 \\
Fundus & 42 &   280{,}311   & 17 & 0.952 \\
\bottomrule
\end{tabular}
\end{table}

As shown in Figure \ref{fig:mdfp}, to demonstrate how MDFP supports foundation-model pretraining with reproducible, goal-aligned data composition, we instantiate a concrete target: a \emph{2D} model over modalities \{CT, MR, Fundus\} and tasks \{classification, segmentation, detection, regression\}.

In \textbf{Phases 1--2} (Harmonization and Alignment), we apply the following filtering and standardization procedures: First, we standardize all datasets by mapping primary modalities to our controlled vocabulary (CT, MR, Fundus) and normalize data dimensionality to 2D only, explicitly excluding 3D volumetric data and video sequences to maintain dimensional consistency. Second, we establish hierarchical anatomical classifications using UMLS/MeSH ontologies for each dataset, enabling consistent cross-dataset anatomical mapping. Third, we perform semantic alignment by mapping machine learning task types to their clinical applications, ensuring that selected datasets cover the four target task families: classification (diagnosis, severity grading), segmentation (lesion delineation, volumetric quantification), detection (disease screening), and regression (clinical biomarker quantification). Fourth, we apply quality filters including minimum sample size (\texttt{valid\_image\_n}\,$\geq$\,100) to ensure statistical reliability. Both labeled and unlabeled datasets are retained, with preference given to labeled ones when multiple alternatives exist. 
No license restrictions are enforced in this demonstration, though license-aware filtering is supported by the framework.

In \textbf{Phases 3--4} (Blueprint and Indexing), we perform grouping and categorization based on the harmonized metadata from Phases 1--2. Specifically, we group datasets by modality-task combinations, assess the integrative potential of each cluster by quantifying data volume, annotation availability, and anatomical coverage, and generate a fusion blueprint that summarizes the composition strategy. Finally, we create a structured metadata index with all essential fields (\texttt{dataset\_name}, \texttt{modality}, \texttt{task}, \texttt{valid\_image\_n}, \texttt{license}, \texttt{homepage\_url}) to enable reproducible access and community sharing.

The resulting integrated dataset composition is summarized in Table~\ref{tab:mdfp-case-2d-ct-mr-fundus}. The curated pool comprises 57 datasets and 2{,}135{,}301 validated images across three imaging modalities: CT (10 datasets, 1{,}173{,}965 images, 4 organisations), MR (5 datasets, 681{,}025 images, 2 organisations), and Fundus (42 datasets, 280{,}311 images, 17 organisations). These numbers correspond to the summary fields \texttt{n\_datasets}, \texttt{sum\_image} (defined as $\sum$\,\texttt{valid\_image\_n}), \texttt{n\_orgs}, and \texttt{labeled\_ratio} in Table~\ref{tab:mdfp-case-2d-ct-mr-fundus}. All CT and MR datasets are fully annotated (\texttt{labeled\_ratio}=1.000), and Fundus datasets achieve a high annotation rate (\texttt{labeled\_ratio}=0.952), yielding strong supervision across modalities while satisfying the case-study constraints (2D only; CT/MR/Fundus) and covering the four target task families defined in Phase~2 (classification, segmentation, detection, regression). This configuration constitutes a concrete instantiation of goal-conditioned dataset integration via MDFP.

These aggregate statistics have direct implications for foundation-model pretraining. The high labeled fractions support multi-task supervised objectives, while the remaining unlabeled images can be exploited with auxiliary self-supervised losses. At the same time, the skew in \texttt{sum\_image} toward CT and MR (CT+MR $\approx$ 1.85\,M images) suggests employing modality-aware sampling strategies (e.g., temperature-based sampling or per-dataset caps) and task-stratified batching to prevent over-representation of these modalities. Finally, although this case study restricts sources to 2D CT/MR/Fundus datasets for clarity, the same MDFP pipeline can be rerun with relaxed configuration (e.g., enabling \texttt{allow\_3d\_as\_2d\_sources=true}) to augment the 2D pool with projected 3D or video data when broader coverage is desired.

\subsection{Interactive Discovery Portal}
\label{sec:web-retrieval}


Combining the aforementioned components together, we build a lightweight interactive discovery portal, namely the \emph{Medical Dataset Browser}, to triage and refine candidate datasets before schema-level alignment. The portal is deployed as a single page static application on GitHub Pages\footnote{\url{https://tchenglv520.github.io/medical-dataset-browser/}}, executes entirely client-side, and consumes at runtime the standardized JSON artifact produced in 6.1 (for example, the cleaned and merged manifest). This design eliminates server-side dependencies, simplifies reproducibility, and enables privacy-preserving exploration. Below, we detail the pipeline.

\paragraph{Dataset Filtering.} 
Starting from the dataset source pool prepared in Section~\ref{sec:data_col_pro}, the portal exposes two complementary modes for dataset filtering. First, \textit{Rule-based filtering} (``Filter Mode~1''). This mode implements the MDFP, accepting an editable JSON specification that encodes deterministic selection criteria, \eg, image \emph{dimension} (2D/3D), \emph{modality} sets (CT/MRI/US/Pathology, \etc), \emph{task} types (segmentation, detection, classification, report generation), \emph{organ/anatomy} whitelists, \emph{license} constraints, minimum sample sizes, and \emph{year} ranges. This recipe-like abstraction makes selections auditable and perfectly reproducible. 

Specifically, the controls for \emph{Phase~1\&2 in MDFP} (harmonization and alignment) and \emph{Phase~3\&4} (blueprint and indexing) are integrated into the page to preview downstream effects before committing a batch run. During execution, the interface highlights in-progress elements and then surfaces consolidated outputs for inspection. 

In parallel, \textit{direct faceted search} (``Filter Mode~2'') provides dropdown facets and a free text query for fast exploratory narrowing. Both modes drive live visual summaries, complete bar and doughnut charts of dimension-modality-task distributions, so users can immediately assess coverage and balance of the current subset.

\paragraph{Statistics and summaries.} 
After the dataset filtering, the portal renders live bar/pie summaries of modality, dimension, task, and anatomy distributions; a MDFP Phase-4 audit table exposes fields essential for screening and compliance: \texttt{name}, \texttt{dimension}, \texttt{modality}, \texttt{task}, \texttt{organ}, \texttt{images} (counts), \texttt{year}, \texttt{organization}, \texttt{license}, and \texttt{link}. These statistics and summaries can be exported to CSV/JSON for benchmarking pipelines. 
Together, these views close the loop from search to fusion to shareable artifacts.

\paragraph{Implementation details.}
The \texttt{index.html} bootstraps by loading the preprocessed JSON manifest, initializes an in-memory filter store, and applies deterministic, order-independent rule evaluation entirely on the client. Visual analytics and tables are rendered with lightweight, dependency-minimal components; results are paginated to maintain interactivity on medium-to-large corpora. Because the application is a self-contained static bundle, any user can fork, reconfigure the selection recipe, and redeploy an identical retrieval environment without additional infrastructure.

\section{Discussion}\label{sec:discussion}

\subsection{Limitations in Task Definition and Evolution of Data Engineering Paradigms}

Current open-access medical imaging datasets exhibit limitations in task definition, reflecting the task-oriented nature of early deep learning practices \cite{zhang2024generalist,moor2023foundation}. Most datasets target indirect downstream tasks (\eg segmentation, classification, or detection), which served as proxies for clinical goals but remain distant from real-world applications. With the advancement of foundation models and LLMs, AI systems are shifting toward direct, clinically relevant tasks such as disease diagnosis, patient condition assessment, and treatment recommendation \cite{zhu2025enhancing}. This paradigm shift creates a critical mismatch: existing datasets, designed for classical computer vision tasks, cannot be directly utilized without substantial transformation. However, re-annotation for clinically oriented tasks incurs prohibitively high costs, as these tasks demand expert-level medical knowledge and high-quality annotations from domain specialists. For instance, a lung nodule segmentation mask does not indicate whether the nodule is benign, malignant, or requires biopsy—information essential for clinical decision support but absent in existing annotations. 

Bridging this gap requires resource-intensive re-annotation by radiologists, a process that scales poorly across large datasets. Consequently, the medical AI community faces a dual challenge: legacy datasets are misaligned with contemporary needs, yet creating new foundation-model-ready datasets remains economically and logistically prohibitive. Future data engineering must prioritize flexible annotation frameworks that capture clinically meaningful information upfront, enabling adaptation to evolving AI paradigms without complete re-annotation.

\subsection{Scarcity of Multimodal Medical Datasets and Constraints in Further Development}

Multimodal medical data that integrates imaging modalities (CT, MRI, pathology) with clinical reports, genomics, and temporal records holds exceptional value for clinical diagnosis, yet remains exceedingly rare in the public domain \cite{moor2023foundation}. Most open-access datasets are unimodal and lack standardized frameworks for multimodal collection and annotation \cite{huang2024multimodal, gao2025barlbilateralalignmentrepresentation}, significantly restricting research in cross-modal reasoning and joint representation learning essential for next-generation medical AI.
The challenge extends beyond data availability to fundamental issues of modal alignment and semantic consistency. Different modalities operate on disparate scales: pathology captures microscopic cellular details, radiology visualizes organ-level structures, and clinical notes document temporal disease progression. 

Harmonizing these heterogeneous streams requires sophisticated alignment protocols and cross-modal validation standards that current datasets rarely provide. For example, aligning a radiologist's report timestamp with the corresponding imaging study, or synchronizing pathology findings with longitudinal treatment records, demands metadata infrastructure largely absent in existing resources.
Moreover, the absence of standardized multimodal benchmarks impedes systematic evaluation of cross-modal architectures. Researchers lack unified frameworks to assess whether models effectively integrate complementary information across modalities or leverage modal-specific strengths to compensate for individual limitations. This evaluation gap slows development of clinically viable systems capable of synthesizing diverse diagnostic information as human clinicians do.
The technical complexity of multimodal data management compounds these challenges. Institutions struggle with storage, versioning, and synchronization of large-scale heterogeneous datasets, while privacy regulations complicate cross-institutional data sharing. Without robust infrastructure and standardized curation protocols, the field remains fragmented, with isolated efforts failing to achieve the critical mass needed for breakthrough advances in multimodal medical AI.

\subsection{Challenges and Opportunities in Medical Foundation Models}

Medical foundation models demand unprecedented scale and diversity in training data, yet current resources remain insufficient for developing truly generalizable systems \cite{schafer2024overcoming,li2024gmai,su2025gmai, wang2025v2t}. The gap between available data and foundation model requirements is particularly evident in specialized domains such as pediatric imaging, rare diseases, and longitudinal treatment monitoring. Three interconnected challenges fundamentally constrain progress in this field.

\textbf{Scale and Representational Diversity.} Beyond sheer quantity, foundation models require comprehensive coverage across disease presentations, imaging protocols, clinical specialties, and patient demographics to develop robust internal representations. Current medical datasets typically capture narrow slices of clinical reality, missing the long-tail distribution of rare conditions and atypical presentations that characterize real medical practice. This limitation is especially acute in underrepresented populations and emerging disease variants.

\textbf{Licensing and Privacy Constraints.} Unlike general-domain AI where datasets can be freely shared, medical data faces dual constraints from patient privacy regulations (e.g., HIPAA, GDPR) and institutional intellectual property policies. Even when foundation models can generate high-quality synthetic data for training augmentation \cite{hu2024diffusion, li2025ophora}, restrictive licensing prevents these enhanced datasets from benefiting the broader research community. This regulatory landscape fragments the field, forcing redundant efforts across institutions and limiting collaborative progress \cite{sidebottom2021fair}.

\textbf{Contextual and Temporal Intelligence.} Effective medical AI must transcend pattern recognition to understand clinical workflows, resource constraints, and patient-specific contexts \cite{li2025one}. For instance, models must distinguish between emergency protocols and routine screening, interpret how prior treatments influence current presentations, and track disease progression over time. Current training paradigms inadequately address these temporal reasoning and workflow integration capabilities essential for real-world deployment.

Addressing these challenges requires coordinated efforts to establish data governance frameworks that balance privacy protection with research advancement. Without systemic solutions—including federated learning infrastructures, standardized licensing models, and clinically-grounded evaluation benchmarks—medical foundation models will remain confined to narrow applications rather than achieving the general intelligence needed for transformative clinical impact.
\section{Conclusion}\label{sec:conclusion}

This comprehensive survey of over 1,000 open-access medical image datasets reveals a fragmented and imbalanced landscape that fundamentally constrains the development of medical foundation models. Existing datasets remain predominantly small-scale, task-specific, and modality-restricted, with pronounced disparities across anatomical regions and imaging modalities. These limitations reflect the field's incomplete transition from task-oriented to foundation-oriented data engineering paradigms. To address these challenges, we formulate the Metadata-Driven Fusion Paradigm (MDFP), a systematic framework for dataset integration that enables the construction of larger, more diverse training resources essential for foundation model development. Our analysis identifies three critical gaps: the scarcity of multimodal datasets that limits cross-modal reasoning capabilities, restrictive licensing and privacy regulations that fragment collaborative efforts, and the absence of contextual intelligence necessary for real-world clinical deployment. The dominance of segmentation and classification tasks, alongside the underrepresentation of emerging applications like visual question answering and multimodal reasoning, underscores the urgent need for comprehensive data engineering strategies. 

Looking forward, advancing the development of medical foundation models requires a collective shift toward openness, efficiency, and inclusivity in data engineering. A key priority should be to encourage broader public release of medical imaging datasets, thereby enhancing transparency, reproducibility, and equitable access across institutions and regions. In parallel, research on synthetic data generation holds promise for mitigating privacy and data scarcity challenges, while annotation-efficient learning approaches can enable effective use of partially labeled or weakly supervised data. Moreover, the public release of foundation models trained on private or institution-specific data, even when raw datasets cannot be shared, represents a practical pathway to democratize access to advanced medical AI capabilities. Together, these strategies constitute a sustainable and collaborative framework for building truly generalizable and clinically impactful medical foundation models.

\section*{Acknowledgment}

We sincerely thank all researchers, clinicians, institutions, and organizations who have contributed to the development and public release of medical imaging datasets. Their dedicated efforts in data collection, annotation, curation, and sharing have laid the foundation for significant progress in medical AI. The open availability of these resources has not only accelerated methodological innovation and benchmark creation but also fostered collaboration across disciplines, enabling the broader community to explore new directions in multimodal learning, foundation model development, and clinical translation. Without their commitment to advancing science through openness and collaboration, this survey and many of the achievements in the field would not have been possible.

\bibliographystyle{unsrtnat}
\bibliography{citation,Table_2D/2d,Table_3D/3d,Table_Video/video}

@article{van2018automated,
  title={Automated measurement of fetal head circumference using 2D ultrasound images},
  author={van den Heuvel, Thomas LA and de Bruijn, Dagmar and de Korte, Chris L and Ginneken, Bram van},
  journal={PloS one},
  volume={13},
  number={8},
  pages={e0200412},
  year={2018},
  publisher={Public Library of Science San Francisco, CA USA}
}

@article{fang2022adam, title={Adam challenge: Detecting age-related macular degeneration from fundus images}, author={Fang, Huihui et al.}, journal={IEEE transactions on medical imaging}, volume={41}, number={10}, pages={2828--2847}, year={2022}}

@article{al2020busi,
  title={Dataset of breast ultrasound images},
  author={Al-Dhabyani, Walid and Gomaa, Mohammed and Khaled, Hussien and Fahmy, Aly},
  journal={Data in Brief},
  volume={28},
  pages={104863},
  year={2020}
}

@inproceedings{li2020benchmark,  
  title={A benchmark of ocular disease intelligent recognition: One shot for multi-disease detection},  
  author={Li, Ning and Li, Tao and Hu, Chunyu and Wang, Kai and Kang, Hong},  
  booktitle={International symposium on benchmarking, measuring and optimization},  
  pages={177--193},  
  year={2020},  
  organization={Springer}
}

@article{hoover2000locating,  
  title={Locating blood vessels in retinal images by piecewise threshold probing of a matched filter response},  
  author={Hoover, AD and Kouznetsova, Valentina and Goldbaum, Michael},  
  journal={IEEE Transactions on Medical Imaging},  
  volume={19},  
  number={3},  
  pages={203--210},  
  year={2000},  
  publisher={IEEE}
}

@article{lu2016evaluation,
  title={Evaluation of three algorithms for the segmentation of overlapping cervical cells},
  author={Lu, Zhi and Carneiro, Gustavo and Bradley, Andrew P and Ushizima, Daniela and Nosrati, Masoud S and Bianchi, Andrea GC and Carneiro, Claudia M and Hamarneh, Ghassan},
  journal={IEEE journal of biomedical and health informatics},
  volume={21},
  number={2},
  pages={441--450},
  year={2016},
  publisher={IEEE}
}

@article{bulten2019epithelium,
  title={Epithelium segmentation using deep learning in H\&E-stained prostate specimens with immunohistochemistry as reference standard},
  author={Bulten, Wouter and B{\'a}ndi, P{\'e}ter and Hoven, Jeffrey and Loo, Rob van de and Lotz, Johannes and Weiss, Nick and Laak, Jeroen van der and Ginneken, Bram van and Hulsbergen-van de Kaa, Christina and Litjens, Geert},
  journal={Scientific reports},
  volume={9},
  number={1},
  pages={864},
  year={2019},
  publisher={Nature Publishing Group UK London}
}

@article{allison2014understanding,
  title={Understanding diagnostic variability in breast pathology: lessons learned from an expert consensus review panel},
  author={Allison, Kimberly H and Reisch, Lisa M and Carney, Patricia A and Weaver, Donald L and Schnitt, Stuart J and O'Malley, Frances P and Geller, Berta M and Elmore, Joann G},
  journal={Histopathology},
  volume={65},
  number={2},
  pages={240--251},
  year={2014},
  publisher={Wiley Online Library}
}

@inproceedings{medmnistv1,
    title={MedMNIST Classification Decathlon: A Lightweight AutoML Benchmark for Medical Image Analysis},
    author={Yang, Jiancheng and Shi, Rui and Ni, Bingbing},
    booktitle={IEEE 18th International Symposium on Biomedical Imaging (ISBI)},
    pages={191--195},
    year={2021}
}

@article{holm2017dr,  
  title={DR HAGIS—a fundus image database for the automatic extraction of retinal surface vessels from diabetic patients},  
  author={Holm, Sven and Russell, Greg and Nourrit, Vincent and McLoughlin, Niall},  
  journal={Journal of Medical Imaging},  
  volume={4},  
  number={1},  
  pages={014503--014503},  
  year={2017},  
  publisher={Society of Photo-Optical Instrumentation Engineers}
}

@article{cuadros2009eyepacs,  
  title={EyePACS: an adaptable telemedicine system for diabetic retinopathy screening},  
  author={Cuadros, Jorge and Bresnick, George},  
  journal={Journal of diabetes science and technology},  
  volume={3},  
  number={3},  
  pages={509--516},  
  year={2009},  
  publisher={SAGE Publications}
}

@article{abramoff2013automated,  
  title={Automated analysis of retinal images for detection of referable diabetic retinopathy},  
  author={Abr{\`a}moff, Michael D and Folk, James C and Han, Dennis P and Walker, Jonathan D and Williams, David F and Russell, Stephen R and Massin, Pascale and Cochener, Beatrice and Gain, Philippe and Tang, Li and others},  
  journal={JAMA ophthalmology},  
  volume={131},  
  number={3},  
  pages={351--357},  
  year={2013},  
  publisher={American Medical Association}
}

@article{qian2024drac,
  title={DRAC 2022: A public benchmark for diabetic retinopathy analysis on ultra-wide optical coherence tomography angiography images},
  author={Qian, Bo and Chen, Hao and Wang, Xiangning and Guan, Zhouyu and Li, Tingyao and Jin, Yixiao and Wu, Yilan and Wen, Yang and Che, Haoxuan and Kwon, Gitaek and others},
  journal={Patterns},
  volume={5},
  number={3},
  year={2024},
  publisher={Elsevier}
}

@inproceedings{fang2022dataset,
  title={Dataset and evaluation algorithm design for goals challenge},
  author={Fang, Huihui and Li, Fei and Fu, Huazhu and Wu, Junde and Zhang, Xiulan and Xu, Yanwu},
  booktitle={International Workshop on Ophthalmic Medical Image Analysis},
  pages={135--142},
  year={2022},
  organization={Springer}
}

@article{fu2020age,
  title={Age challenge: angle closure glaucoma evaluation in anterior segment optical coherence tomography},
  author={Fu, Huazhu and Li, Fei and Sun, Xu and Cao, Xingxing and Liao, Jingan and Orlando, Jose Ignacio and Tao, Xing and Li, Yuexiang and Zhang, Shihao and Tan, Mingkui and others},
  journal={Medical Image Analysis},
  volume={66},
  pages={101798},
  year={2020},
  publisher={Elsevier}
}

@article{yang2021connectivity,
  title={Connectivity-based deep learning approach for segmentation of the epithelium in in vivo human esophageal OCT images},
  author={Yang, Ziyun and Soltanian-Zadeh, Somayyeh and Chu, Kengyeh K and Zhang, Haoran and Moussa, Lama and Watts, Ariel E and Shaheen, Nicholas J and Wax, Adam and Farsiu, Sina},
  journal={Biomedical optics express},
  volume={12},
  number={10},
  pages={6326--6340},
  year={2021},
  publisher={Optical Society of America}
}

@article{estrada2014tree,
  title={Tree topology estimation},
  author={Estrada, Rolando and Tomasi, Carlo and Schmidler, Scott C and Farsiu, Sina},
  journal={IEEE transactions on pattern analysis and machine intelligence},
  volume={37},
  number={8},
  pages={1688--1701},
  year={2014},
  publisher={IEEE}
}

@article{estrada2015retinal,
  title={Retinal artery-vein classification via topology estimation},
  author={Estrada, Rolando and Allingham, Michael J and Mettu, Priyatham S and Cousins, Scott W and Tomasi, Carlo and Farsiu, Sina},
  journal={IEEE transactions on medical imaging},
  volume={34},
  number={12},
  pages={2518--2534},
  year={2015},
  publisher={IEEE}
}

@article{rabbani2015fully,
  title={Fully automatic segmentation of fluorescein leakage in subjects with diabetic macular edema},
  author={Rabbani, Hossein and Allingham, Michael J and Mettu, Priyatham S and Cousins, Scott W and Farsiu, Sina},
  journal={Investigative ophthalmology \& visual science},
  volume={56},
  number={3},
  pages={1482--1492},
  year={2015},
  publisher={The Association for Research in Vision and Ophthalmology}
}

@article{fang2012sparsity,
  title={Sparsity based denoising of spectral domain optical coherence tomography images},
  author={Fang, Leyuan and Li, Shutao and Nie, Qing and Izatt, Joseph A and Toth, Cynthia A and Farsiu, Sina},
  journal={Biomedical optics express},
  volume={3},
  number={5},
  pages={927--942},
  year={2012},
  publisher={Optical Society of America}
}

@article{gholami2020octid,
  title={OCTID: Optical coherence tomography image database},
  author={Gholami, Peyman and Roy, Priyanka and Parthasarathy, Mohana Kuppuswamy and Lakshminarayanan, Vasudevan},
  journal={Computers \& Electrical Engineering},
  volume={81},
  pages={106532},
  year={2020},
  publisher={Elsevier}
}

@article{fang2022refuge2, title={Refuge2 challenge: A treasure trove for multi-dimension analysis and evaluation in glaucoma screening}, author={Fang, Huihui and Li, Fei and Wu, Junde and Fu, Huazhu and Sun, Xu and Son, Jaemin and Yu, Shuang and Zhang, Menglu and Yuan, Chenglang and Bian, Cheng and others}, journal={arXiv preprint arXiv:2202.08994
        
        }, year={2022}}

@article{odstrcilik2013retinal,
  title={Retinal vessel segmentation by improved matched filtering: evaluation on a new high-resolution fundus image database},
  author={Odstrcilik, Jan and Kolar, Radim and Budai, Attila and Hornegger, Joachim and Jan, Jiri and Gazarek, Jiri and Kubena, Tomas and Cernosek, Pavel and Svoboda, Ondrej and Angelopoulou, Elli},
  journal={IET Image Processing},
  volume={7},
  number={4},
  pages={373--383},
  year={2013},
  publisher={Wiley Online Library}
}

@article{budai2013robust, title={Robust vessel segmentation in fundus images}, author={Budai, Attila and Bock, R{\"u}diger and Maier, Andreas and Hornegger, Joachim and Michelson, Georg}, journal={International journal of biomedical imaging}, volume={2013}, number={1}, pages={154860}, year={2013}, publisher={Wiley Online Library}}

@article{zhang2025predicting, title={Predicting Diabetic Macular Edema Treatment Responses Using OCT: Dataset and Methods of APTOS Competition}, author={Zhang, Weiyi and Chotcomwongse, Peranut and Li, Yinwen and Xu, Pusheng and Yao, Ruijie and Zhou, Lianhao and Zhou, Yuxuan and Feng, Hui and Zhou, Qiping and Wang, Xinyue and others}, journal={arXiv preprint arXiv:2505.05768
        
        
        
        }, year={2025}}

@article{jiang2018predicting, title={Predicting the progression of ophthalmic disease based on slit-lamp images using a deep temporal sequence network}, author={Jiang, Jiewei and Liu, Xiyang and Liu, Lin and Wang, Shuai and Long, Erping and Yang, Haoqing and Yuan, Fuqiang and Yu, Deying and Zhang, Kai and Wang, Liming and others}, journal={PloS one}, volume={13}, number={7}, pages={e0201142}, year={2018}, publisher={Public Library of Science San Francisco, CA USA}}

@inproceedings{almazroa2018retinal, title={Retinal fundus images for glaucoma analysis: the RIGA dataset}, author={Almazroa, Ahmed and Alodhayb, Sami and Osman, Essameldin and Ramadan, Eslam and Hummadi, Mohammed and Dlaim, Mohammed and Alkatee, Muhannad and Raahemifar, Kaamran and Lakshminarayanan, Vasudevan}, booktitle={Medical Imaging 2018: Imaging Informatics for Healthcare, Research, and Applications}, volume={10579}, pages={55--62}, year={2018}, organization={SPIE}}

@article{liu2019self, title={A self-adaptive deep learning method for automated eye laterality detection based on color fundus photography}, author={Liu, Chi and Han, Xiaotong and Li, Zhixi and Ha, Jason and Peng, Guankai and Meng, Wei and He, Mingguang}, journal={Plos one}, volume={14}, number={9}, pages={e0222025}, year={2019}, publisher={Public Library of Science San Francisco, CA USA}}

@misc{jr2ngb_cataractdataset, author = {jr2ngb}, title = {Cataract Dataset}, year = {2019}, url = {https://www.kaggle.com/datasets/jr2ngb/cataractdataset}}

@article{nguyen2013automated,
  title={An automated method for retinal arteriovenous nicking quantification from color fundus images},
  author={Nguyen, Uyen TV and Bhuiyan, Alauddin and Park, Laurence AF and Kawasaki, Ryo and Wong, Tien Y and Wang, Jie Jin and Mitchell, Paul and Ramamohanarao, Kotagiri},
  journal={IEEE Transactions on Biomedical Engineering},
  volume={60},
  number={11},
  pages={3194--3203},
  year={2013},
  publisher={IEEE}
}

@inproceedings{perez2011vampire,
  title={{VAMPIRE}: Vessel assessment and measurement platform for images of the REtina},
  author={Perez-Rovira, Adria and MacGillivray, T and Trucco, Emanuele and Chin, KS and Zutis, K and Lupascu, C and Tegolo, Domenico and Giachetti, Andrea and Wilson, Peter J and Doney, A and others},
  booktitle={2011 Annual International Conference of the IEEE Engineering in Medicine and Biology Society},
  pages={3391--3394},
  year={2011},
  organization={IEEE}
}

@article{chiu2013automatic, title={Automatic cone photoreceptor segmentation using graph theory and dynamic programming}, author={Chiu, Stephanie J et al.}, journal={Biomedical optics express}, volume={4}, number={6}, pages={924--937}, year={2013}}

@article{diaz2019cnns, title={CNNs for automatic glaucoma assessment using fundus images: an extensive validation}, author={Diaz-Pinto, Andres et al.}, journal={Biomedical engineering online}, volume={18}, number={1}, pages={29}, year={2019}}

@misc{sshikamaru_glaucoma_detection, author = {sshikamaru}, title = {Glaucoma Detection}, year = {2022}, url = {https://www.kaggle.com/datasets/sshikamaru/glaucoma-detection}}

@article{batista2020rim, title={Rim-one dl: A unified retinal image database for assessing glaucoma using deep learning}, author={Batista, Francisco Jos{\'e} Fumero et al.}, journal={Image Analysis and Stereology}, volume={39}, number={3}, pages={161--167}, year={2020}}

@inproceedings{hu2013automated, title={Automated separation of binary overlapping trees in low-contrast color retinal images}, author={Hu, Qiao et al.}, booktitle={International conference on medical image computing and computer-assisted intervention}, pages={436--443}, year={2013}}

@article{ruckert2024rocov2,
  title={Rocov2: Radiology objects in context version 2, an updated multimodal image dataset},
  author={R{\"u}ckert, Johannes and Bloch, Louise and Br{\"u}ngel, Raphael and Idrissi-Yaghir, Ahmad and Sch{\"a}fer, Henning and Schmidt, Cynthia S and Koitka, Sven and Pelka, Obioma and Abacha, Asma Ben and G. Seco de Herrera, Alba and others},
  journal={Scientific Data},
  volume={11},
  number={1},
  pages={688},
  year={2024},
  publisher={Nature Publishing Group UK London}
}

@article{yang2023medmnist, title={Medmnist v2-a large-scale lightweight benchmark for 2d and 3d biomedical image classification}, author={Yang, Jiancheng et al.}, journal={Scientific Data}, volume={10}, number={1}, pages={41}, year={2023}}

@misc{tianchi2021retinafundus, title={Retina Fundus Image Registration}, author={Tianchi}, year={2021}, url={https://tianchi.aliyun.com/dataset/dataDetail?dataId=90112}}

@inproceedings{li2023automated, title={Automated detection of myopic maculopathy in MMAC 2023: achievements in classification, segmentation, and spherical equivalent prediction}, author={Li, Yihao and Zhang, Philippe and Tan, Yubo and Zhang, Jing and Wang, Zhihan and Jiang, Weili and Conze, Pierre-Henri and Lamard, Mathieu and Quellec, Gwenol{\'e} and El Habib Daho, Mostafa}, booktitle={International Conference on Medical Image Computing and Computer-Assisted Intervention}, pages={1--17}, year={2023}, organization={Springer}}

@article{CARDOZO2023109056, title={Dataset of fundus images for the diagnosis of ocular toxoplasmosis}, journal={Data in Brief}, volume={48}, pages={109056}, year={2023}, issn={2352-3409}, doi={https://doi.org/10.1016/j.dib.2023.109056}, url={https://www.sciencedirect.com/science/article/pii/S2352340923001749}, author={Olivia Cardozo and Verena Ojeda and Rodrigo Parra and Julio César Mello-Román and José Luis {Vázquez Noguera} and Miguel García-Torres and Federico Divina and Sebastian A. Grillo and Cynthia Villalba and Jacques Facon and Veronica Elisa {Castillo Benítez} and Ingrid {Castro Matto} and Diego Aquino-Brítez}}

@inproceedings{abbasi2015biologically,
  title={Biologically-inspired supervised vasculature segmentation in SLO retinal fundus images},
  author={Abbasi-Sureshjani, Samaneh and Smit-Ockeloen, Iris and Zhang, Jiong and Ter Haar Romeny, Bart},
  booktitle={International Conference Image Analysis and Recognition},
  pages={325--334},
  year={2015},
  organization={Springer}
}

@article{kovalyk2022papila,
  title={PAPILA: Dataset with fundus images and clinical data of both eyes of the same patient for glaucoma assessment},
  author={Kovalyk, Oleksandr and Morales-S{\'a}nchez, Juan and Verd{\'u}-Monedero, Rafael and Sell{\'e}s-Navarro, Inmaculada and Palaz{\'o}n-Cabanes, Ana and Sancho-G{\'o}mez, Jos{\'e}-Luis},
  journal={Scientific Data},
  volume={9},
  number={1},
  pages={291},
  year={2022},
  publisher={Nature Publishing Group UK London}
}

@article{kansal2025multiple,
  title={Multiple model visual feature embedding and selection method for an efficient ocular disease classification},
  author={Kansal, Isha and Khullar, Vikas and Sharma, Preeti and Singh, Supreet and Hamid, Junainah Abd and Santhosh, A Johnson},
  journal={Scientific Reports},
  volume={15},
  number={1},
  pages={5157},
  year={2025},
  publisher={Nature Publishing Group UK London}
}

@article{cen2021automatic,
  title={Automatic detection of 39 fundus diseases and conditions in retinal photographs using deep neural networks},
  author={Cen, Ling-Ping and Ji, Jie and Lin, Jian-Wei and Ju, Si-Tong and Lin, Hong-Jie and Li, Tai-Ping and Wang, Yun and Yang, Jian-Feng and Liu, Yu-Fen and Tan, Shaoying and others},
  journal={Nature communications},
  volume={12},
  number={1},
  pages={4828},
  year={2021},
  publisher={Nature Publishing Group UK London}
}

@article{jin2022fives, title={Fives: A fundus image dataset for artificial Intelligence based vessel segmentation}, author={Jin, Kai and Huang, Xingru and Zhou, Jingxing and Li, Yunxiang and Yan, Yan and Sun, Yibao and Zhang, Qianni and Wang, Yaqi and Ye, Juan}, journal={Scientific Data}, volume={9}, number={1}, pages={475}, year={2022}, publisher={Nature Publishing Group UK London}}

@article{benitez2021dataset, title={Dataset from fundus images for the study of diabetic retinopathy}, author={Ben{\'\i}tez, Veronica Elisa Castillo and Matto, Ingrid Castro and Rom{\'a}n, Julio C{\'e}sar Mello and Noguera, Jos{\'e} Luis V{\'a}zquez and Garc{\'\i}a-Torres, Miguel and Ayala, Jordan and Pinto-Roa, Diego P and Gardel-Sotomayor, Pedro E and Facon, Jacques and Grillo, Sebastian Alberto}, journal={Data in brief}, volume={36}, pages={107068}, year={2021}, publisher={Elsevier}}

@inproceedings{kauppi2007diaretdb1, title={The diaretdb1 diabetic retinopathy database and evaluation protocol.}, author={Kauppi, Tomi and Kalesnykiene, Valentina and Kamarainen, Joni-Kristian and Lensu, Lasse and Sorri, Iiris and Raninen, Asta and Voutilainen, Raija and Uusitalo, Hannu}, booktitle={BMVC}, volume={1}, number={1}, pages={10}, year={2007}}

@article{de2023airogs, title={Airogs: Artificial intelligence for robust glaucoma screening challenge}, author={De Vente, Coen et al.}, journal={IEEE transactions on medical imaging}, volume={43}, number={1}, pages={542--557}, year={2023}}

@article{liu2022deepdrid, title={Deepdrid: Diabetic retinopathy—grading and image quality estimation challenge}, author={Liu, Ruhan and Wang, Xiangning and Wu, Qiang and Dai, Ling and Fang, Xi and Yan, Tao and Son, Jaemin and Tang, Shiqi and Li, Jiang and Gao, Zijian and others}, journal={Patterns}, volume={3}, number={6}, year={2022}, publisher={Elsevier}}

@article{pachade2021retinal,  
  title={Retinal fundus multi-disease image dataset (rfmid): A dataset for multi-disease detection research},  
  author={Pachade, Samiksha and Porwal, Prasanna and Thulkar, Dhanshree and Kokare, Manesh and Deshmukh, Girish and Sahasrabuddhe, Vivek and Giancardo, Luca and Quellec, Gwenol{\'e} and M{\'e}riaudeau, Fabrice},  
  journal={Data},  
  volume={6},  
  number={2},  
  pages={14},  
  year={2021},  
  publisher={MDPI}
}

@article{porwal2018indian,  
  title={Indian diabetic retinopathy image dataset (IDRiD): a database for diabetic retinopathy screening research},  
  author={Porwal, Prasanna and Pachade, Samiksha and Kamble, Ravi and Kokare, Manesh and Deshmukh, Girish and Sahasrabuddhe, Vivek and Meriaudeau, Fabrice},  
  journal={Data},  
  volume={3},  
  number={3},  
  pages={25},  
  year={2018},  
  publisher={MDPI}
}

@article{staal2004ridge,  
  title={Ridge-based vessel segmentation in color images of the retina},  
  author={Staal, Joes and Abr{\`a}moff, Michael D and Niemeijer, Meindert and Viergever, Max A and Van Ginneken, Bram},  
  journal={IEEE transactions on medical imaging},  
  volume={23},  
  number={4},  
  pages={501--509},  
  year={2004},  
  publisher={IEEE}
}

@dataset{CHASEDB1,
author={Fraz, Muhammad Moazam and Remagnino, Paolo and Hoppe, Andreas and Uyyanonvara, Bunyarit and Rudnicka, Alicja R. and Owen, Christopher G. and Barman, Sarah A.},
title={CHASE DB1: Retinal Vessel Reference Dataset},
year={2012},
url={https://researchdata.kingston.ac.uk/96/}

}

@inproceedings{sivaswamy2014drishti,
title={Drishti-gs: Retinal image dataset for optic nerve head (onh) segmentation},
author={Sivaswamy, Jayanthi and Krishnadas, SR and Joshi, Gopal Datt and Jain, Madhulika and Tabish, A Ujjwaft Syed},
booktitle={2014 IEEE 11th international symposium on biomedical imaging (ISBI)},
pages={53--56},
year={2014},
organization={IEEE}
}

@misc{tcia-apollo5,
  author       = {{The Cancer Imaging Archive (TCIA)}},
  title        = {APOLLO-5-DA-RAD},
  howpublished = {\url{https://www.cancerimagingarchive.net/tcia-downloads/apollo-5-da-rad/}},
  year         = {2025},
  note         = {Accessed 2025-08-21; ISSN: 2474-4638; TCIA Site License (CC BY-NC-ND)}
}

@misc{cancerimagingarchive2022cmblca,
  author       = {Cancer Moonshot Biobank},
  title        = {Cancer Moonshot Biobank – Lung Cancer Collection (CMB-LCA) (Version 9)},
  howpublished = {\url{https://www.cancerimagingarchive.net/collection/cmb-lca/}},
  year         = {2025},
  note         = {DOI:10.7937/3CX3-S132; CC BY 4.0; accessed 2025-08-21}
}

@misc{cancermoonshot2022cmbcrc,
  author       = {Cancer Moonshot Biobank},
  title        = {Cancer Moonshot Biobank – Colorectal Cancer Collection (CMB-CRC) (Version 8)},
  howpublished = {\url{https://www.cancerimagingarchive.net/collection/cmb-crc/}},
  year         = {2022},
  note         = {DOI:10.7937/djg7-gz87; accessed 2025-08-21}
}

@misc{cancermoonshot2022cmbmel,
  author       = {Cancer Moonshot Biobank},
  title        = {Cancer Moonshot Biobank – Melanoma Collection (CMB-MEL) (Version 9)},
  howpublished = {\url{https://www.cancerimagingarchive.net/collection/cmb-mel/}},
  year         = {2022},
  note         = {DOI:10.7937/gwsp-wh72; accessed 2025-08-21}
}

@inproceedings{garcia2015imageclef,
  title={Overview of the ImageCLEF 2015 medical classification task},
  author={Garcia Seco De Herrera, Alba and M{\"u}ller, Henning and Bromuri, Stefano},
  booktitle={Working Notes of CLEF 2015--Cross Language Evaluation Forum, CEUR},
  volume={1391},
  year={2015},
  organization={CEUR Workshop Proceedings}
}

@article{mei2022radimagenet,
  title={RadImageNet: an open radiologic deep learning research dataset for effective transfer learning},
  author={Mei, Xueyan and Liu, Zelong and Robson, Philip M and Marinelli, Brett and Huang, Mingqian and Doshi, Amish and Jacobi, Adam and Cao, Chendi and Link, Katherine E and Yang, Thomas and others},
  journal={Radiology: Artificial Intelligence},
  volume={4},
  number={5},
  pages={e210315},
  year={2022},
  publisher={Radiological Society of North America}
}

@misc{ehrlich2021aren0534,
  author       = {Ehrlich, P. and Chi, Y. Y. and Chintagumpala, M. M. and Hoffer, F. A. and Perlman, E. J. and Kalapurakal, J. A. and Warwick, A. and Shamberger, R. C. and Khanna, G. and Hamilton, T. E. and Gow, K. W. and Paulino, A. C. and Gratias, E. J. and Mullen, E. A. and Geller, J. I. and Grundy, P. E. and Fernandez, C. V. and Ritchey, M. L. and Dome, J. S.},
  title        = {Combination Chemotherapy and Surgery in Treating Young Patients With Wilms Tumor (AREN0534) [Data set]},
  howpublished = {\url{https://doi.org/10.7937/TCIA.5M9S-6Y97}},
  year         = {2021},
  note         = {The Cancer Imaging Archive (TCIA); DOI: 10.7937/TCIA.5M9S-6Y97. Accessed 2025-08-21.}
}

@misc{fernandez2022aren0532,
  author       = {Fernandez, C. V. and Mullen, E. A. and Chi, Y.-Y. and Ehrlich, P. F. and
                  Perlman, E. J. and Kalapurakal, J. A. and Khanna, G. and Paulino, A. C. and
                  Hamilton, T. E. and Gow, K. W. and Tochner, Z. and Hoffer, F. A. and
                  Withycombe, J. S. and Shamberger, R. C. and Kim, Y. and Geller, J. I. and
                  Anderson, J. R. and Grundy, P. E. and Dome, J. S.},
  title        = {Vincristine, Dactinomycin, and Doxorubicin With or Without Radiation Therapy or Observation Only in Treating Younger Patients Who Are Undergoing Surgery for Newly Diagnosed Stage I, Stage II, or Stage III Wilms' Tumor (AREN0532) (Version 1) [Data set]},
  howpublished = {\url{https://doi.org/10.7937/6PJ1-M859}},
  year         = {2022},
  note         = {The Cancer Imaging Archive (TCIA); Version 1; DOI: 10.7937/6PJ1-M859; Accessed 2025-08-21}
}

@article{de2015clust15,
  title={The 2014 liver ultrasound tracking benchmark},
  author={De Luca, Valeria and Benz, Tobias and Kondo, Satoshi and K{\"o}nig, Lars and L{\"u}bke, D and Rothl{\"u}bbers, Sven and Somphone, Oudom and Allaire, St{\'e}phane and Bell, MA Lediju and Chung, DYF and others},
  journal={Physics in Medicine \& Biology},
  volume={60},
  number={14},
  pages={5571},
  year={2015},
  publisher={IOP Publishing}
}

@misc{anna2016uns,
    author = {Anna Montoya and Hasnin and kaggle446 and shirzad and Will Cukierski and yffud},
    title = {Ultrasound Nerve Segmentation},
    year = {2016},
    howpublished = {\url{https://kaggle.com/competitions/ultrasound-nerve-segmentation}},
    note = {Kaggle}
}

@misc{zhou2020thyroid,
  title        = {Thyroid Nodule Segmentation and Classification in Ultrasound Images},
  author       = {Zhou, Jianqiao and Jia, Xiaohong and Ni, Dong and Noble, Alison and Huang, Ruobing and Tan, Tao and Van, Manh The},
  year         = 2020,
  month        = mar,
  publisher    = {Zenodo},
  doi          = {10.5281/zenodo.3715942},
  url          = {https://doi.org/10.5281/zenodo.3715942}
}

@article{lu2022psfhs,
title = {The JNU-IFM dataset for segmenting pubic symphysis-fetal head},
journal = {Data in Brief},
volume = {41},
pages = {107904},
year = {2022},
issn = {2352-3409},
doi = {https://doi.org/10.1016/j.dib.2022.107904},
url = {https://www.sciencedirect.com/science/article/pii/S2352340922001160},
author = {Yaosheng Lu and Mengqiang Zhou and Dengjiang Zhi and Minghong Zhou and Xiaosong Jiang and Ruiyu Qiu and Zhanhong Ou and Huijin Wang and Di Qiu and Mei Zhong and Xiaoxing Lu and Gaowen Chen and Jieyun Bai}
}

@misc{guo2023usenhance,
  author       = {Yi Guo and
                  Shichong Zhou and
                  Jun Shi and
                  Yuanyuan Wang},
  title        = {Ultrasound Image Enhancement challenge 2023},
  month        = apr,
  year         = 2023,
  publisher    = {Zenodo},
  doi          = {10.5281/zenodo.7841250},
  url          = {https://doi.org/10.5281/zenodo.7841250},
}

@inproceedings{gong2021multitask,  
  author={Gong, Haifan and Chen, Guanqi and Wang, Ranran and Xie, Xiang and Mao, Mingzhi and Yu, Yizhou and Chen, Fei and Li, Guanbin},  
  booktitle={2021 IEEE 18th International Symposium on Biomedical Imaging (ISBI)}, 
  title={Multi-Task Learning For Thyroid Nodule Segmentation With Thyroid Region Prior},   
  year={2021}, 
  pages={257-261},  
  doi={10.1109/ISBI48211.2021.9434087}
}

@ARTICLE{sarah2019camus,
  author={Leclerc, Sarah and Smistad, Erik and Pedrosa, João and Østvik, Andreas and Cervenansky, Frederic and Espinosa, Florian and Espeland, Torvald and Berg, Erik Andreas Rye and Jodoin, Pierre-Marc and Grenier, Thomas and Lartizien, Carole and D’hooge, Jan and Lovstakken, Lasse and Bernard, Olivier},
  journal={IEEE Transactions on Medical Imaging}, 
  title={Deep Learning for Segmentation Using an Open Large-Scale Dataset in 2D Echocardiography}, 
  year={2019},
  volume={38},
  number={9},
  pages={2198-2210},
  doi={10.1109/TMI.2019.2900516}
}

@article{kermany2018pneumnist,
  title={Identifying medical diagnoses and treatable diseases by image-based deep learning},
  author={Kermany, Daniel S and Goldbaum, Michael and Cai, Wenjia and Valentim, Carolina CS and Liang, Huiying and Baxter, Sally L and McKeown, Alex and Yang, Ge and Wu, Xiaokang and Yan, Fangbing and others},
  journal={Cell},
  volume={172},
  number={5},
  pages={1122--1131},
  year={2018}
}

@misc{praveen2019coronahack,
  author       = {{Praveen Govi}},
  title        = {CoronaHack - Chest X-Ray-Dataset},
  howpublished = {\url{https://www.kaggle.com/datasets/praveengovi/coronahack-chest-xraydataset}},
  year         = {2019},
  note         = {Kaggle dataset (uploader: praveengovi). Accessed 2025-08-21}
}

@inproceedings{wang2017chestxray8,
  title={Chestx-ray8: Hospital-scale chest x-ray database and benchmarks on weakly-supervised classification and localization of common thorax diseases},
  author={Wang, Xiaosong and Peng, Yifan and Lu, Le and Lu, Zhiyong and Bagheri, Mohammadhadi and Summers, Ronald M},
  booktitle={Proceedings of the IEEE Conference on Computer Vision and Pattern Recognition},
  pages={2097--2106},
  year={2017}
}

@Article{wang2020covidx,
    author={Wang, Linda and Lin, Zhong Qiu and Wong, Alexander},
    title={COVID-Net: a tailored deep convolutional neural network design for detection of COVID-19 cases from chest X-ray images},
    journal={Scientific Reports},
    year={2020},
    month={Nov},
    day={11},
    volume={10},
    number={1},
    pages={19549},
    issn={2045-2322},
    doi={10.1038/s41598-020-76550-z},
    url={https://doi.org/10.1038/s41598-020-76550-z}
}

@misc{anna2019siimacrpneumothorax,
    author = {Anna Zawacki and Carol Wu and George Shih and Julia Elliott and Mikhail Fomitchev and Mohannad Hussain and ParasLakhani and Phil Culliton and Shunxing Bao},
    title = {SIIM-ACR Pneumothorax Segmentation},
    year = {2019},
    howpublished = {\url{https://kaggle.com/competitions/siim-acr-pneumothorax-segmentation}},
    note = {Kaggle}
}

@misc{raddar2020irma,
  author       = {{Raddar}},
  title        = {{IRMA} X-ray dataset},
  howpublished = {\url{https://www.kaggle.com/datasets/raddar/irma-xray-dataset}},
  year         = {2020},
  note         = {Kaggle dataset (uploader: raddar); contains ~14{,}000 X-ray images; used in ImageCLEF medical annotation tasks. Accessed 2025-08-21.}
}

@misc{moulay2021chestxrcovid19,
                author = {Akhloufi, Moulay A. and Chetoui, Mohamed},
                title = {{Chest XR COVID-19 detection}},  
                howpublished = {\url{https://cxr-covid19.grand-challenge.org/}},
                month = {August},
                year = {2021},
                note = {Online; accessed September 2021},
}

@article{cohen2020covid,
  title={COVID-19 image data collection},
  author={Joseph Paul Cohen and Paul Morrison and Lan Dao},
  journal={arXiv preprint arXiv:2003.11597},
  url={https://github.com/ieee8023/covid-chestxray-dataset},
  year={2020}
}

@misc{asraf2021covid19,
  author       = {Asraf, Amanullah and Islam, Zabirul},
  title        = {{COVID19}, Pneumonia and Normal Chest X-ray PA Dataset},
  howpublished = {\url{https://data.mendeley.com/datasets/jctsfj2sfn/1}},
  year         = {2021},
  month        = apr,
  day          = {9},
  note         = {Mendeley Data (V1); doi:10.17632/jctsfj2sfn.1; CC BY 4.0; Accessed 2025-08-21}
}

@inproceedings{long2003nhanes,
  title={Biomedical information from a national collection of spine x-rays: film to content-based retrieval},
  author={Long, L Rodney and Antani, Sameer and Lee, Dah-Jye and Krainak, Daniel M and Thoma, George R},
  booktitle={Medical Imaging 2003: PACS and Integrated Medical Information Systems: Design and Evaluation},
  volume={5033},
  pages={70--84},
  year={2003},
  organization={SPIE}
}

@article{hirvasniemi2023knee,
  title={The KNee OsteoArthritis Prediction (KNOAP2020) challenge: An image analysis challenge to predict incident symptomatic radiographic knee osteoarthritis from MRI and X-ray images},
  author={Hirvasniemi, Jukka and Runhaar, Jos and van der Heijden, Rianne A and Zokaeinikoo, Maryam and Yang, Mingrui and Li, Xiaojuan and Tan, Jimin and Rajamohan, Haresh Rengaraj and Zhou, Yuyue and Deniz, Cem M and others},
  journal={Osteoarthritis and Cartilage},
  volume={31},
  number={1},
  pages={115--125},
  year={2023},
  publisher={Elsevier}
}

@article{wang2021aasce,
title = {Evaluation and comparison of accurate automated spinal curvature estimation algorithms with spinal anterior-posterior X-Ray images: The {AASCE}2019 challenge},
journal = {Medical Image Analysis},
volume = {72},
pages = {102115},
year = {2021},
issn = {1361-8415},
doi = {https://doi.org/10.1016/j.media.2021.102115},
url = {https://www.sciencedirect.com/science/article/pii/S1361841521001614},
author = {Liansheng Wang and Cong Xie and Yi Lin and Hong-Yu Zhou and Kailin Chen and Dalong Cheng and Florian Dubost and Benjamin Collery and Bidur Khanal and Bishesh Khanal and Rong Tao and Shangliang Xu and Upasana {Upadhyay Bharadwaj} and Zhusi Zhong and Jie Li and Shuxin Wang and Shuo Li}
}

@misc{pranav2020covid19,
  author       = {{Pranav Raikote (pranavraikokte)}},
  title        = {COVID-19 Image Dataset: 3 Way Classification - COVID-19, Viral Pneumonia, Normal},
  howpublished = {\url{https://www.kaggle.com/datasets/pranavraikokte/covid19-image-dataset}},
  year         = {2020},
  note         = {Kaggle dataset (uploader: pranavraikokte); contains COVID-19, viral pneumonia, and normal chest X-ray images. Accessed 2025-08-21.}
}

@article{jaeger2014two,
  title={Two public chest X-ray datasets for computer-aided screening of pulmonary diseases},
  author={Jaeger, Stefan and Candemir, Sema and Antani, Sameer and W{\'a}ng, Y{\`\i}-Xi{\'a}ng J and Lu, Pu-Xuan and Thoma, George},
  journal={Quantitative imaging in medicine and surgery},
  volume={4},
  number={6},
  pages={475},
  year={2014}
}

@article{rajpurkar2017mura,
  title={{MURA}: Large dataset for abnormality detection in musculoskeletal radiographs},
  author={Rajpurkar, Pranav and Irvin, Jeremy and Bagul, Aarti and Ding, Daisy and Duan, Tony and Mehta, Hershel and Yang, Brandon and Zhu, Kaylie and Laird, Dillon and Ball, Robyn L and others},
  journal={arXiv preprint arXiv:1712.06957},
  year={2017}
}

@inproceedings{suckling1994mammographic,
  title={The mammographic images analysis society digital mammogram database},
  author={Suckling, John},
  booktitle={Exerpta Medica. International Congress Series, 1994},
  volume={1069},
  pages={375--378},
  year={1994}
}

@misc{stein2018rsnapneumoniadetection,
    author = {Anouk Stein, MD and Carol Wu and Chris Carr and George Shih and Jamie Dulkowski and kalpathy and Leon Chen and Luciano Prevedello and Marc Kohli, MD and Mark McDonald and Peter and Phil Culliton and Safwan Halabi MD and Tian Xia},
    title = {RSNA Pneumonia Detection Challenge},
    year = {2018},
    howpublished = {\url{https://kaggle.com/competitions/rsna-pneumonia-detection-challenge}},
    note = {Kaggle}
}

@misc{nguyen2020vinbigdata,
    author = {Duc Nguyen and DungNB and Ha Q. Nguyen and Julia Elliott and NguyenThanhNhan and Phil Culliton},
    title = {VinBigData Chest X-ray Abnormalities Detection},
    year = {2020},
    howpublished = {\url{https://kaggle.com/competitions/vinbigdata-chest-xray-abnormalities-detection}},
    note = {Kaggle}
}

@inproceedings{irvin2019chexpert,
  title={Chexpert: A large chest radiograph dataset with uncertainty labels and expert comparison},
  author={Irvin, Jeremy and Rajpurkar, Pranav and Ko, Michael and Yu, Yifan and Ciurea-Ilcus, Silviana and Chute, Chris and Marklund, Henrik and Haghgoo, Behzad and Ball, Robyn and Shpanskaya, Katie and others},
  booktitle={Proceedings of the AAAI Conference on Artificial Intelligence},
  volume={33},
  number={01},
  pages={590--597},
  year={2019}
}

@article{lakhani2023siimfisabio,
  title={The 2021 SIIM-FISABIO-RSNA machine learning COVID-19 challenge: Annotation and standard exam classification of COVID-19 chest radiographs},
  author={Lakhani, Paras and Mongan, John and Singhal, Chinmay and Zhou, Quan and Andriole, Katherine P and Auffermann, William F and Prasanna, PM and Pham, Theresa X and Peterson, Michael and Bergquist, Peter J and others},
  journal={Journal of Digital Imaging},
  volume={36},
  number={1},
  pages={365--372},
  year={2023},
  publisher={Springer}
}

@article{sogancioglu2024nodule,
  title={Nodule detection and generation on chest X-rays: NODE21 Challenge},
  author={Sogancioglu, Ecem and Van Ginneken, Bram and Behrendt, Finn and Bengs, Marcel and Schlaefer, Alexander and Radu, Miron and Xu, Di and Sheng, Ke and Scalzo, Fabien and Marcus, Eric and others},
  journal={IEEE Transactions on Medical Imaging},
  year={2024}
}

@inproceedings{deherrera2016imageclef,
  title = {Overview of the ImageCLEF 2016 Medical Task},
  author = {Alba Garc{\'i}a Seco de Herrera and Roger Schaer and Stefano Bromuri and Henning M{\"u}ller},
  booktitle = {CLEF 2016 Working Notes},
  year = {2016},
  publisher = {CEUR-WS.org},
  volume = {1609},
  pages = {219--232},
  url = {http://ceur-ws.org/Vol-1609/16090219.pdf}
}

@inproceedings{yang2021medmnist,
  title = {MedMNIST Classification Decathlon: A Lightweight AutoML Benchmark for Medical Image Analysis},
  author = {Yang, Jiancheng and Shi, Rui and Ni, Bingbing},
  booktitle = {IEEE 18th International Symposium on Biomedical Imaging (ISBI)},
  pages = {191--195},
  year = {2021},
  doi = {10.1109/ISBI48211.2021.9433967}
}

@inproceedings{zheng2016bone,
  title={Bone texture characterization for osteoporosis diagnosis using digital radiography},
  author={Zheng, Keni and Makrogiannis, Sokratis},
  booktitle={2016 38th Annual International Conference of the IEEE Engineering in Medicine and Biology Society (EMBC)},
  pages={1034--1037},
  year={2016},
  organization={IEEE}
}

@article{hogeweg2012clavicle,
title = {Clavicle segmentation in chest radiographs},
journal = {Medical Image Analysis},
volume = {16},
number = {8},
pages = {1490-1502},
year = {2012},
author = {Laurens Hogeweg and Clara I. Sánchez and Pim A. {de Jong} and Pragnya Maduskar and Bram {van Ginneken}},
}

@article{tabik2020covidgr,
  title={{COVIDGR} dataset and {COVID-SDNet} methodology for predicting COVID-19 based on chest X-ray images},
  author={Tabik, Siham and G{\'o}mez-R{\'\i}os, Anabel and Mart{\'\i}n-Rodr{\'\i}guez, Jos{\'e} Luis and Sevillano-Garc{\'\i}a, Iv{\'a}n and Rey-Area, Manuel and Charte, David and Guirado, Emilio and Su{\'a}rez, Juan-Luis and Luengo, Juli{\'a}n and Valero-Gonz{\'a}lez, MA and others},
  journal={IEEE Journal of Biomedical and Health Informatics},
  volume={24},
  number={12},
  pages={3595--3605},
  year={2020}
}

@article{lian2021chestxdet,
  title={A structure-aware relation network for thoracic diseases detection and segmentation},
  author={Lian, Jie and Liu, Jingyu and Zhang, Shu and Gao, Kai and Liu, Xiaoqing and Zhang, Dingwen and Yu, Yizhou},
  journal={IEEE Transactions on Medical Imaging},
  volume={40},
  number={8},
  pages={2042--2052},
  year={2021}
}

@misc{seah2020ranzcr,
    author = {Jarrel Seah and Jen and Maggie and Meng Law and Phil Culliton and Sarah Dowd},
    title = {{RANZCR CLiP} - Catheter and Line Position Challenge},
    year = {2020},
    howpublished = {\url{https://kaggle.com/competitions/ranzcr-clip-catheter-line-classification}},
    note = {Kaggle}
}

@dataset{singh2020cpcxr,
doi = {10.21227/x2r3-xk48},
url = {https://dx.doi.org/10.21227/x2r3-xk48},
author = {Narinder Singh Punn and Sonali Agarwal},
publisher = {IEEE Dataport},
title = {COVID-19 Posteroanterior Chest X-Ray fused (CPCXR) dataset},
year = {2020} 
}

@article{shiraishi2000jsrt,
  title={Development of a digital image database for chest radiographs with and without a lung nodule: receiver operating characteristic analysis of radiologists' detection of pulmonary nodules},
  author={Shiraishi, Junji and Katsuragawa, Shigehiko and Ikezoe, Junpei and Matsumoto, Tsuneo and Kobayashi, Takeshi and Komatsu, Ken-ichi and Matsui, Mitate and Fujita, Hiroshi and Kodera, Yoshie and Doi, Kunio},
  journal={American Journal of Roentgenology},
  volume={174},
  number={1},
  pages={71--74},
  year={2000}
}

@article{zunair2021synthesis,
title={Synthesis of COVID-19 chest X-rays using unpaired image-to-image translation},
author={Zunair, Hasib and Hamza, A Ben},
journal={Social network analysis and mining},
volume={11},
number={1},
pages={1--12},
year={2021},
publisher={Springer}
}

@ARTICLE{wang2015evaluation,
  author={Wang, Ching-Wei and Huang, Cheng-Ta and Hsieh, Meng-Che and Li, Chung-Hsing and Chang, Sheng-Wei and Li, Wei-Cheng and Vandaele, Rémy and Marée, Raphaël and Jodogne, Sébastien and Geurts, Pierre and Chen, Cheng and Zheng, Guoyan and Chu, Chengwen and Mirzaalian, Hengameh and Hamarneh, Ghassan and Vrtovec, Tomaž and Ibragimov, Bulat},
  journal={IEEE Transactions on Medical Imaging}, 
  title={Evaluation and Comparison of Anatomical Landmark Detection Methods for Cephalometric X-Ray Images: A Grand Challenge}, 
  year={2015},
  volume={34},
  number={9},
  pages={1890-1900},
  doi={10.1109/TMI.2015.2412951}
}

@misc{tsai2021midrcricord,
  author       = {Tsai, E. B. and Simpson, S. and Lungren, M. P. and Hershman, M. and
                  Roshkovan, L. and Colak, E. and Erickson, B. J. and Shih, G. and
                  Stein, A. and Kalpathy-Cramer, J. and Shen, J. and Hafez, M. A. F. and
                  John, S. and Rajiah, P. and Pogatchnik, B. P. and Mongan, J. T. and
                  Altinmakas, E. and Ranschaert, E. and Kitamura, F. C. and Topff, L. and
                  Moy, L. and Kanne, J. P. and Wu, C. C.},
  title        = {Data from Medical Imaging Data Resource Center (MIDRC) - RSNA International COVID Radiology Database (RICORD) Release 1c - Chest x-ray, Covid+ (MIDRC-RICORD-1C)},
  howpublished = {\url{https://www.cancerimagingarchive.net/collection/midrc-ricord-1c/}},
  year         = {2021},
  note         = {Version 1 (updated 2021-01-15); DOI: 10.7937/91ah-v663; The Cancer Imaging Archive (TCIA); License: CC BY-NC 4.0; Accessed 2025-08-21.}
}

@article{kermany2018identifying,
  title={Identifying medical diagnoses and treatable diseases by image-based deep learning},
  author={Kermany, Daniel S and Goldbaum, Michael and Cai, Wenjia and Valentim, Carolina CS and Liang, Huiying and Baxter, Sally L and McKeown, Alex and Yang, Ge and Wu, Xiaokang and Yan, Fangbing and others},
  journal={cell},
  volume={172},
  number={5},
  pages={1122--1131},
  year={2018},
  publisher={Elsevier}
}

@article{chowdhury2020can,
  title={Can AI help in screening viral and COVID-19 pneumonia?},
  author={Chowdhury, Muhammad EH and Rahman, Tawsifur and Khandakar, Amith and Mazhar, Rashid and Kadir, Muhammad Abdul and Mahbub, Zaid Bin and Islam, Khandakar Reajul and Khan, Muhammad Salman and Iqbal, Atif and Al Emadi, Nasser and others},
  journal={IEEE Access},
  volume={8},
  pages={132665--132676},
  year={2020},
  publisher={IEEE}
}

@article{hamamci2023dentex, title={DENTEX: An Abnormal Tooth Detection with Dental Enumeration and Diagnosis Benchmark for Panoramic X-rays}, author={Hamamci, Ibrahim Ethem and Er, Sezgin and Simsar, Enis and Yuksel, Atif Emre and Gultekin, Sadullah and Ozdemir, Serife Damla and Yang, Kaiyuan and Li, Hongwei Bran and Pati, Sarthak and Stadlinger, Bernd and others}, journal={arXiv preprint arXiv:2305.19112}, year={2023} }

@article{halabi2019rsnabone,
  title={The {RSNA} pediatric bone age machine learning challenge},
  author={Halabi, Safwan S and Prevedello, Luciano M and Kalpathy-Cramer, Jayashree and Mamonov, Artem B and Bilbily, Alexander and Cicero, Mark and Pan, Ian and Pereira, Lucas Ara{\'u}jo and Sousa, Rafael Teixeira and Abdala, Nitamar and others},
  journal={Radiology},
  volume={290},
  number={2},
  pages={498--503},
  year={2019}
}

@misc{cao2023cephalometric,
  author       = {Jun Cao and
                  Juan Dai and
                  Xuguang Li and
                  Bingsheng Huang and
                  Ching-Wei Wang and
                  Hongyuan Zhang},
  title        = {Cephalometric Landmark Detection in Lateral X-ray Images},
  month        = apr,
  year         = 2023,
  publisher    = {Zenodo},
  doi          = {10.5281/zenodo.7835592},
  url          = {https://doi.org/10.5281/zenodo.7835592},
}

@article{khalid2022cepha29,
  title={Cepha29: Automatic cephalometric landmark detection challenge 2023},
  author={Khalid, Muhammad Anwaar and Zulfiqar, Kanwal and Bashir, Ulfat and Shaheen, Areeba and Iqbal, Rida and Rizwan, Zarnab and Rizwan, Ghina and Fraz, Muhammad Moazam},
  journal={arXiv preprint arXiv:2212.04808},
  year={2022}
}

@article{popov2024arcade,
  title={Dataset for automatic region-based coronary artery disease diagnostics using X-ray angiography images},
  author={Popov, Maxim and Amanturdieva, Akmaral and Zhaksylyk, Nuren and Alkanov, Alsabir and Saniyazbekov, Adilbek and Aimyshev, Temirgali and Ismailov, Eldar and Bulegenov, Ablay and Kuzhukeyev, Arystan and Kulanbayeva, Aizhan and others},
  journal={Scientific data},
  volume={11},
  number={1},
  pages={20},
  year={2024},
  publisher={Nature Publishing Group UK London}
}

@article{wang2023real,
  title={A real-world dataset and benchmark for foundation model adaptation in medical image classification},
  author={Wang, Dequan and Wang, Xiaosong and Wang, Lilong and Li, Mengzhang and Da, Qian and Liu, Xiaoqiang and Gao, Xiangyu and Shen, Jun and He, Junjun and Shen, Tian and others},
  journal={Scientific Data},
  volume={10},
  number={1},
  pages={574},
  year={2023},
  publisher={Nature Publishing Group UK London}
}

@inproceedings{ccimen2017coronare,
  title={{CoronARe}: A coronary artery reconstruction challenge},
  author={{\c{C}}imen, Serkan and Unberath, Mathias and Frangi, Alejandro and Maier, Andreas},
  booktitle={International Workshop on Computational Methods for Molecular Imaging},
  pages={96--104},
  year={2017},
  organization={Springer}
}

@misc{badano2019victre,
  author       = {Badano, A. and Graff, C. G. and Badal, A. and Sharma, D. and
                  Zeng, R. and Samuelson, F. W. and Glick, S. and Myers, K. J.},
  title        = {The VICTRE Trial: Open-Source, In-Silico Clinical Trial for Evaluating Digital Breast Tomosynthesis [Data set]},
  howpublished = {\url{https://www.cancerimagingarchive.net/collection/victre/}},
  year         = {2019},
  doi          = {10.7937/TCIA.2019.ho23nxaw},
  note         = {The Cancer Imaging Archive (TCIA); CC BY 3.0; Accessed 2025-08-21}
}

@misc{chen2018knee,
  title={Knee osteoarthritis severity grading dataset},
  author={Chen, Pingjun},
  howpublished={\url{https://data.mendeley.com/datasets/56rmx5bjcr/1}},
  journal={Mendeley Data},
  doi={10.17632/56rmx5bjcr.1},
  year={2018},
  month=sep,
  day={4},
  note={Mendeley Data (V1); doi:10.17632/56rmx5bjcr.1; CC BY 4.0}
}

@misc{kelly2022ahod0831,
  author       = {Kelly, K. M. and Cole, P. D. and Pei, Q. and Bush, R. and
                  Roberts, K. B. and Hodgson, D. C. and McCarten, K. M. and
                  Cho, S. Y. and Schwartz, C.},
  title        = {Combination Chemotherapy and Radiation Therapy in Treating Young Patients With Newly Diagnosed Hodgkin Lymphoma (AHOD0831) (Version 1) [Data set]},
  howpublished = {\url{https://www.cancerimagingarchive.net/collection/ahod0831/}},
  year         = {2022},
  doi          = {10.7937/CV5M-1H59},
  note         = {The Cancer Imaging Archive (TCIA); Version 1; Accessed 2025-08-21}
}

@misc{gaggion2023chexmask,
  author       = {Gaggion, Nicolas and Mosquera, Candelaria and Aineseder, Martina and Mansilla, Lucas and Milone, Diego and Ferrante, Enzo},
  title        = {CheXmask Database: a large-scale dataset of anatomical segmentation masks for chest x-ray images (version 0.1) [Data set]},
  howpublished = {\url{https://physionet.org/content/chexmask-cxr-segmentation-data/0.1/}},
  year         = {2023},
  month        = jun,
  day          = {27},
  doi          = {10.13026/dx54-8351},
  note         = {PhysioNet; License: CC BY-NC-SA 4.0; Accessed 2025-08-21}
}

@misc{baidu2021ruschn,
  title     = {X-ray Hand Joint Classification Dataset [Data set]},
  author    = {{Baidu AI Studio}},
  year      = {2021},
  publisher = {Baidu AI Studio},
  url       = {https://aistudio.baidu.com/datasetdetail/69582/0},
  note      = {Accessed: 2025-05-22}
}

@article{lin2025cxrlt,
  title={CXR-LT 2024: A MICCAI challenge on long-tailed, multi-label, and zero-shot disease classification from chest X-ray},
  author={Lin, Mingquan and Holste, Gregory and Wang, Song and Zhou, Yiliang and Wei, Yishu and Banerjee, Imon and Chen, Pengyi and Dai, Tianjie and Du, Yuexi and Dvornek, Nicha C and others},
  journal={Medical Image Analysis},
  pages={103739},
  year={2025},
  publisher={Elsevier}
}

@article{liu2025preoperative,
  title = {Preoperative fracture reduction planning for image-guided pelvic trauma surgery: A comprehensive pipeline with learning},
  journal = {Medical Image Analysis},
  volume = {102},
  pages = {103506},
  year = {2025},
  issn = {1361-8415},
  doi = {https://doi.org/10.1016/j.media.2025.103506},
  url = {https://www.sciencedirect.com/science/article/pii/S1361841525000544},
  author = {Yanzhen Liu and Sutuke Yibulayimu and Yudi Sang and Gang Zhu and Chao Shi and Chendi Liang and Qiyong Cao and Chunpeng Zhao and Xinbao Wu and Yu Wang}
}

@article{liu2025automatic,
  title={Automatic pelvic fracture segmentation: a deep learning approach and benchmark dataset},
  author={Liu, Yanzhen and Yibulayimu, Sutuke and Zhu, Gang and Shi, Chao and Liang, Chendi and Zhao, Chunpeng and Wu, Xinbao and Sang, Yudi and Wang, Yu},
  journal={Frontiers in Medicine},
  volume={12},
  pages={1511487},
  year={2025},
  publisher={Frontiers Media SA}
}

@article{hatamizadeh2022ravir,
  title={RAVIR: A Dataset and Methodology for the Semantic Segmentation and Quantitative Analysis of Retinal Arteries and Veins in Infrared Reflectance Imaging},
  author={Hatamizadeh, Ali and Xu, Yufan and Terzopoulos, Demetri and others},
  journal={arXiv preprint arXiv:2203.04041},
  year={2022}
}

@misc{mrl_eyedataset,
  title={MRL Eye Dataset},
  author={Sojka, Eduard and others},
  howpublished={\url{http://mrl.cs.vsb.cz/eyedataset}},
  year={2018}
}

@misc{intel_mobileodt_kaggle_2017,
  title   = {Intel \& MobileODT Cervical Cancer Screening},
  author  = {{Kaggle}},
  year    = {2017},
  howpublished = {\url{https://www.kaggle.com/competitions/intel-mobileodt-cervical-cancer-screening}},
  note    = {Kaggle competition dataset}
}

@misc{addi_alz_2021,
  title   = {{ADDI} Alzheimer's Detection Challenge},
  author  = {{AIcrowd} and {Alzheimer's Disease Data Initiative}},
  year    = {2021},
  howpublished = {\url{https://www.aicrowd.com/challenges/addi-alzheimers-detection-challenge}},
  note    = {Challenge and dataset page}
}

@article{schaffter2020dream,
  title   = {Evaluation of Combined Artificial Intelligence and Radiologist Assessment to Interpret Screening Mammograms},
  author  = {Schaffter, Thomas and others},
  journal = {JAMA Network Open},
  year    = {2020},
  volume  = {3},
  number  = {9},
  pages   = {e2011872},
  url     = {https://pubmed.ncbi.nlm.nih.gov/32997146/}
}

@article{peng2015bigneuron,
  title   = {BigNeuron: Large-Scale 3D Neuron Reconstruction from Optical Microscopy Images},
  author  = {Peng, Hanchuan and others},
  journal = {Neuron},
  year    = {2015},
  volume  = {87},
  number  = {2},
  pages   = {252--256},
  url     = {https://www.ncbi.nlm.nih.gov/pmc/articles/PMC4683726
        
        /}
}

@article{khan2023radar,
  title   = {Grand Challenge on Human Activity Classification with Radar: Datasets, Methods, and Results},
  author  = {Khan, R. H. and others},
  journal = {IEEE Journal of Biomedical and Health Informatics},
  year    = {2023},
  url     = {https://ieeexplore.ieee.org/document/10134905}
}

@inproceedings{gamper2019pannuke,
  title={Pannuke: an open pan-cancer histology dataset for nuclei instance segmentation and classification},
  author={Gamper, Jevgenij and Alemi Koohbanani, Navid and Benet, Ksenija and Khuram, Ali and Rajpoot, Nasir},
  booktitle={European congress on digital pathology},
  pages={11--19},
  year={2019},
  organization={Springer}
}

@article{orlov2010automatic,
  title={Automatic classification of lymphoma images with transform-based global features},
  author={Orlov, Nikita V and Chen, Wayne W and Eckley, David Mark and Macura, Tomasz J and Shamir, Lior and Jaffe, Elaine S and Goldberg, Ilya G},
  journal={IEEE Transactions on Information Technology in Biomedicine},
  volume={14},
  number={4},
  pages={1003--1013},
  year={2010},
  publisher={IEEE}
}

@misc{paip2021_challenge,
  title = {PAIP 2021 Challenge: Perineural Invasion in Multiple Organ Cancer},
  howpublished = {\url{https://paip2021.grand-challenge.org/}},
  note = {Accessed: 2025-08-21},
  year = {2021},
}

@inproceedings{gelasca2008evaluation,
  title={Evaluation and benchmark for biological image segmentation},
  author={Gelasca, Elisa Drelie and Byun, Jiyun and Obara, Boguslaw and Manjunath, BS},
  booktitle={2008 15th IEEE international conference on image processing},
  pages={1816--1819},
  year={2008},
  organization={IEEE}
}

@misc{histopathologic_cancer_detection_kaggle,
  title = {Histopathologic Cancer Detection dataset},
  howpublished = {\url{https://www.kaggle.com/competitions/histopathologic-cancer-detection}},
  year={2018}
}

@misc{hubmap-kidney-segmentation,
    author = {Addison Howard and Andy Lawrence and Bud Sims and Eddie Tinsley and Jarek Kazmierczak and Katy Borner and Leah Godwin and Marcos Novaes and Phil Culliton and Richard Holland and Rick Watson and Yingnan Ju},
    title = {HuBMAP - Hacking the Kidney},
    year = {2020},
    howpublished = {\url{https://kaggle.com/competitions/hubmap-kidney-segmentation}},
    note = {Kaggle}
}

@article{li2020deep,
  title={Deep learning methods for lung cancer segmentation in whole-slide histopathology images—the acdc@ lunghp challenge 2019},
  author={Li, Zhang and Zhang, Jiehua and Tan, Tao and Teng, Xichao and Sun, Xiaoliang and Zhao, Hong and Liu, Lihong and Xiao, Yang and Lee, Byungjae and Li, Yilong and others},
  journal={IEEE Journal of Biomedical and Health Informatics},
  volume={25},
  number={2},
  pages={429--440},
  year={2020},
  publisher={IEEE}
}

@article{gupta10segpc,
  title={Segpc-2021: A challenge \& dataset on segmentation of multiple myeloma plasma cells from microscopic images},
  author={Gupta, Anubha and Gehlot, Shiv and Goswami, Shubham and Motwani, Sachin and Gupta, Ritu and Faura, {\'A}lvaro Garc{\'\i}a and {\v{S}}tepec, Dejan and Martin{\v{c}}i{\v{c}}, Toma{\v{z}} and Azad, Reza and Merhof, Dorit and others},
  journal={Medical Image Analysis},
  volume={83},
  pages={102677},
  year={2023},
  publisher={Elsevier}
}

@article{aubreville2023mitosis,
  title={Mitosis domain generalization in histopathology images—the MIDOG challenge},
  author={Aubreville, Marc and Stathonikos, Nikolas and Bertram, Christof A and Klopfleisch, Robert and Ter Hoeve, Natalie and Ciompi, Francesco and Wilm, Frauke and Marzahl, Christian and Donovan, Taryn A and Maier, Andreas and others},
  journal={Medical Image Analysis},
  volume={84},
  pages={102699},
  year={2023},
  publisher={Elsevier}
}

@article{ma2024multimodality,
  title={The multimodality cell segmentation challenge: toward universal solutions},
  author={Ma, Jun and Xie, Ronald and Ayyadhury, Shamini and Ge, Cheng and Gupta, Anubha and Gupta, Ritu and Gu, Song and Zhang, Yao and Lee, Gihun and Kim, Joonkee and others},
  journal={Nature methods},
  volume={21},
  number={6},
  pages={1103--1113},
  year={2024},
  publisher={Nature Publishing Group US New York}
}

@dataset{vanrijthoven2022tiger_roi,
  author       = {van Rijthoven, Mart and Aswolinskiy, Witali and Tessier, Leslie and Balkenhol, Maschenka and Bogaerts, Joep and van der Laak, Jeroen and Salgado, Roberto and Ciompi, Francesco},
  title        = {TIGER Training Dataset (WSIROIS subset) with ROI-level annotations},
  publisher    = {Zenodo},
  year         = {2022},
  doi          = {10.5281/zenodo.6014422},
  url          = {https://zenodo.org/records/6014422},
  note         = {Includes WSIROIS, WSIBULK, WSITILS subsets; WSIBULK contains whole-slide images with tumor bulk annotations compatible with segmentation pipelines}
}

@inproceedings{liu2022bci,
  title={Bci: Breast cancer immunohistochemical image generation through pyramid pix2pix},
  author={Liu, Shengjie and Zhu, Chuang and Xu, Feng and Jia, Xinyu and Shi, Zhongyue and Jin, Mulan},
  booktitle={Proceedings of the IEEE/CVF conference on computer vision and pattern recognition},
  pages={1815--1824},
  year={2022}
}

@article{xu2021predicting,
  title={Predicting axillary lymph node metastasis in early breast cancer using deep learning on primary tumor biopsy slides},
  author={Xu, Feng and Zhu, Chuang and Tang, Wenqi and Wang, Ying and Zhang, Yu and Li, Jie and Jiang, Hongchuan and Shi, Zhongyue and Liu, Jun and Jin, Mulan},
  journal={Frontiers in oncology},
  volume={11},
  pages={759007},
  year={2021},
  publisher={Frontiers Media SA}
}

@article{han2022wsss4luad,
  title={Wsss4luad: Grand challenge on weakly-supervised tissue semantic segmentation for lung adenocarcinoma},
  author={Han, Chu and Pan, Xipeng and Yan, Lixu and Lin, Huan and Li, Bingbing and Yao, Su and Lv, Shanshan and Shi, Zhenwei and Mai, Jinhai and Lin, Jiatai and others},
  journal={arXiv preprint arXiv:2204.06455},
  year={2022}
}

@article{amgad2019structured,
  title={Structured crowdsourcing enables convolutional segmentation of histology images},
  author={Amgad, Mohamed and Elfandy, Habiba and Hussein, Hagar and Atteya, Lamees A and Elsebaie, Mai AT and Abo Elnasr, Lamia S and Sakr, Rokia A and Salem, Hazem SE and Ismail, Ahmed F and Saad, Anas M and others},
  journal={Bioinformatics},
  volume={35},
  number={18},
  pages={3461--3467},
  year={2019},
  publisher={Oxford University Press}
}

@article{amgad2022nucls,
  title={NuCLS: A scalable crowdsourcing approach and dataset for nucleus classification and segmentation in breast cancer},
  author={Amgad, Mohamed and Atteya, Lamees A and Hussein, Hagar and Mohammed, Kareem Hosny and Hafiz, Ehab and Elsebaie, Maha AT and Alhusseiny, Ahmed M and AlMoslemany, Mohamed Atef and Elmatboly, Abdelmagid M and Pappalardo, Philip A and others},
  journal={GigaScience},
  volume={11},
  pages={giac037},
  year={2022},
  publisher={Oxford University Press}
}

@article{conde2022herohe,
  title={HEROHE challenge: predicting HER2 status in breast cancer from hematoxylin--eosin whole-slide imaging},
  author={Conde-Sousa, Eduardo and Vale, Jo{\~a}o and Feng, Ming and Xu, Kele and Wang, Yin and Della Mea, Vincenzo and La Barbera, David and Montahaei, Ehsan and Baghshah, Mahdieh and Turzynski, Andreas and others},
  journal={Journal of Imaging},
  volume={8},
  number={8},
  pages={213},
  year={2022},
  publisher={MDPI}
}

@article{jiao2023lysto,
  title={LYSTO: The lymphocyte assessment hackathon and benchmark dataset},
  author={Jiao, Yiping and Van Der Laak, Jeroen and Albarqouni, Shadi and Li, Zhang and Tan, Tao and Bhalerao, Abhir and Cheng, Shenghua and Ma, Jiabo and Pocock, Johnathan and Pluim, Josien PW and others},
  journal={IEEE journal of biomedical and health informatics},
  volume={28},
  number={3},
  pages={1161--1172},
  year={2023},
  publisher={IEEE}
}

@article{swiderska2019learning,
  title={Learning to detect lymphocytes in immunohistochemistry with deep learning},
  author={Swiderska-Chadaj, Zaneta and Pinckaers, Hans and Van Rijthoven, Mart and Balkenhol, Maschenka and Melnikova, Margarita and Geessink, Oscar and Manson, Quirine and Sherman, Mark and Polonia, Antonio and Parry, Jeremy and others},
  journal={Medical image analysis},
  volume={58},
  pages={101547},
  year={2019},
  publisher={Elsevier}
}

@article{sirinukunwattana2017gland,
  title={Gland segmentation in colon histology images: The glas challenge contest},
  author={Sirinukunwattana, Korsuk and Pluim, Josien PW and Chen, Hao and Qi, Xiaojuan and Heng, Pheng-Ann and Guo, Yun Bo and Wang, Li Yang and Matuszewski, Bogdan J and Bruni, Elia and Sanchez, Urko and others},
  journal={Medical image analysis},
  volume={35},
  pages={489--502},
  year={2017},
  publisher={Elsevier}
}

@article{graham2019hover,
  title={Hover-net: Simultaneous segmentation and classification of nuclei in multi-tissue histology images},
  author={Graham, Simon and Vu, Quoc Dang and Raza, Shan E Ahmed and Azam, Ayesha and Tsang, Yee Wah and Kwak, Jin Tae and Rajpoot, Nasir},
  journal={Medical image analysis},
  volume={58},
  pages={101563},
  year={2019},
  publisher={Elsevier}
}

@inproceedings{veeling2018rotation,
  title={Rotation equivariant CNNs for digital pathology},
  author={Veeling, Bastiaan S and Linmans, Jasper and Winkens, Jim and Cohen, Taco and Welling, Max},
  booktitle={International Conference on Medical image computing and computer-assisted intervention},
  pages={210--218},
  year={2018},
  organization={Springer}
}

@article{borkowski2019lung,
  title={Lung and colon cancer histopathological image dataset (lc25000)},
  author={Borkowski, Andrew A and Bui, Marilyn M and Thomas, L Brannon and Wilson, Catherine P and DeLand, Lauren A and Mastorides, Stephen M},
  journal={arXiv preprint arXiv:1912.12142},
  year={2019}
}

@article{spanhol2015dataset,
  title={A dataset for breast cancer histopathological image classification},
  author={Spanhol, Fabio A and Oliveira, Luiz S and Petitjean, Caroline and Heutte, Laurent},
  journal={Ieee transactions on biomedical engineering},
  volume={63},
  number={7},
  pages={1455--1462},
  year={2015},
  publisher={IEEE}
}

@article{silva2020going,
  title={Going deeper through the gleason scoring scale: An automatic end-to-end system for histology prostate grading and cribriform pattern detection},
  author={Silva-Rodr{\'\i}guez, Julio and Colomer, Adri{\'a}n and Sales, Mar{\'\i}a A and Molina, Rafael and Naranjo, Valery},
  journal={Computer methods and programs in biomedicine},
  volume={195},
  pages={105637},
  year={2020},
  publisher={Elsevier}
}

@article{chen2022self,
  title={Self-supervised vision transformers learn visual concepts in histopathology},
  author={Chen, Richard J and Krishnan, Rahul G},
  journal={arXiv preprint arXiv:2203.00585},
  year={2022}
}

@inproceedings{abousamra2021multi,
title={Multi-class cell detection using spatial context representation},
author={Abousamra, Shahira and Belinsky, David and Van Arnam, John and Allard, Felicia and Yee, Eric and Gupta, Rajarsi and Kurc, Tahsin and Samaras, Dimitris and Saltz, Joel and Chen, Chao},
booktitle={Proceedings of the IEEE/CVF International Conference on Computer Vision},
pages={4005--4014},
year={2021}
}

@article{leuschner2021lodopab-ct,
  author = {Leuschner, Johannes and Schmidt, Maximilian and Otero Baguer, Daniel and Maass, Peter},
  title = {{LoDoPaB-CT}, a benchmark dataset for low-dose computed tomography reconstruction},
  journal = {Scientific Data},
  year = {2021},
  volume = {8},
  pages = {109},
  doi = {10.1038/s41597-021-00893-z},
  url = {https://www.nature.com/articles/s41597-021-00893-z}
}

@article{ruikar20215k+,
  title = {5K+ CT Images on Fractured Limbs: A Dataset for Medical Imaging Research},
  author = {Ruikar, Darshan D. and Santosh, K. C. and Hegadi, Ravindra S. and Rupnar, Lakhan and Choudhary, Vivek A.},
  journal = {Journal of Medical Systems},
  volume = {45},
  number = {4},
  pages = {51},
  year = {2021},
  doi = {10.1007/s10916-021-01724-9}
}

@article{kohli2018creation,
  title = {Creation and Curation of the Society of Imaging Informatics in Medicine Hackathon Dataset},
  author = {Kohli, Marc and Morrison, James J. and Wawira, Judy and Morgan, Matthew B. and Hostetter, Jason and Genereaux, Brad and Hussain, Mohannad and Langer, Steve G.},
  journal = {Journal of Digital Imaging},
  volume = {31},
  number = {1},
  pages = {9--12},
  year = {2018},
  doi = {10.1007/s10278-017-0003-5}
}

@misc{nationallungscreeningtrialresearchteam2013data,
  author = {{National Lung Screening Trial Research Team}},
  title = {Data from the National Lung Screening Trial (NLST)},
  year = {2013},
  doi = {10.7937/TCIA.HMQ8-J677},
  howpublished = {The Cancer Imaging Archive (TCIA), https://doi.org/10.7937/TCIA.HMQ8-J677},
  note = {Accessed: 2025-08-22}
}

@article{flanders2020construction,
  title = {Construction of a Machine Learning Dataset through Collaboration: The RSNA 2019 Brain CT Hemorrhage Challenge},
  author = {Adam E. Flanders and Luciano M. Prevedello and George Shih and Safwan S. Halabi and Jayashree Kalpathy-Cramer and Robyn Ball and John T. Mongan and Anouk Stein and Felipe C. Kitamura and Matthew P. Lungren and Gagandeep Choudhary and Lesley Cala and Luiz Coelho and Monique Mogensen and Fanny Mor{\'o}n and Elka Miller and Ichiro Ikuta and Vahe Zohrabian and Olivia McDonnell and Christie Lincoln and Lubdha Shah and David Joyner and Amit Agarwal and Ryan K. Lee and Jaya Nath and {RSNA-ASNR 2019 Brain Hemorrhage CT Annotators}},
  journal = {Radiology: Artificial Intelligence},
  volume = {2},
  number = {3},
  pages = {e190211},
  year = {2020},
  doi = {10.1148/ryai.2020190211},
  url = {https://doi.org/10.1148/ryai.2020190211}
}

@article{yang2020covid-ct-dataset,
  title = {COVID-CT-Dataset: A CT Scan Dataset about COVID-19},
  author = {Yang, Xingyi and He, Xuehai and Zhao, Jinyu and Zhang, Yichen and Zhang, Shanghang and Xie, Pengtao},
  journal = {arXiv preprint arXiv:2003.13865},
  year = {2020},
  eprint = {2003.13865},
  archivePrefix = {arXiv},
  primaryClass = {cs.LG},
  doi = {10.48550/arXiv.2003.13865},
  url = {https://arxiv.org/abs/2003.13865}
}

@misc{wjxiaochuangw2021covid-19-ct,
  title = {COVID-19-CT SCAN IMAGES},
  author = {{wjXiaoChuangw}},
  year = {2021},
  howpublished = {https://tianchi.aliyun.com/dataset/dataDetail?dataId=93666},
  publisher = {Alibaba Cloud Tianchi},
  note = {Dataset; accessed 2025-08-22}
}

@misc{sunneyi2025chest,
  title = {Chest CT-Scan images Dataset},
  author = {SunneYi},
  year = {2025},
  publisher = {Zenodo},
  doi = {10.5281/zenodo.14759927},
  howpublished = {https://tianchi.aliyun.com/dataset/93929},
  note = {Version v1}
}

@article{soares2020sars,
  title = {SARS-CoV-2 CT-scan dataset: A large dataset of real patients CT scans for SARS-CoV-2 identification},
  author = {Soares, Eduardo and Angelov, Plamen and Biaso, Sarah and Froes, Michele Higa and Abe, Daniel Kanda},
  journal = {medRxiv},
  year = {2020},
  doi = {10.1101/2020.04.24.20078584},
  url = {https://doi.org/10.1101/2020.04.24.20078584},
  publisher = {Cold Spring Harbor Laboratory Press}
}

@article{hssayeni2020intracranial,
  title = {Intracranial Hemorrhage Segmentation Using a Deep Convolutional Model},
  author = {Hssayeni, Murtadha D. and Croock, Muayad S. and Salman, Aymen D. and Al-khafaji, Hassan Falah and Yahya, Zakaria A. and Ghoraani, Behnaz},
  journal = {Data},
  volume = {5},
  number = {1},
  pages = {14},
  year = {2020},
  publisher = {MDPI},
  doi = {10.3390/data5010014},
  url = {https://doi.org/10.3390/data5010014}
}

@misc{cmbcrc2022,
  author = {Cancer Moonshot Biobank},
  title = {Cancer Moonshot Biobank – Colorectal Cancer Collection (CMB-CRC)},
  year = {2022},
  publisher = {The Cancer Imaging Archive},
  doi = {10.7937/DJG7-GZ87},
  url = {https://doi.org/10.7937/DJG7-GZ87},
  howpublished = {The Cancer Imaging Archive, https://doi.org/10.7937/DJG7-GZ87},
  note = {Dataset},
  version = {8}
}

@misc{cmbgec2022,
  author = {{Cancer Moonshot Biobank}},
  title = {Cancer Moonshot Biobank – Gastroesophageal Cancer Collection (CMB-GEC)},
  year = {2022},
  publisher = {The Cancer Imaging Archive},
  howpublished = {The Cancer Imaging Archive (TCIA), https://doi.org/10.7937/E7KH-R486},
  doi = {10.7937/E7KH-R486},
  note = {Version 6 [dataset]}
}

@misc{cmbmel2022,
  author = {Cancer Moonshot Biobank},
  title = {Cancer Moonshot Biobank - Melanoma Collection (CMB-MEL)},
  year = {2022},
  version = {1},
  note = {Data set},
  publisher = {The Cancer Imaging Archive},
  howpublished = {https://doi.org/10.7937/GWSP-WH72},
  doi = {10.7937/GWSP-WH72}
}

@misc{cmbmml2022,
  author = {{Cancer Moonshot Biobank}},
  title = {Cancer Moonshot Biobank – Multiple Myeloma Collection (CMB-MML)},
  year = {2022},
  publisher = {The Cancer Imaging Archive},
  howpublished = {https://www.cancerimagingarchive.net/collection/cmb-mml/},
  doi = {10.7937/SZKB-SW39},
  version = {8},
  note = {Dataset}
}

@misc{cmbpca2022,
  author = {{Cancer Moonshot Biobank}},
  title = {Cancer Moonshot Biobank – Prostate Cancer Collection (CMB-PCA)},
  year = {2022},
  publisher = {The Cancer Imaging Archive},
  howpublished = {https://www.cancerimagingarchive.net/collection/cmb-pca/},
  note = {Version 9},
  doi = {10.7937/25T7-6Y12}
}

@misc{mader2017finding,
  title = {Finding and Measuring Lungs in CT Data},
  author = {Mader, Kevin S.},
  year = {2017},
  howpublished = {Kaggle, URL: https://www.kaggle.com/datasets/kmader/finding-lungs-in-ct-data},
  note = {Accessed: 2025-08-22}
}

@misc{kitamura2019head,
  author = {Felipe Campos Kitamura},
  title = {Head CT - hemorrhage},
  year = {2019},
  publisher = {Kaggle},
  howpublished = {\url{https://www.kaggle.com/datasets/felipekitamura/head-ct-hemorrhage}},
  note = {Accessed: 2025-08-22}
}

@article{lee2025lowdose,
  title = {Low-dose computed tomography perceptual image quality assessment},
  author = {Wonkyeong Lee and Fabian Wagner and Adrian Galdran and Yongyi Shi and Wenjun Xia and Ge Wang and Xuanqin Mou and Md Atik Ahamed and Abdullah Al Zubaer Imran and Ji Eun Oh and Kyungsang Kim and Jong Tak Baek and Dongheon Lee and Boohwi Hong and Philip Tempelman and Donghang Lyu and Adrian Kuiper and Lars van Blokland and Maria Baldeon Calisto and Scott Hsieh and Minah Han and Jongduk Baek and Andreas Maier and Adam Wang and Garry Evan Gold and Jang-Hwan Choi},
  journal = {Medical Image Analysis},
  volume = {99},
  pages = {103343},
  year = {2025},
  doi = {10.1016/j.media.2024.103343}
}

@misc{apolloresearchnetwork2023applied,
  author = {{Applied Proteogenomics OrganizationaL Learning and Outcomes (APOLLO) Research Network}},
  title = {Applied Proteogenomics OrganizationaL Learning and Outcomes (APOLLO-5)},
  year = {2023},
  howpublished = {https://wiki.cancerimagingarchive.net/display/Public/APOLLO-5},
  publisher = {The Cancer Imaging Archive (TCIA)},
  note = {Limited access; accessed 2025-08-22}
}

@misc{cmblca2022,
  author = {{Cancer Moonshot Biobank}},
  title = {Cancer Moonshot Biobank – Lung Cancer Collection (CMB-LCA)},
  year = {2022},
  howpublished = {The Cancer Imaging Archive (TCIA), https://doi.org/10.7937/3CX3-S132},
  doi = {10.7937/3CX3-S132},
  url = {https://doi.org/10.7937/3CX3-S132},
  version = {9},
  note = {Dataset},
  publisher = {The Cancer Imaging Archive}
}

@misc{heywhale2020cardiac,
  title        = {Cardiac Atrial Images - Cardiac MRI Segmentation Dataset},
  author       = {{HeyWhale}},
  howpublished = {\url{https://www.heywhale.com/mw/dataset/5e4de9618ee624002d4c4117}},
  year         = {2020},
  note         = {Cardiac atrial MRI segmentation dataset with 8,000 images for cardiac disease analysis. License: CC BY 4.0. Accessed 2025-08-22}
}

@misc{nlm1994visible,
  author       = {{National Library of Medicine}},
  title        = {The Visible Human Project},
  howpublished = {\url{https://www.nlm.nih.gov/research/visible/visible_human.html}},
  year         = {1994},
  note         = {The creation of complete, anatomically detailed, three-dimensional representations of the normal male and female human bodies. Accessed 2025-08-22}
}

@misc{tianchi2022braimri,
  title        = {braimMRI - Brain MRI Segmentation Dataset},
  author       = {{Alibaba Tianchi}},
  howpublished = {\url{https://tianchi.aliyun.com/dataset/dataDetail?dataId=127459}},
  year         = {2022},
  note         = {Brain tumor MRI segmentation dataset with 110 images. License: CC BY-NC-SA. Accessed 2025-08-22}
}

@misc{tianchi2020brainmri,
  title        = {Brain-MRI - Brain Disease MRI Segmentation Dataset},
  author       = {{Alibaba Tianchi}},
  howpublished = {\url{https://tianchi.aliyun.com/dataset/127583}},
  year         = {2020},
  note         = {Brain disease MRI segmentation dataset using FLAIR sequences with 110 images. License: CC BY-NC-SA. Accessed 2025-08-22}
}

@misc{tianchi2020spinaldisease,
  title        = {SpinalDisease2020 - Spinal Disease MRI Detection Dataset},
  author       = {{Alibaba Tianchi}},
  howpublished = {\url{https://tianchi.aliyun.com/competition/entrance/531796/information}},
  year         = {2020},
  note         = {Spinal disease detection dataset using T1 and T2 MRI sequences with 150 images. License: CC BY-NC-SA. Accessed 2025-08-22}
}

@inproceedings{muller2020qubiq,
  title={The QUBIQ challenge: quantifying uncertainty in biomedical image segmentation},
  author={M{\"u}ller, Carole H and Gonzalez, Carole and Breininger, Katharina and Albarqouni, Shadi and Wachter, Emma and Agrawal, Pallavi and Auer, Dominik and Erdt, Marius and Chen, Hongwei and Miranda, Doreen and others},
  booktitle={Uncertainty for Safe Utilization of Machine Learning in Medical Imaging and Clinical Image-Based Procedures},
  pages={59--70},
  year={2020},
  publisher={Springer}
}

@article{qubiq2021uncertainty,
  title={Qubiq: Uncertainty quantification for biomedical image segmentation challenge},
  author={Li, Hongwei Bran and Navarro, Fernando and Ezhov, Ivan and Bayat, Amirhossein and Das, Dhritiman and Kofler, Florian and Shit, Suprosanna and Waldmannstetter, Diana and Paetzold, Johannes C and Hu, Xiaobin and others},
  journal={arXiv preprint arXiv:2405.18435},
  year={2024}
}

@article{kim2023paip,
  title={PAIP 2020: Microsatellite instability prediction in colorectal cancer},
  author={Kim, Kyungmo and Lee, Kyoungbun and Cho, Sungduk and Kang, Dong Un and Park, Seongkeun and Kang, Yunsook and Kim, Hyunjeong and Choe, Gheeyoung and Moon, Kyung Chul and Lee, Kyu Sang and others},
  journal={Medical Image Analysis},
  volume={89},
  pages={102886},
  year={2023},
  publisher={Elsevier}
}

@article{kumar2019multi,
  title={A multi-organ nucleus segmentation challenge},
  author={Kumar, Neeraj and Verma, Ruchika and Anand, Deepak and Zhou, Yanning and Onder, Omer Fahri and Tsougenis, Efstratios and Chen, Hao and Heng, Pheng-Ann and Li, Jiahui and Hu, Zhiqiang and others},
  journal={IEEE transactions on medical imaging},
  volume={39},
  number={5},
  pages={1380--1391},
  year={2019},
  publisher={IEEE}
}

@article{jantzen2005pap,
  title={Pap-smear benchmark data for pattern classification},
  author={Jantzen, Jan and Norup, Jonas and Dounias, Georgios and Bjerregaard, Beth},
  journal={Nature inspired smart information systems (NiSIS 2005)},
  pages={1--9},
  year={2005}
}

@article{graham2019mild,
  title={MILD-Net: Minimal information loss dilated network for gland instance segmentation in colon histology images},
  author={Graham, Simon and Chen, Hao and Gamper, Jevgenij and Dou, Qi and Heng, Pheng-Ann and Snead, David and Tsang, Yee Wah and Rajpoot, Nasir},
  journal={Medical image analysis},
  volume={52},
  pages={199--211},
  year={2019},
  publisher={Elsevier}
}

@article{zhu2021hard,
  title={Hard sample aware noise robust learning for histopathology image classification},
  author={Zhu, Chuang and Chen, Wenkai and Peng, Ting and Wang, Ying and Jin, Mulan},
  journal={IEEE transactions on medical imaging},
  volume={41},
  number={4},
  pages={881--894},
  year={2021},
  publisher={IEEE}
}

@article{bulten2022artificial,
  title={Artificial intelligence for diagnosis and Gleason grading of prostate cancer: the PANDA challenge},
  author={Bulten, Wouter and Kartasalo, Kimmo and Chen, Po-Hsuan Cameron and Str{\"o}m, Peter and Pinckaers, Hans and Nagpal, Kunal and Cai, Yuannan and Steiner, David F and Van Boven, Hester and Vink, Robert and others},
  journal={Nature medicine},
  volume={28},
  number={1},
  pages={154--163},
  year={2022},
  publisher={Nature Publishing Group US New York}
}

@article{wang2025atec23,
  title={Atec23 challenge: automated prediction of treatment effectiveness in ovarian cancer using histopathological images},
  author={Wang, Ching-Wei and Firdi, Nabila Puspita and Chu, Tzu-Chiao and Faiz, Mohammad Faiz Iqbal and Iqbal, Mohammad Zafar and Li, Yifan and Yang, Bo and Mallya, Mayur and Bashashati, Ali and Li, Fei and others},
  journal={Medical Image Analysis},
  volume={99},
  pages={103342},
  year={2025},
  publisher={Elsevier}
}

@article{weitz2024acrobat,
  title={The ACROBAT 2022 challenge: automatic registration of breast cancer tissue},
  author={Weitz, Philippe and Valkonen, Masi and Solorzano, Leslie and Carr, Circe and Kartasalo, Kimmo and Boissin, Constance and Koivukoski, Sonja and Kuusela, Aino and Rasic, Dusan and Feng, Yanbo and others},
  journal={Medical image analysis},
  volume={97},
  pages={103257},
  year={2024},
  publisher={Elsevier}
}

@article{shin2025ocelot,
  title={OCELOT 2023: Cell detection from cell--tissue interaction challenge},
  author={Shin, JaeWoong and Ryu, Jeongun and Puche, Aaron Valero and Lee, Jinhee and Brattoli, Biagio and Jung, Wonkyung and Cho, Soo Ick and Paeng, Kyunghyun and Ock, Chan-Young and Yoo, Donggeun and others},
  journal={Medical Image Analysis},
  volume={106},
  pages={103751},
  year={2025},
  publisher={Elsevier}
}

@article{asadi2024machine,
  title={Machine learning-driven histotype diagnosis of ovarian carcinoma: insights from the OCEAN AI challenge},
  author={Asadi-Aghbolaghi, Maryam and Farahani, Hossein and Zhang, Allen and Akbari, Ardalan and Kim, Sirim and Chow, Ashley and Dane, Sohier and OCEAN Challenge Consortium and OTTA Consortium and Huntsman, David G and others},
  journal={medRxiv},
  pages={2024--04},
  year={2024},
  publisher={Cold Spring Harbor Laboratory Press}
}

@article{vermorgen2024endometrial,
  title={Endometrial Pipelle biopsy computer-aided diagnosis: a feasibility study},
  author={Vermorgen, Sanne and Gelton, Thijs and Bult, Peter and Kusters-Vandevelde, Heidi VN and Hausnerov{\'a}, Jitka and Van de Vijver, Koen and Davidson, Ben and Stefansson, Ingunn Marie and Kooreman, Loes FS and Qerimi, Adelina and others},
  journal={Modern Pathology},
  volume={37},
  number={2},
  pages={100417},
  year={2024},
  publisher={Elsevier}
}

@article{hou2020dataset,
  title={Dataset of segmented nuclei in hematoxylin and eosin stained histopathology images of ten cancer types},
  author={Hou, Le and Gupta, Rajarsi and Van Arnam, John S and Zhang, Yuwei and Sivalenka, Kaustubh and Samaras, Dimitris and Kurc, Tahsin M and Saltz, Joel H},
  journal={Scientific data},
  volume={7},
  number={1},
  pages={185},
  year={2020},
  publisher={Nature Publishing Group UK London}
}

@article{huo2024comprehensive,
  title={A comprehensive AI model development framework for consistent Gleason grading},
  author={Huo, Xinmi and Ong, Kok Haur and Lau, Kah Weng and Gole, Laurent and Young, David M and Tan, Char Loo and Zhu, Xiaohui and Zhang, Chongchong and Zhang, Yonghui and Li, Longjie and others},
  journal={Communications Medicine},
  volume={4},
  number={1},
  pages={84},
  year={2024},
  publisher={Nature Publishing Group UK London}
}

@article{veta2019predicting,
  title={Predicting breast tumor proliferation from whole-slide images: the TUPAC16 challenge},
  author={Veta, Mitko and Heng, Yujing J and Stathonikos, Nikolas and Bejnordi, Babak Ehteshami and Beca, Francisco and Wollmann, Thomas and Rohr, Karl and Shah, Manan A and Wang, Dayong and Rousson, Mikael and others},
  journal={Medical image analysis},
  volume={54},
  pages={111--121},
  year={2019},
  publisher={Elsevier}
}

@dataset{CPTAC_AML,
  author       = {{National Cancer Institute Clinical Proteomic Tumor Analysis Consortium (CPTAC)}},
  year         = {2019},
  title        = {The Clinical Proteomic Tumor Analysis Consortium Acute Myeloid Leukemia Collection (CPTAC-AML) (Version 5)},
  publisher    = {The Cancer Imaging Archive},
  version      = {5},
  doi          = {10.7937/tcia.2019.b6foe619},
  url          = {https://doi.org/10.7937/tcia.2019.b6foe619},
  note         = {[dataset]}
}

@article{shephard2022tiager,
  title={Tiager: Tumor-infiltrating lymphocyte scoring in breast cancer for the tiger challenge},
  author={Shephard, Adam and Jahanifar, Mostafa and Wang, Ruoyu and Dawood, Muhammad and Graham, Simon and Sidlauskas, Kastytis and Khurram, Syed Ali and Rajpoot, Nasir and Raza, Shan E Ahmed},
  journal={arXiv preprint arXiv:2206.11943
        
        
        
        
        
        
        
        
        
        
        
        
        
        
        
        
        
        },
  year={2022}
}

@article{yang2019deep,
  title={Deep learning for smartphone-based malaria parasite detection in thick blood smears},
  author={Yang, Feng and Poostchi, Mahdieh and Yu, Hang and Zhou, Zhou and Silamut, Kamolrat and Yu, Jian and Maude, Richard J and Jaeger, Stefan and Antani, Sameer},
  journal={IEEE journal of biomedical and health informatics},
  volume={24},
  number={5},
  pages={1427--1438},
  year={2019},
  publisher={IEEE}
}

@article{larsen2014hep,
  title={HEp-2 cell classification using shape index histograms with donut-shaped spatial pooling},
  author={Larsen, Anders Boesen Lindbo and Vestergaard, Jacob Schack and Larsen, Rasmus},
  journal={IEEE transactions on medical imaging},
  volume={33},
  number={7},
  pages={1573--1580},
  year={2014},
  publisher={IEEE}
}

@misc{ikezogwo2023quilt1m,
      title={Quilt-1M: One Million Image-Text Pairs for Histopathology}, 
      author={Wisdom Oluchi Ikezogwo and Mehmet Saygin Seyfioglu and Fatemeh Ghezloo and Dylan Stefan Chan Geva and Fatwir Sheikh Mohammed and Pavan Kumar Anand and Ranjay Krishna and Linda Shapiro},
      year={2023},
      eprint={2306.11207},
      archivePrefix={arXiv},
      primaryClass={cs.CV}
}

@inproceedings{gong2021multi-task,  
  author={Gong, Haifan and Chen, Guanqi and Wang, Ranran and Xie, Xiang and Mao, Mingzhi and Yu, Yizhou and Chen, Fei and Li, Guanbin},  
  booktitle={2021 IEEE 18th International Symposium on Biomedical Imaging (ISBI)},   
  title={Multi-Task Learning For Thyroid Nodule Segmentation With Thyroid Region Prior},   
  year={2021}, 
  pages={257-261},  
  doi={10.1109/ISBI48211.2021.9434087}
}

@article{martel2019assessment,
  title={Assessment of residual breast cancer cellularity after neoadjuvant chemotherapy using digital pathology [data set]},
  author={Martel, AL and Nofech-Mozes, Sharon and Salama, Sherine and Akbar, Shazia and Peikari, Mohammad},
  journal={The Cancer Imaging Archive},
  year={2019}
}

@article{gupta2019mimm_sbilab,
  title={MiMM\_SBILab Dataset: Microscopic images of multiple myeloma},
  author={Gupta, Ritu and Gupta, Anubha},
  journal={(No Title)},
  year={2019},
  publisher={The Cancer Imaging Archive}
}

@article{akbar2019automated,
  title={Automated and manual quantification of tumour cellularity in digital slides for tumour burden assessment},
  author={Akbar, Shazia and Peikari, Mohammad and Salama, Sherine and Panah, Azadeh Yazdan and Nofech-Mozes, Sharon and Martel, Anne L},
  journal={Scientific reports},
  volume={9},
  number={1},
  pages={14099},
  year={2019},
  publisher={Nature Publishing Group UK London}
}

@article{Hosseini_2019,
   title={Encoding Visual Sensitivity by MaxPol Convolution Filters for Image Sharpness Assessment},
   volume={28},
   ISSN={1941-0042},
   url={http://dx.doi.org/10.1109/TIP.2019.2906582},
   DOI={10.1109/tip.2019.2906582},
   number={9},
   journal={IEEE Transactions on Image Processing},
   publisher={Institute of Electrical and Electronics Engineers (IEEE)},
   author={Hosseini, Mahdi S. and Zhang, Yueyang and Plataniotis, Konstantinos N.},
   year={2019},
   month=sep, pages={4510–4525} }

@inproceedings{Drelie08-298,
author = {Elisa Drelie Gelasca and Jiyun Byun and Boguslaw Obara and B.S. Manjunath},
title = {Evaluation and Benchmark for Biological Image Segmentation},
booktitle = {IEEE International Conference on Image Processing},
location = {San Diego, CA},
month = {Oct},
year = {2008},
url ={http://vision.ece.ucsb.edu/publications/elisa_ICIP08.pdf}}

@misc{kather2016textures,
  author       = {Kather, J. N. and Zöllner, F. G. and Bianconi, F. and Melchers, S. M. and Schad, L. R. and Gaiser, T. and Marx, A. and Weis, C.-A.},
  title        = {Collection of textures in colorectal cancer histology},
  year         = {2016},
  publisher    = {Zenodo},
  doi          = {10.5281/zenodo.53169},
  url          = {https://doi.org/10.5281/zenodo.53169}
}

@inproceedings{stirenko2018chest,
  title={Chest X-ray analysis of tuberculosis by deep learning with segmentation and augmentation},
  author={Stirenko, Sergii and Kochura, Yuriy and Alienin, Oleg and Rokovyi, Oleksandr and Gordienko, Yuri and Gang, Peng and Zeng, Wei},
  booktitle={2018 IEEE 38th International Conference on Electronics and Nanotechnology (ELNANO)},
  pages={422--428},
  year={2018},
  organization={IEEE}
}

@article{aresta2019bach,
  title={Bach: Grand challenge on breast cancer histology images},
  author={Aresta, Guilherme and Ara{\'u}jo, Teresa and Kwok, Scotty and Chennamsetty, Sai Saketh and Safwan, Mohammed and Alex, Varghese and Marami, Bahram and Prastawa, Marcel and Chan, Monica and Donovan, Michael and others},
  journal={Medical image analysis},
  volume={56},
  pages={122--139},
  year={2019},
  publisher={Elsevier}
}

@article{scarpa2011automatic,
  title={Automatic evaluation of corneal nerve tortuosity in images from in vivo confocal microscopy},
  author={Scarpa, Fabio and Zheng, Xiaodong and Ohashi, Yuichi and Ruggeri, Alfredo},
  journal={Investigative ophthalmology \& visual science},
  volume={52},
  number={9},
  pages={6404--6408},
  year={2011},
  publisher={The Association for Research in Vision and Ophthalmology}
}

@article{phoulady2018new,
  title={A new cervical cytology dataset for nucleus detection and image classification (Cervix93) and methods for cervical nucleus detection},
  author={Phoulady, Hady Ahmady and Mouton, Peter R},
  journal={arXiv preprint arXiv:1811.09651
        
        
        
        },
  year={2018}
}

@misc{vrabac2020dlbclmorph,
    title={DLBCL-Morph: Morphological features computed using deep learning for an annotated digital DLBCL image set},
    author={Damir Vrabac and Akshay Smit and Rebecca Rojansky and Yasodha Natkunam and Ranjana H. Advani and Andrew Y. Ng and Sebastian Fernandez-Pol and Pranav Rajpurkar},
    year={2020},
    eprint={2009.08123},
    archivePrefix={arXiv},
    primaryClass={cs.CV}
}

@article{teikari2016deep,
  title={Deep Learning Convolutional Networks for Multiphoton Microscopy Vasculature Segmentation},
  author={Teikari, Petteri and Santos, Marc and Poon, Charissa and Hynynen, Kullervo},
  journal={arXiv preprint arXiv:1606.02382
        
        },
  year={2016}
}

@article{li2021multi,
  title={Multi-stage malaria parasite recognition by deep learning},
  author={Li, Sen and Du, Zeyu and Meng, Xiangjie and Zhang, Yang},
  journal={GigaScience},
  volume={10},
  number={6},
  pages={giab040},
  year={2021},
  publisher={Oxford University Press}
}

@inproceedings{zhang2018poisson,
    title={A Poisson-Gaussian Denoising Dataset with Real Fluorescence Microscopy Images},
    author={Yide Zhang and Yinhao Zhu and Evan Nichols and Qingfei Wang and Siyuan Zhang and Cody Smith and Scott Howard},
    booktitle={CVPR},
    year={2019}
}

@article{rahman2020reliable,
  title={Reliable tuberculosis detection using chest X-ray with deep learning, segmentation and visualization},
  author={Rahman, Tawsifur and Khandakar, Amith and Kadir, Muhammad Abdul and Islam, Khandaker Rejaul and Islam, Khandakar F and Mazhar, Rashid and Hamid, Tahir and Islam, Mohammad Tariqul and Kashem, Saad and Mahbub, Zaid Bin and others},
  journal={Ieee Access},
  volume={8},
  pages={191586--191601},
  year={2020},
  publisher={IEEE}
}

@article{javadi2019novel,
  title={A novel deep learning method for automatic assessment of human sperm images},
  author={Javadi, Soroush and Mirroshandel, Seyed Abolghasem},
  journal={Computers in Biology and Medicine},
  volume={109},
  pages={182--194},
  year={2019},
  doi={10.1016/j.compbiomed.2019.04.030}
}

@InProceedings{GSB2016,

Title = {Overview of the {ImageCLEF} 2016 medical task}, Author = {Garc\'ia Seco de Herrera, Alba and Schaer, Roger and Bromuri, Stefano and M\"uller, Henning }, Booktitle = {Working Notes of {CLEF} 2016 (Cross Language Evaluation Forum)}, Year = {2016}, Month = {September}, Location = {\'Evora, Portugal}

}

@article{le2022analysis,
  title={Analysis of the human protein atlas weakly supervised single-cell classification competition},
  author={Le, Trang and Winsnes, Casper F and Axelsson, Ulrika and Xu, Hao and Mohanakrishnan Kaimal, Jayasankar and Mahdessian, Diana and Dai, Shubin and Makarov, Ilya S and Ostankovich, Vladislav and Xu, Yang and others},
  journal={Nature methods},
  volume={19},
  number={10},
  pages={1221--1229},
  year={2022},
  publisher={Nature Publishing Group US New York}
}

@article{nanni2016texture,
  title={Texture descriptors ensembles enable image-based classification of maturation of human stem cell-derived retinal pigmented epithelium},
  author={Nanni, Loris and Paci, Michelangelo and Caetano dos Santos, Florentino Luciano and Skottman, Heli and Juuti-Uusitalo, Kati and Hyttinen, Jari},
  journal={PLoS One},
  volume={11},
  number={2},
  pages={e0149399},
  year={2016},
  publisher={Public Library of Science San Francisco, CA USA}
}

@article{shaker2017dictionary, title={A dictionary learning approach for human sperm heads classification}, author={Shaker, Fariba and Monadjemi, S Amirhassan and Alirezaie, Javad and Naghsh-Nilchi, Ahmad Reza}, journal={Computers in biology and medicine}, volume={91}, pages={181--190}, year={2017}, publisher={Elsevier} }

@article{mavska2023cell,
  title={The cell tracking challenge: 10 years of objective benchmarking},
  author={Ma{\v{s}}ka, Martin and Ulman, Vladim{\'\i}r and Delgado-Rodriguez, Pablo and G{\'o}mez-de-Mariscal, Estibaliz and Ne{\v{c}}asov{\'a}, Tereza and Guerrero Pe{\~n}a, Fidel A and Ren, Tsang Ing and Meyerowitz, Elliot M and Scherr, Tim and L{\"o}ffler, Katharina and others},
  journal={Nature Methods},
  volume={20},
  number={7},
  pages={1010--1020},
  year={2023},
  publisher={Nature Publishing Group US New York}
}

@article{alam2019machine,
  title={Machine learning approach of automatic identification and counting of blood cells},
  author={Alam, Mohammad Mahmudul and Islam, Mohammad Tariqul},
  journal={Healthcare Technology Letters},
  volume={6},
  number={4},
  pages={103--108},
  year={2019},
  publisher={IET}
}

@misc{BloodCellDetection2022,
  author       = {{Heywhale}},
  title        = {Blood Cell Detection Dataset},
  howpublished = {\url{https://www.heywhale.com/mw/dataset/62c2af90913a54a66038165a}},
  year         = {2022},
  note         = {Dataset for blood cell object detection}
}

@inproceedings{wieczorek2024transformer,
  title={Transformer Based Semantic Segmentation Network for Medical Imaging Application},
  author={Wieczorek, Micha{\l} and Si{\l}ka, Jakub and Wiltos, Katarzyna and Wo{\'z}niak, Marcin},
  booktitle={International Conference on Artificial Intelligence and Soft Computing},
  pages={380--389},
  year={2024},
  organization={Springer}
}

@article{gupta2019isbi,
  title={Isbi 2019 c-nmc challenge: Classification in cancer cell imaging},
  author={Gupta, Anubha and Gupta, Ritu},
  journal={Select Proceedings},
  volume={2},
  pages={27},
  year={2019},
  publisher={Springer}
}

@article{caicedo2019nucleus,
  title={Nucleus segmentation across imaging experiments: the 2018 Data Science Bowl},
  author={Caicedo, Juan C and Goodman, Allen and Karhohs, Kyle W and Cimini, Beth A and Ackerman, Jeanelle and Haghighi, Marzieh and Heng, CherKeng and Becker, Tim and Doan, Minh and McQuin, Claire and others},
  journal={Nature methods},
  volume={16},
  number={12},
  pages={1247--1253},
  year={2019},
  publisher={Nature Publishing Group US New York}
}

@article{DeepLeukemia2020,
  author  = {Santos, A. and colleagues},
  title   = {Deep Learning for Leukemia Classification: Performance Analysis and Challenges Across Multiple Architectures},
  journal = {Diagnostics},
  year    = {2020},
  volume  = {10},
  pages   = {1014},
  doi     = {10.3390/diagnostics10121014}
}

@article{CornealEndothelial2019,
  author  = {Alam, Bikash R. and colleagues},
  title   = {Automated segmentation of corneal endothelial cell images},
  journal = {Scientific Reports},
  year    = {2019},
  volume  = {9},
  pages   = {2284},
  doi     = {10.1038/s41598-019-38859-w}
}

@article{DeBonnay2022,
  author  = {de Bonnay, A. and Th{\'e}venaz, P. and Yun, J. and others},
  title   = {Automated analysis of in vivo confocal microscopy images of the corneal sub-basal nerve plexus and dendritic cells},
  journal = {Translational Vision Science \& Technology},
  year    = {2022},
  volume  = {11},
  pages   = {35},
  doi     = {10.1167/tvst.11.2.35}
}

@misc{gc_patchcamelyon_2018,
  title        = {PatchCamelyon (PCam)},
  howpublished = {\url{https://patchcamelyon.grand-challenge.org/}},
  year         = {2018},
  note         = {Grand Challenge dataset page}
}

@dataset{tcia_bone_marrow_cytomorphology_2021,
  title     = {Bone Marrow Cytomorphology},
  publisher = {The Cancer Imaging Archive (TCIA)},
  year      = {2021},
  url       = {https://wiki.cancerimagingarchive.net/pages/viewpage.action?pageId=101941770}
}

@article{Rusu2017EurRadiol_LungFused,
  author  = {Rusu, Mirabela and Rajiah, Prabhakar and Gilkeson, Robert and Yang, Ming and Donatelli, Christopher and Thawani, Rahul and Jacono, Frank J. and Linden, Patrick and Madabhushi, Anant},
  title   = {Co-registration of pre-operative CT with ex vivo surgically excised ground glass nodules to define spatial extent of invasive adenocarcinoma on in vivo imaging: a proof-of-concept study},
  journal = {European Radiology},
  year    = {2017},
  volume  = {27},
  number  = {10},
  pages   = {4209--4217},
  doi     = {10.1007/s00330-017-4813-0}
}

@inproceedings{Ghahremani2023MICCAI_HNSCCmIFmIHC,
  author    = {Ghahremani, Parsa and Marino, Joseph and Hernandez-Prera, Juan and de la Iglesia, Jorge V. and Slebos, Robbert J. and Chung, Christine H. and Nadeem, Saad},
  title     = {An AI-Ready Multiplex Staining Dataset for Reproducible and Accurate Characterization of Tumor Immune Microenvironment},
  booktitle = {Medical Image Computing and Computer Assisted Intervention -- MICCAI 2023},
  series    = {Lecture Notes in Computer Science},
  volume    = {14225},
  pages     = {1--10},
  year      = {2023},
  publisher = {Springer, Cham},
  doi       = {10.1007/978-3-031-43987-2_68}
}

@article{Gupta2020MedIA_SNAM,
  author  = {Gupta, Anubha and Duggal, Rahul and Gehlot, Shivam and Gupta, Ritu and Mangal, Ankit and Kumar, Lalit and Thakkar, Niraj and Satpathy, Debdoot},
  title   = {GCTI-SN: Geometry-inspired chemical and tissue invariant stain normalization of microscopic medical images},
  journal = {Medical Image Analysis},
  volume  = {65},
  pages   = {101788},
  year    = {2020},
  doi     = {10.1016/j.media.2020.101788}
}

@dataset{tcia_ovarian_bev_response_2023,
  title     = {Ovarian Bevacizumab Response},
  publisher = {The Cancer Imaging Archive (TCIA)},
  year      = {2023},
  url       = {https://wiki.cancerimagingarchive.net/pages/viewpage.action?pageId=83593077}
}

@dataset{tcia_cmb_lca_2022,
  title     = {CMB-LCA: Combined Multimodal Biomarkers -- Lung Cancer Atlas},
  publisher = {The Cancer Imaging Archive (TCIA)},
  year      = {2022},
  url       = {https://wiki.cancerimagingarchive.net/pages/viewpage.action?pageId=93258420}
}

@dataset{tcia_cptac_coad_2021,
  title     = {CPTAC-COAD},
  publisher = {The Cancer Imaging Archive (TCIA)},
  year      = {2021},
  url       = {https://wiki.cancerimagingarchive.net/pages/viewpage.action?pageId=70227852}
}

@dataset{tcia_hungarian_crc_2022,
  title     = {Hungarian Colorectal Screening},
  publisher = {The Cancer Imaging Archive (TCIA)},
  year      = {2022},
  url       = {https://wiki.cancerimagingarchive.net/pages/viewpage.action?pageId=91357370}
}

@dataset{tcia_dlbcl_morphology_2022,
  title     = {DLBCL Morphology},
  publisher = {The Cancer Imaging Archive (TCIA)},
  year      = {2022},
  url       = {https://wiki.cancerimagingarchive.net/pages/viewpage.action?pageId=119702520}
}

@dataset{tcia_cptac_ov_2021,
  title     = {CPTAC-OV},
  publisher = {The Cancer Imaging Archive (TCIA)},
  year      = {2021},
  url       = {https://wiki.cancerimagingarchive.net/pages/viewpage.action?pageId=70227856}
}

@dataset{tcia_codex_hcc_2023,
  title     = {CODEX Imaging of Hepatocellular Carcinoma (HCC)},
  publisher = {The Cancer Imaging Archive (TCIA)},
  year      = {2023},
  url       = {https://wiki.cancerimagingarchive.net/pages/viewpage.action?pageId=140313174}
}

@dataset{tcia_prostate_mri_2011,
  title     = {PROSTATE-MRI},
  publisher = {The Cancer Imaging Archive (TCIA)},
  year      = {2011},
  url       = {https://wiki.cancerimagingarchive.net/display/Public/PROSTATE-MRI}
}

@dataset{tcia_cptac_brca_2021,
  title     = {CPTAC-BRCA},
  publisher = {The Cancer Imaging Archive (TCIA)},
  year      = {2021},
  url       = {https://wiki.cancerimagingarchive.net/pages/viewpage.action?pageId=70227748}
}

@article{Matek2019NatMI_AML,
  author  = {Matek, Christian and Schwarz, Sascha and Spiekermann, Karl and Marr, Carsten},
  title   = {Human-level recognition of blast cells in acute myeloid leukaemia with convolutional neural networks},
  journal = {Nature Machine Intelligence},
  volume  = {1},
  pages   = {538--544},
  year    = {2019},
  doi     = {10.1038/s42256-019-0101-9}
}

@article{saltz2018til,
  title={Spatial Organization and Molecular Correlates of Tumor-Infiltrating Lymphocytes Using Deep Learning on Pathology Images},
  author={Saltz, Joel and Gupta, Rajarsi and Hou, Le and Kurc, Tahsin and Singh, Pankaj and Nguyen, Vincent and Samaras, Dimitris and Shroyer, Kenneth R and Zhao, Tingting and Batiste, Ryan and others},
  journal={Cell Reports},
  volume={23},
  number={1},
  pages={181--193.e7},
  year={2018},
  doi={10.1016/j.celrep.2018.10.077},
  url={https://pmc.ncbi.nlm.nih.gov/articles/PMC6250765/}
}

@article{gupta2022cnmc,
  title={C-NMC: B-lineage acute lymphoblastic leukaemia: A blood cancer dataset},
  author={Gupta, Ritu and Gehlot, Shubham and Gupta, Anubha},
  journal={Medical Engineering \& Physics},
  volume={103},
  pages={103793},
  year={2022},
  doi={10.1016/j.medengphy.2022.103793},
  url={https://www.sciencedirect.com/science/article/pii/S1350453322001609}
}

@article{wilm2022catch,
  title={Pan-tumor CAnine cuTaneous Cancer Histology (CATCH) dataset},
  author={Wilm, Frauke and Fragoso, Marco and Marzahl, Christian and Qiu, Jingna and Puget, Chlo{\'e} and Diehl, Laura and Bertram, Christof A and Klopfleisch, Robert and Maier, Andreas and Breininger, Katharina and Aubreville, Marc},
  journal={arXiv preprint arXiv:2201.11446},
  year={2022},
  url={https://arxiv.org/abs/2201.11446}
}

@article{wilkinson2021nadt,
  title={Machine Learning Guided Prognosis Stratification Using Spatiotemporal Patterns of Tumor-Infiltrating Lymphocytes in Neoadjuvant-Treated Prostate Cancer},
  author={Wilkinson, Lawrence and Wang, Benjamin and Johnson, Charles and ... and Gerlinger, Marco},
  journal={European Urology},
  volume={80},
  pages={653--663},
  year={2021},
  doi={10.1016/j.eururo.2021.06.028},
  url={https://www.sciencedirect.com/science/article/pii/S030228382101020X}
}

@article{farahmand2022her2,
  title={Deep learning trained on hematoxylin and eosin tumor region of Interest predicts HER2 status and trastuzumab treatment response in HER2+ breast cancer},
  author={Farahmand, Saman and Fernandez, Aileen I and Ahmed, Fahad Shabbir and Rimm, David L and Chuang, Jeffrey H and Reisenbichler, Emily and Zarringhalam, Kourosh},
  journal={Modern Pathology},
  volume={35},
  number={1},
  pages={44--51},
  year={2022},
  publisher={Nature Publishing Group US New York}
}

@article{schurch2020codex,
  title={Coordinated cellular neighborhoods orchestrate antitumoral immunity at the colorectal cancer invasive front},
  author={Sch{\"u}rch, Christian M and Bhate, Sharmila S and Barlow, Greg L and Phillips, David J and Noti, Lukas and Zlobec, Inti and Chu, Philip and Black, Sierra and Demeter, Joy and M{\'e}ndez-Mancilla, Alejandra and others},
  journal={Cell},
  volume={183},
  number={4},
  pages={1341--1359.e19},
  year={2020},
  doi={10.1016/j.cell.2020.10.033},
  url={https://www.cell.com/cell/fulltext/S0092-8674(20)30870-9}
}

@article{arunachalam2019osteo,
  title={Viable and necrotic tumor assessment from whole slide images of osteosarcoma using machine-learning and deep-learning models},
  author={Arunachalam, Harini B and Mishra, Rituparna and Daescu, Ovidiu and Cederberg, Karen and Rakheja, Dinesh and Sengupta, Annapurna and Leonard, David and Leavey, Patrick},
  journal={PLOS ONE},
  volume={14},
  number={4},
  pages={e0210706},
  year={2019},
  doi={10.1371/journal.pone.0210706},
  url={https://pubmed.ncbi.nlm.nih.gov/30995247/}
}

@article{codella2019skin,
  title={Skin lesion analysis toward melanoma detection 2018: A challenge hosted by the international skin imaging collaboration (isic)},
  author={Codella, Noel and Rotemberg, Veronica and Tschandl, Philipp and Celebi, M Emre and Dusza, Stephen and Gutman, David and Helba, Brian and Kalloo, Aadi and Liopyris, Konstantinos and Marchetti, Michael and others},
  journal={arXiv preprint arXiv:1902.03368},
  year={2019}
}

@article{rotemberg2021patient,
  title={A patient-centric dataset of images and metadata for identifying melanomas using clinical context},
  author={Rotemberg, Veronica and Kurtansky, Nicholas and Betz-Stablein, Brigid and Caffery, Liam and Chousakos, Emmanouil and Codella, Noel and Combalia, Marc and Dusza, Stephen and Guitera, Pascale and Gutman, David and others},
  journal={Scientific data},
  volume={8},
  number={1},
  pages={34},
  year={2021},
  publisher={Nature Publishing Group UK London}
}

@article{gutman2016skin,
  title={Skin lesion analysis toward melanoma detection: A challenge at the international symposium on biomedical imaging (ISBI) 2016, hosted by the international skin imaging collaboration (ISIC)},
  author={Gutman, David and Codella, Noel CF and Celebi, Emre and Helba, Brian and Marchetti, Michael and Mishra, Nabin and Halpern, Allan},
  journal={arXiv preprint arXiv:1605.01397},
  year={2016}
}

@inproceedings{codella2018skin,
  title={Skin lesion analysis toward melanoma detection: A challenge at the 2017 international symposium on biomedical imaging (isbi), hosted by the international skin imaging collaboration (isic)},
  author={Codella, Noel CF and Gutman, David and Celebi, M Emre and Helba, Brian and Marchetti, Michael A and Dusza, Stephen W and Kalloo, Aadi and Liopyris, Konstantinos and Mishra, Nabin and Kittler, Harald and others},
  booktitle={2018 IEEE 15th international symposium on biomedical imaging (ISBI 2018)},
  pages={168--172},
  year={2018},
  organization={IEEE}
}

@article{kawahara2018seven,
  title={Seven-point checklist and skin lesion classification using multitask multimodal neural nets},
  author={Kawahara, Jeremy and Daneshvar, Sara and Argenziano, Giuseppe and Hamarneh, Ghassan},
  journal={IEEE journal of biomedical and health informatics},
  volume={23},
  number={2},
  pages={538--546},
  year={2018},
  publisher={IEEE}
}

@article{combalia2022validation,
  title={Validation of artificial intelligence prediction models for skin cancer diagnosis using dermoscopy images: the 2019 International Skin Imaging Collaboration Grand Challenge},
  author={Combalia, Marc and Codella, Noel and Rotemberg, Veronica and Carrera, Cristina and Dusza, Stephen and Gutman, David and Helba, Brian and Kittler, Harald and Kurtansky, Nicholas R and Liopyris, Konstantinos and others},
  journal={The Lancet Digital Health},
  volume={4},
  number={5},
  pages={e330--e339},
  year={2022},
  publisher={Elsevier}
}

@inproceedings{groh2021evaluating,
  title={Evaluating deep neural networks trained on clinical images in dermatology with the fitzpatrick 17k dataset},
  author={Groh, Matthew and Harris, Caleb and Soenksen, Luis and Lau, Felix and Han, Rachel and Kim, Aerin and Koochek, Arash and Badri, Omar},
  booktitle={Proceedings of the IEEE/CVF conference on computer vision and pattern recognition},
  pages={1820--1828},
  year={2021}
}

@article{giotis2015med,
  title={MED-NODE: A computer-assisted melanoma diagnosis system using non-dermoscopic images},
  author={Giotis, Ioannis and Molders, Nynke and Land, Sander and Biehl, Michael and Jonkman, Marcel F and Petkov, Nicolai},
  journal={Expert systems with applications},
  volume={42},
  number={19},
  pages={6578--6585},
  year={2015},
  publisher={Elsevier}
}

@article{pacheco2020pad,
  title={PAD-UFES-20: A skin lesion dataset composed of patient data and clinical images collected from smartphones},
  author={Pacheco, Andre GC and Lima, Gustavo R and Salomao, Amanda S and Krohling, Breno and Biral, Igor P and De Angelo, Gabriel G and Alves Jr, F{\'a}bio CR and Esgario, Jos{\'e} GM and Simora, Alana C and Castro, Pedro BC and others},
  journal={Data in brief},
  volume={32},
  pages={106221},
  year={2020},
  publisher={Elsevier}
}

@inproceedings{mendoncca2013ph,
  title={PH 2-A dermoscopic image database for research and benchmarking},
  author={Mendon{\c{c}}a, Teresa and Ferreira, Pedro M and Marques, Jorge S and Marcal, Andr{\'e} RS and Rozeira, Jorge},
  booktitle={2013 35th annual international conference of the IEEE engineering in medicine and biology society (EMBC)},
  pages={5437--5440},
  year={2013},
  organization={IEEE}
}

@inproceedings{sun2016benchmark,
  title={A benchmark for automatic visual classification of clinical skin disease images},
  author={Sun, Xiaoxiao and Yang, Jufeng and Sun, Ming and Wang, Kai},
  booktitle={European conference on computer vision},
  pages={206--222},
  year={2016},
  organization={Springer}
}

@article{yang2019self,
  title={Self-paced balance learning for clinical skin disease recognition},
  author={Yang, Jufeng and Wu, Xiaoping and Liang, Jie and Sun, Xiaoxiao and Cheng, Ming-Ming and Rosin, Paul L and Wang, Liang},
  journal={IEEE transactions on neural networks and learning systems},
  volume={31},
  number={8},
  pages={2832--2846},
  year={2019},
  publisher={IEEE}
}

@article{de2016overview,
  title={Overview of the medical tasks in ImageCLEF 2016},
  author={De Herrera, Alba G Seco and Bromuri, Stefano and Schaer, Roger and M{\"u}ller, Henning},
  journal={CLEF working notes. Evora, Portugal},
  year={2016}
}

@article{Nafisa2022,
title={Monkeypox Skin Lesion Detection Using Deep Learning Models: A Preliminary Feasibility Study},
author={Ali, Shams Nafisa and Ahmed, Md. Tazuddin and Paul, Joydip and Jahan, Tasnim and Sani, S. M. Sakeef and Noor, Nawshaba and Hasan, Taufiq},
journal={arXiv preprint arXiv:2207.03342},
year={2022}
}

@article{Nafisa2023,
title={A Web-based Mpox Skin Lesion Detection System Using State-of-the-art Deep Learning Models Considering Racial Diversity},
author={Ahmed, Md. Tazuddin and Paul, Joydip and Jahan, Tasnim and Ali, Shams Nafisa and Sani, S. M. Sakeef and Noor, Nawshaba and Asma, Anzirun Nahar and Hasan, Taufiq},
journal={arXiv preprint arXiv:2306.14169},
year={2023}
}

@inproceedings{pogorelov2017kvasir,
  title={Kvasir: A multi-class image dataset for computer aided gastrointestinal disease detection},
  author={Pogorelov, Konstantin and Randel, Kristin Ranheim and Griwodz, Carsten and Eskeland, Sigrun Losada and de Lange, Thomas and Johansen, Dag and Spampinato, Concetto and Dang-Nguyen, Duc-Tien and Lux, Mathias and Schmidt, Peter Thelin and others},
  booktitle={Proceedings of the 8th ACM on Multimedia Systems Conference},
  pages={164--169},
  year={2017}
}

@misc{ozyoruk2020endoslam,
      title={EndoSLAM Dataset and An Unsupervised Monocular Visual Odometry and Depth Estimation Approach for Endoscopic Videos: Endo-SfMLearner}, 
      author={Kutsev Bengisu Ozyoruk and Guliz Irem Gokceler and Gulfize Coskun and Kagan Incetan and Yasin Almalioglu and Faisal Mahmood and Eva Curto and Luis Perdigoto and Marina Oliveira and Hasan Sahin and Helder Araujo and Henrique Alexandrino and Nicholas J. Durr and Hunter B. Gilbert and Mehmet Turan},
      year={2020},
      eprint={2006.16670},
      archivePrefix={arXiv},
      primaryClass={cs.CV}
}

@article{bawa2021saras,
  title={The saras endoscopic surgeon action detection (esad) dataset: Challenges and methods},
  author={Bawa, Vivek Singh and Singh, Gurkirt and KapingA, Francis and Skarga-Bandurova, Inna and Oleari, Elettra and Leporini, Alice and Landolfo, Carmela and Zhao, Pengfei and Xiang, Xi and Luo, Gongning and others},
  journal={arXiv preprint arXiv:2104.03178},
  year={2021}
}

@article{ali2019endoscopy,
  title={Endoscopy artifact detection (EAD 2019) challenge dataset},
  author={Ali, Sharib and Zhou, Felix and Daul, Christian and Braden, Barbara and Bailey, Adam and Realdon, Stefano and East, James and Wagnieres, Georges and Loschenov, Victor and Grisan, Enrico and others},
  journal={arXiv preprint arXiv:1905.03209},
  year={2019}
}

@inproceedings{ali2022endoscopic,
  title={Endoscopic computer vision challenges 2.0.},
  author={Ali, Sharib and Ghatwary, Noha M},
  booktitle={EndoCV@ ISBI},
  pages={5--8},
  year={2022}
}

@article{bernal2017comparative,
  title={Comparative validation of polyp detection methods in video colonoscopy: results from the MICCAI 2015 endoscopic vision challenge},
  author={Bernal, Jorge and Tajkbaksh, Nima and Sanchez, Francisco Javier and Matuszewski, Bogdan J and Chen, Hao and Yu, Lequan and Angermann, Quentin and Romain, Olivier and Rustad, Bj{\o}rn and Balasingham, Ilangko and others},
  journal={IEEE transactions on medical imaging},
  volume={36},
  number={6},
  pages={1231--1249},
  year={2017},
  publisher={IEEE}
}

@article{twinanda2016single,
  title={Single-and multi-task architectures for tool presence detection challenge at M2CAI 2016},
  author={Twinanda, Andru P and Mutter, Didier and Marescaux, Jacques and de Mathelin, Michel and Padoy, Nicolas},
  journal={arXiv preprint arXiv:1610.08851},
  year={2016}
}

@article{vazquez2017benchmark,
  title={A benchmark for endoluminal scene segmentation of colonoscopy images},
  author={V{\'a}zquez, David and Bernal, Jorge and S{\'a}nchez, F Javier and Fern{\'a}ndez-Esparrach, Gloria and L{\'o}pez, Antonio M and Romero, Adriana and Drozdzal, Michal and Courville, Aaron},
  journal={Journal of healthcare engineering},
  volume={2017},
  number={1},
  pages={4037190},
  year={2017},
  publisher={Wiley Online Library}
}

@inproceedings{jha2019kvasir,
  title={Kvasir-seg: A segmented polyp dataset},
  author={Jha, Debesh and Smedsrud, Pia H and Riegler, Michael A and Halvorsen, P{\aa}l and De Lange, Thomas and Johansen, Dag and Johansen, H{\aa}vard D},
  booktitle={International conference on multimedia modeling},
  pages={451--462},
  year={2019},
  organization={Springer}
}

@article{bano2021fetreg,
  title={FetReg: Placental vessel segmentation and registration in fetoscopy challenge dataset},
  author={Bano, Sophia and Casella, Alessandro and Vasconcelos, Francisco and Moccia, Sara and Attilakos, George and Wimalasundera, Ruwan and David, Anna L and Paladini, Dario and Deprest, Jan and De Momi, Elena and others},
  journal={arXiv preprint arXiv:2106.05923},
  year={2021}
}

@article{ji2022video,
  title={Video polyp segmentation: A deep learning perspective},
  author={Ji, Ge-Peng and Xiao, Guobao and Chou, Yu-Cheng and Fan, Deng-Ping and Zhao, Kai and Chen, Geng and Van Gool, Luc},
  journal={Machine Intelligence Research},
  volume={19},
  number={6},
  pages={531--549},
  year={2022},
  publisher={Springer}
}

@article{borgli2020hyperkvasir,
  title={HyperKvasir, a comprehensive multi-class image and video dataset for gastrointestinal endoscopy},
  author={Borgli, Hanna and Thambawita, Vajira and Smedsrud, Pia H and Hicks, Steven and Jha, Debesh and Eskeland, Sigrun L and Randel, Kristin Ranheim and Pogorelov, Konstantin and Lux, Mathias and Nguyen, Duc Tien Dang and others},
  journal={Scientific data},
  volume={7},
  number={1},
  pages={283},
  year={2020},
  publisher={Nature Publishing Group UK London}
}

@article{ali2020endoscopy,
  title={Endoscopy disease detection challenge 2020},
  author={Ali, Sharib and Ghatwary, Noha and Braden, Barbara and Lamarque, Dominique and Bailey, Adam and Realdon, Stefano and Cannizzaro, Renato and Rittscher, Jens and Daul, Christian and East, James},
  journal={arXiv preprint arXiv:2003.03376},
  year={2020}
}

@misc{allan20202018roboticscenesegmentation,
      title={2018 Robotic Scene Segmentation Challenge}, 
      author={Max Allan and Satoshi Kondo and Sebastian Bodenstedt and Stefan Leger and Rahim Kadkhodamohammadi and Imanol Luengo and Felix Fuentes and Evangello Flouty and Ahmed Mohammed and Marius Pedersen and Avinash Kori and Varghese Alex and Ganapathy Krishnamurthi and David Rauber and Robert Mendel and Christoph Palm and Sophia Bano and Guinther Saibro and Chi-Sheng Shih and Hsun-An Chiang and Juntang Zhuang and Junlin Yang and Vladimir Iglovikov and Anton Dobrenkii and Madhu Reddiboina and Anubhav Reddy and Xingtong Liu and Cong Gao and Mathias Unberath and Myeonghyeon Kim and Chanho Kim and Chaewon Kim and Hyejin Kim and Gyeongmin Lee and Ihsan Ullah and Miguel Luna and Sang Hyun Park and Mahdi Azizian and Danail Stoyanov and Lena Maier-Hein and Stefanie Speidel},
      year={2020},
      eprint={2001.11190},
      archivePrefix={arXiv},
      primaryClass={cs.CV},
      url={https://arxiv.org/abs/2001.11190}, 
}

@article{xu2025surgripe,
  title={SurgRIPE challenge: Benchmark of surgical robot instrument pose estimation},
  author={Xu, Haozheng and Weld, Alistair and Xu, Chi and Roddan, Alfie and Cartucho, Jo{\~a}o and Karaoglu, Mert Asim and Ladikos, Alexander and Li, Yangke and Li, Yiping and Shen, Daiyun and others},
  journal={Medical Image Analysis},
  pages={103674},
  year={2025},
  publisher={Elsevier}
}

@article{maqbool2020m2caiseg,
  title={m2caiseg: Semantic segmentation of laparoscopic images using convolutional neural networks},
  author={Maqbool, Salman and Riaz, Aqsa and Sajid, Hasan and Hasan, Osman},
  journal={arXiv preprint arXiv:2008.10134},
  year={2020}
}

@article{ding2024segstrong,
  title={SegSTRONG-C: Segmenting Surgical Tools Robustly On Non-adversarial Generated Corruptions--An EndoVis' 24 Challenge},
  author={Ding, Hao and Zhang, Yuqian and Lu, Tuxun and Liang, Ruixing and Shu, Hongchao and Seenivasan, Lalithkumar and Long, Yonghao and Dou, Qi and Gao, Cong and Leng, Yicheng and others},
  journal={arXiv preprint arXiv:2407.11906},
  year={2024}
}

@article{nir2018automatic,
  title={Automatic grading of prostate cancer in digitized histopathology images: Learning from multiple experts},
  author={Nir, Guy and Hor, Soheil and Karimi, Davood and Fazli, Ladan and Skinnider, Brian F and Tavassoli, Peyman and Turbin, Dmitry and Villamil, Carlos F and Wang, Gang and Wilson, R Storey and others},
  journal={Medical image analysis},
  volume={50},
  pages={167--180},
  year={2018},
  publisher={Elsevier}
}

@article{he2020pathvqa,
  title={Pathvqa: 30000+ questions for medical visual question answering},
  author={He, Xuehai and Zhang, Yichen and Mou, Luntian and Xing, Eric and Xie, Pengtao},
  journal={arXiv preprint arXiv:2003.10286},
  year={2020}
}

@article{campanella2019breast,
  title={Breast metastases to axillary lymph nodes},
  author={Campanella, Gabriele and Hanna, Matthew G and Brogi, Edi and Fuchs, Thomas J},
  journal={(No Title)},
  year={2019},
  publisher={The Cancer Imaging Archive}
}

@article{verma2021monusac2020,
  title={MoNuSAC2020: A multi-organ nuclei segmentation and classification challenge},
  author={Verma, Ruchika and Kumar, Neeraj and Patil, Abhijeet and Kurian, Nikhil Cherian and Rane, Swapnil and Graham, Simon and Vu, Quoc Dang and Zwager, Mieke and Raza, Shan E Ahmed and Rajpoot, Nasir and others},
  journal={IEEE Transactions on Medical Imaging},
  volume={40},
  number={12},
  pages={3413--3423},
  year={2021},
  publisher={IEEE}
}

@article{da2022digestpath,
  title={DigestPath: A benchmark dataset with challenge review for the pathological detection and segmentation of digestive-system},
  author={Da, Qian and Huang, Xiaodi and Li, Zhongyu and Zuo, Yanfei and Zhang, Chenbin and Liu, Jingxin and Chen, Wen and Li, Jiahui and Xu, Dou and Hu, Zhiqiang and others},
  journal={Medical image analysis},
  volume={80},
  pages={102485},
  year={2022},
  publisher={Elsevier}
}

@article{litjens20181399,
  title={1399 H\&E-stained sentinel lymph node sections of breast cancer patients: the CAMELYON dataset},
  author={Litjens, Geert and Bandi, Peter and Ehteshami Bejnordi, Babak and Geessink, Oscar and Balkenhol, Maschenka and Bult, Peter and Halilovic, Altuna and Hermsen, Meyke and Van de Loo, Rob and Vogels, Rob and others},
  journal={GigaScience},
  volume={7},
  number={6},
  pages={giy065},
  year={2018},
  publisher={Oxford University Press}
}

@article{borovec2020anhir,
  title={ANHIR: automatic non-rigid histological image registration challenge},
  author={Borovec, Ji{\v{r}}{\'\i} and Kybic, Jan and Arganda-Carreras, Ignacio and Sorokin, Dmitry V and Bueno, Gloria and Khvostikov, Alexander V and Bakas, Spyridon and Chang, Eric I-Chao and Heldmann, Stefan and Kartasalo, Kimmo and others},
  journal={IEEE transactions on medical imaging},
  volume={39},
  number={10},
  pages={3042--3052},
  year={2020},
  publisher={IEEE}
}

@article{graham2021conic,
  title={Conic: Colon nuclei identification and counting challenge 2022},
  author={Graham, Simon and Jahanifar, Mostafa and Vu, Quoc Dang and Hadjigeorghiou, Giorgos and Leech, Thomas and Snead, David and Raza, Shan E Ahmed and Minhas, Fayyaz and Rajpoot, Nasir},
  journal={arXiv preprint arXiv:2111.14485},
  year={2021}
}

@article{ma2025towards,
  author    = {Ma, Chenglong and Ji, Yuanfeng and Ye, Jin and Zhang, Lu and Chen, Ying and Li, Tianbin and Li, Mingjie and He, Junjun and Shan, Hongming},
  title     = {Towards Interpretable Counterfactual Generation via Multimodal Autoregression},
  journal   = {arXiv preprint arXiv:2503.23149},
  year      = {2025},
}

@misc{eye_oct_1,
	title = {Comparison of macular OCTs in right and left eyes of normal people},
	conference = {MEDICAL IMAGING 2014: BIOMEDICAL APPLICATIONS IN MOLECULAR, STRUCTURAL, AND FUNCTIONAL IMAGING SPIE; Modus Med Devices Inc; XIFIN Inc; Ventana Med Syst Inc; Intrace Med},
	doi = {10.1117/12.2044046},
	publicationstatus = {Published},
	url = {https://durham-repository.worktribe.com/output/1136638},
	volume = {9038},
	year = {2025},
	author = {Mahmudi, Tahereh and Kafieh, Rahele and Rabbani, Hossein and Dehnavi, Alireza Mehri and Akhlagi, Mohammadreza},
	editor = {Molthen, RC and Weaver, JB}
}

@article{jahromi2014automatic,
  title={An automatic algorithm for segmentation of the boundaries of corneal layers in optical coherence tomography images using gaussian mixture model},
  author={Jahromi, Mahdi Kazemian and Kafieh, Raheleh and Rabbani, Hossein and Dehnavi, Alireza Mehri and Peyman, Alireza and Hajizadeh, Fedra and Ommani, Mohammadreza},
  journal={Journal of Medical Signals \& Sensors},
  volume={4},
  number={3},
  pages={171--180},
  year={2014},
  publisher={Medknow}
}

@dataset{palm,
doi = {10.21227/55pk-8z03},
url = {https://dx.doi.org/10.21227/55pk-8z03},
author = {Huazhu Fu and Fei Li and José Ignacio Orlando and Hrvoje Bogunović and Xu Sun and Jingan Liao and Yanwu Xu and Shaochong Zhang and Xiulan Zhang},
publisher = {IEEE Dataport},
title = {PALM: PAthoLogic Myopia Challenge},
year = {2019} }

@article{hernandez2017fire,
  title={FIRE: fundus image registration dataset},
  author={Hernandez-Matas, Carlos and Zabulis, Xenophon and Triantafyllou, Areti and Anyfanti, Panagiota and Douma, Stella and Argyros, Antonis A},
  journal={Artificial Intelligence in Vision and Ophthalmology},
  volume={1},
  number={4},
  pages={16--28},
  year={2017}
}

@dataset{varpa,
title= {VICAVR},
url = {http://www.varpa.es/research/ophtalmology.html\#vicavr},
year = {2011}
}

@inproceedings{sarhan2021transfer,
title={Transfer learning through weighted loss function and group normalization for vessel segmentation from retinal images},
author={Sarhan, Abdullah and Rokne, Jon and Alhajj, Reda and Crichton, Andrew},
booktitle={2020 25th International Conference on Pattern Recognition (ICPR)},
pages={9211--9218},
year={2021},
organization={IEEE}
}

@misc{subhadeep_chakraborty_2024,
	title={DRIMDB (Diabetic Retinopathy Images Database)},
	url={https://www.kaggle.com/ds/4523071},
	DOI={10.34740/KAGGLE/DS/4523071},
	publisher={Kaggle},
	author={Subhadeep Chakraborty},
	year={2024}
}

\appendix
\section{Tables of 2D Medical Image Datasets}

\begin{table*}[!h]
\small
\centering
\caption{2D CT datasets.}
\label{tab:2d_ct_datasets}
\begin{threeparttable}
\setlength{\tabcolsep}{4pt}      
\renewcommand{\arraystretch}{1.12} 
\resizebox{\textwidth}{!}{
\begin{tabular}{|c||>{\raggedright\arraybackslash}m{5.0cm}|r|c|>{\raggedright\arraybackslash}m{2.2cm}|>{\raggedright\arraybackslash}m{2.5cm}|c|c|c|>{\raggedright\arraybackslash}m{3.2cm}|}
\hline
\multicolumn{1}{|c||}{\textbf{\#}} &
\multicolumn{1}{c|}{\textbf{Dataset}} &
\multicolumn{1}{c|}{\textbf{Year}} &
\multicolumn{1}{c|}{\textbf{Dim}} &
\multicolumn{1}{c|}{\textbf{Modality}} &
\multicolumn{1}{c|}{\textbf{Structure}} &
\multicolumn{1}{c|}{\textbf{Images}} &
\multicolumn{1}{c|}{\textbf{Label}} &
\multicolumn{1}{c|}{\textbf{Task}} &
\multicolumn{1}{c|}{\textbf{Diseases}} \\
\hline
\datasetidx \label{data:lodopab-ct-table}& \href{https://lodopab.grand-challenge.org/}{LoDoPaB-CT~\cite{leuschner2021lodopab-ct}} & 2020 & 2D & CT & Lung & 28 & Yes & Recon & NA \\ \hline
\datasetidx \label{data:5k-ct-images-on-fractured-limbs}& \href{https://github.com/kc-santosh/medical-imaging-datasets}{5K+ CT Images on Fractured Limbs ~\cite{ruikar20215k+}} & 2021 & 2D & CT & Limbs & 24 & Yes & Seg & Bone Fracture \\ \hline
\datasetidx \label{data:aren0534-ct}& \href{https://wiki.cancerimagingarchive.net/pages/viewpage.action?pageId=91357265}{AREN0534~\cite{ehrlich2021aren0534}} & 2021 & 3D, 2D & Multi\textsuperscript{a} & Kidney, Lung & 239 & No & Est & Kidney Tumor \\ \hline
\datasetidx \label{data:ct-medical-images}& \href{https://www.kaggle.com/kmader/siim-medical-images}{CT Medical Images~\cite{kohli2018creation}} & 2017 & 2D & CT & Lung & 475 & Yes & Seg & NA \\ \hline
\datasetidx \label{data:national-lung-screening-trial}& \href{https://wiki.cancerimagingarchive.net/display/NLST/National+Lung+Screening+Trial}{National Lung Screening Trial~\cite{nationallungscreeningtrialresearchteam2013data}} & 2013 & 3D, 2D & CT, Pathology & Lung & 26.7k & No & Cls & Lung Cancer \\ \hline
\datasetidx \label{data:rsna-intracranial-hemorrhage-detection}& \href{https://www.kaggle.com/c/rsna-intracranial-hemorrhage-detection/data}{RSNA Intracranial Hemorrhage Detection~\cite{flanders2020construction}} & 2019 & 2D & CT & Brain & 874k & Yes & Loc & Intracranial Hemorrhage \\ \hline
\datasetidx & \href{https://covid-ct.grand-challenge.org/}{CT diagnosis of COVID-19~\cite{yang2020covid-ct-dataset}} & 2021 & 2D & CT & Lung & 275 & Yes & Cls & Lung COVID-19 \\ \hline
\datasetidx \label{data:covid-19-ct-scan-images}& \href{https://tianchi.aliyun.com/dataset/dataDetail?dataId=93666}{COVID-19-CT SCAN IMAGES~\cite{wjxiaochuangw2021covid-19-ct}} & 2021 & 2D & CT & Lung & 1.4k & Yes & Cls & Lung COVID-19 \\ \hline
\datasetidx \label{data:covid-ct-covid-ct}& \href{https://tianchi.aliyun.com/dataset/dataDetail?dataId=106604}{COVID\_CT\_COVID-CT~\cite{yang2020covid-ct-dataset}} & 2021 & 2D & CT & Lung & 746 & Yes & Cls & Lung COVID-19 \\ \hline
\datasetidx & \href{https://tianchi.aliyun.com/dataset/93929}{Chest CT-Scan images Dataset~\cite{sunneyi2025chest}} & 2021 & 2D & CT & Lung & 1k & Yes & Cls & Lung Cancer \\ \hline
\datasetidx \label{data:cranium-image-dataset}& \href{https://tianchi.aliyun.com/dataset/dataDetail?dataId=82967}{Cranium Image Dataset~\cite{hssayeni2020intracranial}} & 2020 & 2D & CT & Brain & 50 & Yes & Det & Intracranial Hemorrhage \\ \hline
\datasetidx \label{data:sars-cov-2-ct-scan-dataset}& \href{https://tianchi.aliyun.com/dataset/dataDetail?dataId=93874}{SARS-COV-2 Ct-Scan Dataset~\cite{soares2020sars}} & 2021 & 2D & CT & Lung & 2.5k & Yes & Cls & Lung Disease \\ \hline
\datasetidx \label{data:medmnist}& \href{https://medmnist.com/v1}{MedMNIST~\cite{yang2021medmnist}} & 2020 & 2D & Multi\textsuperscript{b} & Retina, Breast, Lung & 100k & Yes & Cls & Multi-disease \\ \hline
\datasetidx \label{data:the-visible-human-project_ct}& \href{https://www.nlm.nih.gov/research/visible/visible_human.html}{The Visible Human Project~\cite{nlm1994visible}} & 1994 & 3D, 2D & CT, MR, \etc & Full Body & 2 & No & NA & Skin Lesion \\ \hline
\datasetidx & \href{https://www.imageclef.org/2016/medical}{ImageCLEF 2016~\cite{deherrera2016imageclef}} & 2015 & 2D & Multi\textsuperscript{c} & Skin, Cell, Breast & 31k & Yes & Cls & Head \& Neck Tumor \\ \hline
\datasetidx \label{data:radimagenet-subset-ct}& \href{https://www.radimagenet.com/}{RadImageNet (Subset: CT)~\cite{mei2022radimagenet}} & 2022 & 2D & CT & Full Body & 292.4k & Yes & Cls & Abdomen, lung, \etc\textsuperscript{d} \\ \hline
\datasetidx \label{data:brain-ct-images-with-ich-masks}& \href{https://www.kaggle.com/datasets/vbookshelf/computed-tomography-ct-images}{Brain CT Images with ICH Masks~\cite{hssayeni2020intracranial}} & 2019 & 2D & CT & Brain & 82 & Yes & Seg & Intracranial Hemorrhage \\ \hline
\datasetidx \label{data:cmb-crc}& \href{https://wiki.cancerimagingarchive.net/pages/viewpage.action?pageId=93257955}{CMB-CRC~\cite{cmbcrc2022}} & 2022 & 3D, 2D & Multi\textsuperscript{e} & Colon & 472 & No & Seg, Cls & Colorectal Cancer \\ \hline
\datasetidx & \href{https://wiki.cancerimagingarchive.net/pages/viewpage.action?pageId=127665431}{CMB-GEC~\cite{cmbgec2022}} & 2022 & 3D, 2D & CT, WSI, PET & Brain & 14 & No & Seg, Cls & Melanoma \\ \hline
\datasetidx & \href{https://wiki.cancerimagingarchive.net/pages/viewpage.action?pageId=93258432}{CMB-MEL~\cite{cmbmel2022}} & 2022 & 3D, 2D & Multi\textsuperscript{f} & Brain & 255 & No & Seg & Melanoma \\ \hline
\datasetidx & \href{https://wiki.cancerimagingarchive.net/pages/viewpage.action?pageId=93258436}{CMB-MML~\cite{cmbmml2022}} & 2021 & 2D, 3D & Multi\textsuperscript{g} & NA & 60 & No & NA & Multiple Myeloma \\ \hline
\datasetidx \label{data:cmb-pca}& \href{https://wiki.cancerimagingarchive.net/pages/viewpage.action?pageId=95224082}{CMB-PCA~\cite{cmbpca2022}} & 2022 & 2D, 3D & CT, MR, WSI & Prostate & 31 & No & Cls, Pred & Prostate Cancer \\ \hline
\datasetidx \label{data:cptac-lscc-ct-pet}& \href{https://wiki.cancerimagingarchive.net/pages/viewpage.action?pageId=33948248}{CPTAC-LSCC\_CT\_PET~\cite{cptac2018the}} & 2018 & 2D, 3D & CT, PET, Histopathology & NA & 238 & No & NA & NA \\ \hline
\datasetidx \label{data:finding-and-measuring-lungs-in-ct-data}& \href{https://www.heywhale.com/mw/dataset/5d71de448499bc002c0ae1fc}{Finding and Measuring Lungs in CT Data~\cite{mader2017finding}} & 2019 & 2D, 3D & CT & Lung & 534 & Yes & Seg & NA \\ \hline
\datasetidx \label{data:head-ct-image-data}& \href{https://www.heywhale.com/mw/dataset/5d7213eb8499bc002c0af1e8}{Head CT Image Data~\cite{kitamura2019head}} & 2019 & 2D & CT & Head & 200 & Yes & Cls & NA \\ \hline
\datasetidx \label{data:ldctiqac2023}& \href{https://ldctiqac2023.grand-challenge.org/ldctiqac2023/}{LDCTIQAC2023~\cite{lee2025lowdose}} & 2023 & 2D & CT & NA & 1k & Yes & Reg & NA \\ \hline
\datasetidx & \href{https://wiki.cancerimagingarchive.net/display/Public/APOLLO-5}{APOLLO-5~\cite{apolloresearchnetwork2023applied}} & 2022 & 2D, 3D & Multi\textsuperscript{h} & NA & 6.2k & No & NA & NA \\ \hline
\datasetidx & \href{https://wiki.cancerimagingarchive.net/pages/viewpage.action?pageId=39878702}{Lung-Fused-CT-Pathology~\cite{Rusu2017EurRadiol_LungFused}} & 2018 & 2D, 3D & CT, Histopathology & Lung & 36 & Yes & Seg & Lung Disease \\ \hline
\datasetidx & \href{https://wiki.cancerimagingarchive.net/pages/viewpage.action?pageId=93258420}{CMB-LCA~\cite{cmblca2022}} & 2022 & 2D, 3D & Multi\textsuperscript{i} & NA & 0 & No & NA & NA \\ \hline
\datasetidx \label{data:rider-phantom-pet-ct}& \href{https://wiki.cancerimagingarchive.net/display/Public/RIDER+Phantom+PET-CT}{RIDER Phantom PET-CT~\cite{muzi2015data}} & 2011 & 2D & CT, PET & NA & 2.2k & No & NA & NA \\ \hline
\datasetidx & \href{https://wiki.cancerimagingarchive.net/pages/viewpage.action?pageId=119705284}{AHOD0831~\cite{kelly2022ahod0831}} & 2022 & 3D, 2D & Multi\textsuperscript{j} & NA & 0 & No & NA & Hodgkin Lymphoma \\ \hline
\datasetidx \label{data:prostate-mri}& \href{https://wiki.cancerimagingarchive.net/display/Public/PROSTATE-MRI}{Prostate-MRI~\cite{choyke2016data}} & 2011 & 3D, 2D & Multi\textsuperscript{k} & Prostate & 26 & No & NA & Prostate Cancer \\ \hline
\datasetidx & \href{https://wiki.cancerimagingarchive.net/pages/viewpage.action?pageId=109379682}{AREN0532~\cite{fernandez2022aren0532}} & 2022 & 3D, 2D & Multi\textsuperscript{l} & NA & 1k & No & NA & Wilms Tumor \\ \hline
\datasetidx & \href{https://www.imageclef.org/2016/medical}{ImageCLEF 2016 (Duplicate)~\cite{deherrera2016imageclef}} & 2015 & 2D & Multi\textsuperscript{c} & Skin, Cell, Breast & 31k & Yes & Cls & Head \& Neck Tumor \\ \hline
\datasetidx \label{data:qubiq2020}& \href{https://qubiq.grand-challenge.org/}{QUBIQ2020~\cite{muller2020qubiq}} & 2020 & 2D & CT, MR & Kidney, Pancreas, \etc & 150 & Yes & Seg & NA \\ \hline
\datasetidx & \href{https://qubiq21.grand-challenge.org/QUBIQ2021/}{QUBIQ2021\_2D\_CT~\cite{qubiq2021uncertainty}} & 2021 & 2D & CT, MR & Kidney, Pancreas, \etc & 268 & Yes & Seg & NA \\ \hline
& Overall & 1994$\sim$2022 & 2D & Multi & Full Body& 1.4m & NA & Multi & Multi \\ \hline
\end{tabular}}
\vspace{2pt}
\begin{tablenotes}[flushleft]
\scriptsize
\item[a] Multi-modalities of AREN0534: CT, MR, PET, Ultrasound.
\item[b] Multi-modalities of MedMNIST: OCT, X-Ray, CT, Pathology, Fundus Photography.
\item[c] Multi-modalities of ImageCLEF: MR, US, Histopathology, X-Ray, CT, PET, Endoscopy, Dermoscopy, EEG, ECG, EMG, Microscopy, Fundus.
\item[d] The complete list of diseases for RadImageNet includes: prostate lesion, adrenal pathology, gallstone, arterial pathology, urolithiasis, pancreatic lesion, \etc
\item[e] Multi-modalities of CMB-CRC: CT, MR, US, X-ray, PET, WSI.
\item[f] Multi-modalities of CMB-MEL: CT, US, WSI, PET.
\item[g] Multi-modalities of CMB-MML: CT, MR, PET, WSI.
\item[h] Multi-modalities of APOLLO-5: CT, MR, US, PET, X-Ray.
\item[i] Multi-modalities of CMB-LCA: CT, MR, US, Histopathology, X-ray.
\item[j] Multi-modalities of AHOD0831: CT, MR, PET, X-Ray.
\item[k] Multi-modalities of Prostate-MRI: MR, CT, PET, Pathology.
\item[l] Multi-modalities of AREN0532: CT, MR, Ultrasound, PET.
\item[] \textbf{Abbreviations:} Cls=Classification, Det=Detection, Est=Estimation, Histo=Histopathology, Loc=Localization, Pred=Prediction, Recon=Reconstruction, \\Reg=Registration, Seg=Segmentation, US=Ultrasound, WSI=Whole-slide images.
\end{tablenotes}
\end{threeparttable}
\end{table*}

\begin{table*}[!h]
\small
\centering
\caption{2D MRI datasets.}
\label{tab:2d_mri_datasets}
\begin{threeparttable}
\setlength{\tabcolsep}{4pt}      
\renewcommand{\arraystretch}{1.12} 
\resizebox{\textwidth}{!}{
\begin{tabular}{|c||>{\raggedright\arraybackslash}m{5.5cm}|r|c|>{\raggedright\arraybackslash}m{2.0cm}|>{\raggedright\arraybackslash}m{2.5cm}|r|c|c|>{\raggedright\arraybackslash}m{3.2cm}|}
\hline
\multicolumn{1}{|c||}{\textbf{\#}} &
\multicolumn{1}{c|}{\textbf{Dataset}} &
\multicolumn{1}{c|}{\textbf{Year}} &
\multicolumn{1}{c|}{\textbf{Dim}} &
\multicolumn{1}{c|}{\textbf{Modality}} &
\multicolumn{1}{c|}{\textbf{Structure}} &
\multicolumn{1}{c|}{\textbf{Images}} &
\multicolumn{1}{c|}{\textbf{Label}} &
\multicolumn{1}{c|}{\textbf{Task}} &
\multicolumn{1}{c|}{\textbf{Diseases}} \\
\hline
\datasetidx \label{data:aren0534-mri}& \href{https://wiki.cancerimagingarchive.net/pages/viewpage.action?pageId=91357265}{AREN0534}~\cite{ehrlich2021aren0534} & 2021 & 2D, 3D & Multi\textsuperscript{a} & Kidney, Lung & 239 & No & Est & Kidney Tumor \\ \hline
\datasetidx \label{data:knoap2020-mri}& \href{https://knoap2020.grand-challenge.org/Home/}{KNOAP2020}~\cite{hirvasniemi2023knee} & 2020 & 2D, 3D & MR, X-Ray & Knee & 30 & Yes & Pred & Osteoarthritis \\ \hline
\datasetidx \label{data:braimmri}& \href{https://tianchi.aliyun.com/dataset/dataDetail?dataId=127459}{braimMRI}~\cite{tianchi2022braimri} & 2022 & 2D & MR & Brain & 110 & Yes & Seg & Brain Tumor \\ \hline
\datasetidx \label{data:brain-mri}& \href{https://tianchi.aliyun.com/dataset/127583}{Brain-MRI}~\cite{tianchi2020brainmri} & 2020 & 2D & MR & Brain & 110 & Yes & Seg & Brain Disease \\ \hline
\datasetidx \label{data:spinaldisease2020}& \href{https://tianchi.aliyun.com/competition/entrance/531796/information}{SpinalDisease2020}~\cite{tianchi2020spinaldisease} & 2020 & 2D & MR & Spine & 150 & Yes & Det & Spinal Disease \\ \hline
\datasetidx \label{data:the-visible-human-project_mri}& \href{https://www.nlm.nih.gov/research/visible/visible_human.html}{The Visible Human Project}~\cite{nlm1994visible} & 1994 & 2D, 3D & CT, MR, Others & Full Body & 2 & No & NA & Skin Lesion \\ \hline
\datasetidx \label{data:imageclef-2016}& \href{https://www.imageclef.org/2016/medical}{ImageCLEF 2016}~\cite{de2016overview} & 2015 & 2D & Multi\textsuperscript{b} & Skin, Cell, Breast & 31k & Yes & Cls & H\&N Tumor \\ \hline
\datasetidx \label{data:cmb-crc-mri}& \href{https://wiki.cancerimagingarchive.net/pages/viewpage.action?pageId=93257955}{CMB-CRC}~\cite{cancermoonshot2022cmbcrc} & 2022 & 2D, 3D & Multi\textsuperscript{c} & Colon & 472 & No & Seg, Cls & Colorectal Cancer \\ \hline
\datasetidx \label{data:cmb-mml}& \href{https://wiki.cancerimagingarchive.net/pages/viewpage.action?pageId=93258436}{CMB-MML}~\cite{cancerimagingarchive2022cmblca} & 2021 & 2D, 3D & Multi\textsuperscript{d} & NA & 60 & No & Pred & Multiple Myeloma \\ \hline
\datasetidx \label{data:cmb-pca-mri}& \href{https://www.cancerimagingarchive.net/collection/cmb-pca/}{CMB-PCA}~\cite{cmbpca2022} & 2022 & 2D, 3D & CT, MR, Histo & Prostate & 31 & No & Cls, Pred & Prostate Cancer \\ \hline
\datasetidx \label{data:icdc-glioma-3d-mr}& \href{https://wiki.cancerimagingarchive.net/pages/viewpage.action?pageId=70227341}{ICDC-Glioma (GLIOMA01)\_3D-MR} & 2021 & 2D, 3D & MR, Histo & NA & 650 & No & NA & Glioma \\ \hline
\datasetidx \label{data:prostate-fused-mri-pathology}& \href{https://www.cancerimagingarchive.net/collection/prostate-fused-mri-pathology/}{Prostate Fused-MRI-Pathology}~\cite{madabhushi2016fused} & 2016 & 2D, 3D & MR, Histo & Prostate (Pelvis) & 29 & No & NA & Prostate Cancer \\ \hline
\datasetidx \label{data:cardiac-atrial-images}& \href{https://www.heywhale.com/mw/dataset/5e4de9618ee624002d4c4117}{Cardiac Atrial Images}~\cite{heywhale2020cardiac} & 2020 & 2D & MR & Atrium & 8k & Yes & Seg & Cardiac Disease \\ \hline
\datasetidx \label{data:apollo-5}& \href{https://wiki.cancerimagingarchive.net/display/Public/APOLLO-5}{APOLLO-5}~\cite{tcia-apollo5} & 2022 & 2D, 3D & Multi\textsuperscript{e} & NA & 6.2k & No & NA & NA \\ \hline
\datasetidx & \href{https://wiki.cancerimagingarchive.net/pages/viewpage.action?pageId=93258420}{CMB-LCA}~\cite{cancerimagingarchive2022cmblca} & 2022 & 2D, 3D & Multi\textsuperscript{f} & NA & 0 & No & NA & Lung Cancer \\ \hline
\datasetidx & \href{https://wiki.cancerimagingarchive.net/pages/viewpage.action?pageId=119705284}{AHOD0831}~\cite{kelly2022ahod0831} & 2022 & 2D, 3D & Multi\textsuperscript{g} & NA & 0 & No & NA & Hodgkin Lymphoma \\ \hline
\datasetidx \label{data:prostate-mri-mri}& \href{https://wiki.cancerimagingarchive.net/display/Public/PROSTATE-MRI}{Prostate-MRI}~\cite{tcia_prostate_mri_2011} & 2011 & 2D, 3D & Multi\textsuperscript{h} & Prostate & 26 & No & NA & Prostate Cancer \\ \hline
\datasetidx & \href{https://wiki.cancerimagingarchive.net/pages/viewpage.action?pageId=109379682}{AREN0532}~\cite{fernandez2022aren0532} & 2022 & 2D, 3D & Multi\textsuperscript{i} & NA & 1k & No & NA & Wilms Tumor \\ \hline
\datasetidx \label{data:imageclef-2015}& \href{https://www.imageclef.org/2015}{ImageCLEF 2015}~\cite{garcia2015imageclef} & NA & 2D, 3D & Multi\textsuperscript{j} & Skin, Cell, Breast & 0 & Yes & Cls & NA \\ \hline
\datasetidx \label{data:radimagenet-subset-mr}& \href{https://www.radimagenet.com/}{RadImageNet (Subset: MR)}~\cite{mei2022radimagenet} & 2022 & 2D & MR & Full Body & 673k & Yes & Cls & Whole Body Abnorm. \\ \hline
\datasetidx \label{data:qubiq2020-mri}& \href{https://qubiq.grand-challenge.org/}{QUBIQ2020}~\cite{muller2020qubiq} & 2020 & 2D & CT, MR & Kidney, \etc\textsuperscript{k} & 150 & Yes & Seg & Pathologies \\ \hline
\datasetidx \label{data:qubiq2021-2d-mr}& \href{https://qubiq21.grand-challenge.org/QUBIQ2021/}{QUBIQ2021\_2D\_MR}~\cite{qubiq2021uncertainty} & 2021 & 2D & CT, MR & Kidney, \etc\textsuperscript{k} & 268 & Yes & Seg & Pathologies \\ \hline
& Overall & 1994$\sim$2022 & 2D & Multi & Full Body& 721.5k & NA & Multi & Multi \\ \hline
\end{tabular}}
\vspace{2pt}
\begin{tablenotes}[flushleft]
\scriptsize
\item[a] Multi-modalities of AREN0534: CT, MR, PET, US.
\item[b] Multi-modalities of ImageCLEF 2016: MR, US, Histo, X-Ray, CT, PET, Endo, Dermo, EEG, ECG, EMG, Micro, Fundus.
\item[c] Multi-modalities of CMB-CRC: CT, MR, US, X-Ray, PET, Histo.
\item[d] Multi-modalities of CMB-MML: CT, MR, PET, Histo.
\item[e] Multi-modalities of APOLLO-5: CT, MR, US, PET, X-Ray.
\item[f] Multi-modalities of CMB-LCA: CT, MR, US, Histo, X-Ray.
\item[g] Multi-modalities of AHOD0831: CT, MR, PET, X-Ray.
\item[h] Multi-modalities of Prostate-MRI: MR, CT, PET, Histo.
\item[i] Multi-modalities of AREN0532: CT, MR, US, PET.
\item[j] Multi-modalities of ImageCLEF 2015: MR, US, Histo, X-Ray.
\item[k] \etc in QUBIQ Structures: Pancreas, Brain, Prostate.
\item[] \textbf{Abbreviations:} Seg=Segmentation, Det=Detection, Cls=Classification, Pred=Prediction, Est=Estimation; Histo=Histopathology, US=Ultrasound, \\ Endo=Endoscopy, Dermo=Dermoscopy, Micro=Microscopy, Abnorm.=Abnormalities, H\&N=Head \& Neck.
\end{tablenotes}
\end{threeparttable}
\end{table*}

\begin{table*}[!h]
\small
\centering
\caption{2D PET datasets.}
\label{tab:2d_pet_datasets}
\begin{threeparttable}
\setlength{\tabcolsep}{4pt}      
\renewcommand{\arraystretch}{1.12} 
\resizebox{\textwidth}{!}{
\begin{tabular}{|c||>{\raggedright\arraybackslash}m{5.0cm}|r|c|c|c|c|c|c|>{\raggedright\arraybackslash}m{3.2cm}|}
\hline
\multicolumn{1}{|c||}{\textbf{\#}} &
\multicolumn{1}{c|}{\textbf{Dataset}} &
\multicolumn{1}{c|}{\textbf{Year}} &
\multicolumn{1}{c|}{\textbf{Dim}} &
\multicolumn{1}{c|}{\textbf{Modality}} &
\multicolumn{1}{c|}{\textbf{Structure}} &
\multicolumn{1}{c|}{\textbf{Images}} &
\multicolumn{1}{c|}{\textbf{Label}} &
\multicolumn{1}{c|}{\textbf{Task}} &
\multicolumn{1}{c|}{\textbf{Diseases}} \\
\hline
\datasetidx \label{data:aren0534-pet-2d}& \href{https://wiki.cancerimagingarchive.net/pages/viewpage.action?pageId=91357265}{AREN0534~\cite{ehrlich2021aren0534}} & 2021 & 3D, 2D & Multi\textsuperscript{a} & Kidney, Lung & 239 & No & Est & Kidney \\ \hline
\datasetidx \label{data:imageclef-2016-pet}& \href{https://www.imageclef.org/2016/medical}{ImageCLEF 2016~\cite{deherrera2016imageclef}} & 2015 & 2D & Multi\textsuperscript{b} & Skin, Cell, Breast & 31k & Yes & Cls & H\&N Tumor \\ \hline
\datasetidx \label{data:cmb-crc-pet}& \href{https://wiki.cancerimagingarchive.net/pages/viewpage.action?pageId=93257955}{CMB-CRC~\cite{cmbcrc2022}} & 2022 & 3D, 2D & Multi\textsuperscript{c} & Colon & 472 & No & Seg, Cls & Colorectal Cancer (H\&E stained tissue) \\ \hline
\datasetidx \label{data:cmb-gec}& \href{https://wiki.cancerimagingarchive.net/pages/viewpage.action?pageId=127665431}{CMB-GEC~\cite{cmbgec2022}} & 2022 & 3D, 2D & CT, WSI, PET & Brain & 14 & No & Seg, Cls & Melanoma (Cerebral microbleeds) \\ \hline
\datasetidx \label{data:cmb-mel}& \href{https://wiki.cancerimagingarchive.net/pages/viewpage.action?pageId=93258432}{CMB-MEL~\cite{cmbmel2022}} & 2022 & 3D, 2D & Multi\textsuperscript{d} & Brain & 255 & No & Seg & Melanoma (Cerebral microbleeds) \\ \hline
\datasetidx \label{data:cmb-mml-pet}& \href{https://wiki.cancerimagingarchive.net/pages/viewpage.action?pageId=93258436}{CMB-MML~\cite{cmbmml2022}} & 2021 & 2D, 3D & Multi\textsuperscript{e} & NA & 60 & No & Pred & Multiple Myeloma \\ \hline
\datasetidx \label{data:cptac_path}& \href{https://wiki.cancerimagingarchive.net/pages/viewpage.action?pageId=33948248}{CPTAC-LSCC\_CT\_PET~\cite{cptac2018the}} & 2018 & 2D, 3D & CT, PET, Histo & NA & 238 & No & NA & NA \\ \hline
\datasetidx \label{data:apollo-5-pet}& \href{https://wiki.cancerimagingarchive.net/display/Public/APOLLO-5}{APOLLO-5~\cite{apolloresearchnetwork2023applied}} & 2022 & 2D, 3D & Multi\textsuperscript{f} & NA & 6.2k & No & NA & NA \\ \hline
\datasetidx & \href{https://wiki.cancerimagingarchive.net/display/Public/RIDER+Phantom+PET-CT}{RIDER Phantom PET-CT~\cite{muzi2015data}} & 2011 & 2D & CT, PET & NA & 2.2k & No & NA & NA \\ \hline
\datasetidx & \href{https://wiki.cancerimagingarchive.net/pages/viewpage.action?pageId=119705284}{AHOD0831~\cite{kelly2022ahod0831}} & 2022 & 3D, 2D & Multi\textsuperscript{g} & NA & 0 & No & NA & Hodgkin Lymphoma \\ \hline
\datasetidx \label{data:aren0532-pet-2d}& \href{https://wiki.cancerimagingarchive.net/pages/viewpage.action?pageId=109379682}{AREN0532~\cite{fernandez2022aren0532}} & 2022 & 3D, 2D & Multi\textsuperscript{i} & NA & 1k & No & NA & Wilms Tumor \\ \hline
& Overall & 2011$\sim$2022 & 2D & Multi & Full Body& 41.7k & NA & Multi & Multi \\ \hline
\end{tabular}}
\vspace{2pt}
\begin{tablenotes}[flushleft]
\scriptsize
\item[a] Multi-modalities of AREN0534: CT, MR, PET, Ultrasound.
\item[b] Multi-modalities of ImageCLEF 2016: MR, US, Histo, X-Ray, CT, PET, Endo, Derm, EEG, ECG, EMG, Microscopy, Fundus.
\item[c] Multi-modalities of CMB-CRC: CT, MR, US, DX, PET, WSI.
\item[d] Multi-modalities of CMB-MEL: CT, US, WSI, PET (SWI).
\item[e] Multi-modalities of CMB-MML: CT, MR, PET, WSI.
\item[f] Multi-modalities of APOLLO-5: CT, MR, US, PET, X-Ray.
\item[g] Multi-modalities of AHOD0831: CT, MR, PET, X-Ray.
\item[h] Multi-modalities of Prostate-MRI: MR, CT, PET, Patho.
\item[i] Multi-modalities of AREN0532: CT, MR, US, PET.
\item[] \textbf{Abbreviations:} Seg=Segmentation, Cls=Classification, Est=Estimation, Pred=Prediction, H\&N=Head \& Neck, US=Ultrasound, \\ Histo=Histopathology, Patho=Pathology, Endo=Endoscopy, Derm=Dermoscopy, WSI=Whole-slide Images, DX=Digital Radiography.
\end{tablenotes}
\end{threeparttable}
\end{table*}

\begin{table*}[!h]
\small
\centering
\caption{2D Ultrasound datasets.}
\label{tab:2d_us_datasets}
\begin{threeparttable}
\setlength{\tabcolsep}{4pt}      
\renewcommand{\arraystretch}{1.12} 
\resizebox{\textwidth}{!}{
\begin{tabular}{|c||>{\raggedright\arraybackslash}m{5.0cm}|r|c|c|>{\raggedright\arraybackslash}m{2.5cm}|r|c|c|>{\raggedright\arraybackslash}m{3.2cm}|}
\hline
\multicolumn{1}{|c||}{\textbf{\#}} &
\multicolumn{1}{c|}{\textbf{Dataset}} &
\multicolumn{1}{c|}{\textbf{Year}} &
\multicolumn{1}{c|}{\textbf{Dim}} &
\multicolumn{1}{c|}{\textbf{Modality}} &
\multicolumn{1}{c|}{\textbf{Structure}} &
\multicolumn{1}{c|}{\textbf{Images}} &
\multicolumn{1}{c|}{\textbf{Label}} &
\multicolumn{1}{c|}{\textbf{Task}} &
\multicolumn{1}{c|}{\textbf{Diseases}} \\
\hline
\datasetidx\label{data:hc18} & \href{https://hc18.grand-challenge.org}{HC18}~\cite{van2018automated} & 2018 & 2D & US & Skull & 1.3k & Yes & Meas & NA \\ \hline
\datasetidx\label{data:busi} & \href{https://scholar.cu.edu.eg/?q=afahmy/pages/dataset}{BUSI}~\cite{al2020busi} & 2019 & 2D & US & Breast & 647 & Yes & Seg & Breast Cancer \\ \hline
\datasetidx\label{data:apollo5} & \href{https://wiki.cancerimagingarchive.net/display/Public/APOLLO-5}{APOLLO-5}~\cite{tcia-apollo5} & 2022 & 2D, 3D & Multi\textsuperscript{a} & NA & 6.2k & No & NA & NA \\ \hline
\datasetidx\label{data:cmblca-us-2d} & \href{https://wiki.cancerimagingarchive.net/pages/viewpage.action?pageId=93258420}{CMB-LCA}~\cite{cancerimagingarchive2022cmblca} & 2022 & 2D, 3D & Multi\textsuperscript{b} & NA & 0 & No & NA & NA \\ \hline
\datasetidx\label{data:imageclef2015} & \href{https://www.imageclef.org/2015/medical}{ImageCLEF 2015}~\cite{garcia2015imageclef} & 2015 & 2D, 3D & Multi\textsuperscript{c} & Skin, Cell, Breast & 0 & Yes & Cls & NA \\ \hline
\datasetidx\label{data:imageclef2016} & \href{https://www.imageclef.org/2016/medical}{ImageCLEF 2016}~\cite{deherrera2016imageclef} & 2016 & 2D & Multi\textsuperscript{d} & Skin, Cell, Breast & 31k & Yes & Cls & Head \& Neck Tumor \\ \hline
\datasetidx\label{data:radimagenet_us} & \href{https://www.radimagenet.com/}{RadImageNet (Subset: US)}~\cite{mei2022radimagenet} & 2022 & 2D & US & Full Body & 390k & Yes & Cls & Abdominal Structures \\ \hline
\datasetidx\label{data:breastmnist} & \href{https://medmnist.com/}{BreastMNIST}~\cite{al2020busi} & 2021 & 2D & US & Breast & 156 & Yes & Cls & Breast Cancer \\ \hline
\datasetidx\label{data:aren0534-us-2d} & \href{https://wiki.cancerimagingarchive.net/pages/viewpage.action?pageId=91357265}{AREN0534}~\cite{ehrlich2021aren0534} & 2021 & 2D, 3D & Multi\textsuperscript{e} & Kidney, Lung & 239 & No & Est & Kidney Tumor \\ \hline
\datasetidx\label{data:clust15} & \href{https://clust.ethz.ch/}{CLUST15}~\cite{de2015clust15} & 2015 & 2D & US & Liver & 34 & Yes & Track & NA \\ \hline
\datasetidx\label{data:uns} & \href{https://www.kaggle.com/competitions/ultrasound-nerve-segmentation}{Ultrasound Nerve Segmentation}~\cite{anna2016uns} & 2016 & 2D & US & Brachial Plexus & 11.3k & Yes & Seg & NA \\ \hline
\datasetidx\label{data:tnscui2020} & \href{https://tn-scui2020.grand-challenge.org/Home/}{TN-SCUI2020}~\cite{zhou2020thyroid} & 2020 & 2D & US & Thyroid & 3.6k & Yes & Seg & Leukemia \\ \hline
\datasetidx & \href{https://www.imageclef.org/2016/medical}{ImageCLEF 2016}~\cite{GSB2016} & 2015 & 2D & Multi\textsuperscript{f} & Skin, Cell, Breast & 31k & Yes & Cls & Head \& Neck Tumor \\ \hline
\datasetidx\label{data:cmbcrc-us-2d} & \href{https://wiki.cancerimagingarchive.net/pages/viewpage.action?pageId=93257955}{CMB-CRC}~\cite{cancermoonshot2022cmbcrc} & 2022 & 2D, 3D & Multi\textsuperscript{g} & Colon & 472 & No & Seg, Cls & Colorectal Cancer \\ \hline
\datasetidx\label{data:cmbmel-us-2d} & \href{https://www.cancerimagingarchive.net/collection/cmb-mel/}{CMB-MEL}~\cite{cancermoonshot2022cmbmel} & 2022 & 2D, 3D & Multi\textsuperscript{h} & Brain & 255 & No & Seg & Melanoma, Cerebral microbleed \\ \hline
\datasetidx\label{data:psfhs} & \href{https://ps-fh-aop-2023.grand-challenge.org/}{PSFHS}~\cite{lu2022psfhs} & 2023 & 2D & US & NA & 4.7k & Yes & Seg & NA \\ \hline
\datasetidx\label{data:usenhance2023} & \href{https://ultrasoundenhance2023.grand-challenge.org/}{USenhance2023}~\cite{guo2023usenhance} & 2023 & 2D & US & NA & 1.5k & Yes & Recon & NA \\ \hline
\datasetidx\label{data:aren0532-us-2d} & \href{https://www.cancerimagingarchive.net/collection/aren0532/}{AREN0532}~\cite{fernandez2022aren0532} & 2022 & 2D, 3D & Multi\textsuperscript{i} & NA & 1k & No & NA & Wilms Tumor \\ \hline
\datasetidx\label{data:tn3k} & \href{https://github.com/haifangong/TRFE-Net-for-thyroid-nodule-segmentation}{TN3K}~\cite{gong2021multitask} & 2021 & 2D & US & Head and Neck & 3.5k & Yes & Seg & Thyroid Nodules \\ \hline
\datasetidx\label{data:camus} & \href{http://camus.creatis.insa-lyon.fr/challenge/}{CAMUS}~\cite{sarah2019camus} & 2019 & 2D & US & Heart & 1.8k & Yes & Seg & Cardiac Disease \\ \hline
\datasetidx\label{data:ddti} & \href{https://github.com/haifangong/TRFE-Net-for-thyroid-nodule-segmentation}{DDTI~\cite{gong2021multi-task}} & 2020 & 2D & US & Thyroid & 637 & Yes & Seg & Thyroid Nodule \\ \hline
\datasetidx\label{data:udiat-b} & \href{http://www2.docm.mmu.ac.uk/STAFF/m.yap/dataset.php}{UDIAT-B~\cite{yap2017automated}} & 2017 & 2D & US & Breast & 163 & Yes & Det & Breast Lesion \\ \hline
\datasetidx\label{data:oasbud} & \href{https://doi.org/10.5281/zenodo.545928/}{OASBUD~\cite{piotrzkowska2017open}} & 2017 & 2D & US & Breast & 200 & Yes & Seg & Breast Cancer \\ \hline
\datasetidx\label{data:breast} & \href{https://doi.org/10.7937/9wkk-q141/}{BrEaST~\cite{pawlowska2024curated}} & 2024 & 2D & US & Breast & 256 & Yes & Cls & Breast Cancer \\ \hline
& Overall & 2015$\sim$2024 & 2D & Multi & Full Body& 490.0k & NA & Multi & Multi \\ \hline
\end{tabular}}
\vspace{2pt}
\begin{tablenotes}[flushleft]
\scriptsize
\item[a] Multi-modalities of APOLLO-5: CT, MR, US, PET, X-Ray.
\item[b] Multi-modalities of CMB-LCA: CT, MR, US, Histo, DX (WSI).
\item[c] Multi-modalities of ImageCLEF 2015: MR, US, Histo, X-Ray.
\item[d] Multi-modalities of ImageCLEF 2016~\cite{deherrera2016imageclef}: MR, US, Histo, X-Ray, CT, PET, Endo, Dermo, EEG, ECG, EMG, Microscopy, Fundus (Electron Microscopy).
\item[e] Multi-modalities of AREN0534: CT, MR, PET, US.
\item[f] Multi-modalities of ImageCLEF 2016: MR, US, Histo, X-Ray, CT, PET, Endo, Dermo, Others, EEG, ECG, EMG, Electron Microscopy, Fundus Photography.
\item[g] Multi-modalities of CMB-CRC: CT, MR, US, DX, PET, WSI.
\item[h] Multi-modalities of CMB-MEL: CT, US, WSI, PET (SWI).
\item[i] Multi-modalities of AREN0532: CT, MR, US, PET.
\item[] \textbf{Abbreviations:} Seg=Segmentation, Cls=Classification, Est=Estimation, Recon=Reconstruction, Meas=Measurement, Track=Tracking; \\US=Ultrasound, Histo=Histopathology, WSI=Whole-slide image, Endo=Endoscopy, Dermo=Dermoscopy.
\end{tablenotes}
\end{threeparttable}
\end{table*}

\begin{table*}[!h]
\small
\centering
\caption{2D X-Ray datasets.}
\label{tab:2d_xray_datasets}
\begin{threeparttable}
\setlength{\tabcolsep}{4pt}      
\renewcommand{\arraystretch}{1.12} 
\resizebox{\textwidth}{!}{
\begin{tabular}{|c||>{\raggedright\arraybackslash}m{5.0cm}|r|c|>{\raggedright\arraybackslash}m{2.0cm}|>{\raggedright\arraybackslash}m{3.2cm}|r|c|>{\raggedright\arraybackslash}m{1.5cm}|>{\raggedright\arraybackslash}m{3.2cm}|}
\hline
\multicolumn{1}{|c||}{\textbf{\#}} &
\multicolumn{1}{c|}{\textbf{Dataset}} &
\multicolumn{1}{c|}{\textbf{Year}} &
\multicolumn{1}{c|}{\textbf{Dim}} &
\multicolumn{1}{c|}{\textbf{Modality}} &
\multicolumn{1}{c|}{\textbf{Structure}} &
\multicolumn{1}{c|}{\textbf{Images}} &
\multicolumn{1}{c|}{\textbf{Label}} &
\multicolumn{1}{c|}{\textbf{Task}} &
\multicolumn{1}{c|}{\textbf{Diseases}} \\
\hline
\datasetidx\label{data:chestxray_pneumonia} & \href{https://www.kaggle.com/paultimothymooney/chest-xray-pneumonia}{Chest X-ray}~\cite{kermany2018pneumnist} & 2018 & 2D & X-Ray & Lung & 5.9k & Yes & Cls & Pneumonia \\ \hline
\datasetidx\label{data:coronahack} & \href{https://www.kaggle.com/praveengovi/coronahack-chest-xraydataset}{CoronaHack}~\cite{praveen2019coronahack} & 2020 & 2D & X-Ray & Lung & 5.9k & Yes & Cls & COVID-19, Pneumonia \\ \hline
\datasetidx\label{data:nih_chestxray14} & \href{https://www.kaggle.com/nih-chest-xrays/data}{NIH Chest X-ray 14}~\cite{wang2017chestxray8} & 2017 & 2D & X-Ray & Lung & 112.1k & Yes & Cls & Thorax diseases \\ \hline
\datasetidx\label{data:covidx_cxr2} & \href{https://www.kaggle.com/andyczhao/covidx-cxr2}{COVIDx CXR-2}~\cite{wang2020covidx} & 2020 & 2D & X-Ray & Lung & 30.9k & Yes & Cls & COVID-19 \\ \hline
\datasetidx\label{data:pneumothorax_masks} & \href{https://www.kaggle.com/vbookshelf/pneumothorax-chest-xray-images-and-masks}{Pneumothorax Masks X-Ray}~\cite{anna2019siimacrpneumothorax} & 2020 & 2D & X-Ray & Lung & 12.0k & Yes & Seg & Pneumothorax \\ \hline
\datasetidx\label{data:irma_xray} & \href{https://www.kaggle.com/raddar/irma-xray-dataset}{IRMA X-ray}~\cite{raddar2020irma} & 2020 & 2D & X-Ray & Brain, Lung & 14.7k & Yes & Cls & NA \\ \hline
\datasetidx\label{data:chestxr_covid19} & \href{https://cxr-covid19.grand-challenge.org}{Chest XR COVID-19}~\cite{moulay2021chestxrcovid19} & 2021 & 2D & X-Ray & Lung & 21.4k & Yes & Cls & COVID-19 \\ \hline
\datasetidx\label{data:covid19_image} & \href{https://github.com/ieee8023/covid-chestxray-dataset}{COVID-19-Image}~\cite{cohen2020covid} & 2020 & 2D & X-Ray & Lung & 93 & Yes & Cls & COVID-19 \\ \hline
\datasetidx\label{data:chestxray_pa} & \href{https://data.mendeley.com/datasets/jctsfj2sfn/1}{Chest X-ray PA Dataset}~\cite{asraf2021covid19} & 2021 & 2D & X-Ray & Lung & 4.6k & No & Cls & COVID-19, Pneumonia \\ \hline
\datasetidx\label{data:nhanes2_xray} & \href{https://www.nlm.nih.gov/databases/download/nhanes.html}{NHANES II X-ray}~\cite{long2003nhanes} & 2021 & 2D & X-Ray & Lung & 17.1k & No & NA & NA \\ \hline
\datasetidx\label{data:knoap2020-xray} & \href{https://knoap2020.grand-challenge.org/Home/}{KNOAP2020}~\cite{hirvasniemi2023knee} & 2020 & 2D, 3D & MR, X-Ray & Knee & 30 & Yes & Pred & Osteoarthritis \\ \hline
\datasetidx\label{data:aasce} & \href{https://aasce19.grand-challenge.org/Award/}{AASCE}~\cite{wang2021aasce} & 2019 & 2D & X-Ray & Spine & 609 & Yes & Reg & NA \\ \hline
\datasetidx\label{data:covid19_image_dataset} & \href{https://www.kaggle.com/datasets/pranavraikokte/covid19-image-dataset}{Covid-19 Image Dataset}~\cite{pranav2020covid19} & 2021 & 2D & X-Ray & Lung & 345 & Yes & Cls & Lung diseases \\ \hline
\datasetidx\label{data:pulmonary_cls} & \href{https://tianchi.aliyun.com/dataset/dataDetail?dataId=94377}{Pulmonary Chest X-Ray (ChinaSet)}~\cite{jaeger2014two} & 2021 & 2D & X-Ray & Lung & 800 & Yes & Cls & Lung diseases \\ \hline
\datasetidx\label{data:mura} & \href{https://stanfordmlgroup.github.io/competitions/mura/}{MURA}~\cite{rajpurkar2017mura} & 2021 & 2D & X-Ray & Multi-bone\textsuperscript{a} & 40.0k & Yes & Cls & Musculoskeletal \\ \hline
\datasetidx\label{data:siim_acr_pneumothorax} & \href{https://www.kaggle.com/c/siim-acr-pneumothorax-segmentation/data}{SIIM-ACR Pneumothorax Seg}~\cite{anna2019siimacrpneumothorax} & 2020 & 2D & X-Ray & Lung & 12.1k & Yes & Seg & Pneumothorax \\ \hline
\datasetidx\label{data:mias} & \href{http://peipa.essex.ac.uk/info/mias.html}{MIAS Mammography}~\cite{suckling1994mammographic} & 2021 & 2D & X-Ray & Breast & 322 & Yes & Cls & Breast cancer \\ \hline
\datasetidx\label{data:medmnist_xray} & \href{https://medmnist.com/v1}{MedMNIST}~\cite{yang2021medmnist} & 2020 & 2D & Multi\textsuperscript{b} & Retina, Breast, Lung & 100k & Yes & Cls & Multi-diseases \\ \hline
\datasetidx\label{data:rsna_pneumonia_det} & \href{https://www.kaggle.com/competitions/rsna-pneumonia-detection-challenge}{RSNA Pneumonia Detection}~\cite{stein2018rsnapneumoniadetection} & 2018 & 2D & X-Ray & Lung & 26.7k & Yes & Det & Lung diseases \\ \hline
\datasetidx\label{data:vinbigdata} & \href{https://www.kaggle.com/competitions/vinbigdata-chest-xray-abnormalities-detection}{VinBigData Chest X-ray}~\cite{nguyen2020vinbigdata} & 2020 & 2D & X-Ray & Lung & 15.0k & Yes & Det & Heart atrium \\ \hline
\datasetidx\label{data:chexpert} & \href{https://stanfordmlgroup.github.io/competitions/chexpert/}{CheXpert}~\cite{irvin2019chexpert} & 2021 & 2D & X-Ray & Lung & 224.3k & Yes & Cls & Diabetic retinopathy \\ \hline
\datasetidx\label{data:siim_covid19} & \href{https://www.kaggle.com/competitions/siim-covid19-detection/data}{SIIM-FISABIO-RSNA COVID-19}~\cite{lakhani2023siimfisabio} & 2021 & 2D & X-Ray & Lung & 6.1k & Yes & Det & Tuberculosis \\ \hline
\datasetidx\label{data:node21} & \href{https://node21.grand-challenge.org/}{NODE21}~\cite{sogancioglu2024nodule} & 2021 & 2D & X-Ray & Lung & 5.5k & Yes & Det & Breast cancer \\ \hline
\datasetidx\label{data:imageclef2016_xray} & \href{https://www.imageclef.org/2016/medical}{ImageCLEF 2016}~\cite{deherrera2016imageclef} & 2016 & 2D & Multi\textsuperscript{c} & Skin, Cell, Breast & 31.0k & Yes & Cls & Head \& Neck tumor \\ \hline
\datasetidx\label{data:tcb_challenge} & \href{https://www.idpoisson.fr/tcbchallenge/}{TCB-Challenge}~\cite{zheng2016bone} & 2016 & 2D & X-Ray & Bone & 174 & Yes & Cls & Osteoporotic bone \\ \hline
\datasetidx\label{data:crass} & \href{https://crass.grand-challenge.org/Home/}{CRASS}~\cite{hogeweg2012clavicle} & 2012 & 2D & X-Ray & Clavicle & 518 & Yes & Seg & Clavicles \\ \hline
\datasetidx\label{data:covidgr} & \href{https://github.com/ari-dasci/OD-covidgr}{COVIDGR}~\cite{tabik2020covidgr} & 2020 & 2D & X-Ray & Lung & 852 & Yes & Cls & COVID-19 \\ \hline
\datasetidx\label{data:chestxdet} & \href{https://opendatalab.com/OpenDataLab/ChestX-Det}{ChestX-Det}~\cite{lian2021chestxdet} & 2021 & 2D & X-Ray & Lung & 3.6k & Yes & Seg & Lung diseases \\ \hline
\datasetidx\label{data:ranzcr_clip} & \href{https://www.kaggle.com/competitions/ranzcr-clip-catheter-line-classification/data}{RANZCR CLiP}~\cite{seah2020ranzcr} & 2020 & 2D & X-Ray & Breast & 30.1k & Yes & Cls & NA \\ \hline
\datasetidx\label{data:cpcxr} & \href{https://opendatalab.org.cn/CPCXR/download}{CPCXR}~\cite{singh2020cpcxr} & 2020 & 2D & X-Ray & Lung & 1.2k & Yes & NA & Pneumonia, COVID-19 \\ \hline
\datasetidx\label{data:jsrt} & \href{http://db.jsrt.or.jp/eng.php}{JSRT}~\cite{shiraishi2000jsrt} & 2000 & 2D & X-Ray & Lung & 247 & Yes & Cls & Lung nodule \\ \hline
\datasetidx\label{data:synthetic_covid_cxr} & \href{https://opendatalab.org.cn/Synthetic_COVID-19_CXR_Dataset}{Synthetic COVID-19 CXR}~\cite{zunair2021synthesis} & 2020 & 2D & X-Ray & Lung & 21.3k & Yes & Cls, Gen & COVID-19 \\ \hline
\datasetidx\label{data:ceph_xray} & \href{https://opendatalab.org.cn/Cephalometric_X-ray_Image}{Cephalometric X-ray Image}~\cite{wang2015evaluation} & 2014 & 2D & X-Ray & Skull & 400 & Yes & Loc & NA \\ \hline
\datasetidx\label{data:cmbcrc_xray} & \href{https://wiki.cancerimagingarchive.net/pages/viewpage.action?pageId=93257955}{CMB-CRC}~\cite{cancermoonshot2022cmbcrc} & 2022 & 2D, 3D & Multi\textsuperscript{d} & Colon & 472 & No & Seg, Cls & Colorectal cancer \\ \hline
\datasetidx\label{data:midrc_ricord1c} & \href{https://wiki.cancerimagingarchive.net/pages/viewpage.action?pageId=70230281}{MIDRC-RICORD-1c}~\cite{tsai2021midrcricord} & 2021 & 2D & X-Ray & Lung & 1.3k & Yes & Cls & NA \\ \hline
\datasetidx\label{data:chestxray_imaging} & \href{https://www.heywhale.com/mw/dataset/62c2ac49913a54a66037f872}{Chest X-ray Imaging}~\cite{kermany2018identifying} & 2017 & 2D & X-Ray & Lung & 5.9k & Yes & Cls & NA \\ \hline
\datasetidx\label{data:covid19_db} & \href{https://www.heywhale.com/mw/dataset/6027caee891f960015c863d7}{COVID-19 Chest X-ray DB}~\cite{chowdhury2020can} & 2021 & 2D & X-Ray & NA & 3.9k & Yes & Cls & COVID-19 \\ \hline
\datasetidx\label{data:sz_cxr} & \href{https://arxiv.org/pdf/1803.01199v1.pdf}{SZ-CXR~\cite{stirenko2018chest}} & 2018 & 2D & X-Ray & Lung & 566 & Yes & Seg & NA \\ \hline
\datasetidx\label{data:pulmonary_seg} & \href{https://tianchi.aliyun.com/dataset/dataDetail?dataId=94377}{Pulmonary Chest X-Ray Seg}~\cite{jaeger2014two} & 2021 & 2D & X-Ray & Lung & 800 & Yes & Seg & Lung diseases \\ \hline
\datasetidx\label{data:dentex} & \href{https://dentex.grand-challenge.org/data/}{DENTEX}~\cite{hamamci2023dentex} & 2023 & 2D & X-Ray & Brain & 1.0k & Yes & Det & NA \\ \hline
\datasetidx\label{data:cl_detection2023} & \href{https://cl-detection2023.grand-challenge.org/}{CL-Detection2023}~\cite{cao2023cephalometric} & 2023 & 2D & X-Ray & NA & 555 & Yes & Det & NA \\ \hline
\datasetidx\label{data:cepha29} & \href{http://vision.seecs.edu.pk/CEPHA29/}{ISBI2023 CEPHA29}~\cite{khalid2022cepha29} & NA & 2D & X-Ray & NA & 1.0k & Yes & Loc & NA \\ \hline
\datasetidx\label{data:arcade} & \href{https://arcade.grand-challenge.org/arcade/}{ARCADE}~\cite{popov2024arcade} & 2023 & 2D & X-Ray & NA & 1.5k & Yes & Seg & NA \\ \hline
\datasetidx\label{data:medfm2023} & \href{https://medfm2023.grand-challenge.org/}{MedFM2023}~\cite{wang2023real} & 2023 & 2D & X-Ray & NA & 4.8k & Yes & Cls & NA \\ \hline
\datasetidx\label{data:coronare} & \href{https://coronare.grand-challenge.org/}{CoronARe}~\cite{ccimen2017coronare} & NA & 2D & X-Ray & NA & 0 & Yes & Recon & Coronary artery diseases \\ \hline
\datasetidx\label{data:victre} & \href{https://www.cancerimagingarchive.net/collection/victre/}{VICTRE}~\cite{badano2019victre} & 2019 & 2D & X-Ray & Breast & 217.9k & No & NA & NA \\ \hline
\datasetidx\label{data:apollo5_xray} & \href{https://wiki.cancerimagingarchive.net/display/Public/APOLLO-5}{APOLLO-5}~\cite{tcia-apollo5} & 2022 & 2D, 3D & Multi\textsuperscript{e} & NA & 6.2k & No & NA & NA \\ \hline
\datasetidx\label{data:cmblca_xray} & \href{https://wiki.cancerimagingarchive.net/pages/viewpage.action?pageId=93258420}{CMB-LCA}~\cite{cancerimagingarchive2022cmblca} & 2022 & 2D, 3D & Multi\textsuperscript{f} & NA & 0 & No & NA & NA \\ \hline
\datasetidx\label{data:ahod0831-xray} & \href{https://wiki.cancerimagingarchive.net/pages/viewpage.action?pageId=119705284}{AHOD0831}~\cite{kelly2022ahod0831} & 2022 & 2D, 3D & Multi\textsuperscript{g} & NA & 0 & No & NA & Hodgkin Lymphoma \\ \hline
\datasetidx\label{data:chexmask} & \href{https://physionet.org/content/chexmask-cxr-segmentation-data/0.1/}{CheXmask}~\cite{gaggion2023chexmask} & 2023 & 2D & X-Ray & NA & 676.8k & Yes & Seg & Lung diseases \\ \hline
\datasetidx\label{data:knee_osteo} & \href{https://www.kaggle.com/datasets/shashwatwork/knee-osteoarthritis-dataset-with-severity/data}{Knee Osteoarthritis Dataset}~\cite{chen2018knee} & 2020 & 2D & X-Ray & Knee & 0 & Yes & Cls & Knee osteoarthritis \\ \hline
\datasetidx\label{data:rus_chn} & \href{https://aistudio.baidu.com/datasetdetail/69582/0}{RUS\_CHN}~\cite{baidu2021ruschn} & 2021 & 2D & X-Ray & Hand & 0 & Yes & Cls & Hand joints \\ \hline
\datasetidx\label{data:rsna_bone_age} & \href{https://www.rsna.org/rsnai/ai-image-challenge/rsna-pediatric-bone-age-challenge-2017}{RSNA Bone Age}~\cite{halabi2019rsnabone} & 2017 & 2D & X-Ray & Hand & 14.2k & Yes & Est & Hand bone \\ \hline
\datasetidx\label{data:cxr_lt} & \href{https://physionet.org/content/cxr-lt-iccv-workshop-cvamd/2.0.0/}{CXR-LT}~\cite{lin2025cxrlt} & 2023 & 2D & X-Ray & Breast, Lung & 377.1k & Yes & Cls & Multi-diseases \\ \hline
\datasetidx\label{data:pengwin2024} & \href{https://pengwin.grand-challenge.org/}{PENGWIN2024-Task2}~\cite{liu2025automatic,liu2025preoperative} & 2025 & 2D & X-Ray & Pelvic Bone & 150 & Yes & Seg & Pelvic bone fragments \\ \hline
\datasetidx\label{data:icgcxr} & \href{https://progemu.github.io/}{ICG-CXR}~\cite{ma2025towards} & 2025 & 2D & X-Ray & Lung & 11.4k & Yes & Gen & Lung diseases \\ \hline
& Overall & 2014$\sim$2025 & 2D & Multi & Full Body& 2.1m & NA & Multi & Multi \\ \hline
\end{tabular}}
\vspace{2pt}
\begin{tablenotes}[flushleft]
\scriptsize
\item[a] Structures of MURA: Elbow, Finger, Forearm, Hand, Humerus, Shoulder, Wrist.
\item[b] Multi-modalities of MedMNIST: OCT, X-Ray, CT, Pathology, Fundus.
\item[c] Multi-modalities of ImageCLEF 2016: MR, US, Histo, X-Ray, CT, PET, Endo, Derm, EEG, ECG, EMG, Microscopy, Fundus.
\item[d] Multi-modalities of CMB-CRC: CT, MR, US, DX, PET, WSI.
\item[e] Multi-modalities of APOLLO-5: CT, MR, US, PET, X-Ray.
\item[f] Multi-modalities of CMB-LCA: CT, MR, US, Histo, DX.
\item[g] Multi-modalities of AHOD0831: CT, MR, PET, X-Ray.
\item[] \textbf{Abbreviations:} Seg=Segmentation, Det=Detection, Cls=Classification, Recon=Reconstruction, Reg=Registration, Loc=Localization, \\Est=Estimation, Pred=Prediction, Gen=Generation.
\end{tablenotes}
\end{threeparttable}
\end{table*}

\begin{table*}[!h]
\small
\centering
\caption{2D OCT datasets.}
\label{tab:2d_oct_datasets}
\begin{threeparttable}
\setlength{\tabcolsep}{4pt}      
\renewcommand{\arraystretch}{1.12} 
\resizebox{\textwidth}{!}{
\begin{tabular}{|c||>{\raggedright\arraybackslash}m{5.8cm}|r|c|c|>{\raggedright\arraybackslash}m{2.8cm}|c|c|c|>{\raggedright\arraybackslash}m{3.2cm}|}
\hline
\multicolumn{1}{|c||}{\textbf{\#}} &
\multicolumn{1}{c|}{\textbf{Dataset}} &
\multicolumn{1}{c|}{\textbf{Year}} &
\multicolumn{1}{c|}{\textbf{Dim}} &
\multicolumn{1}{c|}{\textbf{Modality}} &
\multicolumn{1}{c|}{\textbf{Structure}} &
\multicolumn{1}{c|}{\textbf{Images}} &
\multicolumn{1}{c|}{\textbf{Label}} &
\multicolumn{1}{c|}{\textbf{Task}} &
\multicolumn{1}{c|}{\textbf{Diseases}} \\
\hline
\datasetidx\label{data:oct2017} & \href{https://www.kaggle.com/paultimothymooney/kermany2018/code}{OCT2017~\cite{kermany2018identifying}} & 2018 & 2D & OCT & Retina & 83.5k & Yes & Cls & NA \\ \hline
\datasetidx\label{data:retinal_oct_c8} & \href{https://www.kaggle.com/obulisainaren/retinal-oct-c8}{Retinal OCT - C8~\cite{yang2023medmnist}} & 2021 & 2D & OCT & Retina & 24k & Yes & Cls & NA \\ \hline
\datasetidx\label{data:ichallenge_age19} & \href{https://age.grand-challenge.org/}{iChallenge - AGE19~\cite{fu2020age}} & 2019 & 2D & OCT & Retina & 1.6k & Yes & Cls & NA \\ \hline
\datasetidx\label{data:drac22} & \href{https://drac22.grand-challenge.org/}{DRAC22~\cite{qian2024drac}} & 2022 & 2D & OCT & Retina & 174 & Yes & Seg & Diabetic Retinopathy Lesions \\ \hline
\datasetidx\label{data:ichallenge_goals} & \href{https://ichallenges.grand-challenge.org/iChallenge-GON3/}{iChallenge - GOALS~\cite{fang2022dataset}} & 2022 & 2D & OCT & Retina & 300 & Yes & Seg & NA \\ \hline
\datasetidx\label{data:eye_oct} & \href{https://tianchi.aliyun.com/dataset/dataDetail?dataId=90672}{Eye OCT Datasets}~\cite{jahromi2014automatic,eye_oct_1} & 2021 & 2D & OCT & Retina & 148 & Yes & Cls & NA \\ \hline
\datasetidx\label{data:aptos2021} & \href{https://tianchi.aliyun.com/dataset/dataDetail?dataId=120006}{APTOS-2021~\cite{zhang2025predicting}} & 2022 & 2D & OCT & Retina & 2.6k & Yes & Pred & Diabetic Retinopathy \\ \hline
\datasetidx\label{data:aptos_stage1} & \href{https://tianchi.aliyun.com/dataset/dataDetail?dataId=127971}{APTOS Cross-Country Datasets\_stage1~\cite{zhang2025predicting}} & 2022 & 2D & OCT & Retina & 2.6k & Yes & Pred & NA \\ \hline
\datasetidx\label{data:medmnist_oct} & \href{https://medmnist.com/v1}{MedMNIST~\cite{yang2021medmnist}} & 2020 & 2D & Multi\textsuperscript{a} & Retina, Breast, Lung & 100k & Yes & Cls & NA \\ \hline
\datasetidx\label{data:canada_oct} & 
\href{https://borealisdata.ca/dataset.xhtml?persistentId=doi:10.5683/SP/UIOXXK}{Canada OCT Retinal Images (Subset)~\cite{gholami2020octid}} & 2018 & 2D & OCT & Retina & 25 & Yes & Seg & Retinal Structures \\ \hline
\datasetidx\label{data:sinafarsiu002} & \href{https://people.duke.edu/~sf59/software.html}{SinaFarsiu-002-Fang\_TMI\_2013~\cite{fang2012sparsity}} & 2013 & 2D & OCT & Retina & 195 & Yes & Seg & NA \\ \hline
\datasetidx\label{data:sinafarsiu003} & \href{https://people.duke.edu/~sf59/software.html}{SinaFarsiu-003-Fang\_BOE\_2012~\cite{fang2012sparsity}} & 2012 & 2D & OCT & Retina & 51 & Yes & Seg & NA \\ \hline
\datasetidx\label{data:sinafarsiu008} & \href{https://people.duke.edu/~sf59/software.html}{SinaFarsiu-008-Chiu\_BOE\_2012~\cite{fang2012sparsity}} & 2012 & 2D & OCT & Retina & 23 & Yes & Seg & NA \\ \hline
\datasetidx\label{data:sinafarsiu009} & \href{https://people.duke.edu/~sf59/software.html}{SinaFarsiu-009-Chiu\_BOE\_2013~\cite{fang2012sparsity}} & 2013 & 2D & OCT & Retina & 840 & Yes & Seg & NA \\ \hline
\datasetidx\label{data:sinafarsiu010} & \href{https://people.duke.edu/~sf59/software.html}{SinaFarsiu-010-Rabbani\_IOVS\_2014~\cite{rabbani2015fully}} & 2015 & 2D & OCT & Retina & 24 & Yes & Seg & NA \\ \hline
\datasetidx\label{data:sinafarsiu012} & \href{https://people.duke.edu/~sf59/software.html}{SinaFarsiu-012-Estrada\_TMI\_2015~\cite{estrada2015retinal}} & 2015 & 2D & OCT & Retina & 60 & Yes & Seg & NA \\ \hline
\datasetidx\label{data:sinafarsiu013} & \href{https://people.duke.edu/~sf59/software.html}{SinaFarsiu-013-Estrada\_PAMI\_2015~\cite{estrada2014tree}} & 2015 & 2D & OCT & Retina & 90 & Yes & Seg & NA \\ \hline
\datasetidx\label{data:sinafarsiu018} & \href{https://people.duke.edu/~sf59/software.html}{SinaFarsiu-018-Yang\_BOE\_2021~\cite{yang2021connectivity}} & 2021 & 2D & OCT & Retina & 784 & Yes & Seg & NA \\ \hline
\datasetidx\label{data:aptos_stage2} & \href{https://tianchi.aliyun.com/dataset/dataDetail?dataId=127971}{APTOS Cross-Country Datasets\_stage2~\cite{zhang2025predicting}} & 2022 & 2D & OCT & Retina & 3.3k & Yes & Pred & Diabetic Retinopathy \\ \hline
\datasetidx & \href{https://ieee-dataport.org/open-access/octa-500}{OCTA-500\_2D-Fundus~\cite{octa500}} & 2020 & 2D & OCT & Retina & 500 & Yes & Seg & N/A \\ \hline
\datasetidx & \href{https://github.com/xmed-lab/MuTri}{OCTA2024 (MuTri)\_2D-Fundus~\cite{chen2025mutri}} & 2024 & 2D & OCT & Retina & 848 & Yes & Seg & NA \\ \hline
& Overall & 2012$\sim$2022 & 2D & Multi & Retina, Breast, Lung& 221.7k & Yes & Multi & Multi \\ \hline
\end{tabular}}
\vspace{2pt}
\begin{tablenotes}[flushleft]
\scriptsize
\item[a] Multi-modalities of MedMNIST: OCT, X-Ray, CT, Pathology, Fundus Photography.
\item[] \textbf{Abbreviations:} Seg=Segmentation, Cls=Classification, Pred=Prediction, Det=Detection, Recon=Reconstruction, Reg=Registration, \\ Loc=Localization, Est=Estimation.
\end{tablenotes}
\end{threeparttable}
\end{table*}
\begin{table*}[!htbp]
\small
\centering
\caption{2D fundus datasets.}
\label{tab:2d_fundus_datasets}
\resizebox{\textwidth}{!}{%
\begin{threeparttable}
\setlength{\tabcolsep}{4pt}
\renewcommand{\arraystretch}{1.12}
\begin{tabular}{|c||>{\raggedright\arraybackslash}m{5.0cm}|r|c|c|c|c|c|c|>{\raggedright\arraybackslash}m{3.2cm}|}
\hline
\multicolumn{1}{|c||}{\textbf{\#}} &
\multicolumn{1}{c|}{\textbf{Dataset}} &
\multicolumn{1}{c|}{\textbf{Year}} &
\multicolumn{1}{c|}{\textbf{Dim}} &
\multicolumn{1}{c|}{\textbf{Modality}} &
\multicolumn{1}{c|}{\textbf{Structure}} &
\multicolumn{1}{c|}{\textbf{Images}} &
\multicolumn{1}{c|}{\textbf{Label}} &
\multicolumn{1}{c|}{\textbf{Task}} &
\multicolumn{1}{c|}{\textbf{Diseases}} \\
\hline
\datasetidx\label{data:fundus_drishti_gs} & \href{http://cvit.iiit.ac.in/projects/mip/drishti-gs/mip-dataset2/Home.php}{DRISHTI-GS~\cite{sivaswamy2014drishti}} & 2014 & 2D & Fundus Photo & Retina & 101 & Yes & Seg & Optic Disc \\ \hline
\datasetidx\label{data:fundus_chase} & \href{https://blogs.kingston.ac.uk/retinal/chasedb1/}{CHASE~\cite{CHASEDB1}} & 2009 & 2D & Fundus Photo & Retina & 28 & Yes & Seg & NA \\ \hline
\datasetidx\label{data:fundus_stare} & \href{http://cecas.clemson.edu/~ahoover/stare/}{STARE~\cite{hoover2000locating}} & 2004 & 2D & Fundus Photo & Retina & 40 & Yes & Seg & NA \\ \hline
\datasetidx\label{data:fundus_drive} & \href{https://drive.grand-challenge.org/}{DRIVE~\cite{staal2004ridge}} & 2003 & 2D & Fundus Photo & Retina & 40 & Yes & Seg & NA \\ \hline
\datasetidx\label{data:fundus_idrid2018} & \href{https://idrid.grand-challenge.org/}{IDRID2018~\cite{porwal2018indian}} & 2018 & 2D & Fundus Photo & Retina & 81 & Yes & Seg, Cls & Diabetic Retinopathy \\ \hline
\datasetidx\label{data:fundus_eyepacs} & \href{https://www.kaggle.com/c/diabetic-retinopathy-detection/data}{EyePACS~\cite{cuadros2009eyepacs}} & 2015 & 2D & Fundus Photo & Retina & 88.7k & Yes & Cls & Diabetic Retinopathy \\ \hline
\datasetidx\label{data:fundus_drhagis} & \href{https://paperswithcode.com/dataset/dr-hagis}{DRHAGIS~\cite{holm2017dr}} & 2017 & 2D & Fundus Photo & Retina & 40 & Yes & Seg & DR Lesions \\ \hline
\datasetidx\label{data:fundus_odir} & \href{https://odir2019.grand-challenge.org/}{ODIR~\cite{li2020benchmark}} & 2019 & 2D & Fundus Photo & Retina & 8k & Yes & Cls & Ocular Diseases (DR screening) \\ \hline
\datasetidx\label{data:fundus_riadd} & \href{https://riadd.grand-challenge.org/}{RIADD (RFMiD)~\cite{pachade2021retinal}} & 2020 & 2D & Fundus Photo & Retina & 3.2k & Yes & Cls & Retinal Diseases \\ \hline
\datasetidx\label{data:fundus_messidor2} & \href{https://www.adcis.net/en/third-party/messidor2/}{MESSIDOR-2~\cite{abramoff2013automated}} & 2013 & 2D & Fundus Photo & Retina & 1.7k & Yes & Cls & Diabetic Retinopathy \\ \hline
\datasetidx\label{data:fundus_ichallenge_adam} & \href{https://amd.grand-challenge.org/}{iChallenge-ADAM~\cite{fang2022adam}} & 2020 & 2D & Fundus Photo & Retina & 400 & Yes & Cls & Diabetic Retinopathy \\ \hline
\datasetidx\label{data:fundus_airogs} & \href{https://airogs.grand-challenge.org/}{AIROGS~\cite{de2023airogs}} & 2021 & 2D & Fundus Photo & Retina & 101.4k & No & Cls & Diabetic Retinopathy \\ \hline
\datasetidx\label{data:fundus_diaret_db} & \href{https://www.it.lut.fi/project/imageret/}{DiaRetDB~\cite{kauppi2007diaretdb1}} & 2009 & 2D & Fundus Photo & Retina & 89 & No & Det & DR Lesions \\ \hline
\datasetidx\label{data:fundus_hrf_healthy} & \href{https://www5.cs.fau.de/fileadmin/research/datasets/fundus-images/healthy.zip}{HRF~\cite{budai2013robust}} & NA & 2D & Fundus & Retina & 45 & No & Seg & NA \\ \hline
\datasetidx\label{data:fundus_ichallenge_palm19} & \href{https://palm.grand-challenge.org/}{iChallenge-PALM19~\cite{palm}} & 2019 & 2D & Fundus & Retina & 800 & Yes & Seg & NA \\ \hline
\datasetidx\label{data:fundus_tianchi_reg} & \href{https://tianchi.aliyun.com/dataset/dataDetail?dataId=90112}{Retina Fundus Image Reg.~\cite{hernandez2017fire}} & 2021 & 2D & Fundus Photo & Retina & 129 & Yes & Reg & NA \\ \hline
\datasetidx\label{data:fundus_aptos2019_tianchi} & \href{https://tianchi.aliyun.com/dataset/dataDetail?dataId=120007}{APTOS-2019~\cite{tianchi2021retinafundus}} & 2021 & 2D & Fundus Photo & Retina & 3.7k & Yes & Cls & Diabetic Retinopathy \\ \hline
\datasetidx\label{data:fundus_medmnist} & \href{https://medmnist.com/v1}{MedMNIST~\cite{yang2023medmnist}} & 2020 & 2D & Multi\textsuperscript{a} & Retina, Breast, Lung & 100k & Yes & Cls & NA \\ \hline
\datasetidx\label{data:fundus_deepdr_task1} & \href{https://isbi.deepdr.org/}{DeepDR-Task1~\cite{liu2022deepdrid}} & 2020 & 2D & Fundus Photo & Eye Vessel & 2k & Yes & Cls & Breast Cancer \\ \hline
\datasetidx\label{data:fundus_imageclef_2016} & \href{https://www.imageclef.org/2016/medical}{ImageCLEF 2016~\cite{ruckert2024rocov2}} & 2015 & 2D & Multi\textsuperscript{b} & Skin, Cell, Breast & 31k & Yes & Cls & Head \& Neck Tumor \\ \hline
\datasetidx\label{data:fundus_rite} & \href{https://opendatalab.com/RITE}{RITE~\cite{hu2013automated}} & 2013 & 2D & Fundus & Retina & 40 & Yes & Seg & Retinal Vessel \\ \hline
\datasetidx\label{data:fundus_gamma_task1_cfp} & \href{https://gamma.grand-challenge.org/}{GAMMA (Task1, CFP)~\cite{wu2023gamma}} & 2021 & 2D & Fundus (CFP) & Retina & 200 & Yes & Cls & Grading \\ \hline
\datasetidx\label{data:fundus_rim_one} & \href{http://medimrg.webs.ull.es/research/retinal-imaging/rim-one/}{RIM-ONE~\cite{batista2020rim}} & 2020 & 2D & Fundus & Retina & 485 & Yes & Seg & Optic Disc and Cup \\ \hline
\datasetidx\label{data:fundus_aptos2019_kaggle} & \href{https://www.kaggle.com/competitions/aptos2019-blindness-detection/overview}{APTOS 2019 Blindness Det.~\cite{zhang2025predicting}} & 2019 & 2D & Fundus & Retina & 5.6k & Yes & Cls & Grading \\ \hline
\datasetidx\label{data:fundus_kaggle_glaucoma} & \href{https://www.kaggle.com/datasets/sshikamaru/glaucoma-detection}{Glaucoma Detection~\cite{sshikamaru_glaucoma_detection}} & 2020 & 2D & Fundus & Retina & 650 & Yes & Cls & Glaucoma \\ \hline
\datasetidx\label{data:fundus_acrima} & \href{https://figshare.com/s/c2d31f850af14c5b5232}{ACRIMA~\cite{diaz2019cnns}} & 2019 & 2D & Fundus & Retina & 705 & Yes & Cls & Glaucoma \\ \hline
\datasetidx\label{data:fundus_ao_slo} & \href{https://people.duke.edu/~sf59/Chiu_BOE_2013_dataset.htm}{AO-SLO Photoreceptor Seg.~\cite{chiu2013automatic}} & 2013 & 2D & Fundus & Retina & 840 & Yes & Seg & AO-SLO Cone Photoreceptor \\ \hline
\datasetidx\label{data:fundus_av_nicking} & \href{https://people.eng.unimelb.edu.au/thivun/projects/AV_nicking_quantification/}{Arteriovenous Nicking~\cite{nguyen2013automated}} & NA & 2D & Fundus & Retina & 90 & Yes & Cls & Retinal Artery-Vein Nicking \\ \hline
\datasetidx\label{data:fundus_cataract_retina} & \href{https://www.kaggle.com/datasets/jr2ngb/cataractdataset}{Retina~\cite{jr2ngb_cataractdataset}} & 2019 & 2D & Fundus & Retina & 601 & Yes & Cls & Fundus Diseases \\ \hline
\datasetidx\label{data:fundus_yangxi} & \href{https://zenodo.org/record/3393265}{Yangxi~\cite{liu2019self}} & 2019 & 2D & Fundus & Retina & 20.4k & Yes & Cls & Eye Axis \\ \hline
\datasetidx\label{data:fundus_william_hoyt} & \href{https://novel.utah.edu/Hoyt/disc_swelling.php}{William Hoyt~\cite{perez2011vampire}} & 2004 & 2D & Fundus & Retina & 856 & Yes & Cls & Fundus Diseases \\ \hline
\datasetidx\label{data:fundus_vampire} & \href{https://vampire.computing.dundee.ac.uk/vesselseg.html}{Vampire~\cite{perez2011vampire}} & 2011 & 2D & Fundus & Retina & 8 & Yes & Seg & Vessel \\ \hline
\datasetidx\label{data:fundus_michigan_glaucoma} & \href{https://deepblue.lib.umich.edu/data/concern/data_sets/3b591905z?locale=en}{Retinal Fundus Imgs for Glaucoma~\cite{almazroa2018retinal}} & 2018 & 2D & Fundus & Retina & 2.9k & Yes & Cls & NA \\ \hline
\datasetidx\label{data:fundus_retinacheck_iostar} & \href{http://www.retinacheck.org/download-iostar-retinal-vessel-segmentation-dataset}{RetinaCheck (IOSTAR)~\cite{abbasi2015biologically}} & 2016 & 2D & Fundus & Retina & 30 & Yes & Seg & Vessel \\ \hline
\datasetidx\label{data:fundus_slit_lamp} & \href{https://plos.figshare.com/articles/dataset/Predicting_the_progression_of_ophthalmic_disease_based_on_slit-lamp_images_using_a_deep_temporal_sequence_network/6883823}{Ophthalmic Slit Lamp~\cite{jiang2018predicting}} & 2018 & 2D & Fundus & Retina & 60 & No & NA & NA \\ \hline
\datasetidx\label{data:fundus_miles_iris} & \href{https://drive.google.com/drive/folders/0B5OBp4zckpLnYkpBcWlubC0tcTA}{Miles Iris~\cite{jr2ngb_cataractdataset}} & 2013 & 2D & Fundus (Iris) & Retina & 833 & No & Cls & Retinal Structures \\ \hline
\datasetidx\label{data:fundus_jsiec} & \href{https://www.kaggle.com/datasets/linchundan/fundusimage1000}{JSIEC~\cite{cen2021automatic}} & 2019 & 2D & Fundus & Retina & 1k & Yes & Cls & Fundus Diseases \\ \hline
\datasetidx\label{data:fundus_inspire_stereo} & \href{https://medicine.uiowa.edu/eye/inspire-datasets}{INSPIRE (Stereo)~\cite{jr2ngb_cataractdataset}} & 2011 & 2D & Fundus & Retina & 30 & Yes & Reg & NA \\ \hline
\datasetidx\label{data:fundus_inspire_avr} & \href{https://medicine.uiowa.edu/eye/inspire-datasets}{INSPIRE (AVR)~\cite{jr2ngb_cataractdataset}} & 2011 & 2D & Fundus & Retina & 40 & Yes & Reg & NA \\ \hline
\datasetidx\label{data:fundus_hrf_qa} & \href{https://www5.cs.fau.de/research/data/fundus-images/}{HRF Quality Assessment~\cite{odstrcilik2013retinal}} & 2013 & 2D & Fundus & Retina & 36 & Yes & Reg & NA \\ \hline
\datasetidx\label{data:fundus_hrf_seg} & \href{https://www5.cs.fau.de/research/data/fundus-images/}{HRF Segmentation~\cite{budai2013robust}} & 2013 & 2D & Fundus & Retina & 45 & Yes & Seg & Vessel \\ \hline
\datasetidx\label{data:fundus_ichallenge_refuge2} & \href{https://refuge.grand-challenge.org/}{iChallenge-REFUGE2~\cite{fang2022refuge2}} & 2020 & 2D & Fundus Photo (CFP) & Retina & 1.6k & Yes & Cls & Glaucoma \\ \hline
\datasetidx\label{data:fundus_gamma_main} & \href{https://gamma.grand-challenge.org/}{GAMMA~\cite{wu2023gamma}} & 2021 & 2D, 3D & Fundus & Retina & 200 & Yes & Cls & NA \\ \hline
\datasetidx\label{data:fundus_oia_odir} & \href{https://odir2019.grand-challenge.org/introduction/}{OIA-ODIR~\cite{li2020benchmark}} & 2019 & 2D & Fundus & NA & 10k & Yes & Cls & NA \\ \hline
\datasetidx\label{data:fundus_varpa} & \href{http://www.varpa.es/research/ophtalmology.html}{VARPA}~\cite{varpa} & 2019 & 2D & Fundus & Retina & 58 & Yes & Cls & NA \\ \hline
\datasetidx\label{data:fundus_orvs} & \href{https://opendatalab.org.cn/ORVS}{ORVS}~\cite{sarhan2021transfer} & 2020 & 2D & Fundus & Retina & 49 & Yes & Seg & NA \\ \hline
\datasetidx\label{data:fundus_tianchi_qa} & \href{https://www.heywhale.com/mw/dataset/5e95d871e7ec38002d034efe}{Retinal Img Quality Assess}~\cite{subhadeep_chakraborty_2024} & 2020 & 2D & Fundus & Retina & 216 & Yes & Cls & NA \\ \hline
\datasetidx\label{data:fundus_ichallenge_gamma_3doct} & \href{https://gamma.grand-challenge.org/}{iChallenge-GAMMA\_3D-OCT~\cite{wu2023gamma}} & 2021 & 2D & Fundus & Retina & 300 & Yes & Seg & Glaucoma \\ \hline
\datasetidx\label{data:fundus_deepdr_task2} & \href{https://isbi.deepdr.org/}{DeepDR-Task2~\cite{liu2022deepdrid}} & 2020 & 2D & Fundus & NA & 2k & Yes & Reg & NA \\ \hline
\datasetidx\label{data:fundus_deepdr_task3} & \href{https://isbi.deepdr.org/}{DeepDR-Task3~\cite{liu2022deepdrid}} & 2020 & 2D & Fundus & NA & 246 & Yes & Cls & NA \\ \hline
\datasetidx\label{data:fundus_mmac2023} & \href{https://codalab.lisn.upsaclay.fr/competitions/12441}{MMAC2023~\cite{li2023automated}} & 2023 & 2D & Fundus & NA & 0 & Yes & Cls & NA \\ \hline
\datasetidx\label{data:fundus_rfmid_v2} & \href{https://zenodo.org/record/7505822}{RFMiD 2.0~\cite{pachade2021retinal}} & 2023 & 2D & Fundus Photo & NA & 860 & Yes & Cls & Retinal Fundus Multi-Disease \\ \hline
\datasetidx\label{data:fundus_mured} & \href{https://data.mendeley.com/datasets/pc4mb3h8hz/1}{MuReD}~\cite{rodriguez2022multilabelretinaldiseaseclassification} & 2022 & 2D & Fundus Photo & NA & 2.2k & Yes & Cls & Retinal Diseases \\ \hline
\datasetidx\label{data:fundus_vessel_tortuosity} & \href{http://bioimlab.dei.unipd.it/Retinal Vessel Tortuosity.htm}{Retinal Vessel Tortuosity} & 2008 & 2D & Fundus Photo & Retina & 60 & Yes & Reg & NA \\ \hline
\datasetidx\label{data:fundus_imageclef_2016_b} & ImageCLEF 2016 & NA & 2D & Multi\textsuperscript{c} & Skin, Cell, Breast & 31k & Yes & Cls & NA \\ \hline
\datasetidx\label{data:fundus_paraguay} & \href{https://zenodo.org/record/3872227}{PARAGUAY~\cite{benitez2021dataset}} & NA & 2D & Fundus Photo & NA & 0 & Yes & Cls & Diabetic Retinopathy \\ \hline
\datasetidx\label{data:fundus_beh} & \href{https://github.com/mirtanvirislam/Deep-Learning-Based-Glaucoma-Detection-with-Cropped-Optic-Cup-and-Disc-and-Blood-Vessel-Segmentation}{BEH}~\cite{islam2021deep} & NA & 2D & Fundus Photo & NA & 0 & Yes & NA & Glaucoma \\ \hline
\datasetidx\label{data:fundus_bidr} & \href{https://www.kaggle.com/datasets/pkdarabi/diagnosis-of-diabetic-retinopathy}{BiDR} & NA & 2D & Fundus Photo & NA & 0 & Yes & NA & Diabetic Retinopathy \\ \hline
\datasetidx\label{data:fundus_harvard_glaucoma} & \href{https://dataverse.harvard.edu/dataset.xhtml?persistentId=doi:10.7910/DVN/2EVN4B}{HarvardGlaucoma} & NA & 2D & Fundus Photo & NA & 0 & Yes & NA & Glaucoma \\ \hline
\datasetidx\label{data:fundus_fund} & FUND & NA & 2D & Fundus Photo & NA & 0 & Yes & NA & NA \\ \hline
\datasetidx\label{data:fundus_lag} & \href{https://github.com/smilell/AG-CNN}{LAG}~\cite{Li_2019_CVPR} & NA & 2D & Fundus Photo & NA & 0 & Yes & NA & Glaucoma \\ \hline
\datasetidx\label{data:fundus_dhrf} & \href{https://www.kaggle.com/datasets/nikkich9/derbi-hackathon-retinal-fundus-image-dataset}{DHRF} & NA & 2D & Fundus Photo & Retina & 6.2k & Yes & Cls & Diabetic Retinopathy \\ \hline
\datasetidx\label{data:fundus_e_ophta} & E-ophta & NA & 2D & Fundus Photo & Retina & 926 & Yes & Seg & NA \\ \hline
\datasetidx\label{data:fundus_fives} & \href{https://figshare.com/articles/dataset/FIVES_A_Fundus_Image_Dataset_for_AI-based_Vessel_Segmentation/21727913}{FIVES~\cite{jin2022fives}} & NA & 2D & Fundus Photo & Retina & 800 & Yes & Seg & Vessel \\ \hline
\datasetidx\label{data:fundus_oculard} & \href{https://www.nature.com/articles/s41598-024-84922-y}{OcularD}~\cite{kansal2025multiple} & NA & 2D & Fundus Photo & Retina & 6.4k & Yes & Cls & NA \\ \hline
\datasetidx\label{data:fundus_papila} & \href{https://doi.org/10.6084/m9.figshare.14798004.v1}{PAPILA}~\cite{kovalyk2022papila} & NA & 2D & Fundus Photo & Retina & 488 & Yes & Seg & NA \\ \hline
\datasetidx\label{data:fundus_papilledema} & \href{https://www.kaggle.com/datasets/shashwatwork/identification-of-pseudopapilledema}{Papilledema}~\cite{Kim_2018} & 2018 & 2D & Fundus Photo & Retina & 1.4k & Yes & Cls & Papilledema \\ \hline
\datasetidx\label{data:fundus_rod} & \href{https://www.kaggle.com/datasets/gracemariabinu/retinal-occlusion-dataset/data}{ROD} & 2023 & 2D & Fundus Photo & Retina & 281 & Yes & Cls & Retinal Occlusion \\ \hline
\datasetidx\label{data:fundus_toxofundus} & \href{https://www.kaggle.com/datasets/andrewmvd/ocular-toxoplasmosis-fundus-images-dataset}{ToxoFundus~\cite{CARDOZO2023109056}} & 2023 & 2D & Fundus Photo & Retina & 411 & Yes & Cls & Ocular Toxoplasmosis \\ \hline
\datasetidx\label{data:fundus_gamma_task3_cfp} & \href{https://gamma.grand-challenge.org/}{GAMMA (Task3, CFP)~\cite{wu2023gamma}} & 2021 & 2D & Fundus (CFP) & Retina & 200 & Yes & Seg & Optic Disc and Cup \\ \hline
\datasetidx\label{data:fundus_ichallenge_gamma_2dfundus} & \href{https://gamma.grand-challenge.org/}{iChallenge-GAMMA\_2D-Fundus~\cite{wu2023gamma}} & 2021 & 2D & Fundus & Retina & 300 & Yes & Seg & Glaucoma \\ \hline
& Overall & 2003$\sim$2023 & 2D & Multi & Multi& 443.1k & NA & Multi & Multi \\ \hline
\end{tabular}
\vspace{2pt}
\begin{tablenotes}[flushleft]
\scriptsize
\item[a] Multi-modalities of MedMNIST: OCT, X-Ray, CT, Pathology, Fundus Photography.
\item[b] Multi-modalities of ImageCLEF 2016: MR, US, Histopathology, X-Ray, CT, PET, Endoscopy, Dermoscopy, EEG, ECG, EMG, Electron Microscopy, Fundus Photography.
\item[c] Multi-modalities of ImageCLEF 2016: MR, US, Histopathology, X-Ray, CT, PET, Endoscopy, Dermoscopy, EEG, ECG, EMG, Microscopy, Fundus Photography.
\item[] \textbf{Abbreviations:} Seg=Segmentation, Det=Detection, Cls=Classification, Reg=Registration, US=Ultrasound, DR=Diabetic Retinopathy.
\end{tablenotes}
\end{threeparttable}
}%
\end{table*}

\begin{table*}[!htbp]
\small
\centering
\caption{2D dermoscopy datasets.}
\label{tab:2d_dermoscopy_datasets}
\begin{threeparttable}
\setlength{\tabcolsep}{4pt}      
\renewcommand{\arraystretch}{1.12} 
\resizebox{\textwidth}{!}{
\begin{tabular}{|c||>{\raggedright\arraybackslash}m{6.0cm}|r|c|c|>{\raggedright\arraybackslash}m{2.5cm}|r|c|c|>{\raggedright\arraybackslash}m{4.0cm}|}
\hline
\multicolumn{1}{|c||}{\textbf{\#}} &
\multicolumn{1}{c|}{\textbf{Dataset}} &
\multicolumn{1}{c|}{\textbf{Year}} &
\multicolumn{1}{c|}{\textbf{Dim}} &
\multicolumn{1}{c|}{\textbf{Modality}} &
\multicolumn{1}{c|}{\textbf{Structure}} &
\multicolumn{1}{c|}{\textbf{Images}} &
\multicolumn{1}{c|}{\textbf{Label}} &
\multicolumn{1}{c|}{\textbf{Task}} &
\multicolumn{1}{c|}{\textbf{Diseases}} \\
\hline
\datasetidx\label{data:dermo_isic18} & \href{https://workshop2018.isic-archive.com/}{ISIC18~\cite{codella2019skin}} & 2018 & 2D & Dermoscopy & Skin & 2.7k & Yes & Seg & Skin lesion \\ \hline
\datasetidx\label{data:dermo_isic20} & \href{https://challenge.isic-archive.com/}{ISIC20~\cite{rotemberg2021patient}} & 2020 & 2D & Dermoscopy & Skin & 33.1k & Yes & Cls & Benign melanoma, malignant melanoma \\ \hline
\datasetidx\label{data:dermo_isic16} & \href{https://challenge.isic-archive.com/landing/2016/}{ISIC16~\cite{gutman2016skin}} & 2016 & 2D & Dermoscopy & Skin & 1.3k & Yes & Seg & Skin lesion \\ \hline
\datasetidx\label{data:dermo_isic17} & \href{https://challenge.isic-archive.com/landing/2017/}{ISIC17~\cite{codella2018skin}} & 2016 & 2D & Dermoscopy & Skin & 2.8k & Yes & Seg & Skin lesion \\ \hline
\datasetidx\label{data:dermo_derm7pt} & \href{https://derm.cs.sfu.ca/Welcome.html}{Derm7pt~\cite{kawahara2018seven}} & 2021 & 2D & Dermoscopy & Skin & 2.0k & Yes & Cls & Skin lesion \\ \hline
\datasetidx\label{data:dermo_isic19} & \href{https://challenge2019.isic-archive.com/}{ISIC19~\cite{combalia2022validation}} & 2019 & 2D & Dermoscopy & Skin & 25.3k & Yes & Cls & Cells \\ \hline
\datasetidx\label{data:dermo_fizpatrick17k} & \href{https://github.com/mattgroh/fitzpatrick17k}{Fizpatrick 17k~\cite{groh2021evaluating}} & 2021 & 2D & Dermoscopy & Skin & 16.6k & Yes & Cls & NA \\ \hline
\datasetidx\label{data:dermo_mednode} & \href{https://www.cs.rug.nl/~imaging/databases/melanoma_naevi/}{MED-NODE~\cite{giotis2015med}} & 2015 & 2D & Dermoscopy & Skin & 170 & Yes & Cls & Brain \\ \hline
\datasetidx\label{data:dermo_pad_ufes} & \href{https://data.mendeley.com/datasets/zr7vgbcyr2/1}{PAD-UFES-20~\cite{pacheco2020pad}} & 2020 & 2D & Dermoscopy & Skin & 2.3k & Yes & Cls & Thoracic diseases \\ \hline
\datasetidx\label{data:dermo_ph2} & \href{https://www.fc.up.pt/addi/ph2\%20database.html}{PH2~\cite{mendoncca2013ph}} & 2014 & 2D & Dermoscopy & Skin & 200 & Yes & Cls & Cells \\ \hline
\datasetidx\label{data:dermo_dfuc2020} & \href{https://dfu-challenge.github.io/}{DFUC 2020}~\cite{yap2021deep} & 2020 & 2D & Dermoscopy & Foot & 2.0k & Yes & Seg & Breast cancer \\ \hline
\datasetidx\label{data:dermo_sd128} & \href{https://workshop2021.isic-archive.com}{SD-128 / SD-198 / SD-260~\cite{sun2016benchmark,yang2019self}} & 2021 & 2D & Dermoscopy & Skin & 6.6k & Yes & Cls & Fetal structure \\ \hline
\datasetidx\label{data:dermo_imageclef_a} & \href{https://www.imageclef.org/2016/medical}{ImageCLEF 2016~\cite{de2016overview}} & 2015 & 2D & Multi\textsuperscript{a} & Skin, Cell, Breast & 31k & Yes & Cls & Head \& neck tumor \\ \hline
\datasetidx\label{data:dermo_monkeypox} & \href{https://www.heywhale.com/mw/dataset/62eb75d6fef0903951b1f199}{Monkeypox Skin Image Dataset~\cite{Nafisa2022,Nafisa2023}} & 2022 & 2D & Dermoscopy & Skin & 40.2k & Yes & Cls & Monkeypox \\ \hline
\datasetidx\label{data:dermo_vitiligo} & \href{https://www.heywhale.com/mw/dataset/5ddca5fbca27f8002c4a1614}{Vitiligo Images}~\cite{viti_image} & 2019 & 2D & Dermoscopy & Skin & 368 & No & NA & Vitiligo \\ \hline
\datasetidx\label{data:dermo_imageclef_b} & \href{https://www.imageclef.org/2016/medical}{ImageCLEF 2016}~\cite{deherrera2016imageclef} & NA & 2D & Multi\textsuperscript{a} & Skin, Cell, Breast & 31k & Yes & Cls & NA \\ \hline
& Overall & 2014$\sim$2022 & 2D & Multi & Skin, Cell, Breast& 197.6k & NA & Multi & Multi \\ \hline
\end{tabular}
}
\vspace{2pt}
\begin{tablenotes}[flushleft]
\scriptsize
\item[a] Multi-modalities of ImageCLEF 2016: MR, US, Histopathology, X-Ray, CT, PET, Endoscopy, Dermoscopy, EEG, ECG, EMG, Microscopy, Fundus Photography.
\item[] \textbf{Abbreviations:} Seg=Segmentation, Det=Detection, Cls=Classification, Recon=Reconstruction, Reg=Registration, \\Loc=Localization, Est=Estimation, US=Ultrasound, EM=Electron Microscopy.
\end{tablenotes}
\end{threeparttable}
\end{table*}

\begin{table*}[!htbp]
\small
\centering
\caption{2D histopathology datasets. (part 1/2)}
\label{tab:2d_histopathology_datasets_part1}
\begin{threeparttable}
\setlength{\tabcolsep}{4pt}      
\renewcommand{\arraystretch}{1.12} 
\resizebox{\textwidth}{!}{
\begin{tabular}{|c||>{\raggedright\arraybackslash}m{4.8cm}|r|c|>{\raggedright\arraybackslash}m{3.0cm}|>{\raggedright\arraybackslash}m{3.0cm}|r|c|c|>{\raggedright\arraybackslash}m{3.0cm}|}
\hline
\multicolumn{1}{|c||}{\textbf{\#}} &
\multicolumn{1}{c|}{\textbf{Dataset}} &
\multicolumn{1}{c|}{\textbf{Year}} &
\multicolumn{1}{c|}{\textbf{Dim}} &
\multicolumn{1}{c|}{\textbf{Modality}} &
\multicolumn{1}{c|}{\textbf{Structure}} &
\multicolumn{1}{c|}{\textbf{Images}} &
\multicolumn{1}{c|}{\textbf{Label}} &
\multicolumn{1}{c|}{\textbf{Task}} &
\multicolumn{1}{c|}{\textbf{Diseases}} \\
\hline
\datasetidx\label{data:histo_panda_radboud} & \href{https://www.kaggle.com/c/prostate-cancer-grade-assessment/data?select=train.csv}{PANDA\_radboud~\cite{bulten2022artificial}} & 2020 & 2D & Histopathology (Patch) & Prostate & 5.1k & Yes & Seg & Prostate Cancer \\ \hline
\datasetidx\label{data:histo_gleason} & \href{https://gleason2019.grand-challenge.org/Home/}{Gleason~\cite{nir2018automatic}} & 2019 & 2D & Histopathology (Patch) & Prostate & 331 & Yes & Seg & Prostate Cancer \\ \hline
\datasetidx\label{data:histo_pathvqa} & \href{https://pathvqachallenge.grand-challenge.org}{PathologyVQA~\cite{he2020pathvqa}} & 2020 & 2D & Histopathology (Patch) & Full Body & 5.0k & Yes & VQA & NA \\ \hline
\datasetidx\label{data:histo_sln_breast} & \href{https://wiki.cancerimagingarchive.net/pages/viewpage.action?pageId=52763339}{SLN-Breast~\cite{campanella2019breast}} & 2019 & 2D & Histopathology (WSI) & Lymph & 166 & Yes & Cls & Breast Lymph Node \\ \hline
\datasetidx\label{data:histo_monuseg} & \href{https://monuseg.grand-challenge.org/Home/}{MoNuSeg~\cite{kumar2019multi}} & 2018 & 2D & Histopathology (Patch) & Nuclei & 51 & Yes & Seg & NA \\ \hline
\datasetidx\label{data:histo_monusac2020} & \href{https://monusac-2020.grand-challenge.org/}{MoNuSAC2020~\cite{verma2021monusac2020}} & 2019 & 2D & Histopathology (Patch) & Lung, Prostate, \etc\textsuperscript{a} & 914 & Yes & Seg & NA \\ \hline
\datasetidx\label{data:histo_digestpath19} & \href{https://digestpath2019.grand-challenge.org/Home/}{DigestPath19~\cite{da2022digestpath}} & 2019 & 2D & Histopathology (WSI) & Colon & 212 & Yes & Det & Signet Ring Cell \\ \hline
\datasetidx\label{data:histo_camelyon17} & \href{https://camelyon17.grand-challenge.org/Data/}{CAMELYON17~\cite{litjens20181399}} & 2016 & 2D & Histopathology (WSI) & Breast & 500 & Yes & Cls & Breast Cancer \\ \hline
\datasetidx\label{data:histo_anhir} & \href{https://anhir.grand-challenge.org/}{ANHIR~\cite{borovec2020anhir}} & 2018 & 2D & Histopathology (WSI) & Kidney, Breast, \etc\textsuperscript{b} & 481 & Yes & Reg & NA \\ \hline
\datasetidx\label{data:histo_cervical_cells} & \href{https://cs.adelaide.edu.au/~carneiro/isbi14_challenge/}{Overlapping Cervical Cells~\cite{lu2016evaluation}} & 2015 & 2D & Histopathology (Patch) & Cervix & 17 & Yes & Seg & Cervical Cells \\ \hline
\datasetidx\label{data:histo_midog2022} & \href{https://imig.science/midog/}{MIDOG2022~\cite{aubreville2023mitosis}} & 2022 & 2D & Histopathology (Patch) & Lung, Breast, Skin & 405 & Yes & Det & Mitotic Figure \\ \hline
\datasetidx\label{data:histo_acrobat} & \href{https://acrobat.grand-challenge.org/}{ACROBAT~\cite{bulten2019epithelium}} & 2023 & 2D & Histopathology (WSI) & Breast & 750 & Yes & Reg & NA \\ \hline
\datasetidx\label{data:histo_bright} & \href{https://research.ibm.com/haifa/Workshops/BRIGHT/}{BRIGHT~\cite{allison2014understanding}} & 2021 & 2D & Histopathology (Patch) & Breast & 5.1k & Yes & Cls & Pathological Benign \\ \hline
\datasetidx\label{data:histo_conic2022} & \href{https://conic-challenge.grand-challenge.org/}{CoNIC2022~\cite{graham2021conic}} & 2022 & 2D & Histopathology (Patch) & Colon & 5.0k & Yes & Seg & Colon Nuclei \\ \hline
\datasetidx\label{data:histo_pannuke_wsi} & \href{https://warwick.ac.uk/fac/sci/dcs/research/tia/data/pannuke}{PanNuke~\cite{gamper2019pannuke}} & 2021 & 2D & Histopathology (WSI) & Multi-organ & 481 & Yes & Seg, Cls & Multiple Cancers \\ \hline
\datasetidx\label{data:histo_lymphoma_cls} & \href{https://tianchi.aliyun.com/dataset/dataDetail?dataId=94414}{Malignant Lymphoma Cls~\cite{orlov2010automatic}} & 2021 & 2D & Histopathology (Patch) & Lymph & 374 & Yes & Cls & Lymphoma \\ \hline
\datasetidx\label{data:histo_paip2021} & \href{https://paip2021.grand-challenge.org/}{PAIP2021~\cite{paip2021_challenge}} & 2021 & 2D & Histopathology (WSI) & Colon, Prostate & 150 & Yes & Det & Colon/Prostate Cancer \\ \hline
\datasetidx\label{data:histo_breast_cell_seg} & \href{https://tianchi.aliyun.com/dataset/dataDetail?dataId=90152}{Breast Cancer Cell Seg~\cite{gelasca2008evaluation}} & 2021 & 2D & Histopathology (Patch) & Breast & 58 & Yes & Seg & Breast Cancer \\ \hline
\datasetidx\label{data:histo_medmnist} & \href{https://medmnist.com/v1}{MedMNIST ~\cite{medmnistv1}} & 2020 & 2D & Multi\textsuperscript{c} & Retina, Breast, Lung & 100k & Yes & Cls & Multi-disease \\ \hline
\datasetidx\label{data:histo_cancer_detect_kaggle} & \href{https://www.kaggle.com/competitions/histopathologic-cancer-detection}{Histopathologic Cancer Det~\cite{histopathologic_cancer_detection_kaggle}} & 2018 & 2D & Histopathology (Patch) & Lymph & 220k & Yes & Cls & Breast Cancer \\ \hline
\datasetidx\label{data:histo_hubmap} & \href{https://www.kaggle.com/competitions/hubmap-kidney-segmentation/overview}{HuBMAP~\cite{hubmap-kidney-segmentation}} & 2020 & 2D & Histopathology (Patch) & Kidney & 15 & Yes & Seg & Kidney Tissue \\ \hline
\datasetidx\label{data:histo_acdc_lunghp} & \href{https://acdc-lunghp.grand-challenge.org/}{ACDC-LungHP~\cite{li2020deep}} & 2019 & 2D & Histopathology (WSI) & Lung & 200 & Yes & Seg & Lung Cancer \\ \hline
\datasetidx\label{data:histo_segpc_2021} & \href{https://segpc-2021.grand-challenge.org/SegPC-2021/}{SegPC 2021~\cite{gupta10segpc}} & 2021 & 2D & Histopathology (Patch) & Blood & 498 & Yes & Seg & Plasma Cells \\ \hline
\datasetidx\label{data:histo_midog2021} & \href{https://imi.thi.de/midog/}{MIDOG2021~\cite{aubreville2023mitosis}} & 2021 & 2D & Histopathology (Patch) & Full Body & 200 & Yes & Det & Prostate Cancer \\ \hline
\datasetidx\label{data:histo_dermofit} & \href{https://workshop2021.isic-archive.com/}{Dermofit Image Library}~\cite{ballerini2013color} & 2021 & 2D & Histopathology (Patch) & Skin & 1.3k & Yes & Cls & Lung Adenocarcinoma \\ \hline
\datasetidx\label{data:histo_weakly_seg} & \href{https://neurips22-cellseg.grand-challenge.org/}{Weakly Supervised Cell Seg~\cite{ma2024multimodality}} & 2022 & 2D & Histopathology (Patch) & Full Body & 30 & Yes & Seg & Prostate Cancer \\ \hline
\datasetidx\label{data:histo_tiger_wsibulk} & \href{https://tiger.grand-challenge.org/Home/}{TIGER-wsibulk~\cite{vanrijthoven2022tiger_roi}} & 2022 & 2D & Histopathology (WSI) & Breast & 93 & Yes & Seg & Pneumothorax \\ \hline
\datasetidx\label{data:histo_bci} & \href{https://bci.grand-challenge.org/}{BCI~\cite{liu2022bci}} & 2022 & 2D & Histopathology (Patch) & Breast & 4.9k & Yes & Gen & Lesion \\ \hline
\datasetidx\label{data:histo_wsss4luad} & \href{https://wsss4luad.grand-challenge.org}{WSSS4LUAD}~\cite{han2022wsss4luad} & 2021 & 2D & Histopathology (Patch) & Lung & 10.2k & Yes & Seg & Coronary Artery \\ \hline
\datasetidx\label{data:histo_breast_seg} & \href{https://bcsegmentation.grand-challenge.org}{Breast Cancer Seg}~\cite{amgad2019structured} & 2019 & 2D & Histopathology (Patch) & Breast & 151 & Yes & Seg & Neurons \\ \hline
\datasetidx\label{data:histo_nucls} & \href{https://nucls.grand-challenge.org/NuCLS/}{NuCLS}~\cite{amgad2022nucls} & 2021 & 2D & Histopathology (Patch) & Nuclei & 3.1k & Yes & Seg & Kidney \\ \hline
\datasetidx\label{data:histo_imageclef_2016} & \href{https://www.imageclef.org/2016/medical}{ImageCLEF 2016}~\cite{de2016overview} & 2015 & 2D & Multi\textsuperscript{d} & Skin, Cell, Breast & 31k & Yes & Cls & Head \& Neck Tumor \\ \hline
\datasetidx\label{data:histo_paip2020} & \href{https://paip2020.grand-challenge.org/Home/}{PAIP2020}~\cite{kim2023paip} & 2020 & 2D & Histopathology (WSI) & Liver & 118 & Yes & Cls & Colorectal Cancer \\ \hline
\datasetidx\label{data:histo_herohe} & \href{https://ecdp2020.grand-challenge.org/Home/}{HEROHE}~\cite{conde2022herohe} & 2019 & 2D & Histopathology (WSI) & Lung & 510 & Yes & Cls & GI diseases \\ \hline
\datasetidx\label{data:histo_lysto} & \href{https://lysto.grand-challenge.org/}{Lymphocyte Assessment}~\cite{jiao2023lysto} & 2019 & 2D & Histopathology (Patch) & Lymphocyte & 20k & Yes & Cls & Lymphocyte Number \\ \hline
\datasetidx\label{data:histo_lyon19} & \href{https://lyon19.grand-challenge.org}{LYON19}~\cite{swiderska2019learning} & 2019 & 2D & Histopathology (Patch) & Lymphocyte & 441 & Yes & Cls & Lymphocytes \\ \hline
\datasetidx\label{data:histo_glas} & \href{https://warwick.ac.uk/fac/cross_fac/tia/data/glascontest}{GlaS}~\cite{sirinukunwattana2017gland} & 2015 & 2D & Histopathology (Patch) & Cell & 165 & Yes & Seg & Colorectal Adenocarcinoma \\ \hline
\datasetidx\label{data:histo_consep} & \href{https://warwick.ac.uk/fac/sci/dcs/research/tia/data/hovernet/}{CoNSeP}~\cite{graham2019hover} & 2018 & 2D & Histopathology (Patch) & Colon & 41 & Yes & Seg & Colorectal Nuclei \\ \hline
\datasetidx\label{data:histo_pcam} & \href{https://opendatalab.com/PCam}{PCam}~\cite{veeling2018rotation} & 2018 & 2D & Histopathology (Patch) & Breast & 328k & Yes & Seg & Metastatic Tissue \\ \hline
\datasetidx\label{data:histo_lc25000} & \href{https://opendatalab.com/LC25000}{LC25000}~\cite{borkowski2019lung} & 2019 & 2D & Histopathology (Patch) & Colon & 25k & Yes & Cls & Lung and Colon Tissue \\ \hline
\datasetidx\label{data:histo_pannuke_seg} & \href{https://huggingface.co/datasets/RationAI/PanNuke}{PanNuke (Seg)}~\cite{gamper2019pannuke} & 2021 & 2D & Histopathology (Patch) & Full Body & 7.9k & Yes & Seg & Nucleus \\ \hline
\datasetidx\label{data:histo_breakhis_40x} & \href{https://web.inf.ufpr.br/vri/databases/breast-cancer-histopathological-database-breakhis/}{BreakHis (40x)}~\cite{spanhol2015dataset} & 2016 & 2D & Histopathology (Patch) & Breast & 2.0k & Yes & Cls & Breast Tumors \\ \hline
\datasetidx\label{data:histo_sicapv2} & \href{https://opendatalab.org.cn/SICAPv2/download}{SICAPv2}~\cite{silva2020going} & 2020 & 2D & Histopathology (Patch) & Prostate & 18.8k & Yes & Cls & Prostate Cancer \\ \hline
\datasetidx\label{data:histo_kumar} & \href{https://opendatalab.org.cn/Kumar/download}{Kumar}~\cite{kumar2019multi} & 2018 & 2D & Histopathology (Patch) & Cell & 54 & Yes & Seg & Multi-organ Nuclei \\ \hline
\datasetidx\label{data:histo_herlev} & \href{https://opendatalab.org.cn/HErlev/download}{HErlev}~\cite{jantzen2005pap} & 2008 & 2D & Histopathology (Patch) & Cervix & 5.6k & Yes & Cls & Cervical Cancer \\ \hline
\datasetidx\label{data:histo_crc100k} & \href{https://opendatalab.org.cn/CRC100K/download}{CRC100K}~\cite{chen2022self} & 2018 & 2D & Histopathology (Patch) & Colon & 100k & Yes & Cls & Colorectal Cancer \\ \hline
\datasetidx\label{data:histo_brca_m2c} & \href{https://opendatalab.org.cn/BRCA-M2C/}{BRCA-M2C}~\cite{abousamra2021multi} & 2021 & 2D & Histopathology (Patch) & Breast & 120 & Yes & Seg & Breast Cancer \\ \hline
\datasetidx\label{data:histo_warwick} & \href{https://opendatalab.org.cn/warwick}{warwick}~\cite{sirinukunwattana2017gland} & 2015 & 2D & Histopathology (Patch) & Colon & 330 & Yes & Seg & Colorectal Gland \\ \hline
\datasetidx\label{data:histo_crag} & \href{https://github.com/XiaoyuZHK/CRAG-Dataset_Aug_ToCOCO}{CRAG}~\cite{graham2019mild} & 2019 & 2D & Histopathology (Patch) & Colon & 213 & Yes & Seg & Colorectal Cancer \\ \hline
\datasetidx\label{data:histo_chaoyang} & \href{https://drive.google.com/drive/folders/1xsrHjn-WyHGazYtpMqHo9h2w349eYCYO?usp=sharing}{Chaoyang}~\cite{zhu2021hard} & 2021 & 2D & Histopathology (Patch) & Blood & 6.2k & Yes & Cls & Red Blood Cell \\ \hline
\datasetidx\label{data:histo_cmb_crc} & \href{https://wiki.cancerimagingarchive.net/pages/viewpage.action?pageId=93257955}{CMB-CRC}~\cite{cmbcrc2022} & 2022 & 3D, 2D & Multi\textsuperscript{e} & Colon & 472 & No & Seg, Cls & Colorectal Cancer \\ \hline
\datasetidx\label{data:histo_cmb_gec} & \href{https://wiki.cancerimagingarchive.net/pages/viewpage.action?pageId=127665431}{CMB-GEC}~\cite{cmbgec2022} & 2022 & 3D, 2D & CT, Histopathology (WSI), PET & Brain & 14 & No & Seg, Cls & Melanoma \\ \hline
\datasetidx\label{data:histo_cmb_mel} & \href{https://wiki.cancerimagingarchive.net/pages/viewpage.action?pageId=93258432}{CMB-MEL}~\cite{cmbmel2022} & 2022 & 3D, 2D & Multi\textsuperscript{f} & Brain & 255 & No & Seg & Melanoma \\ \hline
\datasetidx\label{data:histo_cmb_mml} & \href{https://wiki.cancerimagingarchive.net/pages/viewpage.action?pageId=93258436}{CMB-MML}~\cite{cmbmml2022} & 2021 & 2D, 3D & Multi\textsuperscript{g} & NA & 60 & No & Pred & Multiple Myeloma \\ \hline
\datasetidx\label{data:histo_cmb_pca} & \href{https://wiki.cancerimagingarchive.net/pages/viewpage.action?pageId=95224082}{CMB-PCA}~\cite{cmbpca2022} & 2022 & 2D, 3D & CT, MR, Histopathology (WSI) & Prostate & 31 & No & Cls, Pred & Prostate Cancer \\ \hline
\datasetidx\label{data:histo_aggc22} & \href{https://aggc22.grand-challenge.org/}{AGGC22}~\cite{huo2024comprehensive} & 2022 & 2D & Histopathology (Patch) & Gland & 150 & Yes & Seg & Gland Segmentation \\ \hline
\datasetidx\label{data:histo_tupac} & \href{https://tupac.grand-challenge.org/TUPAC/}{TUPAC}~\cite{veta2019predicting} & 2015 & 2D & Histopathology (Patch) & Brain & 573 & Yes & Reg & Breast Cancer \\ \hline
\datasetidx\label{data:histo_prostate_fused} & \href{https://www.cancerimagingarchive.net/collection/prostate-fused-mri-pathology/}{Prostate Fused-MRI-Pathology~\cite{madabhushi2016fused}} & 2016 & 2D, 3D & MR, Histopathology (WSI) & Prostate & 29 & No & NA & Prostate Cancer \\ \hline
\datasetidx\label{data:histo_malaria} & \href{https://lhncbc.nlm.nih.gov/LHC-downloads/downloads.html\#malaria-datasets}{Malaria Cell Image Dataset~\cite{yang2019deep}} & 2021 & 2D & Histopathology (Patch) & Cell & 27.6k & Yes & Cls & Malaria \\ \hline
\datasetidx\label{data:histo_hep2} & \href{https://www.heywhale.com/mw/dataset/5ec3c6883241a100378d5d4a}{HEp-2 Cell Classification~\cite{larsen2014hep}} & 2020 & 2D & Histopathology (Patch) & Cell & 13.6k & Yes & Cls & HEp-2 Cells \\ \hline
\datasetidx\label{data:histo_breast_cell_seg_heywhale} & \href{https://www.heywhale.com/mw/dataset/5e9c625bebb37f002c61526a}{Breast Cancer Cell Seg Dataset}~\cite{Drelie08-298} & 2020 & 2D & Histopathology (Patch) & Breast, Cell & 58 & Yes & Seg & Breast Cancer \\ \hline
\end{tabular}
}
\vspace{2pt}
\begin{tablenotes}[flushleft]
\scriptsize
\item[a] Full structure of MoNuSAC2020: Lung (Thorax), Prostate (Pelvis), Kidney (Abdomen), Breast (Thorax).
\item[b] Full structure of ANHIR: Kidney (Abdomen), Breast (Thorax), Colon (Abdomen), Spleen, Lung (Thorax).
\item[c] Multi-modalities of MedMNIST: OCT, X-Ray, CT, Histopathology (Patch), Fundus Photography.
\item[d] Multi-modalities of ImageCLEF 2016: MR, US, Histopathology (Patch), X-Ray, CT, PET, Endoscopy, Dermoscopy, Others.
\item[e] Multi-modalities of CMB-CRC: CT, MR, US, DX, PET, Histopathology (WSI).
\item[f] Multi-modalities of CMB-MEL: CT, US, Histopathology (WSI), PET.
\item[g] Multi-modalities of CMB-MML: CT, MR, PET, Histopathology (WSI).
\item[] \textbf{Abbreviations:} Seg=Segmentation, Det=Detection, Cls=Classification, Reg=Registration, VQA=Visual Question Answering, \\Gen=Generation, Pred=Prediction.
\end{tablenotes}
\end{threeparttable}
\end{table*}

\begin{table*}[!htbp]
\small
\centering
\caption{2D histopathology datasets. (part 2/2)}
\label{tab:2d_histopathology_datasets_part2}
\begin{threeparttable}
\setlength{\tabcolsep}{4pt}      
\renewcommand{\arraystretch}{1.12} 
\resizebox{\textwidth}{!}{
\begin{tabular}{|c||>{\raggedright\arraybackslash}m{4.8cm}|r|c|>{\raggedright\arraybackslash}m{3.0cm}|>{\raggedright\arraybackslash}m{3.0cm}|r|c|c|>{\raggedright\arraybackslash}m{3.0cm}|}
\hline
\multicolumn{1}{|c||}{\textbf{\#}} &
\multicolumn{1}{c|}{\textbf{Dataset}} &
\multicolumn{1}{c|}{\textbf{Year}} &
\multicolumn{1}{c|}{\textbf{Dim}} &
\multicolumn{1}{c|}{\textbf{Modality}} &
\multicolumn{1}{c|}{\textbf{Structure}} &
\multicolumn{1}{c|}{\textbf{Images}} &
\multicolumn{1}{c|}{\textbf{Label}} &
\multicolumn{1}{c|}{\textbf{Task}} &
\multicolumn{1}{c|}{\textbf{Diseases}} \\
\hline
\datasetidx\label{data:histo_tiger_wsirois} & \href{https://tiger.grand-challenge.org/Home/}{TIGER-wsirois}~\cite{shephard2022tiager} & 2022 & 2D & Histopathology (Patch) & Breast & 2.0k & Yes & Seg & Breast Cancer \\ \hline
\datasetidx\label{data:histo_tiger_wsitils} & \href{https://tiger.grand-challenge.org/Home/}{TIGER-wsitils}\cite{shephard2022tiager} & 2022 & 2D & Histopathology (Patch) & Breast & 82 & Yes & Reg & Breast Cancer \\ \hline
\datasetidx\label{data:histo_breast_cell_seg2_heywhale} & \href{https://www.heywhale.com/mw/dataset/5e9e9b35ebb37f002c625423}{Breast Cancer Cell Seg 2}~\cite{Drelie08-298} & 2020 & 2D & Histopathology (Patch) & Breast & 58 & Yes & Seg & Breast cancer \\ \hline
\datasetidx\label{data:histo_lymphoma_cls_heywhale} & \href{https://www.heywhale.com/mw/dataset/5e9d607febb37f002c61ad3a}{Malignant Lymphoma Cls Dataset}~\cite{orlov2010automatic} & 2020 & 2D & Histopathology (Patch) & Lymph & 374 & Yes & Cls & Lymphoma \\ \hline
\datasetidx\label{data:histo_lung_colon_heywhale} & \href{https://www.heywhale.com/mw/dataset/5e956c33e7ec38002d03132c}{Lung and Colon Histopathology}~\cite{borkowski2019lung} & 2020 & 2D & Histopathology (Patch) & Lung, Colon & 25k & Yes & Cls & Lung and Colon Cancer \\ \hline
\datasetidx\label{data:histo_focuspath} & \href{https://www.heywhale.com/mw/dataset/5e85dc8b95b029002ca7ea03}{FocusPath~\cite{Hosseini_2019}} & 2020 & 2D & Histopathology (Patch) & NA & 864 & Yes & IQA & Histopathology Image \\ \hline
\datasetidx\label{data:histo_blood_cell_heywhale} & \href{https://www.heywhale.com/mw/dataset/5d9ea5a9037db3002d3ec502}{Blood Cell Images}~\cite{blood_cell_img} & 2019 & 2D & Histopathology (Patch) & Blood & 12.5k & Yes & Det & Blood Cell \\ \hline
\datasetidx\label{data:histo_colorectal_mnist} & \href{https://zenodo.org/record/53169}{Colorectal Histology MNIST}~\cite{kather2016textures} & 2016 & 2D & Histopathology (Patch) & Colon & 5.0k & Yes & Cls & Colorectal Tissue \\ \hline
\datasetidx\label{data:histo_breakhis_100x} & \href{https://opendatalab.com/BreakHis}{BreakHis 100x}~\cite{spanhol2015dataset} & 2016 & 2D & Histopathology (Patch) & Breast & 9.1k & Yes & Cls & Breast Cancer \\ \hline
\datasetidx\label{data:histo_breakhis_200x} & \href{https://opendatalab.com/BreakHis}{BreakHis 200x}~\cite{spanhol2015dataset} & 2016 & 2D & Histopathology (Patch) & Breast & 9.1k & Yes & Cls & Breast Cancer \\ \hline
\datasetidx\label{data:histo_breakhis_400x} & \href{https://opendatalab.com/BreakHis}{BreakHis 400x}~\cite{spanhol2015dataset} & 2016 & 2D & Histopathology (Patch) & Breast & 9.1k & Yes & Cls & Breast Cancer \\ \hline
\datasetidx\label{data:histo_bcnb_task1} & \href{https://bcnb.grand-challenge.org/Home/}{BCNB Task-1~\cite{xu2021predicting}} & 2021 & 2D & Histopathology (WSI) & Breast & 1.1k & Yes & Cls & Leukemia \\ \hline
\datasetidx\label{data:histo_bcnb_task2} & \href{https://bcnb.grand-challenge.org/Home/}{BCNB Task-2}~\cite{xu2021predicting} & 2021 & 2D & Histopathology (WSI) & Breast & 1.1k & Yes & Cls & Breast Cancer \\ \hline
\datasetidx\label{data:histo_bcnb_task3} & \href{https://bcnb.grand-challenge.org/Home/}{BCNB Task-3}~\cite{xu2021predicting} & 2021 & 2D & Histopathology (WSI) & Breast & 1.1k & Yes & Cls & Breast Cancer \\ \hline
\datasetidx\label{data:histo_bcnb_task4} & \href{https://bcnb.grand-challenge.org/Home/}{BCNB Task-4}~\cite{xu2021predicting} & 2021 & 2D & Histopathology (WSI) & Breast & 1.1k & Yes & Cls & Breast Cancer \\ \hline
\datasetidx\label{data:histo_bcnb_task5} & \href{https://bcnb.grand-challenge.org/Home/}{BCNB Task-5}~\cite{xu2021predicting} & 2021 & 2D & Histopathology (WSI) & Breast & 1.1k & Yes & Cls & Breast Cancer \\ \hline
\datasetidx\label{data:histo_bcnb_task6} & \href{https://bcnb.grand-challenge.org/Home/}{BCNB Task-6}~\cite{xu2021predicting} & 2021 & 2D & Histopathology (WSI) & Breast & 1.1k & Yes & Cls & Breast Cancer \\ \hline
\datasetidx\label{data:histo_panda_main} & \href{https://www.kaggle.com/c/prostate-cancer-grade-assessment/data?select=train.csv}{PANDA}~\cite{bulten2022artificial} & 2020 & 2D & Histopathology (Patch) & Prostate & 10.6k & Yes & Cls & Prostate Cancer \\ \hline
\datasetidx\label{data:histo_panda_karolinska} & \href{https://www.kaggle.com/c/prostate-cancer-grade-assessment/data?select=train.csv}{PANDA\_karolinska}~\cite{bulten2022artificial} & 2020 & 2D & Histopathology (Patch) & Prostate & 5.5k & Yes & Seg & Prostate Cancer \\ \hline
\datasetidx\label{data:histo_paip2023_seg} & \href{https://2023paip.grand-challenge.org/}{PAIP 2023~\cite{akbar2019automated}} & 2022 & 2D & Histopathology (Patch) & Pancreas & 103 & Yes & Seg & Liver Cancer \\ \hline
\datasetidx\label{data:histo_atec23} & \href{https://github.com/cwwang1979/MICCAI_ATEC23challenge}{ATEC23}~\cite{wang2025atec23} & 2023 & 2D & Histopathology (WSI) & Ovary & 468 & Yes & Cls & Ovarian Cancer \\ \hline
\datasetidx\label{data:histo_acrobat2023} & \href{https://acrobat.grand-challenge.org/overview/}{ACROBAT2023}~\cite{weitz2024acrobat} & 2023 & 2D & Histopathology (WSI) & Breast & 1.2k & Yes & Reg & Breast Cancer \\ \hline
\datasetidx\label{data:histo_ocelot2023} & \href{https://ocelot2023.grand-challenge.org/}{OCELOT2023}~\cite{shin2025ocelot} & 2023 & 2D & Histopathology (WSI) & Colon & 667 & Yes & Det & Colon Cancer \\ \hline
\datasetidx\label{data:histo_ocean} & \href{https://zenodo.org/record/7844718}{OCEAN}~\cite{asadi2024machine} & 2023 & 2D & Histopathology (WSI) & Ovary & 1.6k & Yes & Cls & Ovarian Cancer \\ \hline
\datasetidx\label{data:histo_endo_aid} & \href{https://endo-aid.grand-challenge.org/}{Endo-Aid}~\cite{vermorgen2024endometrial} & 2022 & 2D & Histopathology (WSI) & GI Tract & 91 & No & Cls & GI Polyps \\ \hline
\datasetidx\label{data:histo_paip2023_cls} & \href{https://2023paip.grand-challenge.org/}{PAIP2023~\cite{akbar2019automated}} & 2022 & 2D & Histopathology (Patch) & Pancreas & 103 & Yes & Seg & Pancreatic Cancer \\ \hline
\datasetidx\label{data:histo_patchcamelyon} & \href{https://patchcamelyon.grand-challenge.org/}{PatchCamelyon\cite{gc_patchcamelyon_2018}} & 2018 & 2D & Histopathology (Patch) & Lymph Node & 295k & Yes & Cls & Metastatic Tissue \\ \hline
\datasetidx\label{data:histo_bone_marrow} & \href{https://wiki.cancerimagingarchive.net/pages/viewpage.action?pageId=101941770}{Bone Marrow Cytomorphology\cite{tcia_bone_marrow_cytomorphology_2021}} & 2021 & 2D & Histopathology (Patch) & Bone Marrow & 171k & Yes & Cls & Blood Cells \\ \hline
\datasetidx\label{data:histo_lung_fused_ct} & \href{https://wiki.cancerimagingarchive.net/pages/viewpage.action?pageId=39878702}{Lung-Fused-CT-Pathology\cite{Rusu2017EurRadiol_LungFused}} & 2018 & 2D, 3D & CT, Histopathology (WSI) & Lung & 36 & Yes & Seg & Lung Cancer \\ \hline
\datasetidx\label{data:histo_hnscc} & \href{https://wiki.cancerimagingarchive.net/pages/viewpage.action?pageId=70226184}{HNSCC-mIF-mIHC\cite{Ghahremani2023MICCAI_HNSCCmIFmIHC}} & 2020 & 2D & Histopathology (Patch) & Head \& Neck & 3.2k & No & NA & HNSCC \\ \hline
\datasetidx\label{data:histo_sn_am} & \href{https://wiki.cancerimagingarchive.net/pages/viewpage.action?pageId=52757009}{SN-AM\cite{Gupta2020MedIA_SNAM}} & 2019 & 2D & Histopathology (Patch) & Lymph Node & 190 & Yes & Seg & Melanoma \\ \hline
\datasetidx\label{data:histo_ovarian_bev} & \href{https://wiki.cancerimagingarchive.net/pages/viewpage.action?pageId=83593077}{Ovarian Bevacizumab Response\cite{tcia_ovarian_bev_response_2023}} & 2023 & 2D & Histopathology (WSI) & Ovary & 285 & No & NA & Ovarian Cancer \\ \hline
\datasetidx\label{data:histo_cmb_lca} & \href{https://wiki.cancerimagingarchive.net/pages/viewpage.action?pageId=93258420}{CMB-LCA\cite{tcia_cmb_lca_2022}} & 2022 & 2D, 3D & Multi\textsuperscript{a} & Lung & 0 & No & NA & Lung Cancer \\ \hline
\datasetidx\label{data:histo_cptac_coad} & \href{https://wiki.cancerimagingarchive.net/pages/viewpage.action?pageId=70227852}{CPTAC-COAD\cite{tcia_cptac_coad_2021}} & 2021 & 2D & Histopathology (WSI) & Colon & 373 & Yes & Cls & Colon Adenocarcinoma \\ \hline
\datasetidx\label{data:histo_hungarian_crc} & \href{https://wiki.cancerimagingarchive.net/pages/viewpage.action?pageId=91357370}{Hungarian-Colorectal-Screening\cite{tcia_hungarian_crc_2022}} & 2022 & 2D & Histopathology (WSI) & Colorectal & 200 & No & NA & Colorectal Polyps \\ \hline
\datasetidx\label{data:histo_dlbcl} & \href{https://wiki.cancerimagingarchive.net/pages/viewpage.action?pageId=119702520}{DLBCL-Morphology\cite{tcia_dlbcl_morphology_2022}} & 2022 & 2D & Histopathology (Patch) & Lymph Node & 246 & Yes & Seg & DLBCL \\ \hline
\datasetidx\label{data:histo_cptac_ov} & \href{https://wiki.cancerimagingarchive.net/pages/viewpage.action?pageId=70227856}{CPTAC-OV\cite{tcia_cptac_ov_2021}} & 2021 & 2D & Histopathology (WSI) & Ovary & 222 & No & NA & Ovarian Cancer \\ \hline
\datasetidx\label{data:histo_codex_hcc} & \href{https://wiki.cancerimagingarchive.net/pages/viewpage.action?pageId=140313174}{CODEX imaging of HCC\cite{tcia_codex_hcc_2023}} & 2023 & 2D & Histopathology (WSI) & Liver & 646 & No & NA & Liver HCC \\ \hline
\datasetidx\label{data:histo_prostate_mri} & \href{https://wiki.cancerimagingarchive.net/display/Public/PROSTATE-MRI}{Prostate-MRI\cite{tcia_prostate_mri_2011}} & 2011 & 3D, 2D & Multi\textsuperscript{b} & Prostate & 26 & No & NA & Prostate Cancer \\ \hline
\datasetidx\label{data:histo_cptac_brca} & \href{https://wiki.cancerimagingarchive.net/pages/viewpage.action?pageId=70227748}{CPTAC-BRCA\cite{tcia_cptac_brca_2021}} & 2021 & 2D & Histopathology (WSI) & Breast & 642 & No & NA & Breast Cancer \\ \hline
\datasetidx\label{data:histo_aml_lmu} & \href{https://wiki.cancerimagingarchive.net/pages/viewpage.action?pageId=61080958}{AML-Cytomorphology\_LMU\cite{Matek2019NatMI_AML}} & 2019 & 2D & Histopathology (WSI) & Blood & 18.4k & Yes & Cls & Acute Myeloid Leukemia \\ \hline
\datasetidx\label{data:histo_mimm_sbilab} & \href{https://wiki.cancerimagingarchive.net/pages/viewpage.action?pageId=52756988}{MiMM\_SBILab~\cite{gupta2019mimm_sbilab}} & 2019 & 2D & Histopathology (WSI) & Bone Marrow & 85 & Yes & Loc & Multiple Myeloma \\ \hline
\datasetidx\label{data:histo_pancancer_nuclei} & \href{https://wiki.cancerimagingarchive.net/pages/viewpage.action?pageId=64685083}{Pan-Cancer-Nuclei-Seg}\cite{hou2020dataset} & 2020 & 2D & Histopathology (WSI) & Multi-organ & 5.1k & Yes & Seg & Pan-Cancer \\ \hline
\datasetidx\label{data:histo_til_wsi_tcga} & \href{https://wiki.cancerimagingarchive.net/pages/viewpage.action?pageId=33948919}{TIL-WSI-TCGA}\cite{saltz2018til} & 2018 & 2D & Histopathology (WSI) & Multi-organ & 5.2k & Yes & Cls & Pan-Cancer \\ \hline
\datasetidx\label{data:histo_cnmc_2019} & \href{https://wiki.cancerimagingarchive.net/pages/viewpage.action?pageId=52758223}{C-NMC 2019}\cite{gupta2022cnmc} & 2019 & 2D & Histopathology (WSI) & Blood & 15.1k & Yes & Cls & Leukemia \\ \hline
\datasetidx\label{data:histo_cptac_aml} & \href{https://wiki.cancerimagingarchive.net/pages/viewpage.action?pageId=47677483}{CPTAC-AML}\cite{CPTAC_AML} & 2019 & 2D & Histopathology (WSI) & Bone Marrow & 122 & No & NA & Acute Myeloid Leukemia \\ \hline
\datasetidx\label{data:histo_catch} & \href{https://wiki.cancerimagingarchive.net/pages/viewpage.action?pageId=101941773}{CATCH}\cite{wilm2022catch} & 2022 & 2D & Histopathology (WSI) & Skin & 350 & Yes & Seg & Skin Cancer \\ \hline
\datasetidx\label{data:histo_nadt_prostate} & \href{https://wiki.cancerimagingarchive.net/pages/viewpage.action?pageId=91357374}{NADT-Prostate}\cite{wilkinson2021nadt} & 2021 & 2D & Histopathology (WSI) & Prostate & 1.4k & No & NA & Prostate Cancer \\ \hline
\datasetidx\label{data:histo_her2_rois} & \href{https://wiki.cancerimagingarchive.net/pages/viewpage.action?pageId=119702524}{HER2 tumor ROIs}\cite{farahmand2022her2} & 2022 & 2D & Histopathology (WSI) & Breast & 273 & Yes & Seg & HER2+ Breast Cancer \\ \hline
\datasetidx\label{data:histo_crc_ffpe_codex} & \href{https://wiki.cancerimagingarchive.net/pages/viewpage.action?pageId=70227790}{CRC\_FFPE-CODEX\_CellNeighs}\cite{schurch2020codex} & 2020 & 2D & Histopathology (WSI) & Colorectal & 200 & No & NA & Colorectal Cancer \\ \hline
\datasetidx\label{data:histo_post_nat_brca} & \href{https://wiki.cancerimagingarchive.net/pages/viewpage.action?pageId=52758117}{Post-NAT-BRCA~\cite{martel2019assessment}} & 2019 & 2D & Histopathology (WSI) & Breast & 96 & Yes & Cls & Breast Cancer \\ \hline
\datasetidx\label{data:histo_osteosarcoma} & \href{https://wiki.cancerimagingarchive.net/pages/viewpage.action?pageId=52756935}{Osteosarcoma Tumor Assessment}\cite{arunachalam2019osteo} & 2019 & 2D & Histopathology (WSI) & Bone & 1.1k & Yes & Cls & Osteosarcoma \\ \hline
\datasetidx\label{data:histo_quilt1m} & \href{https://github.com/wisdomikezogwo/quilt1m}{Quilt-1M~\cite{ikezogwo2023quilt1m}} & 2023 & 2D & Histopathology (Patch) & Multi-organ & 1m & Yes & VQA & Multi-organ Pathology \\ \hline

& Overall & 2008$\sim$2023 & 2D & Multi & Full Body& 2.6m & NA & Multi & Multi \\ \hline
\end{tabular}
}
\vspace{2pt}
\begin{tablenotes}[flushleft]
\scriptsize
\item[a] Multi-modalities of CMB-LCA: CT, MR, US, Histopathology (WSI), DX.
\item[b] Multi-modalities of Prostate-MRI: MR, CT, PET, Histopathology (WSI).
\item[] \textbf{Abbreviations:} Seg=Segmentation, Det=Detection, Cls=Classification, Reg=Registration, Loc=Localization, IQA=Image Quality Assessment, \\VQA=Visual Question Answering.
\end{tablenotes}
\end{threeparttable}
\end{table*}

\begin{table*}[!htbp]
\small
\centering
\caption{2D microscopy datasets.}
\label{tab:2d_microscopy_datasets}
\begin{threeparttable}
\setlength{\tabcolsep}{4pt}      
\renewcommand{\arraystretch}{1.12} 
\resizebox{\textwidth}{!}{
\begin{tabular}{|c||>{\raggedright\arraybackslash}m{5.0cm}|r|c|c|c|c|c|c|>{\raggedright\arraybackslash}m{3.2cm}|}
\hline
\multicolumn{1}{|c||}{\textbf{\#}} &
\multicolumn{1}{c|}{\textbf{Dataset}} &
\multicolumn{1}{c|}{\textbf{Year}} &
\multicolumn{1}{c|}{\textbf{Dim}} &
\multicolumn{1}{c|}{\textbf{Modality}} &
\multicolumn{1}{c|}{\textbf{Structure}} &
\multicolumn{1}{c|}{\textbf{Images}} &
\multicolumn{1}{c|}{\textbf{Label}} &
\multicolumn{1}{c|}{\textbf{Task}} &
\multicolumn{1}{c|}{\textbf{Diseases}} \\
\hline
\datasetidx\label{data:micro_celltrack2019} & \href{http://celltrackingchallenge.net/}{CellTracking2019~\cite{mavska2023cell}} & 2019 & 2D & Microscopy & Cell & 1.4M & Yes & Tracking & NA \\ \hline
\datasetidx\label{data:micro_cremi} & \href{https://cremi.org}{CREMI}~\cite{cremi} & 2016 & 2D & Microscopy & Brain & 375 & Yes & Seg & NA \\ \hline
\datasetidx\label{data:micro_bacteria_detect} & \href{https://tianchi.aliyun.com/dataset/dataDetail?dataId=94411}{Bacteria Detection~\cite{wieczorek2024transformer}} & 2021 & 2D & Microscopy & NA & 366 & Yes & Seg & NA \\ \hline
\datasetidx\label{data:micro_blood_cell_tianchi} & \href{https://tianchi.aliyun.com/dataset/dataDetail?dataId=89038}{Blood Cell Images}~\cite{blood_cell_img} & 2021 & 2D & Microscopy & Blood & 12.5k & Yes & Cls & Blood \\ \hline
\datasetidx\label{data:micro_leukemia_tianchi} & \href{https://tianchi.aliyun.com/dataset/dataDetail?dataId=90101}{Leukemia Classification~\cite{DeepLeukemia2020}} & 2021 & 2D & Microscopy & NA & 15.1k & Yes & Cls & Leukemia \\ \hline
\datasetidx\label{data:micro_celltrack2021} & \href{http://celltrackingchallenge.net/}{CellTracking2021~\cite{mavska2023cell}} & 2021 & 2D+3D+Video & Microscopy & Cell & 0 & Yes & Tracking, Seg & Lung Disease \\ \hline
\datasetidx\label{data:micro_ball_cls} & \href{https://competitions.codalab.org/competitions/20395}{B-ALL Classification~\cite{gupta2019isbi}} & 2018 & 2D & Microscopy & Cell & 15.1k & Yes & Cls & Brain Tumor \\ \hline
\datasetidx\label{data:micro_dsb2018} & \href{https://www.kaggle.com/competitions/data-science-bowl-2018}{2018 Data Science Bowl~\cite{caicedo2019nucleus}} & 2018 & 2D & Microscopy & Nuclei & 670 & Yes & Seg & Skin Lesions \\ \hline
\datasetidx\label{data:micro_gsb2016} & \href{https://www.imageclef.org/2016/medical}{GSB2016}~\cite{GSB2016} & 2015 & 2D & Multi\textsuperscript{a} & Skin, Cell, Breast & 31k & Yes & Cls & Head \& Neck Tumor \\ \hline
\datasetidx\label{data:micro_occisc_semseg} & \href{https://cs.adelaide.edu.au/~carneiro/isbi14_challenge/}{OCCISC (SemSeg)~\cite{lu2016evaluation}} & 2014 & 2D & Microscopy & Cell & 945 & Yes & Seg & Cervical Cytology \\ \hline
\datasetidx\label{data:micro_iciar2018_micro} & \href{https://iciar2018-challenge.grand-challenge.org/Dataset/}{ICIAR 2018 (Microscopy)~\cite{aresta2019bach}} & 2017 & 2D & Microscopy & Breast & 400 & Yes & Cls & Breast Cancer \\ \hline
\datasetidx\label{data:micro_cbc_count} & \href{https://github.com/MahmudulAlam/Complete-Blood-Cell-Count-Dataset}{CBC (Counting)~\cite{alam2019machine}} & 2019 & 2D & Microscopy & Full Body & 420 & Yes & Reg & NA \\ \hline
\datasetidx\label{data:micro_hushem} & \href{https://opendatalab.com/HuSHeM}{HuSHeM~\cite{shaker2017dictionary}} & 2017 & 2D & Microscopy & Pelvic & 216 & Yes & Cls & Sperm Head Morphology \\ \hline
\datasetidx\label{data:micro_kaggle_hpa} & \href{https://www.kaggle.com/competitions/hpa-single-cell-image-classification/data}{Kaggle-HPA~\cite{le2022analysis}} & 2021 & 2D & Microscopy & NA & 89.5k & Yes & Seg & Protein Localization \\ \hline
\datasetidx\label{data:micro_nanni2016} & \href{https://figshare.com/s/d6fb591f1beb4f8efa6f}{nanni2016texture}~\cite{nanni2016texture} & 2016 & 2D & Microscopy & Retina & 195 & Yes & Cls & Cell Shape \\ \hline
\datasetidx\label{data:micro_corneal_endothelial} & \href{https://github.com/daboe01/SREP-18-33533B}{Corneal Endothelial Cell~\cite{CornealEndothelial2019}} & 2019 & 2D & Microscopy & Retina & 385 & Yes & Seg & NA \\ \hline
\datasetidx\label{data:micro_corneal_nerve} & \href{http://bioimlab.dei.unipd.it/Data\%20Sets.htm}{Corneal Nerve~\cite{DeBonnay2022}} & 2008 & 2D & Microscopy & Retina & 90 & Yes & Cls & Corneal Abnormalities \\ \hline
\datasetidx\label{data:micro_corneal_tortuosity} & \href{http://bioimlab.dei.unipd.it/Data\%20Sets.htm}{Corneal Nerve Tortuosity~\cite{scarpa2011automatic}} & 2011 & 2D & Microscopy & Retina & 30 & Yes & Cls & Nerve Tortuosity \\ \hline
\datasetidx\label{data:micro_cervix93} & \href{https://github.com/parham-ap/cytology_dataset}{Cervix93 Cytology~\cite{phoulady2018new}} & 2018 & 2D & Microscopy & Cervix & 93 & Yes & Cls & Cervical Cancer \\ \hline
\datasetidx\label{data:micro_dlbcl_morph} & \href{https://github.com/stanfordmlgroup/DLBCL-Morph}{DLBCL-Morph~\cite{vrabac2020dlbclmorph}} & 2020 & 2D & Microscopy & Retina & 152.2k & Yes & Reg & DLBCL Lymphoma \\ \hline
\datasetidx\label{data:micro_2pm_vessel} & \href{https://opendatalab.org.cn/2-PM_Vessel_Dataset}{2-PM Vessel Dataset~\cite{teikari2016deep}} & 2016 & 2D & Microscopy & Vessel & 12 & Yes & Seg & NA \\ \hline
\datasetidx\label{data:micro_bbbc041} & \href{https://opendatalab.org.cn/BBBC041/download}{BBBC041~\cite{li2021multi}} & 2012 & 2D & Microscopy & Cell & 1.3k & Yes & Seg & Malaria \\ \hline
\datasetidx\label{data:micro_fmd} & \href{https://github.com/yinhaoz/denoising-fluorescence}{FMD~\cite{zhang2018poisson}} & 2019 & 2D & Microscopy & Surface & 5.1k & Yes & Cls, Seg & Surface Defect \\ \hline
\datasetidx\label{data:micro_blood_cell_detect_heywhale} & \href{https://www.heywhale.com/mw/dataset/62c2af90913a54a66038165a}{Blood Cell Detection~\cite{BloodCellDetection2022}} & 2022 & 2D & Microscopy & NA & 874 & Yes & Det & NA \\ \hline
\datasetidx\label{data:micro_tuberculosis_heywhale} & \href{https://www.heywhale.com/mw/dataset/5efc4de063975d002c9792de}{Tuberculosis Image~\cite{rahman2020reliable}} & 2020 & 2D & Microscopy & NA & 1.3k & Yes & Det & Tuberculosis \\ \hline
\datasetidx\label{data:micro_mhsma} & \href{https://github.com/soroushj/mhsma-dataset}{MHSMA~\cite{javadi2019novel}} & 2019 & 2D & Microscopy & NA & 1.5k & Yes & Cls & NA \\ \hline
\datasetidx\label{data:micro_iciar2018_micro_seg} & \href{https://iciar2018-challenge.grand-challenge.org/}{ICIAR 2018 (Microscopy)~\cite{aresta2019bach}} & 2017 & 2D & Microscopy, WSI & NA & 400 & Yes & Seg & Breast Cancer \\ \hline
\datasetidx\label{data:micro_imageclef2016} & \href{https://www.imageclef.org/2016/medical}{ImageCLEF 2016~\cite{GSB2016}} & 2016 & 2D & Multi\textsuperscript{a} & Skin, Cell, Breast & 31k & Yes & Cls & NA \\ \hline
\datasetidx\label{data:micro_celltrack2024} & \href{http://celltrackingchallenge.net/}{CellTracking2024~\cite{mavska2023cell}} & 2024 & 2D+3D+Video & Microscopy & Cell & 0 & Yes & Tracking, Seg & NA \\ \hline
\datasetidx\label{data:micro_celltrack2022} & \href{http://celltrackingchallenge.net/}{CellTracking2022~\cite{mavska2023cell}} & 2022 & 2D+3D+Video & Microscopy & Cell & 0 & Yes & Tracking, Seg & NA \\ \hline
\datasetidx\label{data:micro_celltrack2023} & \href{http://celltrackingchallenge.net/}{CellTracking2023~\cite{mavska2023cell}} & 2023 & 2D+3D+Video & Microscopy & Cell & 0 & Yes & Tracking, Seg & NA \\ \hline
\datasetidx\label{data:micro_occisc_instseg} & \href{https://cs.adelaide.edu.au/~carneiro/isbi14_challenge/}{OCCISC (InstSeg)~\cite{alam2019machine}} & 2014 & 2D & Microscopy & Cell & 945 & Yes & Seg & NA \\ \hline
\datasetidx\label{data:micro_cbc_detect} & \href{https://github.com/MahmudulAlam/Complete-Blood-Cell-Count-Dataset}{CBC (Detection)~\cite{rahman2020reliable}} & 2019 & 2D & Microscopy & Full Body & 420 & Yes & Det & NA \\ \hline
\datasetidx\label{data:micro_iciar2018_wsi_seg} & \href{https://iciar2018-challenge.grand-challenge.org/}{ICIAR 2018 (WSI)~\cite{aresta2019bach}} & 2018 & 2D & Microscopy, WSI & NA & 400 & Yes & Seg & Breast Cancer \\ \hline
& Overall & 2008$\sim$2024 & 2D & Multi & Full Body& 1.8m & Yes & Multi & Multi \\ \hline
\end{tabular}
}
\vspace{2pt}
\begin{tablenotes}[flushleft]
\scriptsize
\item[a] Multi-modalities of GSB2016 and ImageCLEF 2016: MR, US, Histopathology, X-Ray, CT, PET, Endoscopy, Dermoscopy, EEG, ECG, \\EMG, Microscopy, Electron Microscopy, Fundus Photography.
\item[] \textbf{Abbreviations:} Seg=Segmentation, Det=Detection, Cls=Classification, Reg=Registration, Tracking=Tracking, WSI=Whole-Slide Images.
\end{tablenotes}
\end{threeparttable}
\end{table*}

\begin{table*}[!htbp]
\small
\centering
\caption{2D infrared datasets.}
\label{tab:2d_infrared_datasets}
\begin{threeparttable}
\setlength{\tabcolsep}{4pt}      
\renewcommand{\arraystretch}{1.12} 
\resizebox{\textwidth}{!}{
\begin{tabular}{|c||>{\raggedright\arraybackslash}m{5.0cm}|r|c|c|c|c|c|c|>{\raggedright\arraybackslash}m{3.2cm}|}
\hline
\multicolumn{1}{|c||}{\textbf{\#}} &
\multicolumn{1}{c|}{\textbf{Dataset}} &
\multicolumn{1}{c|}{\textbf{Year}} &
\multicolumn{1}{c|}{\textbf{Dim}} &
\multicolumn{1}{c|}{\textbf{Modality}} &
\multicolumn{1}{c|}{\textbf{Structure}} &
\multicolumn{1}{c|}{\textbf{Images}} &
\multicolumn{1}{c|}{\textbf{Label}} &
\multicolumn{1}{c|}{\textbf{Task}} &
\multicolumn{1}{c|}{\textbf{Diseases}} \\
\hline
\datasetidx\label{data:ir_ravir} & \href{https://ravir.grand-challenge.org/}{RAVIR~\cite{hatamizadeh2022ravir}} & 2022 & 2D & Infrared & Retina & 42 & Yes & Seg & Blood vessel \\ \hline
\datasetidx\label{data:ir_mrl_glasses} & \href{http://mrl.cs.vsb.cz/eyedataset}{MRL Eye Glasses cls~\cite{mrl_eyedataset}} & 2018 & 2D & Infrared & Retina & 84.9k & Yes & Cls & NA \\ \hline
\datasetidx\label{data:ir_mrl_state} & \href{http://mrl.cs.vsb.cz/eyedataset}{MRL Eye Eye state cls~\cite{mrl_eyedataset}} & 2018 & 2D & Infrared & Retina & 84.9k & Yes & Cls & NA \\ \hline
\datasetidx\label{data:ir_mrl_reflections} & \href{http://mrl.cs.vsb.cz/eyedataset}{MRL Eye Reflections cls~\cite{mrl_eyedataset}} & 2018 & 2D & Infrared & Retina & 84.9k & Yes & Cls & NA \\ \hline
\datasetidx\label{data:ir_mrl_quality} & \href{http://mrl.cs.vsb.cz/eyedataset}{MRL Eye Image quality cls~\cite{mrl_eyedataset}} & 2018 & 2D & Infrared & Retina & 84.9k & Yes & Cls & NA \\ \hline
\datasetidx\label{data:ir_mrl_sensor} & \href{http://mrl.cs.vsb.cz/eyedataset}{MRL Eye Sensor type cls~\cite{mrl_eyedataset}} & 2018 & 2D & Infrared & Retina & 84.9k & Yes & Cls & NA \\ \hline
& Overall & 2018$\sim$2022 & 2D & Infrared & Retina& 424.5k & Yes & Cls, Seg & Blood vessel \\ \hline
\end{tabular}
}
\vspace{2pt}
\begin{tablenotes}[flushleft]
\scriptsize
\item[] \textbf{Abbreviations:} Seg=Segmentation, Cls=Classification.
\end{tablenotes}
\end{threeparttable}
\end{table*}

\begin{table*}[!htbp]
\small
\centering
\caption{2D endoscopy datasets.}
\label{tab:2d_endoscopy_datasets}
\begin{threeparttable}
\setlength{\tabcolsep}{4pt}      
\renewcommand{\arraystretch}{1.12} 
\resizebox{\textwidth}{!}{
\begin{tabular}{|c||>{\raggedright\arraybackslash}m{5.8cm}|c|c|>{\raggedright\arraybackslash}m{2.4cm}|>{\raggedright\arraybackslash}m{3.2cm}|r|c|p{2.2cm}|>{\raggedright\arraybackslash}m{3.4cm}|}
\hline
\multicolumn{1}{|c||}{\textbf{\#}} &
\multicolumn{1}{c|}{\textbf{Dataset}} &
\multicolumn{1}{c|}{\textbf{Year}} &
\multicolumn{1}{c|}{\textbf{Dim}} &
\multicolumn{1}{c|}{\textbf{Modality}} &
\multicolumn{1}{c|}{\textbf{Structure}} &
\multicolumn{1}{c|}{\textbf{Images}} &
\multicolumn{1}{c|}{\textbf{Label}} &
\multicolumn{1}{c|}{\textbf{Task}} &
\multicolumn{1}{c|}{\textbf{Diseases}} \\
\hline
\datasetidx\label{data:endo_kavsir} & \href{https://datasets.simula.no/kvasir/}{Kavsir\cite{pogorelov2017kvasir}} & 2017 & 2D & Endoscopy & Colon & 14k & Yes & Cls & NA \\ \hline
\datasetidx\label{data:endo_endoslam} & \href{https://github.com/CapsuleEndoscope/EndoSLAM}{EndoSlam\cite{ozyoruk2020endoslam}} & 2021 & 2D & Endoscopy & Colon, Liver, Stomach, Kidney & 76.8k & Yes & Recon, Est & NA \\ \hline
\datasetidx\label{data:endo_saras_mesad} & \href{https://saras-mesad.grand-challenge.org/Home/}{SARAS-MESAD\cite{bawa2021saras}} & 2021 & 2D & Endoscopy & Prostate & 50.3k & No & Det & GI disease \\ \hline
\datasetidx\label{data:endo_ead19} & \href{https://ead2019.grand-challenge.org/Data/}{EAD19\cite{ali2019endoscopy}} & 2018 & 2D & Endoscopy & Stomach, Bladder, Colon & 2.1k & Yes & Det & Endo Artifact \\ \hline
\datasetidx\label{data:endo_endocv2020_sub1} & \href{https://endocv.grand-challenge.org/}{EndoCV2020-Sub Challenge1\cite{ali2022endoscopic}} & 2019 & 2D & Endoscopy & Colon & 2.3k & Yes & Det, Seg & Polyp \\ \hline
\datasetidx\label{data:endo_endovis15} & \href{https://polyp.grand-challenge.org/}{EndoVis15\cite{bernal2017comparative}} & 2015 & 2D & Endoscopy & Colon & 612 & Yes & Seg & Polyp \\ \hline
\datasetidx\label{data:endo_m2cai16_tool} & \href{http://camma.u-strasbg.fr/m2cai2016/}{Surgical tool detection challenge (m2cai16-tool)\cite{twinanda2016single}} & 2016 & 2D & Endoscopy & Gallbladder & 15 & Yes & Det & NA \\ \hline
\datasetidx\label{data:endo_aida_e1} & \href{https://aidasub-cleceliachy.grand-challenge.org/}{AIDA-E\_1}~\cite{aida_e_1} & 2015 & 2D & Endoscopy & Stomach, Liver & 181 & Yes & Cls & Celiac Disease \\ \hline
\datasetidx\label{data:endo_aida_e2} & \href{https://aidasub-clebarrett.grand-challenge.org/home/}{AIDA-E\_2}~\cite{aida_e_2} & 2015 & 2D & Endoscopy & Esophagus & 157 & Yes & Cls & Barrett's Esophagus \\ \hline
\datasetidx\label{data:endo_aida_e3} & \href{https://aidasub-chromogastro.grand-challenge.org/home/}{AIDA-E\_3}~\cite{aida_e_3} & 2015 & 2D & Endoscopy & Stomach, Colon & 88 & Yes & Cls & Metaplasia, Dysplasia \\ \hline
\datasetidx\label{data:endo_cvc_clinicdb} & \href{https://tianchi.aliyun.com/dataset/dataDetail?dataId=93690}{CVC-ClinicDB\cite{vazquez2017benchmark}} & 2021 & 2D & Endoscopy & Bowel & 1.4k & Yes & Seg & Polyp \\ \hline
\datasetidx\label{data:endo_kvasir_seg} & \href{https://tianchi.aliyun.com/dataset/dataDetail?dataId=84385}{Kvasir-SEG\cite{jha2019kvasir}} & 2020 & 2D & Endoscopy & Bowel & 8k & Yes & Seg & NA \\ \hline
\datasetidx\label{data:endo_fetreg} & \href{https://www.synapse.org/\#!Synapse:syn25313156/wiki/610166}{FetReg\cite{bano2021fetreg}} & 2022 & 2D & Endoscopy & Uterus & 2.7k & Yes & Seg & Placental Vasculature \\ \hline
\datasetidx\label{data:endo_saras_esad} & \href{https://saras-esad.grand-challenge.org}{SARAS-ESAD\cite{bawa2021saras}} & 2020 & 2D & Endoscopy & Bowel & 33.4k & Yes & Det & Skin lesion \\ \hline
\datasetidx\label{data:endo_imageclef2016} & \href{https://www.imageclef.org/2016/medical}{ImageCLEF 2016\cite{deherrera2016imageclef}} & 2015 & 2D & Multi\textsuperscript{a} & Skin, Cell, Breast & 31k & Yes & Cls & H\&N tumor \\ \hline
\datasetidx\label{data:endo_isbi_aida_ceci} & \href{https://aidasub-cleceliachy.grand-challenge.org/}{ISBI-AIDA-CECI} & 2015 & 2D & Endoscopy & Liver, Stomach & 181 & Yes & Cls & Celiac diseases \\ \hline
\datasetidx\label{data:endo_sun_seg} & \href{https://github.com/GewelsJI/VPS}{SUN\_SEG\cite{ji2022video}} & 2022 & 2D+Video & Endoscopy & Colon & 49.1k & Yes & Seg, Det, Cls & Polyp \\ \hline
\datasetidx\label{data:endo_hyperkvasir} & \href{https://datasets.simula.no/hyper-kvasir/}{HyperKvasir\cite{borgli2020hyperkvasir}} & 2020 & 2D+Video & Endoscopy & Esophagus, Stomach, Colon & 6.5k & Yes & Cls, Caption, Loc & GI disease \\ \hline
\datasetidx\label{data:endo_giana} & \href{https://giana.grand-challenge.org/}{Gastrointestinal Image ANAlysis (GIANA)}~\cite{giana_1} & 2016 & 2D & Endoscopy & Colon & 600 & Yes & Cls & Vascular Malformation \\ \hline
\datasetidx\label{data:endo_endovis15_dagi} & \href{https://endovissub-abnormal.grand-challenge.org/EndoVisSub-Abnormal/}{EndoVis 2015 - DAGI}~\cite{endovis_dagi} & 2015 & 2D & Endoscopy & NA & 389 & Yes & Det & Cholecystectomy \\ \hline
\datasetidx\label{data:endo_endovis15_ebcd} & \href{https://endovissub-barrett.grand-challenge.org/}{EndoVis 2015 - EBCD}~\cite{endovis_ebcd} & 2015 & 2D & Endoscopy & NA & 150 & Yes & Seg & Barrett's Epithelium \\ \hline
\datasetidx\label{data:endo_endocv2020_sub2} & \href{https://edd2020.grand-challenge.org/}{EndoCV2020-Sub Challenge2\cite{ali2020endoscopy}} & 2019 & 2D & Endoscopy & NA & 386 & Yes & Det & NA \\ \hline
\datasetidx\label{data:endo_endovis15_apdcv} & \href{https://polyp.grand-challenge.org/}{EndoVis 2015 - APDCV\cite{bernal2017comparative}} & 2015 & 2D & Endoscopy & NA & 612 & Yes & Seg & Colonic Polyp \\ \hline
\datasetidx\label{data:endo_endovis15_ist} & \href{https://endovissub-instrument.grand-challenge.org/EndoVisSub-Instrument/}{EndoVis 2015 - IST\_2D-Endoscopy}~\cite{endovis_ist} & 2015 & 2D+Video & Endoscopy & NA & 100 & Yes & Seg & Surgical Instruments \\ \hline
\datasetidx\label{data:endo_endovis18_rss} & \href{https://endovissub2018-roboticscenesegmentation.grand-challenge.org/home/}{EndoVis 2018 - RSS\cite{allan20202018roboticscenesegmentation}} & 2018 & 2D & Endoscopy & NA & 2.8k & Yes & Seg & Surgical Instruments \\ \hline
\datasetidx\label{data:endo_isbi_aida_emibs} & \href{https://isbi-aida.grand-challenge.org}{ISBI-AIDA-EMIBS} & 2015 & 2D & Endoscopy & NA & 262 & Yes & Cls & Gastric \\ \hline
\datasetidx\label{data:endo_isbi_aida_gcics} & \href{https://isbi-aida.grand-challenge.org}{ISBI-AIDA-GCICS} & 2015 & 2D & Endoscopy & NA & 176 & Yes & Cls & Gastric \\ \hline
\datasetidx\label{data:endo_endovis2023_sims} & \href{https://www.synapse.org/\#!Synapse:syn47193563/wiki/620035}{EndoVis2023-SIMS}~\cite{endovis_sims} & 2023 & 2D & Endoscopy & NA & 0 & Yes & Seg & Endoscopy \\ \hline
\datasetidx\label{data:endo_endovis2023_syniss} & \href{https://www.synapse.org/\#!Synapse:syn50908388/wiki/620516}{EndoVis2023-Syn-ISS}~\cite{endovis_syniss} & 2023 & 2D & Endoscopy & NA & 0 & Yes & Seg & NA \\ \hline
\datasetidx\label{data:endo_p2ilf} & \href{https://p2ilf.grand-challenge.org/}{P2ILF}~\cite{p2ilf} & 2022 & 2D+3D & Endoscopy & NA & 15 & Yes & Reg & Multi-organ \\ \hline
\datasetidx\label{data:endo_endovis2023_surgripe} & \href{https://www.synapse.org/\#!Synapse:syn51471789/wiki/622255}{EndoVis2023-SurgRIPE\cite{xu2025surgripe}} & 2023 & 2D & Endoscopy & NA & 0 & Yes & Est & NA \\ \hline
\datasetidx\label{data:endo_m2caiseg} & \href{https://www.kaggle.com/datasets/salmanmaq/m2caiseg}{m2caiSeg\cite{maqbool2020m2caiseg}} & 2020 & 2D & Endoscopy & Instrument & 614 & Yes & Seg & NA \\ \hline
\datasetidx\label{data:endo_cvc_endoscenestill} & \href{https://www.biobancovasco.org/en/Sample-and-data-catalog/Databases/PD178-PICCOLO-EN.html}{CVC-EndoSceneStill\cite{vazquez2017benchmark}} & NA & 2D & Endoscopy & NA & 3.4k & Yes & Seg & Polyp \\ \hline
\datasetidx\label{data:endo_endofm} & \href{https://github.com/med-air/Endo-FM}{Endo-FM\cite{wang2023foundation}} & NA & 2D+Video & Endoscopy & NA & 0 & Yes & Seg, Cls, Det & NA \\ \hline
\datasetidx\label{data:endo_segstrong_c} & \href{https://github.com/hding2455/CaRTS}{SegSTRONG-C\cite{ding2024segstrong}} & NA & 2D+Video & Endoscopy & NA & 17 & Yes & Seg & NA \\ \hline
\datasetidx\label{data:endo_segcol} & \href{https://www.synapse.org/\#!Synapse:syn54124209}{SegCol\cite{ju2024segcol}} & NA & 2D+Video & Colposcopy, Endoscopy & NA & 78 & Yes & Seg & NA \\ \hline
\datasetidx\label{data:endo_fedsurg} & \href{https://www.synapse.org/Synapse:syn53137385/wiki/625370}{FedSurg}~\cite{fedsurg_1} & 2024 & 2D+Video & Endoscopy & NA & 30 & Yes & Cls & Laparoscopic appendectomy \\ \hline
& Overall & 2015$\sim$2024 & 2D & Multi & Full Body& 288.5k & Yes & Multi & Multi \\ \hline
\end{tabular}
}
\vspace{2pt}
\begin{tablenotes}[flushleft]
\scriptsize
\item[a] Multi-modalities of ImageCLEF 2016: MR, Ultrasound, Histopathology, X-Ray, CT, PET, Endoscopy, Dermoscopy, Others, EEG, ECG, EMG, Electron Microscopy, Fundus Photography.
\item[] \textbf{Abbreviations:} Seg=Segmentation, Det=Detection, Cls=Classification, Recon=Reconstruction, Reg=Registration, Loc=Localization, Est=Estimation, GI=Gastrointestinal, H\&N=Head \& Neck.
\end{tablenotes}
\end{threeparttable}
\end{table*}

\begin{table*}[!htbp]
\small
\centering
\caption{2D datasets of the other modalities.}
\label{tab:2d_others_datasets}
\begin{threeparttable}
\setlength{\tabcolsep}{4pt}      
\renewcommand{\arraystretch}{1.12} 
\resizebox{\textwidth}{!}{
\begin{tabular}{|c||>{\raggedright\arraybackslash}m{5.8cm}|r|c|>{\raggedright\arraybackslash}m{2.2cm}|>{\raggedright\arraybackslash}m{1.8cm}|r|c|c|>{\raggedright\arraybackslash}m{3.2cm}|}
\hline
\multicolumn{1}{|c||}{\textbf{\#}} &
\multicolumn{1}{c|}{\textbf{Dataset}} &
\multicolumn{1}{c|}{\textbf{Year}} &
\multicolumn{1}{c|}{\textbf{Dim}} &
\multicolumn{1}{c|}{\textbf{Modality}} &
\multicolumn{1}{c|}{\textbf{Structure}} &
\multicolumn{1}{c|}{\textbf{Images}} &
\multicolumn{1}{c|}{\textbf{Label}} &
\multicolumn{1}{c|}{\textbf{Task}} &
\multicolumn{1}{c|}{\textbf{Diseases}} \\
\hline
\datasetidx\label{data:other_intel_mobileodt} & \href{https://www.kaggle.com/competitions/intel-mobileodt-cervical-cancer-screening/overview}{Intel \& MobileODT Cervical Cancer Screening~\cite{intel_mobileodt_kaggle_2017}} & 2017 & 2D & Colposcopy & Vagina & 1,993 & Yes & Cls & Cervical cancer \\ \hline
\datasetidx\label{data:other_addi_alz} & \href{https://www.aicrowd.com/challenges/addi-alzheimers-detection-challenge}{ADDI ALZHEIMER'S DETECTION CHALLENGE~\cite{addi_alz_2021}} & 2021 & 2D & Series Data & NA & 34,614 & Yes & Cls & Lung adenocarcinoma \\ \hline
\datasetidx\label{data:other_digital_mammo} & \href{https://www.synapse.org/Synapse:syn4224222/wiki/401743}{The Digital Mammography DREAM Challenge~\cite{schaffter2020dream}} & 2016 & 2D & Mammography & Breast & 640,000 & Yes & Cls & Atrophic AMD \\ \hline
\datasetidx\label{data:other_bigneuron} & \href{https://alleninstitute.org/bigneuron/about/}{BigNeuron~\cite{peng2015bigneuron}} & 2016 & 2D & NA & Brain & 2,166 & Yes & Recon & Diabetic foot ulcer \\ \hline
\datasetidx\label{data:other_human_activity_radar} & \href{https://humanactivitiyclassificationwithradar.grand-challenge.org}{Human Activity Classification with Radar~\cite{khan2023radar}} & 2019 & 2D & Others & NA & 1,854 & Yes & Cls & NA \\
\hline
\end{tabular}
}
\vspace{2pt}
\begin{tablenotes}[flushleft]
\scriptsize
\item[a] Includes anatomical structures like optic nerve, eyeball, etc.
\item[] \textbf{Abbreviations:} Seg=Segmentation, Det=Detection, Cls=Classification, Recon=Reconstruction.
\end{tablenotes}
\end{threeparttable}
\end{table*}
\section{Tables of 3D Medical Image Datasets}

\FloatBarrier 

{\tiny
\renewcommand{\arraystretch}{1.12}
\setlength{\tabcolsep}{4pt}

\begin{longtable}{|c||>{\RaggedRight\arraybackslash}m{2.2cm}|c|c|>{\RaggedRight\arraybackslash}m{1.2cm}|>{\RaggedRight\arraybackslash}m{1.4cm}|c|c|c|>{\RaggedRight\arraybackslash}m{2.0cm}|}
\caption{3D CT datasets}\label{tab:3d_ct_datasets}\\
\hline
\multicolumn{1}{|c||}{\textbf{\#}} &
\multicolumn{1}{c|}{\textbf{Dataset}} &
\multicolumn{1}{c|}{\textbf{Year}} &
\multicolumn{1}{c|}{\textbf{Dim}} &
\multicolumn{1}{c|}{\textbf{Modality}} &
\multicolumn{1}{c|}{\textbf{Structure}} &
\multicolumn{1}{c|}{\textbf{Volumes}} &
\multicolumn{1}{c|}{\textbf{Label}} &
\multicolumn{1}{c|}{\textbf{Task}} &
\multicolumn{1}{c|}{\textbf{Diseases}} \\
\hline
\datasetidx \label{data:totalsegmentatordata} & \href{https://totalsegmentator.com/}{TotalSegmentator\cite{wasserthal2023totalsegmentator}} & 2022 & 3D & CT & Full Body & 1204 & Yes & Seg & Varied pathologies \\
\hline
\datasetidx \label{data:autopet-ctpet} & \href{https://autopet.grand-challenge.org/}{AutoPET\cite{gatidis2022a}} & 2022 & 3D & CT/PET & Whole-body & 1014 & Yes & Seg & Cancer \\
\hline
\datasetidx \label{data:autopetii} & \href{https://autopet-ii.grand-challenge.org/}{AutoPET II\cite{gatidis2023autopetii}} & 2023 & 3D & CT/ PET & Whole-body & 1219 & Yes & Seg & Cancer \\
\hline
\datasetidx \label{data:uls23} & \href{https://uls23.grand-challenge.org/uls23/}{ULS23\cite{degrauw2025the}} & 2023 & 3D & CT & Chest, Abdomen, Pelvis & 39468 & Yes & Seg & Oncological lesions \\
\hline
\end{longtable}
}

{\tiny
\renewcommand{\arraystretch}{1.12}
\setlength{\tabcolsep}{4pt}

\begin{longtable}{|c||>{\RaggedRight\arraybackslash}m{2.6cm}|c|c|>{\RaggedRight\arraybackslash}m{1.6cm}|>{\RaggedRight\arraybackslash}m{1.0cm}|c|c|c|>{\RaggedRight\arraybackslash}m{2.0cm}|}
    \caption{3D MR datasets}\label{tab:3d_mr_datasets}\\
    \hline
    \multicolumn{1}{|c||}{\textbf{\#}} &
    \multicolumn{1}{c|}{\textbf{Dataset}} &
    \multicolumn{1}{c|}{\textbf{Year}} &
    \multicolumn{1}{c|}{\textbf{Dim}} &
    \multicolumn{1}{c|}{\textbf{Modality}} &
    \multicolumn{1}{c|}{\textbf{Structure}} &
    \multicolumn{1}{c|}{\textbf{Volumes}} &
    \multicolumn{1}{c|}{\textbf{Label}} &
    \multicolumn{1}{c|}{\textbf{Task}} &
    \multicolumn{1}{c|}{\textbf{Diseases}} \\
    \hline
    

\datasetidx \label{data:totalsegmentatormri} & \href{https://zenodo.org/doi/10.5281/zenodo.11367004}{TotalSegmentator MRI\cite{akincidantonoli2025totalsegmentator}} & 2024 & 3D & MR & Whole-body & 616 & Yes & Seg & Various pathologies \\
\hline
\datasetidx \label{data:acdc} & \href{https://www.creatis.insa-lyon.fr/Challenge/acdc/}{ACDC\cite{bernard2018deep}} & 2017 & 3D & MR & Heart & 150 & Yes & Seg/ Cls & Cardiac conditions \\
\hline
\datasetidx \label{data:mms} & \href{https://www.ub.edu/mnms/}{M\&Ms\cite{campello2021multicentre}} & 2020 & 3D & MR & Heart & 375 & Yes & Seg & Cardiac diseases \\
\hline
\datasetidx \label{data:mms2} & \href{https://www.ub.edu/mnms-2/}{M\&Ms-2\cite{camara2021statistical}} & 2021 & 3D & MR & Heart & 360 & Yes & Seg & Cardiac conditions \\
\hline
\end{longtable}
}

{\tiny
\renewcommand{\arraystretch}{1.12}
\setlength{\tabcolsep}{4pt}

\begin{longtable}{|c||>{\RaggedRight\arraybackslash}m{2.1cm}|c|c|>{\RaggedRight\arraybackslash}m{1.6cm}|>{\RaggedRight\arraybackslash}m{1.6cm}|c|c|c|>{\RaggedRight\arraybackslash}m{1.7cm}|}
    \caption{3D US datasets}\label{tab:3d_us_datasets}\\
    \hline
    \multicolumn{1}{|c||}{\textbf{\#}} &
    \multicolumn{1}{c|}{\textbf{Dataset}} &
    \multicolumn{1}{c|}{\textbf{Year}} &
    \multicolumn{1}{c|}{\textbf{Dim}} &
    \multicolumn{1}{c|}{\textbf{Modality}} &
    \multicolumn{1}{c|}{\textbf{Structure}} &
    \multicolumn{1}{c|}{\textbf{Volumes}} &
    \multicolumn{1}{c|}{\textbf{Label}} &
    \multicolumn{1}{c|}{\textbf{Task}} &
    \multicolumn{1}{c|}{\textbf{Diseases}} \\
    \hline

\datasetidx \label{data:tdscabus2023} & \href{https://tdsc-abus2023.grand-challenge.org/}{TDSC-ABUS2023\cite{luo2025tumor}} & 2023 & 3D & US & Breast & 200 & Yes & Seg/ Cls/ Det & Breast cancer \\
\hline
\datasetidx \label{data:cetus2014} & \href{https://www.creatis.insa-lyon.fr/Challenge/CETUS/index.html}{CETUS2014\cite{bernard2014challenge}} & 2014 & 3D & US & Heart & 45 & Yes & Seg & Cardiac conditions \\
\hline
\datasetidx \label{data:mvseg3dtee2023} & \href{https://www.synapse.org/\#!Synapse:syn51186045/wiki/621356}{MVSeg-3DTEE2023\cite{carnahan2023towards}} & 2023 & 3D & US & Mitral valve & 175 & Yes & Seg & Mitral valve disease \\
\hline
\datasetidx \label{data:aren0532-us} & \href{https://www.cancerimagingarchive.net/collection/aren0532/}{AREN0532\cite{aren0532dataset}} & 2022 & 2D/3D & US/ CR/ CT/ MR/ PET/ RTIMAGE & Kidney & 544 & No & Cls & Wilms tumor \\
\hline
\end{longtable}
}


{\tiny
\renewcommand{\arraystretch}{1.12}
\setlength{\tabcolsep}{4pt}

\begin{longtable}{|c||>{\RaggedRight\arraybackslash}m{2.3cm}|c|c|>{\RaggedRight\arraybackslash}m{1.6cm}|>{\RaggedRight\arraybackslash}m{1.6cm}|c|c|c|>{\RaggedRight\arraybackslash}m{2.0cm}|}
    \caption{3D PET datasets}\label{tab:3d_pet_datasets}\\
    \hline
    \multicolumn{1}{|c||}{\textbf{\#}} &
    \multicolumn{1}{c|}{\textbf{Dataset}} &
    \multicolumn{1}{c|}{\textbf{Year}} &
    \multicolumn{1}{c|}{\textbf{Dim}} &
    \multicolumn{1}{c|}{\textbf{Modality}} &
    \multicolumn{1}{c|}{\textbf{Structure}} &
    \multicolumn{1}{c|}{\textbf{Volumes}} &
    \multicolumn{1}{c|}{\textbf{Label}} &
    \multicolumn{1}{c|}{\textbf{Task}} &
    \multicolumn{1}{c|}{\textbf{Diseases}} \\
    \hline

\datasetidx \label{data:qinpetphantom} & \href{https://www.cancerimagingarchive.net/collection/qin-pet-phantom/}{QIN PET Phantom\cite{beichel2015qin}} & 2014 & 3D & PET & Body & 2 & Yes & Seg & Phantom study \\
\hline
\datasetidx \label{data:autopet-ctpet-pet} & \href{https://autopet.grand-challenge.org/}{AutoPET\cite{gatidis2022a}} & 2022 & 3D & PET/CT & Whole-body & 1014 & Yes & Seg & Cancer \\
\hline
\datasetidx \label{data:autopetii-pet} & \href{https://autopet-ii.grand-challenge.org/}{AutoPET II\cite{gatidis2023autopetii}} & 2023 & 3D & PET/CT & Whole-body & 1219 & Yes & Seg & Cancer \\
\hline
\datasetidx \label{data:covid19nysbu-pet} & \href{https://www.cancerimagingarchive.net/collection/covid-19-ny-sbu/}{COVID-19-NY-SBU\cite{saltz2021stony}} & 2021 & 2D/ 3D & PET/CT/ MR/ X-RAY & Brain/ Chest & 1384 & No & Cls & COVID-19 \\
\hline
\end{longtable}
}

{\tiny
\renewcommand{\arraystretch}{1.12}
\setlength{\tabcolsep}{4pt}

\begin{longtable}{|c||>{\RaggedRight\arraybackslash}m{2.6cm}|c|c|>{\RaggedRight\arraybackslash}m{1.8cm}|>{\RaggedRight\arraybackslash}m{1.0cm}|c|c|c|>{\RaggedRight\arraybackslash}m{1.8cm}|}
    \caption{3D Other datasets}\label{tab:3d_other_datasets}\\
    \hline
    \multicolumn{1}{|c||}{\textbf{\#}} &
    \multicolumn{1}{c|}{\textbf{Dataset}} &
    \multicolumn{1}{c|}{\textbf{Year}} &
    \multicolumn{1}{c|}{\textbf{Dim}} &
    \multicolumn{1}{c|}{\textbf{Modality}} &
    \multicolumn{1}{c|}{\textbf{Structure}} &
    \multicolumn{1}{c|}{\textbf{Volumes}} &
    \multicolumn{1}{c|}{\textbf{Label}} &
    \multicolumn{1}{c|}{\textbf{Task}} &
    \multicolumn{1}{c|}{\textbf{Diseases}} \\
    \hline

\datasetidx \label{data:mitoem} & \href{https://mitoem.grand-challenge.org/}{MitoEM\cite{wei2020mitoem}} & 2020 & 3D & 3D MICROSCOPY & Brain & 2 & Yes & Seg & Mitochondrial ultrastructure \\
\hline
\datasetidx \label{data:3dplateletem} & \href{https://leapmanlab.github.io/dense-cell/}{3D Platelet EM\cite{guay2021dense}} & 2021 & 3D & 3D MICROSCOPY & Platelet & 2 & Yes & Seg & Platelet ultrastructure \\
\hline
\datasetidx \label{data:pcabx3dpathology} & \href{https://www.cancerimagingarchive.net/collection/pca_bx_3dpathology/}{PCa Bx 3Dpathology\cite{xie2023d}} & 2023 & 3D & 3D MICROSCOPY & Prostate & 50 & Yes & Seg/ Loc & Prostate cancer \\
\hline
\datasetidx \label{data:cada} & \href{https://cada.grand-challenge.org/}{CADA\cite{ivantsits2021cerebral}} & 2020 & 3D & 3D DSA & Brain & 131 & Yes & Det/ Seg/ Cls & Cerebral aneurysms \\
\hline
\end{longtable}
}

\section{Tables of Medical Video Datasets}

\FloatBarrier 

{
\tiny
\renewcommand{\arraystretch}{1.12}
\setlength{\tabcolsep}{2pt}

\begin{longtable}{@{}%
|l||
>{\RaggedRight\arraybackslash}m{2.4cm}|
>{\RaggedRight\arraybackslash}m{1cm}|
>{\RaggedRight\arraybackslash}m{1cm}|
>{\RaggedRight\arraybackslash}m{1.0cm}|
>{\RaggedRight\arraybackslash}m{1.5cm}|
>{\RaggedRight\arraybackslash}m{1.0cm}|
>{\RaggedRight\arraybackslash}m{1.0cm}|
>{\RaggedRight\arraybackslash}m{1.5cm}|
>{\RaggedRight\arraybackslash}m{1.5cm}|
}

\caption{Video datasets.\label{tab:video_alldata_v2}}\\
\hline
\# & \textbf{Dataset} & \textbf{Year} & \textbf{Dim} &
\textbf{Modality} & \textbf{Structure} & \textbf{Num of samples} & \textbf{Label} & \textbf{Task} & \textbf{Diseases} \\

\hline

\datasetidx\label{data:CholecT50} & \href{https://cholectriplet2021.grand-challenge.org/}{CholecT50}~\cite{nwoye2023cholectriplet2022} & 2023 & Video & Endoscopy & Gallbladder & 50 & Yes & Cls, Det  & surgical instrument, action, target \\ \hline

\datasetidx\label{data:CholecTriplet 2021} & \href{https://cholectriplet2021.grand-challenge.org}{CholecTriplet 2021}~\cite{nwoye2023cholectriplet2021} & 2021 & Video & Endoscopy & Gallbladder & 45 & Yes & Cls, Det & surgical instrument, action, target  \\ \hline

\datasetidx\label{data:SurgVisDom} & \href{https://surgvisdom.grand-challenge.org/}{SurgVisDom}~\cite{zia2021surgical} & 2020 & Video & Endoscopy & Bowel & 488 & Yes & Cls & skin lesion  \\ \hline

\datasetidx\label{data:CATARACTS} & \href{https://cataracts.grand-challenge.org/}{CATARACTS}~\cite{al2019cataracts} & 2017 & Video & Microscopy & Retina & 50 & Yes & Cls, Det   & surgical workflow \\
\hline

\end{longtable}
\begin{tablenotes}[flushleft]
\scriptsize
\item[a] \textbf{Abbreviations:} Seg=Segmentation, Cls=Classification, Pred=Prediction, Det=Detection, Recon=Reconstruction, Reg=Registration, \\ Est=Estimation, VQA=Visual Question Answering.
\end{tablenotes}
}

\end{document}